\documentclass[preprint,12pt,3p]{elsarticle}
\usepackage[utf8]{inputenc}
\usepackage{multirow}
\usepackage{array}
\usepackage[lofdepth,lotdepth]{subfig}
\usepackage[ruled]{algorithm2e}
\usepackage{times,amsmath,amssymb}
\usepackage[T1]{fontenc}
\usepackage[polutonikogreek,english]{babel}
\usepackage{float}
 \usepackage{appendix}

\newenvironment{Sub-algorithm}[1][htb]
  {% Update algorithm name
   \begin{algorithm}[#1]%
  }{\end{algorithm}}
\usepackage{setspace}
\usepackage{amssymb}
\journal{pattern recognition}

\begin{document}

\begin{frontmatter}

\title{Graph-Embedded Subspace Support Vector Data Description}

\address[label1]{Faculty of Information Technology and Communication Sciences, Tampere University, FI-33720 Tampere, Finland}
\address[label3]{Department of Electrical and Computer Engineering, Aarhus University, DK-8200 Aarhus, Denmark}
\address[label2]{Programme for Environmental Information, Finnish Environment Institute, FI-40500 Jyväskylä, Finland}

\author[label1]{Fahad Sohrab\corref{cor1}}
\cortext[cor1]{corresponding author}
\ead{fahad.sohrab@tuni.fi}
\author[label3]{Alexandros Iosifidis}
\ead{ai@ece.au.dk}
\author[label1]{Moncef Gabbouj}
\ead{moncef.gabbouj@tuni.fi}
\author[label2]{Jenni Raitoharju}
\ead{jenni.raitoharju@syke.fi}

\begin{abstract}
In this paper, we propose a novel subspace learning framework for one-class classification. The proposed framework presents the problem in the form of graph embedding. It includes the previously proposed subspace one-class techniques as its special cases and provides further insight on what these techniques actually optimize. The framework allows to incorporate other meaningful optimization goals via the graph preserving criterion and reveals a spectral solution and a spectral regression-based solution as alternatives to the previously used gradient-based technique. We combine the subspace learning framework iteratively with Support Vector Data Description applied in the subspace to formulate Graph-Embedded Subspace Support Vector Data Description. We experimentally analyzed the performance of newly proposed different variants. We demonstrate improved performance against the baselines and the recently proposed subspace learning methods for one-class classification.
\end{abstract}

\begin{keyword}
%% keywords here, in the form: keyword \sep keyword
One-Class Classification \sep Support Vector Data Description \sep Subspace Learning \sep Spectral Regression
%% MSC codes here, in the form: \MSC code \sep code
%% or \MSC[2008] code \sep code (2000 is the default)
\end{keyword}

\end{frontmatter}

%%
%% Start line numbering here if you want
%%
% \linenumbers

%% main text

\section{Introduction}
Dimensionality reduction has been an important and active research area in the field of machine learning and data science. The aim is to enhance the performance of a specific application by transforming the data from its original feature space to a lower-dimensional subspace. Dimensionality reduction has been used effectively as a tool in applications ranging from traditional data analysis and classification to many modern applications such as video analytics, recommendation system design, and detecting anomalies in computer and social networks \cite{vaswani2018robust}. 

The three main application domains of dimensionality reduction algorithms are feature matching, model interpretation, and data representation \cite{xu2016sliced}. In feature matching, the aim is to find the similarity between two or more objects via a distance metric such as the Euclidean distance \cite{guo2009based}. The model interpretation is enhanced by reducing the number of variables in the subspace by dimensionality reduction methods \cite{rodriguez2014sequential}. In data representation applications, dimensionality reduction methods are used to better represent the data in a lower dimensional space for the task at hand \cite{he2015optimal}.

The approaches used for dimensionality reduction can be either supervised or unsupervised. In supervised learning, the algorithm relies mainly on the structure of data, and the mapping function is inferred from a set of labeled training samples. For example, Fisher’s Linear Discriminant Analysis (LDA) is an example of a supervised method that exhibits good discrimination qualities. LDA maximizes the between-class scatter and minimizes the within-class scatter. In unsupervised learning, the algorithm does not leverage the information of pre-existing labels. For example, Principal Component Analysis (PCA) is a well-known unsupervised method for dimensionality reduction. PCA extracts the dominant features of a high-dimensional data and represents it by a small number of orthogonal basis vectors, i.e., the principal components. Numerous extensions and applications of PCA and LDA have been proposed in the literature \cite{xu2021saliency,lim2021principal}, and it has been shown that LDA can outperform PCA when the training data set is large \cite{sheikh2020recognizing}. However, for large-scale datasets, the computation and memory problems, particularly for the eigen-decomposition step of LDA, can be cumbersome. The spectral regression-based technique was proposed in \cite{cai2007srda} for speeding up the eigen-decomposition step of LDA. The spectral regression-based technique consolidates spectral graph analysis and regression to provide an efficient solution to LDA.

In general, the supervised dimensionality reduction approaches work better than unsupervised algorithms if sufficient data are available \cite{xu2016sliced}. However, in real case scenarios, the labeled data may be scarce, noisy, or expensive to collect. In such situations, semi-supervised learning algorithms are preferred \cite{berthelot2019mixmatch}. Semi-supervised learning mitigates the necessity for labeled data by allowing a model to leverage unlabeled data. Semi-supervised algorithms can extend the learning strategies of either supervised or unsupervised learning algorithms. If the data are available from only one class during the training, one-class classification algorithms are used to determine the predictive model \cite{kefi2019novel}. In one-class classification, the decision function is inferred using training data from a single class only \cite{lenz2021average}. The class used to obtain the data description is referred to as the positive class, while all other classes are referred to as the negative class.

\begin{figure}[ht]
	\centering
	\includegraphics[scale=0.65]{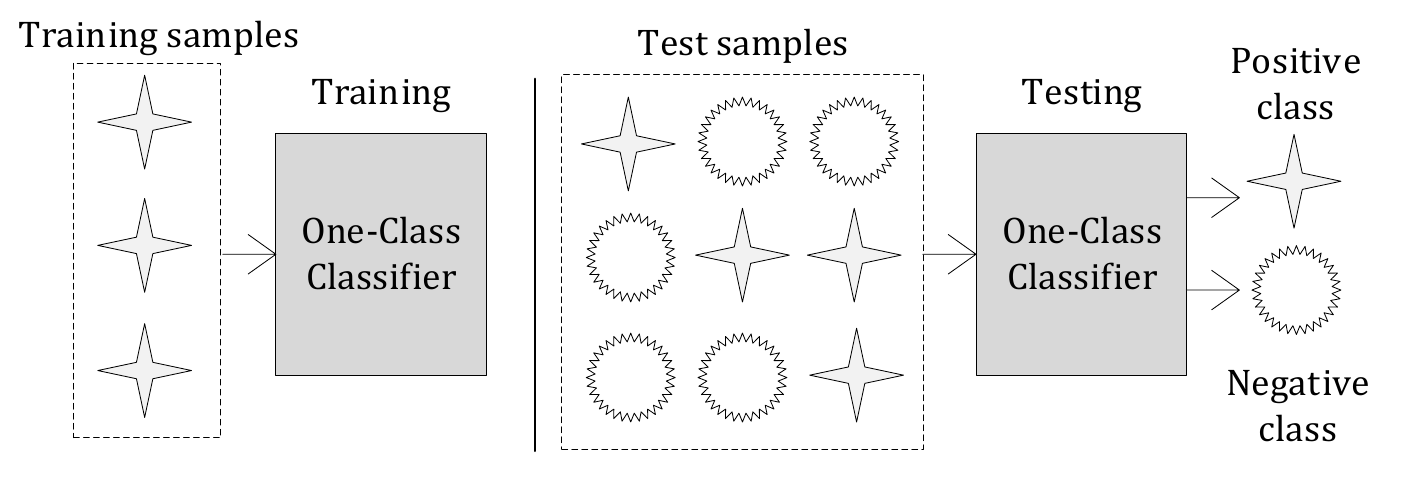}
		%\vspace{-2cm}
		%\rule{35em}{0.5pt}
	\caption{In one-class classification, a data model is learned by using samples of a positive class only. During inference, the model is used to detect objects also from the negative class.}
	\label{occ}
\end{figure}

One-class classifiers have been extensively studied and improved for several technology-driven applications \cite{alam2020one}. One-class classification techniques are found suitable for a specific target class detection in applications such as document classification \cite{manevitz2001one}, disease diagnosis \cite{cohen2008novelty}, fraud detection \cite{hejazi2013one}, rare species identification \cite{sohrab2020boosting}, intrusion detection \cite{khreich2017anomaly}, or novelty detection \cite{yin2018active}. Figure \ref{occ} depicts the basic idea of one-class classification.

Most one-class classification techniques operate in the original feature space and suffer from the curse of dimensionality \cite{xing2021robust}. In this paper, we propose a general subspace learning framework for one-class classification. We pose the subspace learning for one-class classification as a graph embedding problem. We show that the previously proposed subspace one-class techniques can be reformulated through the proposed framework, while the framework brings more insight into their optimization process. The framework also allows to integrate other data relations to the optimization process and highlights the similarities to other subspace learning techniques. The framework motivates a novel spectral solution as well as a spectral regression-based solution as alternatives to the previously used gradient-based approach. Finally we integrate the subspace learning framework with the Support Vector Data Description (SVDD) applied in the subspace into an iterative Graph-Embedded Subspace Support Vector Data Description (GESSVDD) method.

The rest of the paper is organized as follows. In Section \ref{relatedwork}, we review the related work. In Section \ref{formulation}, we formulate the proposed framework, describe the full GESSVDD algorithm, and discuss the new insights obtained from the framework. Details of the experiments and the results are provided in Section \ref{experiments}. We finally deduce the conclusions in Section \ref{Conclusions}.

\section{Related work and background}\label{relatedwork}

In this work, we focus on support vector (SV)-based one-class classification methods, which form a decision boundary represented by so-called support vectors by solving an optimization problem. The support vectors are selected from the training data points to define the boundary maximizing the considered criterion uniquely. One-class Support Vector Machine (OCSVM) \cite{scholkopfu1999sv} and SVDD \cite{tax2004support} are classic examples of SV-based one-class classification methods. OCSVM
constructs a hyperplane that separates the positive class by maximizing the distance of the hyperplane from the origin. In SVDD, a hypersphere with minimum volume is formed around the positive class. Numerous extensions of OCSVM and SVDD have been proposed in the literature \cite{mygdalis2016graph,turkoz2020generalized}. Traditionally, the SV-based one-class classification models data in the initially given feature space, but we have recently proposed one-class classification algorithms operating in an optimized lower-dimensional subspace~\cite{9133428,sohrab2018subspace}. 

\subsection{Support Vector Data Description}
SVDD \cite{tax2004support} finds a hyperspherical boundary around the positive class data in the original feature space by minimizing the volume of the hypersphere. Let us denote the training samples to be encapsulated inside a closed boundary by a matrix $\mathbf{X}=[\mathbf{x}_{1},\mathbf{x}_{2},\dots, \mathbf{x}_{N}],\mathbf{x}_{i} \in \mathbb{R}^{D}$, where $N$ is total number of samples and $D$ is the dimensionality of data.
The optimization problem of SVDD is formulated as follows:
\begin{align}\label{erfuncSVDD2}
\min \quad & F(R,\mathbf{a}) = R^2 + C\sum_{i=1}^{N} \xi_i \nonumber\\
\textrm{s.t.} \quad & \|\mathbf{x}_i - \mathbf{a}\|_2^2 \le R^2 + \xi _i,\nonumber\\
&\xi_i \ge 0, \:\: \forall i\in\{1,\dots,N\},
\end{align}
where $R$ is the radius and $\mathbf{a}\in \mathbb{R}^{D}$ is the center of the hypersphere. The slack variables $\xi_i,\:i=1,\dots, N$ are introduced to allow the possibility of data being outliers and the hyperparameter $C>0$ controls the trade-off between the volume of the hypersphere and the amount of data outside the hypersphere. The Lagrangian of SVDD can be given as
\begin{align}\label{langSVDD}
L = \sum_{i=1}^{N} \alpha_i \mathbf{x}_i^{\intercal} \mathbf{x}_i - \sum_{i}^{N}\sum_{j}^{N} \alpha_i \alpha_j \mathbf{x}_i^{\intercal} \mathbf{x}_j,
\end{align}
subject to the constraint that $0 \le \alpha_i \le C$ \cite{tax2004support}. Maximizing \eqref{langSVDD} gives a set of $\alpha_i$ values corresponding to each data points. The data points with $0<\alpha_i<C$ are called \textit{support vectors} and define the data description. A test sample $\mathbf{x}_{*}$ is classified to the positive class if the distance of the test sample from the center of the hypersphere is smaller than or equal to the radius:
\begin{align}\label{svddtest}
\|\mathbf{x}_{*} - \mathbf{a}\|_2 \le R,
\end{align}
where $R$ is the distance from the center of hypersphere to any sample with $0 < \alpha_i < C$.

\subsection{Subspace Support Vector Data Description}\label{SSVDD}
SSVDD \cite{sohrab2018subspace} optimizes a data mapping to a lower-dimensional subspace along with data description in the subspace. The optimization function is as follows:
\begin{align}\label{ssvdd}
\min \quad F(R,\mathbf{a}) = R^2 + C\sum_{i=1}^{N} \xi_i \nonumber\\
\textrm{s.t.} \quad  \|\mathbf{Qx}_i - \mathbf{a}\|_2^2 \le R^2 + \xi _i,\nonumber\\
\xi_i \ge 0, \:\: \forall i\in\{1,\dots,N\},
\end{align}  
where $\mathbf{Q} \in \mathbb{R}^{d \times D}$ is the projection matrix for mapping the data from original \textit{D}-dimensional feature space to an optimized lower \textit{d}-dimensional space. In SSVDD, an iterative process is followed: at each iteration, a set of $\alpha_i$ values is obtained by solving SVDD in the subspace, and then an augmented Lagrangian is optimized to update the projection matrix. The augmented Lagrangian is given as follows:
\begin{equation}\label{langSSVDD}
L= \sum_{i=1}^{N} \alpha _i  \mathbf{x}_i^{\intercal} \mathbf{Q}^{\intercal} \mathbf{Q} \mathbf{x}_i - \sum_{i=1}^{N}\sum_{j=1}^{N} \alpha_i \mathbf{x}_i^{\intercal} \mathbf{Q}^{\intercal} \mathbf{Q} \mathbf{x}_j \alpha_j + \beta\psi,
\end{equation} 
where $\psi$ is an optional regularization term expressing the class variance in the lower \textit{d}-dimensional space and $\beta$ is the regularization parameter which controls the weight of $\psi$. The regularization term $\psi$ has the following form:
\begin{equation}
\label{generalconstraintpsi} 
\psi = \text{Tr}(\mathbf{Q}\mathbf{X} \boldsymbol{\lambda}\boldsymbol{\lambda}^\intercal \mathbf{X}^\intercal\mathbf{Q}^\intercal),
\end{equation} 
where $\text{Tr}$ is the trace operator and different values of $\boldsymbol{\lambda}$ lead to different variants of SSVDD. The projection matrix $\mathbf{Q}$ is updated by using the gradient of \eqref{langSSVDD}, i.e.,
\begin{align}\label{Quopdate}
\mathbf{Q}\leftarrow \mathbf{Q}-\eta\Delta L,
\end{align}
where $\eta$ is the learning rate parameter. The projection matrix is orthogonalized after every update.

Recently, Ellipsoidal Subspace Support Vector
Data Description (ESSVDD) was proposed in \cite{9133428}. ESSVDD considers the covariance of the data in the subspace and the optimization problem is given as \begin{align}\label{primalobj_gesvdd}
\min \quad & R^2+ C\sum_{i=1}^{N}\xi_i\nonumber\\
\textrm{s.t.} \quad & (\mathbf{Qx}_i-\mathbf{a})^\intercal\mathbf{E}^{-1}(\mathbf{Qx}_i-\mathbf{a}) \le R^2 + \xi_i, \nonumber\\
&\xi_i \ge 0, \forall i\in \{1,\dots,N\},
\end{align}
where 
\begin{align}\label{Eessvdd}
\mathbf{E}=\mathbf{Q}\mathbf{X}\mathbf{X}^\intercal\mathbf{Q}^\intercal
\end{align}
is a covariance matrix of the data in \textit{d}-dimensional subspace. The rest of the ESSVDD solution follows the main principles of SSVDD explained above, while including the covariance matrix yields are more generalized solutions compared to SSVDD.

\subsection{Graph embedding}
Let $\mathcal{G}=\{\mathbf{X},\mathbf{A}\}$ be an undirected weighted graph, where the data points in $\mathbf{X}$ are the graph nodes and $\mathbf{A}\in \mathbb{R}^{N\times N}$ is the graph weight matrix that can measure different relations between the data points. The Laplacian matrix $\mathbf{L}$ of the graph and the diagonal degree matrix $\mathbf{D}$ are defined as~follows:
\begin{align}\label{graph} 
\mathbf{L}=\mathbf{D}-\mathbf{A},\;\;\;
[\mathbf{D}]_{ii}=\sum_{j\neq i}[\mathbf{A}]_{ij},\forall i\in \{1,\dots,N\}.
\end{align}

Graph embedding \cite{yan2006graph} was proposed as a general framework for encapsulating several subspace learning algorithms under the graph preserving criterion
\begin{equation} \label{eq:graphpreserving}
\begin{split}
\textbf{Q}^* &= {\underset{\text{Tr}(\mathbf{QX}\mathbf{L_p}\mathbf{X}^\intercal\mathbf{Q}^\intercal)=m}{\arg\min }\:\: \sum_{i\neq j} ( \mathbf{Qx}_i - \mathbf{Qx}_j )^2\textbf{A}_{ij} }\\
&= {\underset{\text{Tr}(\mathbf{QX}\mathbf{L}_p\mathbf{X}^\intercal\mathbf{Q}^\intercal)=m}{\arg\min }\:\: \text{Tr}(\mathbf{QX}\mathbf{L}\mathbf{X}^\intercal\mathbf{Q}^\intercal)},
\\
&= \arg\min \:\: \frac{\text{Tr}(\mathbf{QX}\mathbf{L}\mathbf{X}^\intercal\mathbf{Q}^\intercal)}{\text{Tr}(\mathbf{QX}\mathbf{L}_p\mathbf{X}^\intercal\mathbf{Q}^\intercal)},
\end{split}
\end{equation}
where $\mathbf{L}$ and $\mathbf{L}_p$ are the graph Laplacian matrices of the \emph{intrinsic} and \emph{penalty} graphs that correspond to data relations to be preserved or penalized, respectively. With different formulations of $\mathbf{L}$ and $\mathbf{L}_p$, \eqref{eq:graphpreserving} can represent different subspace learning algorithms. If there are no data-dependent penalty criteria to consider, the constraint $\text{Tr}(\mathbf{QX}\mathbf{L_p}\mathbf{X}^\intercal\mathbf{Q}^\intercal)=m$ can be replaced with the orthogonality constraint $\text{Tr}(\mathbf{Q}\mathbf{Q}^\intercal)=m$.

The solution to the \emph{trace ratio} optimization in \eqref{eq:graphpreserving} is typically approximated by the corresponding \emph{ratio trace} problem
\begin{equation} \label{eq:rt_ge}
\begin{split}
\textbf{Q}^* = \arg\min \:\: \text{Tr}\left((\mathbf{QX}\mathbf{L}_p\mathbf{X}^\intercal\mathbf{Q}^\intercal)^{-1}\mathbf{QX}\mathbf{L}\mathbf{X}^\intercal\mathbf{Q}^\intercal\right).
\end{split}
\end{equation}
The solution to \eqref{eq:rt_ge} can be obtained by solving the generalized eigenvalue value problem
\begin{equation}
\textbf{X}\textbf{L}\textbf{X}^T\textbf{q} = \lambda \textbf{X} \textbf{L}_p \textbf{X}^T \textbf{q} 
\label{eq:EigGE}
\end{equation}
and keeping the eigenvectors corresponding to the $d$ smallest non-zero eigenvalues as the rows of~$\mathbf{Q}$.

The total scatter, within-class, and between-classes matrices commonly used in subspace learning can be expressed in the graph embedding framework as follows: 
\begin{align}
\mathbf{S}_{t}&=\mathbf{X}\Big(\mathbf{I}-\frac{1}{N}\mathbf{1}\mathbf{1}^{\intercal}\Big)\mathbf{X}^{\intercal}=\mathbf{X}\mathbf{L}_{t}\mathbf{X}^\intercal \label{eq:st}\\
\mathbf{S}_{w}&=\mathbf{X}\Big(\mathbf{I}-\sum_{c=1}^\mathcal{C}\frac{1}{N_c}\mathbf{1}_c\mathbf{1}_c^{\intercal}\Big)\mathbf{X}^{\intercal}=\mathbf{X}\mathbf{L}_{w}\mathbf{X}^\intercal \label{eq:sw}\\
\mathbf{S}_{b}&=\mathbf{X}\Big(\sum_{c=1}^\mathcal{C}N_c(\frac{1}{N_c}\mathbf{1}_c-\frac{1}{N}\mathbf{1})(\frac{1}{N_c}\mathbf{1}_c-\frac{1}{N}\mathbf{1})^\intercal
\Big)\mathbf{X}^{\intercal}=\mathbf{X}\mathbf{L}_{b}\mathbf{X}^\intercal\label{eq:sb}
\end{align}
where $\mathbf{I}$ is an identity matrix, $\mathbf{1}$ is a vector of ones, $N_c$ is the total number of instances belonging to class $c$ and
$\mathbf{1}_c$ represents a vector with ones corresponding to instances which belongs to class $c$ and zeros elsewhere. For centered data $\mathbf{S}_t$ reduces to $\mathbf{S}_t = \mathbf{XX}^\intercal$. Using these Laplacians, LDA can be expressed in the graph embedding framework by setting $\mathbf{L} = \mathbf{L}_w$ and $\mathbf{L}_p =\mathbf{L}_b$ in \eqref{eq:graphpreserving}. In a similar manner, PCA can be expressed in the graph embedding framework by setting $\mathbf{L} = \frac{1}{N}\mathbf{L}_t$, and replacing the constraint $\text{Tr}(\mathbf{QX}\mathbf{L_p}\mathbf{X}^\intercal\mathbf{Q}^\intercal)=m$ with the orthogonality constraint $\text{Tr}(\mathbf{Q}\mathbf{Q}^\intercal)=m$. Since PCA seeks the projection directions with maximal variances, the criterion is maximized in the case of PCA.

Graph-Embedded Support Vector Data Description \cite{mygdalis2016graph} was proposed to solve the following optimization problem
\begin{align}\label{primalobj_gesvdd1}
\min \quad & R^2+ C\sum_{i=1}^{N}\xi_i\nonumber\\
\textrm{s.t.} \quad & (\mathbf{x}_i-\mathbf{a})^\intercal\mathbf{S}_x^{-1}(\mathbf{x}_i-\mathbf{a}) \le R^2 + \xi_i, \nonumber\\
&\xi_i \ge 0, \forall i\in \{1,\dots,N\},
\end{align}
where $\mathbf{S}_x = \mathbf{X}\mathbf{L}_x\mathbf{X}^\intercal$ and $\mathbf{L}_x$ is the graph Laplacian of any graph expressing geometric data~relationship.

\subsection{Spectral regression}\label{ssec:sr}

Spectral regression \cite{cai2007spectral} is an alternative way to solve the generalized eigen-decomposition in \eqref{eq:EigGE}.
If $\mathbf{X}^\intercal\mathbf{q}=\mathbf{t}$, and $\mathbf{t}$ and $\lambda$ are an eigenvector and eigenvalue solving the eigenproblem
\begin{eqnarray}\label{jj}
\mathbf{L}\mathbf{t}=\lambda\mathbf{L}_p\mathbf{t}, 
\end{eqnarray}  
$\mathbf{q}$ is the eigenvector of \eqref{eq:EigGE} with the same eigenvalue, because $\mathbf{X}\mathbf{L}\mathbf{X}^\intercal\mathbf{q}=\mathbf{X}\mathbf{L}\mathbf{t}=\lambda\mathbf{X}\mathbf{L}_p\mathbf{t}=\lambda\mathbf{X}\mathbf{L}_p\mathbf{X}^\intercal\mathbf{q}$. In order to find $\mathbf{Q}$, first the target vectors $\mathbf{t}$ can be obtained from \eqref{jj} and then vectors $\mathbf{q}$ satisfying $\mathbf{X}^\intercal\mathbf{q}=\mathbf{t}$ found. An exact solution may not exists but it can be estimated using regularized least squares also known as ridge regression \cite{hastie2009elements}:
\begin{eqnarray}\label{leastsquare1}
\mathbf{q}&=&\arg\min\Big(\|\mathbf{X}^\intercal\mathbf{q}-\mathbf{t}\|^2+\eta\|\mathbf{q}\|^2\Big) \nonumber\\
&=&(\mathbf{X}\mathbf{X}^\intercal+\epsilon \mathbf{I})^{-1}\mathbf{X}\mathbf{t},
\end{eqnarray}  
where $\epsilon$ is a tiny constant. The above technique combines the spectral analysis and the regression, hence the approach is named as spectral regression. The main benefit of spectral regression approach is that most graph Laplacian are sparse and, thus, the approach bypasses the need of computing the eigen-decomposition of dense matrices. The least squares problem can be solved efficiently and, in some cases \cite{cai2007srda} it is also possible to compute the target vectors $\mathbf{t}$ directly without using eigen-decomposition at all, which makes the process much faster.

\section{Graph Embedded Subspace Support Vector Data Description}\label{formulation}
In subspace one-class classification, the aim is to determine a projection matrix $\mathbf{Q} \in \mathbb{R}^{d \times D}$ for mapping data $\mathbf{X}\in \mathbb{R}^{D \times N}$ from the \textit{D}-dimensional original feature space to a lower \textit{d}-dimensional subspace optimized for one-class classification. In this work, we assume that the data has been centered by setting $\mathbf{X}\leftarrow \mathbf{X}-\boldsymbol{\mu}$, where $\boldsymbol{\mu}$ represents the mean of the training data. The mapped data in the subspace is represented by
\begin{equation}\label{subspacedata}
\mathbf{y}_i = \mathbf{Q} \mathbf{x}_i, \:\:i=1,\dots,N.
\end{equation}
After the transformation, the data is encapsulated inside a closed boundary to obtain an optimized data description in the subspace. In order to obtain a generalized solution, we consider the following optimization criterion:
\begin{align}\label{primalobj_gesvdd2}
\min \quad & R^2+ C\sum_{i=1}^{N}\xi_i\nonumber\\
\textrm{s.t.} \quad & (\mathbf{Qx}_i-\mathbf{a})^\intercal\mathbf{S}_Q^{-1}(\mathbf{Qx}_i-\mathbf{a}) \le R^2 + \xi_i, \nonumber\\
&\xi_i \ge 0, \forall i\in \{1,\dots,N\},
\end{align}
where the matrix $\mathbf{S}_Q$ encodes geometric data relationships in the subspace as 
\begin{eqnarray}\label{Einv}
\mathbf{S}_Q=\mathbf{Q}\mathbf{X}\mathbf{L}_x\mathbf{X}^\intercal\mathbf{Q}^\intercal =\mathbf{Q}\mathbf{S}_x\mathbf{Q}^\intercal,
\end{eqnarray}
where $\mathbf{L}_x$ is a graph Laplacian. It can take different forms depending on the graph type used. By defining a new vector $\mathbf{u}=\mathbf{S}_Q^{-\frac{1}{2}}\mathbf{a}$, \eqref{primalobj_gesvdd2} can be written as
\begin{align}\label{primalobj_gesvdd3}
\min \quad & R^2+ C \sum_{i=1}^{N}\xi_i\nonumber\\
\textrm{s.t.} \quad &  \| \mathbf{S}_Q^{-\frac{1}{2}}\mathbf{Qx}_i-\mathbf{u} \|^2_{2} \le R^2 + \xi_i, \nonumber\\
&\xi_i \ge 0, \forall i\in \{1,\dots,N\}.
\end{align}
This shows that we can consider $\mathbf{S}_Q^{-\frac{1}{2}}\mathbf{Q}$ as a new projection matrix to a subspace, where SVDD is to be applied. We denote the mapped input vectors as $\mathbf{z}_i = \mathbf{S}_Q^{-\frac{1}{2}}\mathbf{Qx}_i$.

The constraints in \eqref{primalobj_gesvdd3} can be incorporated into a corresponding dual objective function by using Lagrange multipliers: 
\begin{align}\label{lagrangian0}
L= R^2+C\sum_{i=1}^{N}\xi_i-\sum_{i=1}^{N}\alpha_i\big(R^2+\xi_i 
-\nonumber\\(\mathbf{S}_Q^{-\frac{1}{2}}\mathbf{Q} \mathbf{x}_i)^\intercal
\mathbf{S}_Q^{-\frac{1}{2}}\mathbf{Q} \mathbf{x}_i + 
2\mathbf{u}^\intercal\mathbf{S}_Q^{-\frac{1}{2}}\mathbf{Q} \mathbf{x}_i-\mathbf{u}^\intercal\mathbf{u}\big)-\sum_{i=1}^{N}\gamma_i\xi_i,
\end{align}
where $\alpha_i\ge0$ and $\gamma_i\ge0$ are the Lagrange multipliers. The Lagrangian \eqref{lagrangian0} should be minimized with respect to $R$, $\mathbf{u}$, and $\xi_i$ and maximized with respect to Lagrange multipliers $\alpha_i$ and $\gamma_i$. By setting partial derivative to zero, we get
\begin{eqnarray}
\frac{\partial L}{\partial R}=0 &\Rightarrow& \sum_{i=1}^{N} \alpha_i = 1, \label{der1} \\
\frac{\partial L}{\partial {\mathbf{u}}}=0 &\Rightarrow& {\mathbf{u}} = \sum_{i=1}^{N} \alpha_i \mathbf{S}_Q^{-\frac{1}{2}}\mathbf{Qx}_i, \label{der2} \\
\frac{\partial L}{\partial \xi _i}=0 &\Rightarrow& C- \alpha _i - \xi _i  = 0. \label{der3}
\end{eqnarray} 
By substituting \eqref{der1}-\eqref{der3} into \eqref{lagrangian0}, we get
\begin{align}\label{LL}
L &= \sum_{i=1}^{N} \alpha _i {\mathbf{x}}_i^\intercal {\mathbf{Q}}^\intercal \mathbf{S}_Q^{-1} \mathbf{Qx}_i -
\sum_{i=1}^{N}\sum_{j=1}^{N} \alpha _i \mathbf{x}_i^\intercal \mathbf{Q}^\intercal\mathbf{S}_Q^{-1} \mathbf{Qx}_j \alpha _j \nonumber\\
&=\sum_{i=1}^{N} \alpha _i \mathbf{z}_i^\intercal \mathbf{z}_i - 
\sum_{i=1}^{N}\sum_{j=1}^{N} \alpha _i\alpha _j \mathbf{z}_i^\intercal \mathbf{z}_j.
\end{align} 
Maximizing \eqref{LL} corresponds to solving SVDD in the new subspace and will give us $\alpha_i$ values for all instances, which will define their position in the data description. The samples in the subspace corresponding to values $0 < \alpha{_i} < C$ will lie on the boundary, while those outside the boundary will correspond to values $\alpha_i = C$. For the samples inside the closed boundary, the corresponding values of $\alpha_i$ will be equal to zero:
\begin{align}\label{alphaEQ}
&\|\mathbf{z}_i - \mathbf{u}\|_2 < R \rightarrow \alpha_i=0,\gamma_i=0, \\
&\|\mathbf{z}_i - \mathbf{u}\|_2 = R\rightarrow 0 < \alpha{_i} < C,\gamma_i=0,\\
&\|\mathbf{z}_i - \mathbf{u}\|_2 > R \rightarrow\alpha{_i} = C,\gamma_i>0.
\end{align}

\begin{figure}[t]
	\centering
	\includegraphics[scale=0.60]{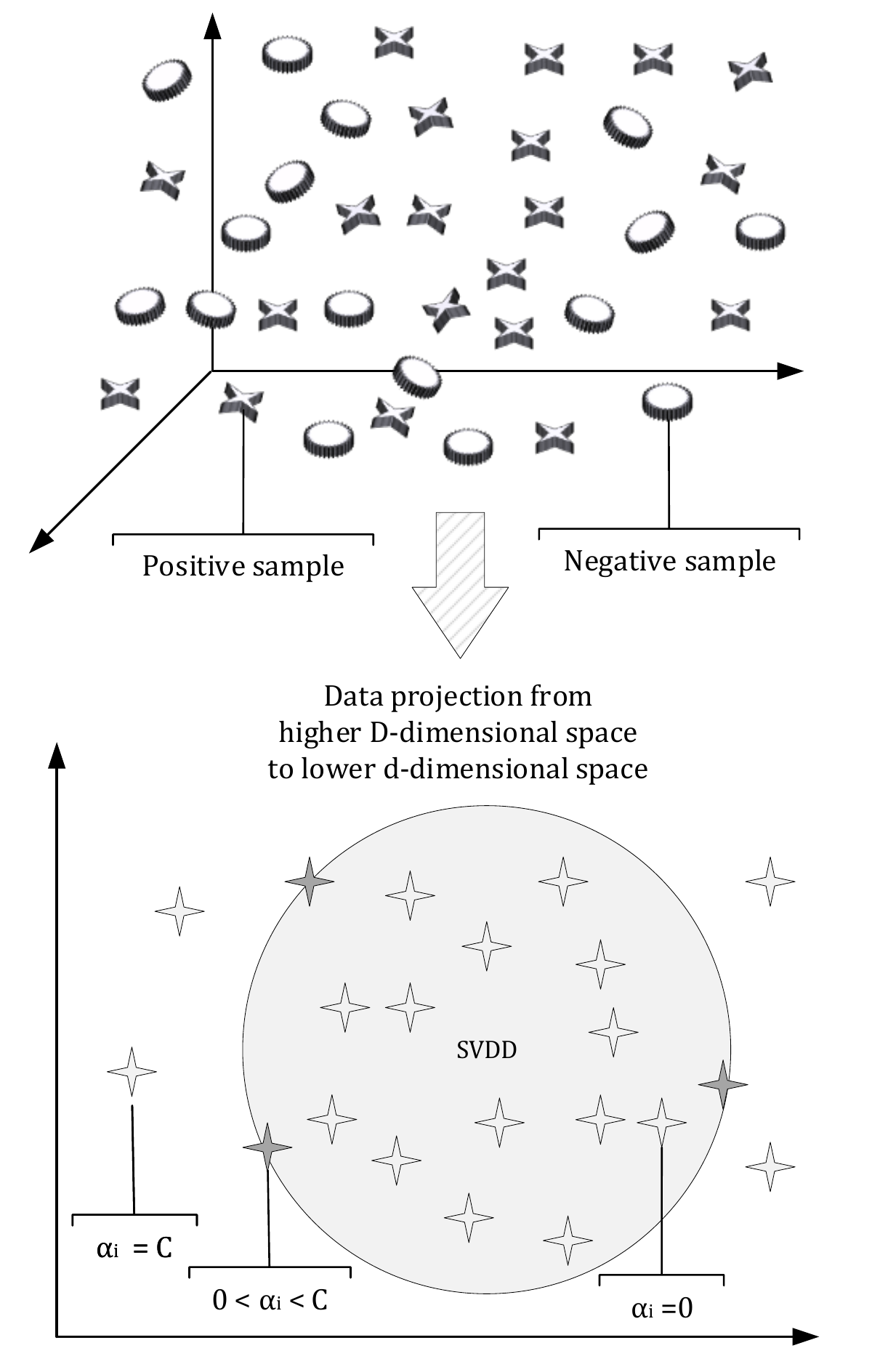}
		%\vspace{-2cm}
		%\rule{35em}{0.5pt}
	\caption{Depiction of data projection to a lower \textit{d}-dimensional space optimized for one-class classification with corresponding $\alpha_i$ values}
	\label{spectralocc2}
\end{figure}

Figure \ref{spectralocc2} depicts the idea of projecting data into an optimized subspace along with the positions of instances according to $\alpha$ values. The negative class samples are not considered in the process; hence, it is not guaranteed that they will be outside the obtained closed boundary.

The Lagrangian in \eqref{LL} can be written in a trace form as
\begin{equation}\label{simplifiedL}
\begin{split}
L&=\text{Tr}(\mathbf{S}_Q^{-1}\mathbf{QX}\mathbb{A}\mathbf{X}^\intercal\mathbf{Q}^\intercal)-\text{Tr}(\mathbf{S}_Q^{-1}\mathbf{QX}\boldsymbol{\alpha} \boldsymbol{\alpha}^\intercal\mathbf{X}^\intercal\mathbf{Q}^\intercal)\\ &= \text{Tr}((\mathbf{QX}\mathbf{L}_x\mathbf{X}^{\intercal}\mathbf{Q}^{\intercal})^{-1}\mathbf{QX}(\mathbb{A}-\boldsymbol{\alpha} \boldsymbol{\alpha}^\intercal)\mathbf{X}^\intercal\mathbf{Q}^\intercal),
\end{split}
\end{equation}  
where the matrix $\mathbb{A}\in \mathbb{R}^{N \times N}$ contains $\alpha_i$ values in its diagonal and zeros elsewhere, $\boldsymbol{\alpha}$ is a vector of $\alpha_i$ values. Now by defining the matrices 
\begin{eqnarray}\label{st}
\mathbf{L}_{\alpha}&=&\mathbb{A}-\boldsymbol{\alpha} \boldsymbol{\alpha}^\intercal\label{Jterm}\\
\mathbf{S}_{\alpha}&=&\mathbf{X}\mathbf{L}_{\alpha}\mathbf{X}^\intercal \label{salpha},
\end{eqnarray}  
we can simplify \eqref{simplifiedL} to
\begin{eqnarray}\label{Ltrace}
L=\text{Tr}\big((\mathbf{Q}\mathbf{S}_x\mathbf{Q}^\intercal)^{-1}\mathbf{Q}\mathbf{S}_{\alpha}\mathbf{Q}^\intercal\big).
\end{eqnarray}  

We note that \eqref{Ltrace} is in a ratio trace form that resembles the trace ratio in \eqref{eq:graphpreserving}. As mentioned, the trace ratio in \eqref{eq:graphpreserving} is typically approximated by the corresponding ratio trace to be able to solve the optimization using eigen-decomposition.
We also note that $\mathbf{L}_{\alpha}$ is a graph Laplacian (see Section \ref{ssec:analysis}). Thus, we have presented the subspace learning for SVDD in the general graph embedding framework for subspace learning with its own fixed intrinsic graph $\mathbf{L}_\alpha$. Different graphs $\mathbf{L}_x$ create different variants and can be selected to enforce different constraints for the data. We will get back to different insights offered by the new framework in Section~\ref{ssec:analysis}, but first we will introduce the full Graph-Embedded Subspace Support Vector Data Description (GESSVDD) algorithm.

\subsection{GESSVDD algorithm}
\label{ssec:algo}

We can directly see from \eqref{Ltrace} that it can be minimized/maximized by solving the generalized eigenproblem in \eqref{eq:EigGE} and keeping the eigenvectors corresponding to the smallest/largest non-zero eigenvalues as projection vectors. We can also formulate a spectral regression-based solution as explained in Section~\ref{ssec:sr}. While earlier subspace SVDD variants \cite{sohrab2018subspace, 9133428, SOHRAB2021107648} have only used gradient-based solution, we now have three alternatives: 1) gradient-based, 2) spectral, and 3) spectral regression-based updates. Furthermore, we can pick any desired graph as $\mathbf{L}_x$ and we note that it can be meaningful to also maximize \eqref{der3} (see further discussion in Section~\ref{ssec:analysis}). With this we can give the main GESSVDD algorithm in Algorithm~\ref{mainalgo} and the three update options in Sub-algorithms 1-3. The gradient of \eqref{simplifiedL} used in the gradient-based update can be obtained using identity 126 in \cite{IMM2012-03274}. 

\begin{algorithm}[p]
\SetAlgoLined
  \caption{GESSVDD optimization }\label{mainalgo}
\SetAlgoLined
\SetKwInOut{Input}{Input}
\SetKwInOut{Output}{Output}
\Input{$\mathbf{X}$, // Input data\\
$\mathbf{L}_x$ // Selected Laplacian\\
$\eta$, // Learning rate parameter \\$d$, // Dimensionality of subspace \\$C$, // Regularization parameter in SVDD\\
min or max // Either minimize or maximize the criterion}
\vspace{2mm}
\Output{$\mathbf{Q}$ // Projection matrix \\$R$, // Radius of hypersphere \\$\mbox{\boldmath$\alpha$}$ // Defines the data description }  
 \vspace{3mm}
Initialize $\mathbf{Q}$ via PCA; // Select $d$-vectors corresponding to $d$ largest eigenvalues.\\
Compute $\mathbf{S}_x=\mathbf{X}\mathbf{L}_x\mathbf{X}^\intercal$\;
 \vspace{2mm}
  \For{$iter=1:max\_iter$}{
  
  Calculate $\mathbf{S}_{inv}=\mathbf{S}_{Q}^{-1}=(\mathbf{Q}\mathbf{S}_x \mathbf{Q}^\intercal)^{-1}$\;
  Project data to subspace  $\mathbf{z}_i=\mathbf{S}_Q^{-\frac{1}{2}}\mathbf{Qx}_i = (\mathbf{S}_{inv})^{\frac{1}{2}}\mathbf{Qx}_i$\;
   Calculate $\alpha$ values by maximizing 
   $L=\sum_{i=1}^{N} \alpha _i \mathbf{z}_i^\intercal \mathbf{z}_i - 
\sum_{i=1}^{N}\sum_{j=1}^{N} \alpha _i\alpha _j \mathbf{z}_i^\intercal \mathbf{z}_j$\;
     \: Compute $\mathbf{L}_{\alpha}=\mathbb{A}-\boldsymbol{\alpha} \boldsymbol{\alpha}^\intercal$\;
\vspace{3mm}
\textbf{if} \textit{gradient-based update}\\
Call Sub-algorithm \ref{gradientalgo} to obtain $\mathbf{Q}$\; 
\textbf{elseif} \textit{spectral update}\\
Call Sub-algorithm \ref{eigenvaluealgo} to obtain $\mathbf{Q}$\; 
\textbf{elseif} \textit{spectral regression-based update}:\\
Call Sub-algorithm \ref{spectralalgo} to obtain $\mathbf{Q}$\;  
 \textbf{endif}\\
    \vspace{3mm}
    Orthogonalize $\mathbf{Q}$ using QR decomposition\;
   }
   Project data to subspace  $\mathbf{z}_i =\mathbf{S}_Q^{-\frac{1}{2}}\mathbf{Qx}_i$\;
   Calculate $\alpha$ values by maximizing 
   $L=\sum_{i=1}^{N} \alpha _i \mathbf{z}_i^\intercal \mathbf{z}_i - 
\sum_{i=1}^{N}\sum_{j=1}^{N} \alpha _i\alpha _j \mathbf{z}_i^\intercal \mathbf{z}_j$\;
Compute center of data description in the subspace as $\mathbf{u} = \sum_{i=1}^{N} \alpha_i \mathbf{S}_Q^{-\frac{1}{2}}\mathbf{Qx}_i$\;
   Identify any support vector $\mathbf{s}$ having $0 < \alpha_s < C$\; 
   Compute radius
   $R= \sqrt{(\mathbf{S}_Q^{-\frac{1}{2}}\mathbf{Qs})^\intercal\mathbf{S}_Q^{-\frac{1}{2}}\mathbf{Qs}- 2(\mathbf{S}_Q^{-\frac{1}{2}}\mathbf{Qs})^\intercal\mathbf{u}
+ \mathbf{u}^\intercal\mathbf{u}}$\;
   \end{algorithm}

\setcounter{algocf}{0}
\begin{Sub-algorithm}[h!]
\caption{Gradient-based update}
\label{gradientalgo}
\SetAlgoLined
\SetKwInOut{Input}{Input}
\SetKwInOut{Output}{Output}
\Input{$\mathbf{Q}$, $\mathbf{X}$, $\mathbf{S}_x$, $\mathbf{S}_{inv}$ $\mathbf{L}_\alpha$, $\eta$, \emph{min/max} //Input from Algorithm \ref{mainalgo}} 
 \Output{$\mathbf{Q}$ //Return output to Algorithm \ref{mainalgo}} 
 \vspace{2mm}
\:\:Compute $\mathbf{S}_{\alpha}=\mathbf{X}\mathbf{L}_{\alpha}\mathbf{X}^\intercal$\;
\:\:Compute $\Delta L=  2\mathbf{S}_{inv} \mathbf{Q}  \mathbf{S}_{\alpha} -2\mathbf{S}_{inv}\mathbf{Q}\mathbf{S}_{\alpha} \mathbf{Q} ^\intercal 
\mathbf{S}_{inv}\mathbf{Q}\mathbf{S}_x^\intercal$\;
\vspace{1mm}
\textbf{if} \textit{minimization}\\
\:\:Update $\mathbf{Q} \leftarrow \mathbf{Q} - \eta \Delta L$\;
\textbf{elseif} \textit{maximization}\\
\:\:Update $\mathbf{Q} \leftarrow \mathbf{Q} + \eta \Delta L$\;
\end{Sub-algorithm}

\begin{Sub-algorithm}[h!]
\caption{Spectral update}
\label{eigenvaluealgo}
\SetAlgoLined
\SetKwInOut{Input}{Input}
\SetKwInOut{Output}{Output}
\Input{$\mathbf{X}$, $\mathbf{S}_x$, $\mathbf{L}_\alpha$, \emph{min/max} //Input from Algorithm~\ref{mainalgo}}
 \Output{$\mathbf{Q}$ //Return output to Algorithm \ref{mainalgo}}  
 \vspace{2mm}
\:\:Compute $\mathbf{S}_{\alpha}=\mathbf{X}\mathbf{L}_{\alpha}\mathbf{X}^\intercal$\;
\:\:Solve generalized eigenvalue problem
$\mathbf{S}_{\alpha}\mathbf{q}=\upsilon\mathbf{S}_x\mathbf{q} $\; 
\vspace{1mm}
\textbf{if} \textit{minimization}\\
 Select the eigenvectors corresponding to $d$ smallest positive eigenvalues as rows of $\mathbf{Q}$\;
\textbf{elseif} \textit{maximization}\\
 Select the eigenvectors corresponding to $d$ largest eigenvalues as rows of $\mathbf{Q}$\;
\end{Sub-algorithm}

\begin{Sub-algorithm}[h!]
\caption{Spectral regression-based update}
\label{spectralalgo}
\SetAlgoLined
\SetKwInOut{Input}{Input}
\SetKwInOut{Output}{Output}
\Input{$\mathbf{X}$, $\mathbf{L}$, $\mathbf{L}_\alpha$, \emph{min/max} //Input from Algorithm \ref{mainalgo}}
 \Output{$\mathbf{Q}$ //Return output to Algorithm \ref{mainalgo}}  
 \vspace{2mm}
   Solve generalized eigenvalue problem:
$\mathbf{L}_{\alpha}\mathbf{t}=\upsilon\mathbf{L}_x\mathbf{t}$\;
\vspace{1mm}
\textbf{if} \textit{minimization} then\\Select the eigenvectors corresponding to $d$ smallest positive eigenvalues as columns of $\mathbf{T}$\;
\textbf{elseif} \textit{maximization} then\\Select the eigenvectors corresponding to $d$ largest eigenvalues to as columns of $\mathbf{T}$\;
\vspace{1mm}
Obtain $\mathbf{Q}=\mathbf{T}^\intercal\mathbf{X}^\intercal(\mathbf{X}\mathbf{X}^\intercal + \eta \mathbf{I})^{-1}$\; 
\end{Sub-algorithm} 

\subsubsection{Non-linear data description}\label{SS:KernelSSVDD}
To obtain a non-linear mapping with the proposed method, we employ a non-linear projection trick (NPT) \cite{kwak2013nonlinear}. NPT is equivalent to applying the well-known kernel trick, while allows using the linear variant of the method. In NPT, the data $\mathbf{X}$ is mapped from the original \textit{D}-dimensional space to $\mathbf{\Phi}$ in \textit{F}-dimensional space as follows: The kernel matrix is obtained as
\begin{equation}\label{RBFkernel}
\mathbf{K}_{ij} = \exp  \left( \frac{ -\| \mathbf{x}_{i} - \mathbf{x}_{j}\|_2^2 }{ 2\sigma^2 } \right),
\end{equation} 
where $\sigma$ is a hyperparameter scaling the distance between $\mathbf{x}_i$ and $\mathbf{x}_j$. The kernel matrix is centered~as
\begin{align}\label{centerK}
\mathbf{\Hat{K}} = \big(\mathbf{I}- \frac{1}{N}\mathbf{1} \mathbf{1}^\intercal\big) \mathbf{K} \big( \mathbf{I}-\frac{1}{N}\mathbf{1} \mathbf{1}^\intercal\big),
\end{align}
The centered kernel matrix $\mathbf{\Hat{K}}$ is decomposed by using eigen-decomposition:
\begin{align}\label{eigen}
\mathbf{\Hat{K}} = \mathbf{U}\mathbf{\Lambda}\mathbf{U}^\intercal, 
\end{align}
where $\mathbf{\Lambda}$ contains the non-negative eigenvalues of $\mathbf{\Hat{K}}$ in its diagonal and the columns of $\mathbf{U}$ contain the corresponding eigenvectors. Finally, the data representation $\mathbf{\Phi}$ is obtained as
\begin{align}\label{nptdata}
\mathbf{\Phi} = \mathbf{\Lambda}^{\frac{1}{2}} \mathbf{U}^\intercal.
\end{align}
Now we consider the obtained data transformation $\mathbf{\Phi}$ as the input to the linear algorithm, which is equivalent to applying the kernel method on $\mathbf{X}$.

\subsubsection{Test phase}\label{SS:Test}
During testing, a test instance $\mathbf{x}_*$ is first mapped to an optimized \textit{d}-dimensional space as
\begin{align}\label{test}
\mathbf{z}_* = \mathbf{S}_Q^{-\frac{1}{2}}\mathbf{Q} \mathbf{x}_*.
\end{align}
The distance of the test instance to the center of the data description in the subspace is calculated. The test instance is classified as a positive instance if the distance is equal to or smaller than the~radius:
\begin{align}\label{test1}
\|\mathbf{z}_{*} - \mathbf{u}\|_2^2 \le R^2, 
\end{align}
where $\mathbf{u}$ is obtained by solving \eqref{der2}, and $R^2$ is calculated as
 \begin{align}\label{rsquare}
R^2= (\mathbf{S}_Q^{-\frac{1}{2}}\mathbf{Qs})^\intercal\mathbf{S}_Q^{-\frac{1}{2}}\mathbf{Qs}- 2(\mathbf{S}_Q^{-\frac{1}{2}}\mathbf{Qs})^\intercal\mathbf{u}
+ \mathbf{u}^\intercal\mathbf{u},
\end{align}
and $\mathbf{s}$ is any support vector with $0 < \alpha_s < C$. Otherwise, the test instance is classified as a negative~instance.

In the non-linear approach, we first find the kernel vector
\begin{align}\label{kvector}
\mathbf{k}_{*} = \mathbf{\Phi}^\intercal \phi(\mathbf{x}_{*}).
\end{align}
The kernel vector is centered as
\begin{align}\label{centerKtest}
\mathbf{{\Hat{k}}}_{*}= (\mathbf{I}-  \frac{1}{N}\mathbf{1} \mathbf{1}^\intercal) [  \mathbf{{{k}}}_{*}-\frac{1}{N}\mathbf{K} \mathbf{1}].
\end{align}
Finally, the NPT representation of the test instance is obtained as
\begin{align}\label{npttest}
{\boldsymbol{\phi}}_{*} = \mathbf{(\Phi}^T)^{+}\mathbf{\Hat{k}}_{*},
\end{align}
where $(.)^{+}$ is a pseudo-inverse.
Now $\boldsymbol{\phi}_{*}$ is classified similar to the linear case, which is equivalent to applying a kernel method on $\mathbf{x}_*$. 

\subsubsection{Different variants}
\label{sssec:variants}

While any suitable graph can be used as $\mathbf{L}_x$, we list here some reasonable choices, which are also used in our experiments. In the first option, GESSVDD-0, we have no data-dependent constraint, but $\mathbf{S}_x$ in \eqref{Ltrace} is replaced by an identity matrix $\mathbf{I}$, which corresponds to the orthogonality constraint. In the second option GESSVDD-I, we use $\mathbf{L}_x = \mathbf{I}$. The third option GESSVDD-PCA uses the PCA graph: $\mathbf{S}_x = \frac{1}{N}\mathbf{S}_t$.

While we only have samples from the positive class, it may include several clusters. To consider this option, we cluster the positive training samples using k-means and then define options GESSVDD-Sw and GESSVDD-Sb with $\mathbf{S}_x = \mathbf{S}_w$ and $\mathbf{S}_x = \mathbf{S}_b$, respectively. Here, $\mathbf{S}_w$ and $\mathbf{S}_b$ are solved as in \eqref{eq:sw} and \eqref{eq:sb}, but $c$ now refers to a cluster, not a class. 

We also exploit the local geometric information by employing k-Nearest Neighbor (kNN) and~setting
\begin{align}\label{kNNscatter}
\mathbf{S}_{x} = \mathbf{S}_{kNN}=\mathbf{X}\Big(\mathbf{D}_{kNN}-\mathbf{A}_{kNN}\Big)\mathbf{X}^{\intercal}=\mathbf{X}\mathbf{L}_{kNN}\mathbf{X}^\intercal,
\end{align}
where $[\mathbf{A}]_{ij}=1,$ if $\mathbf{x}_{i}\in \mathcal{N}_j$ or $\mathbf{x}_{j}\in \mathcal{N}_i$ and 0, otherwise. 
$\mathcal{N}_i$ denotes the nearest neighbors of $\mathbf{x}_i$. This gives our last option denoted as GESSVDD-kNN.

Each of these options using different $\mathbf{S}_x$ can be solved using one of the update choices: gradient-based (GR), spectral ($\mathcal{S}$), or spectral regression-based (SR). Furthermore, in each case it is possible to either minimize or maximize the criterion in \eqref{Ltrace}. To refer all these variants, we denote them as GESSVDD-0-GR-min, GESSVDD-0-GR-max, GESSVDD-0-$\mathcal{S}$-min and so on.

\subsection{Framework analysis}
\label{ssec:analysis}

Now we will get back to our main result, the general subspace learning framework for SVDD expressed as follows (repeated from \eqref{simplifiedL}):
\begin{equation}
\label{eq:framework}
 \text{Tr}((\mathbf{QX}\mathbf{L}_x\mathbf{X}^{\intercal}\mathbf{Q}^{\intercal})^{-1}\mathbf{QX}(\mathbb{A}-\boldsymbol{\alpha} \boldsymbol{\alpha}^\intercal)\mathbf{X}^\intercal\mathbf{Q}^\intercal),
\end{equation} 
where $\mathbf{L}_x$ can be used to enforce local/global data relations relevant for the task. Let us consider a graph with a weight matrix $[\mathbf{A}_{\alpha}]_{ij} = \alpha_i \alpha_j \; \forall i \neq j$ and $[\mathbf{A}_{\alpha}]_{ii} = 0$. With the constraint $\sum_{i=1}^{N} \alpha_i = 1$ \eqref{der1}, we get
$[\mathbf{D_\alpha}]_{ii}=\sum_{j\neq i}[\mathbf{A_\alpha}]_{ij} = \sum_{j=1}^{N} \alpha_j \alpha_i - \alpha_i^2 = \alpha_i - \alpha_i^2$ and $\mathbf{L}_\alpha = \mathbf{D}_\alpha - \mathbf{A}_\alpha = \text{diag}(\boldsymbol{\alpha}) - \boldsymbol{\alpha}\boldsymbol{\alpha}^\intercal = \mathbb{A}-\boldsymbol{\alpha} \boldsymbol{\alpha}^\intercal$. This shows that $\mathbb{A}-\boldsymbol{\alpha} \boldsymbol{\alpha}^\intercal$ is a graph Laplacian of a graph that connects the samples $i$ and $j$ with a weight $\alpha_i \alpha_j$. As $\alpha_i$ values are zero for any samples inside the hypersphere, the resulting graph has only connections between the support vectors and outliers.

\begin{figure}[t]
	\centering
	\includegraphics[scale=0.30]{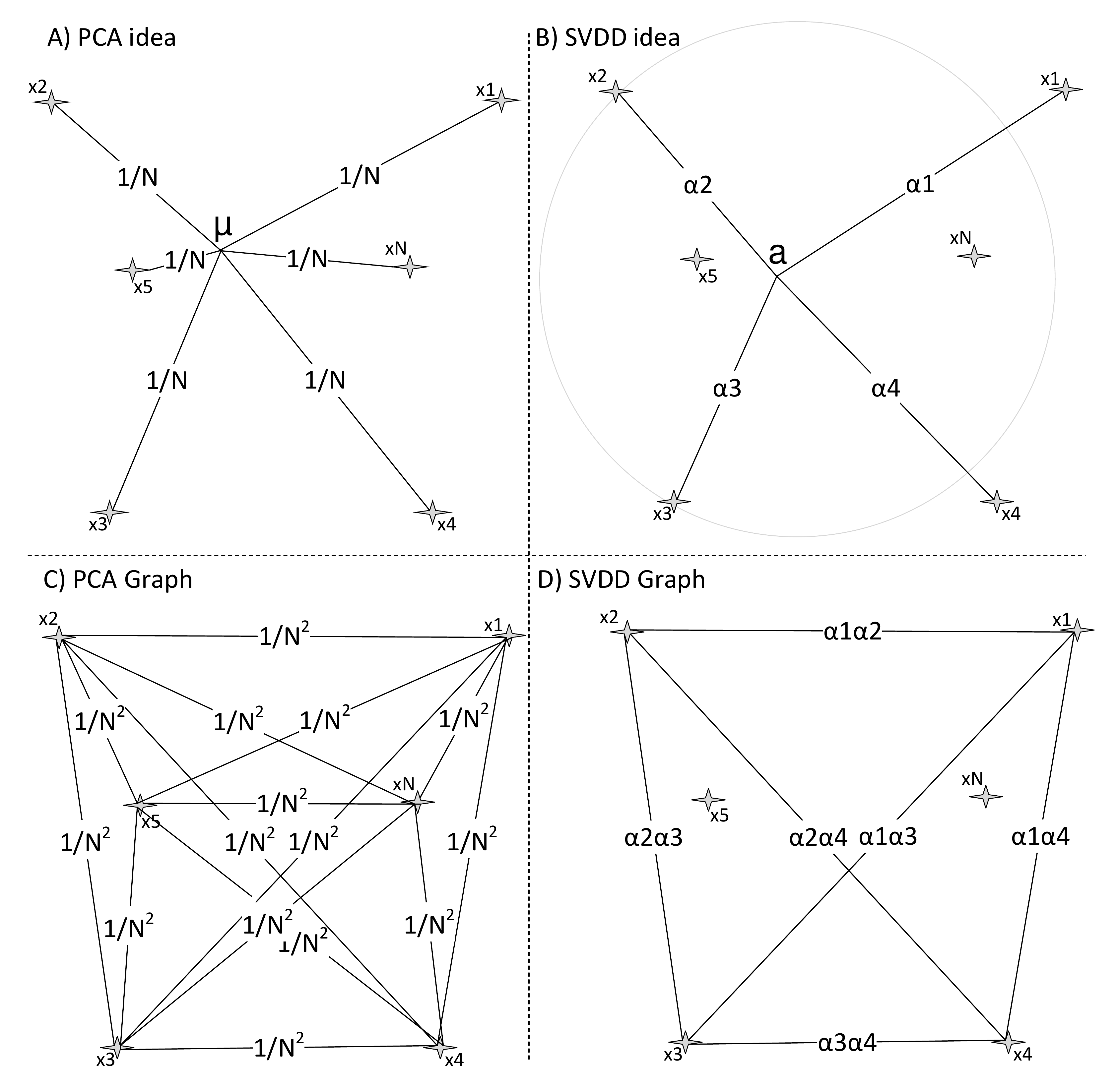}
		%\vspace{-2cm}
		%\rule{35em}{0.45pt}
	\caption{A) PCA considers the (unweighted) variance of all the points from the center $\boldsymbol{\mu}$. B) SVDD considers weighted variance of support vectors and outliers from the SVDD center $\mathbf{a}$. C) PCA graph is fully-connected with equal weights. D) SVDD graph is sparse (only the support vectors and outliers are connected) and has varying weights. }
	\label{pcasvdd}
\end{figure}

We also see that the graph of $\mathbf{L}_\alpha$ has a strong similarity with the PCA graph. PCA maximizes the variance of the samples to their center $\mbox{\boldmath$\mu$} = \frac{1}{N}\sum_{i=1}^{N} \mathbf{x}_i$, i.e.,
\begin{equation}
\begin{split}
\mathbf{S}_{pca}&=\frac{1}{N}\sum_{i=1}^{N} (\mathbf{x}_i-\mbox{\boldmath$\mu$})(\mathbf{x}_i-\mbox{\boldmath$\mu$})^\intercal \\
&= \frac{1}{N}\sum_{i=1}^{N}(\mathbf{x}_i \mathbf{x}_i^\intercal - 2\mathbf{x}_i\mbox{\boldmath$\mu$}^\intercal + \mbox{\boldmath$\mu$}\mbox{\boldmath$\mu$}^\intercal)\\
&= \frac{1}{N}\sum_{i=1}^{N}(\mathbf{x}_i \mathbf{x}_i^\intercal) - 2\mbox{\boldmath$\mu$}\mbox{\boldmath$\mu$}^\intercal + \mbox{\boldmath$\mu$}\mbox{\boldmath$\mu$}^\intercal = \frac{1}{N}\sum_{i=1}^{N}(\mathbf{x}_i \mathbf{x}_i^\intercal) - \mbox{\boldmath$\mu$}\mbox{\boldmath$\mu$}^\intercal \\
&= \frac{1}{N}\mathbf{XX}^\intercal - \frac{1}{N^2}\mathbf{X}\mathbf{1}\mathbf{1}^\intercal\mathbf{X}^\intercal = \frac{1}{N}\mathbf{X}(\mathbf{I}-\frac{1}{N}\mathbf{1}\mathbf{1}^\intercal)\mathbf{X}^\intercal\\
&= \mathbf{X}\mathbf{L}_{pca}\mathbf{X}^\intercal,
\end{split}
\end{equation}
where $\mathbf{L}_{pca} = \mathbf{D}_{pca} - \mathbf{A}_{pca}$ and $[\mathbf{A}_{pca}]_{ij} = 1/N^2 \;\; \forall i\neq j$ and $[\mathbf{A}_{pca}]_{ii} = 0$. With an analogous derivation using the constraint $\sum_{i=1}^{N} \alpha_i = 1$, we see that $\mathbf{L}_{\alpha}$ represents the weighted variance of the support vectors and outliers to the center of SVDD defined as $\mathbf{a} = \sum_{i=1}^{N} \alpha_i \mathbf{x}_i$:
\begin{equation}
\begin{split}
\mathbf{S}_{\alpha}&=\sum_{i=1}^{N} (\mathbf{x}_i-\mathbf{a})(\mathbf{x}_i-\mathbf{a})^\intercal \alpha_i \\
&= \sum_{i=1}^{N}(\alpha_i \mathbf{x}_i \mathbf{x}_i^\intercal - 2\alpha_i\mathbf{x}_i\mathbf{a}^\intercal + \alpha_i\mathbf{a}\mathbf{a}^\intercal)\\
&= \sum_{i=1}^{N}(\alpha_i\mathbf{x}_i \mathbf{x}_i^\intercal) - 2\mathbf{a}\mathbf{a}^\intercal + \mathbf{a}\mathbf{a}^\intercal = \sum_{i=1}^{N}(\alpha_i\mathbf{x}_i \mathbf{x}_i^\intercal) - \mathbf{a}\mathbf{a}^\intercal \\
&= \mathbf{X}\text{diag}(\mbox{\boldmath$\alpha$})\mathbf{X}^\intercal - \mathbf{X}\mbox{\boldmath$\alpha$}\mbox{\boldmath$\alpha$}^\intercal\mathbf{X}^\intercal = \mathbf{X}(\mathbb{A}-\boldsymbol{\alpha} \boldsymbol{\alpha}^\intercal)\mathbf{X}^\intercal\\
&= \mathbf{X}\mathbf{L}_{\alpha}\mathbf{X}^\intercal.
\end{split}
\end{equation}
The main idea of PCA and SVDD along with graphs $\mathbf{L}_{pca}$ and $\mathbf{L}_{\alpha}$ are illustrated in
Figure \ref{pcasvdd}.

By approximating the ratio trace in \eqref{eq:framework} with the corresponding trace ratio, we obtain a general subspace learning graph embedding framework with the graph preserving criterion 
\begin{equation} \label{eq:graphpreserving2}
\begin{split}
\textbf{Q}^* &= {\underset{\text{Tr}(\mathbf{QX}\mathbf{L_x}\mathbf{X}^\intercal\mathbf{Q}^\intercal)=m}{\arg\min }\:\: \sum_{i\neq j} ( \mathbf{Qx}_i - \mathbf{Qx}_j )^2\alpha_{i}\alpha_j }\\
&= \arg\min \:\: \frac{\text{Tr}(\mathbf{QX}\mathbf{L}_\alpha\mathbf{X}^\intercal\mathbf{Q}^\intercal)}{\text{Tr}(\mathbf{QX}\mathbf{L}_x\mathbf{X}^\intercal\mathbf{Q}^\intercal)}.
\end{split}
\end{equation}
The criteria minimized in the previously proposed SSVDD \cite{sohrab2018subspace, SOHRAB2021107648} and ESSVDD \cite{9133428} are special cases of the proposed framework and correspond to variants GESSVDD-0-GR-min and GESSVDD-I-GR-min. We conclude that SSVDD minimizes the weighted variance of the support vectors and outliers, while having an orthogonality constraint. ESSVDD also minimizes the weighted variance of the support vectors and outliers, while simultaneously maximizing the total scatter of the centered inputs.

Previously, SSVDD and ESSVDD used the gradient-based update of the projection vector. It should be noted that while the gradient-based approach moves only a single step toward the optimum of \eqref{LL}, the spectral and spectral regression-based updates proposed in Section \ref{ssec:algo} directly jump to the optimum. This may help the overall iterative GESSVDD process converge faster, but it may also introduce some instability, because the objectives of the iteration steps may be contradictory.

To summarize, the new framework in \eqref{eq:framework} places subspace learning for SVDD in the general graph embedding framework with a fixed data-dependent SVDD graph $\mathbf{L}_\alpha$, which resembles PCA on the support vectors and outliers, and an additional constraint graph $\mathbf{L}_x$, which allows to incorporate other meaningful data relationships to the subspace learning step. When the overall objective function in \eqref{eq:framework} is minimized, $\mathbf{L}_\alpha$ represents data relationships to be minimized and $\mathbf{L}_x$ represents data relationships to be maximized. In the earlier works, the overall objective function has been minimized via gradient-descent. However, the new framework hints that it can also make sense to reverse the objective and maximize instead of minimizing. Also this approach has been previously followed in the literature in \cite{hoffmann2007kernelpca}, where kernel PCA was successfully applied for novelty detection. 

Intuitively, the original minimization of $\mathbf{L}_\alpha$ focuses on dimensions where the target class samples are the most similar, which indeed may help to discriminate the class from (unseen) other classes. On the other hand,
from the similarity to PCA, we understand that these dimensions may be the dimensions that are not providing useful information in general (the corresponding PCA would discard them). Therefore, it is necessary to combine the criterion on $\mathbf{L}_\alpha$ with another criterion so that the combination can help to preserve the overall variance and minimize intra-class similarity simultaneously. In general, it may not be clear which criterion to minimize and which to maximize, but when considering the intra-cluster based graphs $\mathbf{L}_w$ and $\mathbf{L}_b$, an intuitive assumption is that within-cluster scatter $\mathbf{L}_w$ should be minimized (i.e., \eqref{eq:framework} maximized), while the between-cluster scatter $\mathbf{L}_b$ is more reasonable to be maximized (i.e., \eqref{eq:framework} minimized).

\subsection{Complexity analysis}\label{complexityanalysis}
The proposed GESSVDD comprises three solutions: 1) gradient-based, 2) spectral, and 3) spectral regression-based updates.
 We first carry out the complexity analysis of the main algorithm (\ref{mainalgo}), which contains the shared steps for all the updates, and then proceed to the steps different in each solution update. The following steps contribute to the overall complexity of the Algorithm \ref{mainalgo}: 
\begin{enumerate}
\item Initializing of the projection matrix $\mathbf{Q}$ via PCA comprises two steps, i.e., computing the covariance matrix and then the eigenvalue decomposition. The complexity of these steps is $\mathcal{O}\big(ND\times min(N,D)\big)$ and $\mathcal{O}\big(D^3\big)$, respectively.
\item Computing $\mathbf{S}_x=\mathbf{X}\mathbf{L}_x\mathbf{X}^\intercal$ for a given $\mathbf{L}_x$ has the complexity of $\mathcal{O}(DN^2+ND^2)$.
\item Computing $\mathbf{S}_Q=\mathbf{Q}\mathbf{S}_x\mathbf{Q}^\intercal$ has the complexity of $\mathcal{O}(dD^2+d^2D)$. Since, $D>d$, the complexity becomes $\mathcal{O}(dD^2)$. 
\item Computing $\mathbf{S}_{inv}$ and the square-root of the matrix $\mathbf{S}_Q$ have the complexity of $\mathcal{O}(N^3)$. 
\item SVDD has the complexity of $\mathcal{O}\big({N}^3\big)$ for $N$ data points \cite{zheng2016smoothly}. \item The complexity of QR decomposition is $\mathcal{O}(d{D}^2)$ \cite{sharma2013qr}.
\end{enumerate}

Dropping relatively lower computational costs and adding the rest, the complexity becomes $\mathcal{O}\big({N}^3+{D}^3\big)$. The total number of samples is assumed to be always greater than the dimensionality; hence the complexity becomes $\mathcal{O}\big({N}^3\big)$. The complexity of each Sub-algorithm \ref{gradientalgo}, \ref{eigenvaluealgo}, and \ref{spectralalgo} is $\mathcal{O}\big({N}^3\big)$. We provide the details of complexity analysis of Sub-algorithms 1, 2, and 3 in Sections \textit{1.1. Complexity analysis of gradient-based update}, \textit{1.2. Complexity analysis of spectral-based update}, and
\textit{1.3. Complexity analysis of spectral regression-based update} respectively in the supplementary material. Adding the complexity of each Sub-algorithm to the main algorithm, the overall complexity remains at $\mathcal{O}(N^3)$, which is the same as for the original SVDD \cite{zheng2016smoothly}. Moreover, in the non-linear case, the steps involved in NPT have the complexity of $\mathcal{O}(N^3)$; thus, the complexity in terms of the big $\mathcal{O}$ notation still stays as $\mathcal{O}(N^3)$.

\section{Experiments}\label{experiments}
\subsection{Datasets and experimental setup}
To evaluate the proposed method's performance, we used nine different datasets. The datasets used in the experiments are Seeds, Qualitative bankruptcy, Somerville happiness, Liver, Iris, Ionosphere, Sonar, Heart (from UCI\footnote{http://archive.ics.uci.edu/ml} machine learning repository) and MNIST \cite{deng2012mnist} with original dimensionality $D$ of 7, 6, 6, 6, 4, 34, 60, 13, and 784 respectively. MNIST  has 10 classes, Seeds and Iris datasets are ternary, while the rest of the datasets are binary.

In Seeds dataset, the classes are named as Kama (S-K), Rosa (S-R), and Canadian (S-C) with 70 samples from each class. In Qualitative bankruptcy, the class labels are bankruptcy (QB-B) and non-bankruptcy (QB-N) with 107 and 143 samples, respectively. The Somerville happiness dataset contains 77 samples from the happy (SH-H) category and 66 from the unhappy (SH-U) category. Liver contain 145 samples from Disorder Present (DP) category and 200 samples from Disorder Absent (DA) category. Iris dataset contains 50 samples from each category of Setosa (I-S), Versicolor (S-VC), and Virginica (S-V). The Ionosphere dataset contains samples categorized as Bad (I-B) and Good (I-G). It contains 126 and 225 samples from bad and good categories, respectively. Sonar dataset has Rock (S-R) and Mines (S-M) as its two classes with 97 samples from Rock and 111 samples from Mines category. Heart dataset contain 139 samples from disease present and 164 samples from disease absent categories, respectively. 

MNIST  dataset contains 5923, 6742, 5958, 6131, 5842, 5421, 5918, 6265, 5851, 5949 samples in the training set for classes 0-9, respectively. In the test set, it contains 980, 1135, 1032, 1010, 982, 892, 958, 1028, 974, and 1009 from corresponding classes (0-9). In our experiments, we select 10$\%$ of the data from MNIST  while keeping the representation of each class in train and test set similar to the original train and test split in the dataset.

We manually created a corrupted version of the heart dataset to report the impact of noise. We added the noise in the manner described in \cite{zhu2004class}. The corrupted data were created by adding pseudo-random values drawn from the standard normal distribution to the features. We bound the range of added noise for the corresponding attribute to the maximum and minimum value of each feature of the target class in the training set.

We converted these datasets into one-class classification datasets by considering a single class at a time as the positive class and the rest as the negative class. For MNIST, the train and test sets are given, so we used the original train and test splits for the experiments. We divided the rest of the datasets into train and test sets by considering 70$\%$ of data as training data and the remaining 30$\%$ as test data. We selected the 70-30 splits randomly by keeping the representation of each class similar to the original dataset. We performed the 70-30$\%$ selection five times; hence we repeated the experiment 5 times for a single scenario where each class is considered a positive class. Note that at this point, both the training and test sets contained samples from both positive and negative classes. We did not use the negative samples in the training set in optimizing the models but only to select the hyperparameters by using five-fold cross-validation within the training set. To this end, four of the folds (only positive items) at a time were used for optimizing the model, and the fifth fold (both positive and negative items) was used to evaluate the performance. Finally, we used the best-performing hyperparameter values to optimize the model with the entire training set (only positive items) and reported the performance over the test set. We used a similar setup for all the competing methods. During the five-fold cross-validation over the training set, we found the best hyperparameters from the following values: $C\in\{0.1,0.2,0.3,0.4,0.5,0.6\}$, $\sigma\in\{10^{-1},10^{0},10^{1},10^{2},10^{3}\}$, $d\in\{1,2,3,4,5,10,20\}$, $\eta\in\{10^{-1},10^{0},10^{1},10^{2},10^{3}\}$.
The number of iterations for all the iterative methods was set to 5.

As our evaluation metrics, we report Geometric Mean $Gmean$, True Positive Rate ($TPR$), True Negative Rate ($TNR$), False Positive Rate ($FPR$), and False Negative Rate ($FNR$), where $TPR=\frac{TP}{P}$, $TNR=\frac{TN}{\mathcal{N}}$,
$FPR=\frac{FP}{\mathcal{N}}$, and $FNR=\frac{FN}{P}$. $TP$, $TN$, $FP$, $FN$, $P$, $\mathcal{N}$ denote true positives, true negatives, false positives, false negatives, and number of positive samples, and number of negative samples, respectively. We use $Gmean$ as the main performance metric as it takes into account both $TPR$ and $TNR$. We also report the standard deviations over the five data~splittings. 

For the proposed method, we consider all the variants introduced in Section~\ref{sssec:variants}: GESSVDD-0, GESSVDD-I, GESSVDD-PCA, GESSVDD-Sw, GESSVDD-Sb, and GESSVDD-kNN. For each, we consider all the alternative solutions (GR-gradient-based, $\mathcal{S}$-spectral, SR-spectral regression-based). The criterion in \eqref{Ltrace} is maximized and minimized in a separate set of experiments respectively for each variant and alternative solution. In order to construct the Laplacians $\mathbf{L}_w$ and $\mathbf{L}_b$, the number of clusters $\mathcal{C}$ was fixed to 5. Moreover, the numbers of neighbours for defining $\mathbf{L}_{kNN}$ was also fixed to 5.

We also carried out sensitivity analysis for the model for the range of hyperparameters. We followed the approach mentioned in \cite{SOHRAB2021107648} for sensitivity analysis. In order to analyze the sensitivity of the model for the corresponding hyperparameter, we fix other hyperparameters to their optimal values found over the training set and record the performance with all the hyperparameter values considered in the given range. 

To evaluate whether the observed differences between different methods are statistically significant, we follow the recommendations of \cite{demsar2006statistical}. We perform Wilcoxon Sign-Ranks test over the average results for the nine datasets to evaluate the pair-wise differences between the methods. The test ranks the differences between each pair of classifiers ignoring the signs and uses the ranks to determine value $T$ as described, e.g., in \cite{demsar2006statistical}. Finally, the $T$ value is compared to a critical value which depends on the number of datasets. In our experiments, we used 9 datasets, which means that the null hypothesis can be rejected at 0.05 significance level level if $T \leq 5$.

\subsection{Experimental results and discussion}

We report the results of the best performing linear and non-linear variants among the proposed variants compared against the previously proposed SSVDD \cite{sohrab2018subspace} and ESSVDD \cite{9133428}, and the competing methods GESVM \cite{mygdalis2016graph}, GESVDD \cite{mygdalis2016graph}, OCSVM \cite{scholkopfu1999sv}, SVDD \cite{tax2004support}, and ESVDD for all datasets in Table~\ref{bestperforming}. In each experiment, a single class is used as a target class and the rest of the data as outliers. The average performance over each dataset is reported in the average (Av.) column. We report the average test results of different variants of the proposed framework over the five splittings of the Seeds, Qualitative bankruptcy, Somerville happiness, Iris, Ionosphere, and Sonar datasets in Table \ref{nonlinearresults} for the non-linear data description, while the results over MNIST, Liver, and Heart datasets are reported in Section S2 of the supplementary material along with the results of all variants in the proposed framework in case of linear data description. We also provide $TPR$, $TNR$, $FPR$, and $FNR$ results in S3 of the supplementary material. The corresponding standard deviations of $Gmean$ over five splits are provided in Section S4 and Section S5 for linear and non-linear cases, respectively, in the supplementary material. Implementations of the proposed framework are available online in GitHub\footnote{https://github.com/fahadsohrab/gessvdd}.
\begin{table*}[h!]
  \footnotesize\setlength{\tabcolsep}{2.2pt}\renewcommand{\arraystretch}{0.70}
  \centering \scriptsize
       \caption{\textit{Gmean} results for linear and non-linear data description over different datasets, selected variants from the proposed framework vs. other one-class classification methods}
\begin{tabular}{llccccccccccccllll}
Dataset                            &                       & \multicolumn{4}{c}{Seeds}                                                                                                                     &                                   & \multicolumn{3}{c}{Qualitative bankruptcy}                                                                &                                   & \multicolumn{3}{c}{Somerville happiness}                                                         &                       & \multicolumn{3}{c}{Liver}                                                        \\ \cline{1-1} \cline{3-6} \cline{8-10} \cline{12-14} \cline{16-18} 
\multicolumn{1}{|l|}{Target class} & \multicolumn{1}{l|}{} & \multicolumn{1}{l}{S-K}           & \multicolumn{1}{l}{S-R}           & \multicolumn{1}{l}{S-C}           & \multicolumn{1}{l|}{Av.}          & \multicolumn{1}{l|}{}             & \multicolumn{1}{l}{QB-B}          & \multicolumn{1}{l}{QB-N}          & \multicolumn{1}{l|}{Av.}          & \multicolumn{1}{l|}{}             & \multicolumn{1}{l}{SH-H}          & \multicolumn{1}{l}{SH-U}          & \multicolumn{1}{l|}{Av.} & \multicolumn{1}{l|}{} & DP             & DA                                   & \multicolumn{1}{l|}{Av.} \\ \cline{1-1} \cline{3-6} \cline{8-10} \cline{12-14} \cline{16-18} 
Linear                             &                       &                                   &                                   &                                   &                                   &                                   &                                   &                                   &                                   &                                   &                                   &                                   &                          &                       &                &                                      &                          \\
\end{tabular}\label{bestperforming}
\end{table*}

From the experimental results comparing different variants of GESSVDD, we observe that in both linear and non-linear methods, the gradient-based solution performs better than the spectral and spectral regression-based solutions in the majority of the cases. The spectral approaches are typically more unstable over iterations as discussed in Section~\ref{ssec:analysis}. When comparing the minimization/maximization, we see that our claim that $\mathbf{L}_w$ should be used with maximization and $\mathbf{L}_b$ with minimization seems to be valid in most cases. Overall, minimization typically leads to better results. Moreover, the performance of kNN graph is better than that of other variants for both min and max cases and for both linear and non-linear~methods.

\begin{table*}[h!]
  \footnotesize\setlength{\tabcolsep}{4.5pt}\renewcommand{\arraystretch}{0.70}
  \centering \scriptsize
         \caption{\textit{Gmean} results for non-linear data description in the proposed framework}
\begin{tabular}{llllllllllllll}
Dataset                            &                       & \multicolumn{4}{c}{Seeds}                                                & \multicolumn{1}{c}{}  & \multicolumn{3}{c}{Qualitative bankruptcy}               & \multicolumn{1}{c}{}  & \multicolumn{3}{c}{Somerville happiness}                 \\ \cline{1-1} \cline{3-6} \cline{8-10} \cline{12-14} 
\multicolumn{1}{|l|}{Target class} & \multicolumn{1}{l|}{} & S-K          & S-R          & S-C      & \multicolumn{1}{l|}{Av.} & \multicolumn{1}{l|}{} & QB-B            & QB-N        & \multicolumn{1}{l|}{Av.} & \multicolumn{1}{l|}{} & SH-H         & SH-U      & \multicolumn{1}{l|}{Av.} \\ \cline{1-1} \cline{3-6} \cline{8-10} \cline{12-14} 
GESSVDD-Sb-$\mathcal{S}$-max                  &                       & 0.75          & 0.91          & 0.88          & 0.85                     &                       & 0.58          & 0.47          & 0.52                     &                       & 0.37          & 0.33          & 0.35                     \\
GESSVDD-Sb-GR-max                  &                       & 0.83          & 0.72          & 0.89          & 0.81                     &                       & 0.82          & 0.36          & 0.59                     &                       & 0.41          & 0.40          & 0.41                     \\
GESSVDD-Sb-SR-max                  &                       & 0.77          & 0.81          & 0.93          & 0.83                     &                       & 0.56          & 0.29          & 0.43                     &                       & 0.44          & 0.30          & 0.37                     \\
GESSVDD-Sb-$\mathcal{S}$-min                  &                       & 0.79          & 0.90          & 0.84          & 0.84                     &                       & 0.61          & 0.42          & 0.52                     &                       & 0.43          & 0.39          & 0.41                     \\
\end{tabular}
\label{nonlinearresults}
\end{table*}
\begin{table*}[h!]
  \footnotesize\setlength{\tabcolsep}{6.2pt}\renewcommand{\arraystretch}{0.60}
  \centering \scriptsize
       \caption{\textit{Gmean} results for linear and non-linear data description over manually created corrupted versions of heart dataset, selected variants from the proposed framework vs. other one-class classification methods}
%
} \\ \cline{1-1} \cline{3-5} \cline{7-9} \cline{11-13} 
\multicolumn{1}{|l|}{Target class} & \multicolumn{1}{l|}{} & \multicolumn{1}{l}{DP}           & \multicolumn{1}{l}{DA}           & \multicolumn{1}{l|}{Av.}          & \multicolumn{1}{l|}{} & \multicolumn{1}{l}{DP}           & \multicolumn{1}{l}{DA}          & \multicolumn{1}{l|}{Av.}          & \multicolumn{1}{l|}{} & \multicolumn{1}{l}{DP}            & \multicolumn{1}{l}{DA}            & \multicolumn{1}{l|}{Av.}           \\ \cline{1-1} \cline{3-5} \cline{7-9} \cline{11-13} 
Linear                             &                       &                                  &                                  &                                   &                       &                                  &                                 &                                   &                       &                                   &                                   &                                    \\
GESSVDD-Sb-GR-max                  &                       & 0.17                             & 0.22                             & 0.20                              &                       & 0.24                             & \textbf{0.36}                   & 0.30                              &                       & 0.36                              & 0.30                              & 0.33                               \\
GESSVDD-kNN-GR-min                 &                       & 0.09                             & 0.00                             & 0.04                              &                       & 0.32                             & 0.08                            & 0.20                              &                       & 0.35                              & 0.44                              & 0.39                               \\
GESSVDD-I-GR-min (ESSVDD)                      &                       & 0.00                             & 0.00                             & 0.00                              &                       & 0.00                             & 0.00                            & 0.00                              &                       & 0.35                              & 0.39                              & 0.37                               \\
GESSVDD-0-GR-min (SSVDD)                       &                       & 0.00                             & 0.17                             & 0.09                              &                       & 0.00                             & 0.00                            & 0.00                              &                       & 0.36                              & 0.40                              & 0.38                               \\
ESVDD                              &                       & 0.00                             & 0.00                             & 0.00                              &                       & 0.00                             & 0.00                            & 0.00                              &                       & 0.43                              & 0.49                              & 0.46                               \\
SVDD                               &                       & \textbf{0.38}                    & \textbf{0.41}                    & \textbf{0.39}                     &                       & \textbf{0.42}                    & 0.27                            & \textbf{0.35}                     &                       & \textbf{0.48}                     & \textbf{0.51}                     & \textbf{0.49}                      \\
OCSVM                              &                       & 0.00                             & 0.00                             & 0.00                              &                       & 0.00                             & 0.00                            & 0.00                              &                       & 0.34                              & 0.41                              & 0.38                               \\ \cline{1-1} \cline{3-5} \cline{7-9} \cline{11-13} 
Non-Linear                         &                       &                                  &                                  &                                   &                       &                                  &                                 &                                   &                       &                                   &                                   &                                    \\
GESSVDD-Sb-GR-max                  &                       & \textbf{0.35}                    & 0.23                             & \textbf{0.29}                     &                       & \textbf{0.33}                    & 0.31                            & \textbf{0.32}                     &                       & 0.27                              & 0.45                              & 0.36                               \\
GESSVDD-kNN-SR-max                 &                       & 0.04                             & 0.03                             & 0.03                              &                       & 0.00                             & 0.00                            & 0.00                              &                       & 0.47                              & 0.40                              & 0.44                               \\
GESSVDD-I-GR-min (ESSVDD)                      &                       & 0.37                             & 0.15                             & 0.26                              &                       & 0.00                             & 0.11                            & 0.06                              &                       & 0.38                              & 0.46                              & 0.42                               \\
GESSVDD-0-GR-min (SSVDD)                       &                       & 0.12                             & \textbf{0.24}                    & 0.18                              &                       & 0.00                             & 0.09                            & 0.04                              &                       & 0.37                              & 0.47                              & 0.42                               \\
ESVDD                              &                       & 0.00                             & 0.00                             & 0.00                              &                       & 0.00                             & 0.03                            & 0.02                              &                       & 0.49                              & 0.52                              & 0.51                               \\
SVDD                               &                       & 0.07                             & 0.15                             & 0.11                              &                       & 0.22                             & 0.07                            & 0.15                              &                       & 0.47                              & 0.45                              & 0.46                               \\
OCSVM                              &                       & 0.00                             & 0.00                             & 0.00                              &                       & 0.00                             & 0.07                            & 0.03                              &                       & \textbf{0.50}                     & 0.49                              & 0.50                               \\
GESVDD-PCA                         &                       & 0.00                             & 0.00                             & 0.00                              &                       & 0.00                             & 0.00                            & 0.00                              &                       & 0.51                              & \textbf{0.53}                     & \textbf{0.52}                      \\
GESVDD-Sw                          &                       & 0.00                             & 0.00                             & 0.00                              &                       & 0.00                             & 0.00                            & 0.00                              &                       & 0.53                              & 0.52                              & \textbf{0.52}                      \\
GESVDD-kNN                         &                       & 0.00                             & 0.00                             & 0.00                              &                       & 0.00                             & 0.00                            & 0.00                              &                       & 0.48                              & 0.52                              & 0.50                               \\
GESVM-PCA                          &                       & 0.00                             & 0.00                             & 0.00                              &                       & 0.00                             & 0.00                            & 0.00                              &                       & 0.49                              & 0.49                              & 0.49                               \\
GESVM-Sw                           &                       & 0.00                             & 0.00                             & 0.00                              &                       & 0.00                             & 0.05                            & 0.03                              &                       & \textbf{0.50}                     & 0.48                              & 0.49                               \\
GESVM-kNN                          &                       & 0.00                             & 0.00                             & 0.00                              &                       & 0.00                             & 0.03                            & 0.02                              &                       & 0.49                              & 0.50                              & 0.49                              
\end{tabular}\label{corruptheartresults}
\end{table*}

Overall in linear methods, it is noted that employing the kNN graph for encoding geometric information in the subspace yields better results also compared to the competing methods in the majority of the cases. Linear GESSVDD-kNN-GR-min variant performs best over 5 and second-best over 2 out of 9 datasets. For non-linear methods, the different variants of GESSVDD have a more varying performance suggesting that finding a suitable graph for the task at hand may be more important. For comparisons, we report the results of a single variant GESSVDD-kNN-SR-max in the non-linear section of Table \ref{bestperforming}. It performs best over the Qualitative Bankruptcy and second-best over Seeds and Ionosphere datasets. For MNIST, we see that some other methods outperform the proposed variants by a clear margin. As the maximum dimensionality allowed for our proposed methods in our experiments is 20, whereas the original dimensionality of MNIST data is 784, we can conclude that the reduction in dimensionality is likely too dramatic for preserving the significant information.    

We applied Wilcoxon Sign-Ranks separately for linear and non-linear methods. We compared all other linear methods in Table~\ref{bestperforming} against our proposed GESSVDD-kNN-GR-min variant and all other non-linear methods Table~\ref{bestperforming} against our proposed GESSVDD-kNN-SR-max variant. We give the $T$-values in Table~\ref{bestperforming} and bold the values if they show that the difference between the methods is statistically significant at 0.05 significance level. Negative values indicate that the other method was performing better than our proposed variant GESSVDD-kNN-GR-min or GESSVDD-kNN-SR-max. We see that GESSVDD-kNN-GR-min outperforms ESVDD and SVDD in a statistically significant manner for linear data description and GESSVDD-kNN-SR-max outperforms ESVDD in a statistically significant manner for non-linear data description. All other differences are statistically insignificant. However, it should be noted that for individual datasets the differences in both ways can be still significant due to different reasons, such as our proposed variant failing with MNIST due to the drastic dimensionality reduction, and it cannot be concluded that the selection of the method is insignificant.

In evaluating the effect of added noise on the features of the heart dataset, it can be noticed that GESSVDD-Sb-GR-max performs second-best in the linear case when only the train or test set is corrupted. In the non-linear case of adding noise to either train or test set, GESSVDD-Sb-GR-max performs best on average. While the competing methods perform better than the proposed methods when both train and test datasets are corrupted, the competing methods underperform severely if the only train or test set is corrupted. There is not a single case where the proposed method would severely underperform. We report the performance of the selected variants of our method along with the competing methods in Table \ref{corruptheartresults}. We provide the $Gmean$ results for all proposed variants over the Heart dataset and its manually created corrupted versions in Section S2 and $TPR$, $TNR$ $FPR$, and $FNR$ results in Section S3 of the supplementary material.

\begin{figure}[t]
	\centering
	\includegraphics[scale=0.38]{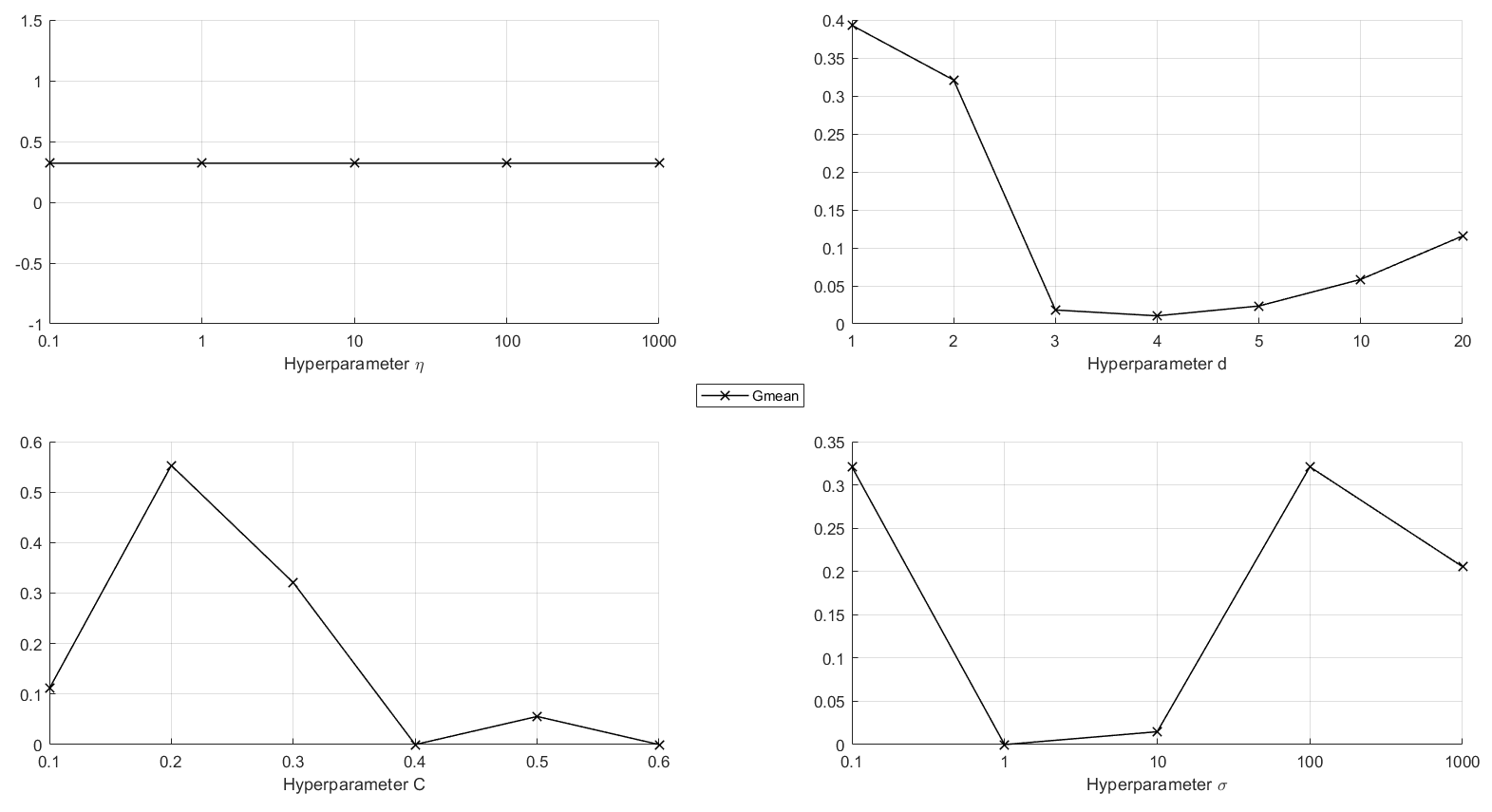}
		%\vspace{-2cm}
		%\rule{35em}{0.5pt}
	\caption{Sensitivity Analysis for non-linear GESSVDD-kNN-SR-max trained over MNIST dataset with target class~0}
	\label{sensitivityplot}
\end{figure}

We carried out a sensitivity analysis of different hyperparameters. Figure \ref{sensitivityplot} shows the sensitivity plot for non-linear GESSVDD-kNN-SR-max trained over MNIST dataset with target class 0. For all other variants, we provide the plots of sensitivity analysis in Section S6 of the supplementary material. We observe that the performance of GESSVDD-kNN-SR-max is not sensitive to the hyperparameter $\eta$. In the case of increasing the value of hyperparameter $d$, a sudden drop and then a steady rise in the performance is observed over the range of values. We also notice the poor performance of the model at higher values of hyperparameter $C$; moreover, a varying performance is noticed at different values for hyperparameter $\sigma$. 

\section{Conclusion}\label{Conclusions}
In this paper, we formulated subspace learning for one-class classification in the graph embedding framework and discussed the novel insights obtained from this formulation. In particular, we showed that subspace learning for SVDD applies a weighted PCA over the support vectors and outliers to define the projection matrix and we discussed how this information can be combined with other data relationships in the optimization process via an adaptable graph. We also formulated a novel Graph-Embedded Subspace Support Vector Data Description with gradient-based, spectral, and spectral regression-based solutions and different adaptable graphs. 
We reported the experimental results over nine different datasets by considering each class of a dataset as a target class at a time. The results showed that the proposed framework with the kNN graph as the adaptable graph had the best overall performance, while the gradient-based solution was more stable than the spectral and spectral regression-based solutions.

While the proposed framework showed promising results over different datasets and can be applied on different domain applications, there are some limitations that can be taken into account in the future. The methods exploit only a single Laplacian $\mathbf{L}_x$ to enforce local/global data relations relevant to the task. This can be enhanced by exploiting multiple graphs by combining the geometric data relationships using a weight parameter. 

In the future, we plan to extend the proposed methods in the framework by investigating other kernel types in the non-linear case. The proposed framework can also be extended to multimodal one-class classification, where data is projected from multiple modalities to a joint subspace.

\section{Acknowledgement}\label{Acknowledgement}
This work has been supported by NSF IUCRC CVDI, project AMALIA funded by Business Finland and DSB, as well as projects Mad@work and Stroke-Data funded by Haltian. The work of Jenni Raitoharju was supported by Academy of Finland project 324475.
\bibliographystyle{elsarticle-num}
\bibliography{bibliography}

\renewcommand{\thesection}{S\arabic{section}}
%% The amsthm package provides extended theorem environments
%% \usepackage{amsthm}

%% The lineno packages adds line numbers. Start line numbering with
%% \begin{linenumbers}, end it with \end{linenumbers}. Or switch it on
%% for the whole article with \linenumbers after \end{frontmatter}.
%% \usepackage{lineno}

%% natbib.sty is loaded by default. However, natbib options can be
%% provided with \biboptions{...} command. Following options are
%% valid:

%%   round  -  round parentheses are used (default)
%%   square -  square brackets are used   [option]
%%   curly  -  curly braces are used      {option}
%%   angle  -  angle brackets are used    <option>
%%   semicolon  -  multiple citations separated by semi-colon
%%   colon  - same as semicolon, an earlier confusion
%%   comma  -  separated by comma
%%   numbers-  selects numerical citations
%%   super  -  numerical citations as superscripts
%%   sort   -  sorts multiple citations according to order in ref. list
%%   sort&compress   -  like sort, but also compresses numerical citations
%%   compress - compresses without sorting
%%
%% \biboptions{comma,round}

% \biboptions{}

\journal{pattern recognition}

\newpage
\setcounter{figure}{0}  
\setcounter{table}{0} 
\setcounter{section}{0}  
\setcounter{page}{1}  

\begin{center}\textbf{Graph-Embedded Subspace Support Vector Data Description\\Supplementary Material}
\end{center}

\begin{center}
Fahad Sohrab, Alexandros Iosifidis, Moncef Gabbouj, Jenni Raitoharju
\end{center}

This document contains supplementary material for the proposed Graph-Embedded Subspace Support Vector Data Description (GESSVDD). In Section \ref{complexity}, we provide the complexity analysis of gradient-based, spectral-based, and spectral regression-based updates. In the other sections, we report more extensive results and analysis for different variants of the proposed GESSVDD: In Section \ref{gmeanresults}, we report the Geometric Mean $Gmean$ results of linear data description for all datasets along with $Gmean$ results for \textbf{non-linear} data description for MNIST and Liver datasets. The non-linear $Gmean$ results for all other datasets besides MNIST, Liver, and Heart are given in the manuscript. Section \ref{gmeanresults} also provides $Gmean$ results of the linear and non-linear data description in the proposed framework for different variants over the Heart data set with added noise to train and test sets. In Section \ref{tpretc}, we provide results in terms of True Positive Rate ($TPR$), True Negative Rate ($TNR$), False Positive Rate ($FPR$), and False Negative Rate ($FNR$) for all proposed variants along with the competing methods for all datasets for both linear and non-linear cases. We provide the standard deviation of $Gmean$ over test sets for the datasets where the 70-30$\%$ selection is made five times for forming the train and test sets in Section \ref{sdlinear} for the linear case and Section \ref{sdnonlinear} for non-linear cases. Finally, all the plots of sensitivity analysis over MNIST data with target class 0 are provided for all the variants in the proposed framework for the non-linear case in Section \ref{sensitivityanalysis}.

\section{Complexity analysis of GESSVDD}\label{complexity}
GESSVDD can be solved via spectral, spectral regression-based, and gradient-based techniques. We provide the detailed complexity analysis of the main algorithm in the manuscript (see Section 3.3. Complexity analysis) and focus on the different solution updates in the following subsections.
\subsection{Complexity analysis of gradient-based update}
The gradient-based update has the following steps involved in the sub-algorithm.
\begin{enumerate}

\item Computing $\mathbf{S}_\alpha=\mathbf{X}\mathbf{L}_\alpha\mathbf{X}^\intercal$ for a given $\mathbf{L}_\alpha$ has a complexity of multiplying three matrices as well. Thus, computing $\mathbf{S}_\alpha$ has the complexity of $\mathcal{O}(DN^2+ND^2)$.
\item Computing $\Delta L=  2\mathbf{S}_{inv} \mathbf{Q}  \mathbf{S}_{\alpha} -2\mathbf{S}_{inv}\mathbf{Q}\mathbf{S}_{\alpha} \mathbf{Q} ^\intercal 
\mathbf{S}_{inv}\mathbf{Q}\mathbf{S}_x^\intercal$ requires several matrix multiplications with the highest complexity being $\mathcal{O}(dD^2)$ and, thus, the overall complexity of this step also becomes $\mathcal{O}(dD^2)$.
\item The complexity of updating the $\mathbf{Q}$ is $\mathcal{O}\big(dD\big)$. 
\end{enumerate}
After adding all complexities of the gradient-based updates along with the complexity of the main algorithm, the gradient-based solution has a complexity of $\mathcal{O}(N^3)$. In a non-linear case, the kernel matrix $\mathbf{K}$ is formed, centralized, and decomposed via eigendecomposition. These steps have the complexity of $\mathcal{O}(N^3)$. The dimensionality of data in non-linear case changes from $D$ to $N$ for all corresponding steps. The total complexity of the non-linear version stays at $\mathcal{O}(N^3)$.

\subsection{Complexity analysis of spectral-based update}
The main relatively intensive computational steps in the spectral-based update are computing $\mathbf{S}_\alpha$ and the generalized eigenvalue problem.
\begin{enumerate}
\item Computing $\mathbf{S}_\alpha$ has the complexity of $\mathcal{O}(DN^2+ND^2)$.
\item Solving the eigenvalue problem $\mathbf{S}_{\alpha}\mathbf{q}=\upsilon\mathbf{S}_x\mathbf{q} $ has the complexity of $\mathcal{O}(D^3)$.
\end{enumerate}
By adding the above complexities, the complexity of spectral-based update becomes $\mathcal{O}(DN^2+ND^2+D^3)$. By adding this to the complexity of the main algorithm, the complexity becomes $\mathcal{O}\big({N}^3+DN^2+ND^2+D^3\big)$. Hence for \textbf{linear} case, the complexity is $\mathcal{O}\big({N}^3\big)$. In a non-linear case, the dimensionality of the data changes from $D$ to $N$. In this case, the complexity in terms $\mathcal{O}\big(4{N}^3\big)$. In terms of $\mathcal{O}$ notation, the complexity in non-linear case stays at $\mathcal{O}\big({N}^3\big)$.

\subsection{Complexity analysis of spectral regression-based update}
The main computational steps in the spectral regression-based update are computing the generalized eigenvalue problem and obtaining the projection matrix $\mathbf{Q}$ in a least-square sense.

\begin{enumerate}

\item Solving the generalized eigenvalue problem $\mathbf{L}_{\alpha}\mathbf{t}=\upsilon\mathbf{L}_x\mathbf{t}$ has the complexity of $\mathcal{O}(N^3)$.

\item Solving $\mathbf{Q}=\mathbf{T}^\intercal\mathbf{X}^\intercal(\mathbf{X}\mathbf{X}^\intercal + \eta \mathbf{I})^{-1}$ involves the following steps. Computing $\mathbf{X}\mathbf{X}^\intercal$ has the complexity of $\mathcal{O}(D^2N)$. The complexity of multiplying $\eta$ with each element of $\mathbf{I}$ is  $\mathcal{O}(D^2)$. Adding $\mathbf{X}\mathbf{X}^\intercal$ with $\eta\mathbf{I}$ has the complexity of $\mathcal{O}(D^2)$. Taking inverse of ($\mathbf{X}\mathbf{X}^\intercal+\eta\mathbf{I}$) has the complexity of $D^3$ and multiplying the rest of matrices has the complexity of $dND+D^3$. Adding the complexities of all these steps, the complexity of $\mathbf{Q}=\mathbf{T}^\intercal\mathbf{X}^\intercal(\mathbf{X}\mathbf{X}^\intercal + \eta \mathbf{I})^{-1}$ becomes $\mathcal{O}(D^2N+2D^2+D^3+dND+D^3)$.
\end{enumerate}
Adding the above two complexities and the complexity of the main algorithm and assuming that the dimensionality $D$ of data is always lower than the number of samples $N$, the total complexity in terms of big $\mathcal{O}$ notation becomes $\mathcal{O}(N^3)$. In non-linear case, the steps involved in npt have the complexity of $\mathcal{O}(N^3)$, moreover the dimensionality changes from $D$ to $N$. Thus, the complexity increases but in terms of the big $\mathcal{O}$ notation still stays as $\mathcal{O}(N^3)$.

\clearpage
\section{GESSVDD Gmean results}\label{gmeanresults}
\subsection{Linear GESSVDD Gmean results}
\begin{table*}[h!]
  \footnotesize\setlength{\tabcolsep}{4.2pt}\renewcommand{\arraystretch}{1.30}
  \centering \scriptsize
         \caption{\textit{Gmean} results for \textbf{linear} data description in the proposed framework for MNIST and Liver datasets}
\begin{tabular}{lllllllllllllllll}
Dataset                            &                       & \multicolumn{11}{c}{MNIST}                                                                                                                                                               & \multicolumn{1}{c}{}  & \multicolumn{3}{c}{Liver}                                \\ \cline{1-1} \cline{3-13} \cline{15-17} 
\multicolumn{1}{|l|}{Target class} & \multicolumn{1}{l|}{} & 0             & 1             & 2             & 3             & 4             & 5             & 6             & 7             & 8             & 9             & \multicolumn{1}{l|}{Av.} & \multicolumn{1}{l|}{} & DP            & DA            & \multicolumn{1}{l|}{Av.} \\ \cline{1-1} \cline{3-13} \cline{15-17} 
GESSVDD-Sb-$\mathcal{S}$-max                   &                       & 0.48          & 0.67          & 0.45          & 0.30          & 0.55          & 0.49          & 0.58          & 0.45          & 0.45          & 0.55          & 0.50                     &                       & 0.38          & 0.45          & 0.41                     \\
GESSVDD-Sb-GR-max                  &                       & \textbf{0.62} & 0.64          & 0.28          & 0.45          & 0.54          & 0.34          & 0.49          & 0.33          & 0.44          & 0.50          & 0.46                     &                       & 0.47          & 0.46          & 0.47                     \\
GESSVDD-Sb-SR-max                  &                       & 0.48          & 0.59          & \textbf{0.48} & 0.44          & \textbf{0.63} & 0.53          & 0.12          & 0.65          & \textbf{0.56} & 0.57          & 0.50                     &                       & \textbf{0.51} & 0.38          & 0.45                     \\
\end{tabular}
\end{table*}

\begin{table*}[h!]
\footnotesize\setlength{\tabcolsep}{4.7pt}\renewcommand{\arraystretch}{0.80}
  \centering \scriptsize
         \caption{\textit{Gmean} results for \textbf{linear} data description in the proposed framework for Seeds, Qualitative bankruptcy, Somerville happiness, Iris, Ionosphere and Sonar datasets}
\begin{tabular}{llllllllllllll}
Dataset                            &                       & \multicolumn{4}{c}{Seeds}                                                & \multicolumn{1}{c}{}  & \multicolumn{3}{c}{Qualitative bankruptcy}               & \multicolumn{1}{c}{}  & \multicolumn{3}{c}{Somerville happiness}                 \\ \cline{1-1} \cline{3-6} \cline{8-10} \cline{12-14} 
\multicolumn{1}{|l|}{Target class} & \multicolumn{1}{l|}{} & S-K          & S-R          & S-C      & \multicolumn{1}{l|}{Av.} & \multicolumn{1}{l|}{} & QB-B            & QB-N        & \multicolumn{1}{l|}{Av.} & \multicolumn{1}{l|}{} & SH-H         & SH-U      & \multicolumn{1}{l|}{Av.} \\ \cline{1-1} \cline{3-6} \cline{8-10} \cline{12-14} 
GESSVDD-Sb-$\mathcal{S}$-max                  &                       & 0.71          & 0.60          & 0.85          & 0.72                     &                       & 0.64          & 0.31          & 0.47                     &                       & 0.35          & 0.33          & 0.34                     \\
GESSVDD-Sb-GR-max                  &                       & 0.74          & 0.69          & 0.67          & 0.70                     &                       & 0.65          & 0.44          & 0.55                     &                       & 0.45          & \textbf{0.49} & \textbf{0.47}            \\
GESSVDD-Sb-SR-max                  &                       & 0.76          & 0.90          & 0.91          & 0.86                     &                       & 0.59          & 0.30          & 0.44                     &                       & 0.49          & 0.31          & 0.40                     \\
GESSVDD-Sb-$\mathcal{S}$-min                  &                       & 0.76          & 0.85          & 0.85          & 0.82                     &                       & 0.60          & 0.44          & 0.52                     &                       & 0.45          & 0.37          & 0.41                     \\
\end{tabular}
\label{linearresults}
\end{table*}

\begin{table*}[h!]
  \footnotesize\setlength{\tabcolsep}{4.7pt}\renewcommand{\arraystretch}{1.20}
  \centering \scriptsize
         \caption{\textit{Gmean} results for \textbf{linear} data description over Heart dataset and its manually created corrupted versions}
%
} \\ \cline{1-1} \cline{3-5} \cline{7-9} \cline{11-13} \cline{15-17} 
\multicolumn{1}{|l|}{Target class} & \multicolumn{1}{l|}{} & DP                          & DA                          & \multicolumn{1}{l|}{Av.}               & \multicolumn{1}{l|}{} & DP                            & DA                            & \multicolumn{1}{l|}{Av.}                & \multicolumn{1}{l|}{} & DP                            & DA                           & \multicolumn{1}{l|}{Av.}                & \multicolumn{1}{l|}{} & DP                             & DA                             & \multicolumn{1}{l|}{Av.}                 \\ \cline{1-1} \cline{3-5} \cline{7-9} \cline{11-13} \cline{15-17} 
GESSVDD-Sb-$\mathcal{S}$-max                   &                       & 0.52                        & 0.44                        & 0.48                                   &                       & 0.20                          & 0.22                          & 0.21                                    &                       & 0.12                          & 0.07                         & 0.09                                    &                       & 0.42                           & 0.36                           & 0.39                                     \\
GESSVDD-Sb-GR-max                  &                       & 0.49                        & 0.48                        & 0.49                                   &                       & 0.17                          & 0.22                          & 0.20                                    &                       & 0.24                          & \textbf{0.36}                & \textbf{0.30}                           &                       & 0.36                           & 0.30                           & 0.33                                     \\
GESSVDD-Sb-SR-max                  &                       & 0.52                        & 0.43                        & 0.47                                   & \textbf{}             & 0.15                          & 0.11                          & 0.13                                    &                       & \textbf{0.35}                 & 0.17                         & 0.26                                    &                       & 0.44                           & 0.41                           & 0.43                                     \\
GESSVDD-Sb-$\mathcal{S}$-min                   &                       & 0.51                        & 0.48                        & 0.49                                   &                       & 0.26                          & \textbf{0.26}                 & \textbf{0.26}                           &                       & 0.05                          & 0.06                         & 0.05                                    &                       & 0.35                           & 0.42                           & 0.38                                     \\
GESSVDD-Sb-GR-min                  &                       & 0.59                        & 0.68                        & \textbf{0.64}                          &                       & 0.14                          & 0.23                          & 0.18                                    &                       & 0.21                          & 0.23                         & 0.22                                    &                       & 0.32                           & \textbf{0.46}                  & 0.39                                     \\
GESSVDD-Sb-SR-min                  &                       & 0.55                        & 0.53                        & 0.54                                   & \textbf{}             & \textbf{0.27}                 & 0.11                          & 0.19                                    &                       & 0.24                          & 0.09                         & 0.16                                    &                       & 0.44                           & 0.43                           & 0.44                                     \\
GESSVDD-Sw-$\mathcal{S}$-max                   &                       & 0.51                        & 0.49                        & 0.50                                   &                       & 0.13                          & 0.14                          & 0.13                                    &                       & 0.00                          & 0.00                         & 0.00                                    &                       & 0.26                           & 0.40                           & 0.33                                     \\
GESSVDD-Sw-GR-max                  &                       & 0.54                        & 0.66                        & 0.60                                   &                       & 0.21                          & 0.19                          & 0.20                                    &                       & 0.00                          & 0.00                         & 0.00                                    &                       & 0.30                           & 0.40                           & 0.35                                     \\
GESSVDD-Sw-SR-max                  &                       & 0.52                        & 0.57                        & 0.55                                   &                       & 0.03                          & 0.00                          & 0.01                                    &                       & 0.00                          & 0.00                         & 0.00                                    &                       & 0.34                           & 0.42                           & 0.38                                     \\
GESSVDD-Sw-$\mathcal{S}$-min                   &                       & 0.49                        & 0.61                        & 0.55                                   &                       & 0.09                          & 0.12                          & 0.10                                    &                       & 0.00                          & 0.00                         & 0.00                                    &                       & 0.30                           & 0.43                           & 0.37                                     \\
GESSVDD-Sw-GR-min                  &                       & \textbf{0.62}               & 0.60                        & 0.61                                   &                       & 0.21                          & 0.20                          & 0.20                                    &                       & 0.00                          & 0.00                         & 0.00                                    &                       & 0.33                           & 0.37                           & 0.35                                     \\
GESSVDD-Sw-SR-min                  &                       & 0.50                        & 0.58                        & 0.54                                   &                       & 0.00                          & 0.03                          & 0.01                                    &                       & 0.03                          & 0.00                         & 0.01                                    &                       & 0.29                           & 0.33                           & 0.31                                     \\
GESSVDD-kNN-$\mathcal{S}$-max                  &                       & 0.56                        & 0.55                        & 0.55                                   &                       & 0.04                          & 0.13                          & 0.08                                    &                       & 0.04                          & 0.00                         & 0.02                                    &                       & 0.41                           & 0.44                           & 0.42                                     \\
GESSVDD-kNN-GR-max                 &                       & 0.53                        & 0.60                        & 0.57                                   &                       & 0.08                          & 0.19                          & 0.14                                    &                       & 0.14                          & 0.08                         & 0.11                                    &                       & 0.40                           & 0.38                           & 0.39                                     \\
GESSVDD-kNN-SR-max                 &                       & 0.47                        & 0.57                        & 0.52                                   &                       & 0.00                          & 0.06                          & 0.03                                    &                       & 0.00                          & 0.00                         & 0.00                                    &                       & 0.39                           & 0.43                           & 0.41                                     \\
GESSVDD-kNN-$\mathcal{S}$-min                  &                       & 0.53                        & 0.54                        & 0.54                                   &                       & 0.08                          & 0.03                          & 0.05                                    &                       & 0.08                          & 0.00                         & 0.04                                    &                       & \textbf{0.48}                  & 0.43                           & \textbf{0.45}                            \\
GESSVDD-kNN-GR-min                 &                       & 0.54                        & 0.61                        & 0.58                                   &                       & 0.09                          & 0.00                          & 0.04                                    &                       & 0.32                          & 0.08                         & 0.20                                    &                       & 0.35                           & 0.44                           & 0.39                                     \\
GESSVDD-kNN-SR-min                 &                       & 0.53                        & 0.58                        & 0.55                                   &                       & 0.00                          & 0.13                          & 0.06                                    &                       & 0.00                          & 0.00                         & 0.00                                    &                       & 0.35                           & 0.42                           & 0.38                                     \\
GESSVDD-PCA-$\mathcal{S}$-max                  &                       & 0.50                        & 0.58                        & 0.54                                   &                       & 0.05                          & 0.00                          & 0.02                                    & \textbf{}             & 0.00                          & 0.00                         & 0.00                                    &                       & 0.33                           & 0.38                           & 0.35                                     \\
GESSVDD-PCA-GR-max                 &                       & 0.53                        & 0.53                        & 0.53                                   &                       & 0.00                          & 0.00                          & 0.00                                    &                       & 0.00                          & 0.00                         & 0.00                                    &                       & 0.38                           & 0.43                           & 0.40                                     \\
GESSVDD-PCA-SR-max                 &                       & 0.48                        & 0.54                        & 0.51                                   &                       & 0.00                          & 0.12                          & 0.06                                    &                       & 0.00                          & 0.00                         & 0.00                                    &                       & 0.31                           & 0.39                           & 0.35                                     \\
GESSVDD-PCA-$\mathcal{S}$-min                  &                       & 0.54                        & 0.66                        & 0.60                                   &                       & 0.00                          & 0.13                          & 0.07                                    & \textbf{}             & 0.00                          & 0.00                         & 0.00                                    &                       & 0.30                           & 0.45                           & 0.37                                     \\
GESSVDD-PCA-GR-min                 &                       & 0.54                        & 0.62                        & 0.58                                   &                       & 0.12                          & 0.00                          & 0.06                                    &                       & 0.00                          & 0.00                         & 0.00                                    &                       & 0.38                           & 0.45                           & 0.41                                     \\
GESSVDD-PCA-SR-min                 &                       & 0.54                        & 0.55                        & 0.54                                   &                       & 0.07                          & 0.12                          & 0.10                                    &                       & 0.00                          & 0.00                         & 0.00                                    &                       & 0.31                           & 0.36                           & 0.34                                     \\
GESSVDD-I-$\mathcal{S}$-max                    &                       & 0.53                        & 0.54                        & 0.53                                   &                       & 0.06                          & 0.06                          & 0.06                                    & \textbf{}             & 0.00                          & 0.00                         & 0.00                                    &                       & 0.25                           & 0.34                           & 0.30                                     \\
GESSVDD-I-GR-max                   &                       & 0.56                        & 0.60                        & 0.58                                   &                       & 0.20                          & 0.10                          & 0.15                                    &                       & 0.00                          & 0.00                         & 0.00                                    &                       & 0.31                           & 0.42                           & 0.36                                     \\
GESSVDD-I-SR-max                   &                       & 0.48                        & 0.57                        & 0.53                                   &                       & 0.03                          & 0.03                          & 0.03                                    &                       & 0.00                          & 0.00                         & 0.00                                    &                       & 0.41                           & 0.36                           & 0.39                                     \\
GESSVDD-I-$\mathcal{S}$-min                    &                       & 0.56                        & 0.57                        & 0.57                                   &                       & 0.07                          & 0.03                          & 0.05                                    & \textbf{}             & 0.00                          & 0.00                         & 0.00                                    &                       & 0.38                           & 0.43                           & 0.41                                     \\
GESSVDD-I-GR-min (ESSVDD)          &                       & 0.53                        & \textbf{0.69}               & 0.61                                   &                       & 0.00                          & 0.00                          & 0.00                                    &                       & 0.00                          & 0.00                         & 0.00                                    &                       & 0.35                           & 0.39                           & 0.37                                     \\
GESSVDD-I-SR-min                   &                       & 0.50                        & 0.49                        & 0.49                                   &                       & 0.03                          & 0.00                          & 0.01                                    &                       & 0.00                          & 0.00                         & 0.00                                    &                       & 0.35                           & 0.32                           & 0.34                                     \\
GESSVDD-0-$\mathcal{S}$-max                    &                       & 0.48                        & 0.47                        & 0.47                                   &                       & 0.10                          & 0.19                          & 0.14                                    &                       & 0.00                          & 0.00                         & 0.00                                    &                       & 0.14                           & 0.28                           & 0.21                                     \\
GESSVDD-0-GR-max                   &                       & 0.55                        & 0.64                        & 0.59                                   &                       & 0.14                          & 0.18                          & 0.16                                    &                       & 0.00                          & 0.00                         & 0.00                                    &                       & 0.25                           & 0.31                           & 0.28                                     \\
GESSVDD-0-$\mathcal{S}$-min                    &                       & 0.53                        & 0.62                        & 0.57                                   &                       & 0.18                          & 0.00                          & 0.09                                    &                       & 0.00                          & 0.00                         & 0.00                                    &                       & 0.31                           & 0.40                           & 0.36                                     \\
GESSVDD-0-GR-min (SSVDD)           &                       & 0.59                        & 0.62                        & 0.61                                   &                       & 0.00                          & 0.17                          & 0.09                                    &                       & 0.00                          & 0.00                         & 0.00                                    &                       & 0.36                           & 0.40                           & 0.38                                    
\end{tabular}
\end{table*}
\clearpage
\subsection{Non-linear GESSVDD Gmean results}
\begin{table*}[h!]
  \footnotesize\setlength{\tabcolsep}{4.5pt}\renewcommand{\arraystretch}{1.30}
  \centering \scriptsize
         \caption{\textit{Gmean} results for \textbf{non-linear} data description in the proposed framework for MNIST and Liver datasets}
\begin{tabular}{lllllllllllllllll}
Dataset                            &                       & \multicolumn{11}{c}{MNIST}                                                                                                                                                               & \multicolumn{1}{c}{}  & \multicolumn{3}{c}{Liver}                                \\ \cline{1-1} \cline{3-13} \cline{15-17} 
\multicolumn{1}{|l|}{Target class} & \multicolumn{1}{l|}{} & 0             & 1             & 2             & 3             & 4             & 5             & 6             & 7             & 8             & 9             & \multicolumn{1}{l|}{Av.} & \multicolumn{1}{l|}{} & DP            & DA            & \multicolumn{1}{l|}{Av.} \\ \cline{1-1} \cline{3-13} \cline{15-17} 
GESSVDD-Sb-$\mathcal{S}$-max                   &                       & 0.16          & 0.16          & 0.06          & 0.08          & 0.39          & 0.29          & 0.31          & 0.00          & 0.12          & 0.09          & 0.17                     &                       & 0.37          & 0.35          & 0.36                     \\
GESSVDD-Sb-GR-max                  &                       & \textbf{0.62} & 0.32          & 0.48          & 0.41          & 0.20          & 0.33          & 0.21          & 0.38          & 0.11          & \textbf{0.58} & 0.36                     &                       & 0.29          & 0.35          & 0.32                     \\
GESSVDD-Sb-SR-max                  &                       & 0.12          & 0.50          & 0.26          & \textbf{0.51} & 0.19          & 0.08          & 0.29          & 0.28          & 0.29          & 0.47          & 0.30                     &                       & 0.29          & 0.27          & 0.28                     \\
GESSVDD-Sb-$\mathcal{S}$-min                   &                       & 0.16          & 0.16          & 0.06          & 0.08          & 0.39          & 0.29          & 0.31          & 0.00          & 0.12          & 0.09          & 0.17                     &                       & 0.34          & 0.33          & 0.33                     \\
GESSVDD-Sb-GR-min                  &                       & \textbf{0.62} & 0.32          & 0.48          & 0.41          & 0.20          & 0.33          & 0.21          & 0.38          & 0.11          & \textbf{0.58} & 0.36                     &                       & 0.35          & 0.24          & 0.30                     \\
GESSVDD-Sb-SR-min                  &                       & 0.12          & 0.50          & 0.26          & \textbf{0.51} & 0.19          & 0.08          & 0.29          & 0.28          & 0.29          & 0.47          & 0.30                     &                       & 0.34          & 0.29          & 0.31                     \\
GESSVDD-Sw-$\mathcal{S}$-max                   &                       & 0.07          & 0.05          & 0.25          & 0.14          & 0.28          & 0.11          & 0.18          & 0.33          & 0.05          & 0.10          & 0.16                     &                       & \textbf{0.45} & 0.32          & 0.38                     \\
GESSVDD-Sw-GR-max                  &                       & 0.25          & 0.22          & 0.09          & 0.44          & 0.12          & \textbf{0.52} & 0.23          & 0.43          & 0.20          & 0.17          & 0.27                     &                       & 0.36          & 0.40          & 0.38                     \\
GESSVDD-Sw-SR-max                  &                       & 0.24          & \textbf{0.54} & 0.15          & 0.27          & 0.46          & 0.32          & 0.45          & 0.43          & 0.19          & 0.31          & 0.34                     &                       & 0.39          & 0.47          & 0.43                     \\
GESSVDD-Sw-$\mathcal{S}$-min                   &                       & 0.07          & 0.05          & 0.25          & 0.14          & 0.28          & 0.11          & 0.18          & 0.33          & 0.05          & 0.10          & 0.16                     &                       & 0.24          & 0.33          & 0.28                     \\
GESSVDD-Sw-GR-min                  &                       & 0.25          & 0.22          & 0.09          & 0.44          & 0.12          & \textbf{0.52} & 0.23          & 0.43          & 0.20          & 0.17          & 0.27                     &                       & 0.33          & 0.28          & 0.30                     \\
GESSVDD-Sw-SR-min                  &                       & 0.24          & \textbf{0.54} & 0.15          & 0.27          & 0.46          & 0.32          & 0.45          & 0.43          & 0.19          & 0.31          & 0.34                     &                       & 0.31          & 0.40          & 0.36                     \\
GESSVDD-kNN-$\mathcal{S}$-max                  &                       & 0.08          & 0.23          & \textbf{0.53} & 0.06          & \textbf{0.55} & 0.31          & 0.31          & 0.26          & 0.27          & 0.25          & 0.28                     &                       & 0.39          & 0.32          & 0.36                     \\
GESSVDD-kNN-GR-max                 &                       & \textbf{0.62} & 0.32          & 0.49          & 0.49          & 0.24          & 0.47          & 0.18          & 0.40          & 0.37          & 0.55          & \textbf{0.41}            &                       & 0.25          & 0.45          & 0.35                     \\
GESSVDD-kNN-SR-max                 &                       & 0.38          & 0.53          & 0.16          & 0.34          & 0.49          & 0.46          & \textbf{0.48} & 0.43          & 0.31          & 0.50          & \textbf{0.41}            &                       & 0.41          & 0.42          & 0.41                     \\
GESSVDD-kNN-$\mathcal{S}$-min                  &                       & 0.08          & 0.23          & \textbf{0.53} & 0.06          & \textbf{0.55} & 0.31          & 0.31          & 0.26          & 0.27          & 0.25          & 0.28                     &                       & 0.33          & 0.31          & 0.32                     \\
GESSVDD-kNN-GR-min                 &                       & \textbf{0.62} & 0.32          & 0.49          & 0.49          & 0.24          & 0.47          & 0.18          & 0.40          & 0.37          & 0.55          & \textbf{0.41}            &                       & 0.39          & 0.39          & 0.39                     \\
GESSVDD-kNN-SR-min                 &                       & 0.38          & 0.53          & 0.16          & 0.34          & 0.49          & 0.46          & \textbf{0.48} & 0.43          & 0.31          & 0.50          & \textbf{0.41}            &                       & 0.40          & 0.44          & 0.42                     \\
GESSVDD-PCA-$\mathcal{S}$-max                  &                       & 0.19          & 0.33          & 0.15          & 0.17          & 0.18          & 0.31          & 0.05          & \textbf{0.54} & 0.24          & 0.17          & 0.23                     &                       & 0.28          & 0.37          & 0.32                     \\
GESSVDD-PCA-GR-max                 &                       & 0.45          & 0.25          & 0.19          & 0.29          & 0.27          & 0.27          & 0.41          & 0.08          & 0.28          & 0.28          & 0.28                     &                       & 0.39          & 0.34          & 0.37                     \\
GESSVDD-PCA-SR-max                 &                       & 0.60          & 0.51          & 0.20          & 0.26          & 0.47          & 0.16          & 0.43          & 0.16          & 0.30          & 0.39          & 0.35                     &                       & 0.36          & 0.27          & 0.32                     \\
GESSVDD-PCA-$\mathcal{S}$-min                  &                       & 0.19          & 0.33          & 0.15          & 0.17          & 0.18          & 0.31          & 0.05          & \textbf{0.54} & 0.24          & 0.17          & 0.23                     &                       & 0.34          & 0.28          & 0.31                     \\
GESSVDD-PCA-GR-min                 &                       & 0.45          & 0.25          & 0.19          & 0.29          & 0.27          & 0.27          & 0.41          & 0.08          & 0.28          & 0.28          & 0.28                     &                       & 0.31          & 0.34          & 0.32                     \\
GESSVDD-PCA-SR-min                 &                       & 0.60          & 0.51          & 0.20          & 0.26          & 0.47          & 0.16          & 0.43          & 0.16          & 0.30          & 0.39          & 0.35                     &                       & 0.32          & 0.26          & 0.29                     \\
GESSVDD-I-$\mathcal{S}$-max                    &                       & 0.30          & 0.33          & 0.17          & 0.17          & 0.30          & 0.32          & 0.42          & \textbf{0.54} & 0.21          & 0.17          & 0.29                     &                       & 0.42          & 0.27          & 0.34                     \\
GESSVDD-I-GR-max                   &                       & 0.36          & 0.34          & 0.18          & 0.09          & 0.19          & \textbf{0.52} & 0.46          & 0.43          & 0.36          & 0.21          & 0.31                     &                       & 0.35          & 0.42          & 0.39                     \\
GESSVDD-I-SR-max                   &                       & 0.60          & 0.51          & 0.21          & 0.28          & 0.47          & 0.30          & 0.43          & 0.25          & 0.30          & 0.39          & 0.37                     &                       & 0.35          & 0.42          & 0.39                     \\
GESSVDD-I-$\mathcal{S}$-min                    &                       & 0.30          & 0.33          & 0.17          & 0.17          & 0.30          & 0.32          & 0.42          & \textbf{0.54} & 0.21          & 0.17          & 0.29                     &                       & 0.39          & 0.26          & 0.32                     \\
GESSVDD-I-GR-min (ESSVDD)          &                       & 0.36          & 0.34          & 0.18          & 0.09          & 0.19          & \textbf{0.52} & 0.46          & 0.43          & 0.36          & 0.21          & 0.31                     &                       & 0.40          & \textbf{0.49} & \textbf{0.45}            \\
GESSVDD-I-SR-min                   &                       & 0.60          & 0.51          & 0.21          & 0.28          & 0.47          & 0.30          & 0.43          & 0.25          & 0.30          & 0.39          & 0.37                     &                       & 0.30          & 0.37          & 0.33                     \\
GESSVDD-0-$\mathcal{S}$-max                    &                       & 0.22          & 0.09          & 0.06          & 0.28          & 0.23          & 0.40          & 0.08          & 0.28          & 0.11          & 0.05          & 0.18                     &                       & 0.36          & 0.37          & 0.36                     \\
GESSVDD-0-GR-max                   &                       & 0.60          & 0.34          & 0.48          & 0.39          & 0.43          & 0.49          & 0.43          & 0.35          & \textbf{0.44} & 0.17          & \textbf{0.41}            &                       & 0.36          & 0.44          & 0.40                     \\
GESSVDD-0-$\mathcal{S}$-min                    &                       & 0.22          & 0.09          & 0.06          & 0.28          & 0.23          & 0.40          & 0.08          & 0.28          & 0.11          & 0.05          & 0.18                     &                       & 0.37          & 0.37          & 0.37                     \\
GESSVDD-0-GR-min (SSVDD)           &                       & 0.60          & 0.34          & 0.48          & 0.39          & 0.43          & 0.49          & 0.43          & 0.35          & \textbf{0.44} & 0.17          & \textbf{0.41}            &                       & 0.37          & 0.39          & 0.38                    
\end{tabular}
\end{table*}
\begin{table*}[h!]
  \footnotesize\setlength{\tabcolsep}{4.9pt}\renewcommand{\arraystretch}{1.20}
  \centering \scriptsize
         \caption{\textit{Gmean} results for \textbf{non-linear} data description over Heart dataset and its manually created corrupted versions}
\begin{tabular}{lllllllllllllllll}
Dataset                            &                       & \multicolumn{3}{c}{\begin{tabular}[c]{@{}c@{}}Heart\\ Clean train set\\ Clean test set\end{tabular}} & \multicolumn{1}{c}{}  & \multicolumn{3}{c}{\begin{tabular}[c]{@{}c@{}}Heart \\ Clean train set\\ Corrupted test set\end{tabular}} & \multicolumn{1}{c}{}  & \multicolumn{3}{c}{\begin{tabular}[c]{@{}c@{}}Heart\\ Corrupted train set\\ Clean test set\end{tabular}} & \multicolumn{1}{c}{}  & \multicolumn{3}{c}{\begin{tabular}[c]{@{}c@{}}Heart\\ Corrupted train set\\ Corrupted test set\end{tabular}} \\ \cline{1-1} \cline{3-5} \cline{7-9} \cline{11-13} \cline{15-17} 
\multicolumn{1}{|l|}{Target class} & \multicolumn{1}{l|}{} & DP                          & DA                          & \multicolumn{1}{l|}{Av.}               & \multicolumn{1}{l|}{} & DP                            & DA                            & \multicolumn{1}{l|}{Av.}                & \multicolumn{1}{l|}{} & DP                            & DA                           & \multicolumn{1}{l|}{Av.}                & \multicolumn{1}{l|}{} & DP                             & DA                             & \multicolumn{1}{l|}{Av.}                 \\ \cline{1-1} \cline{3-5} \cline{7-9} \cline{11-13} \cline{15-17} 
GESSVDD-Sb-$\mathcal{S}$-max                   &                       & 0.29                        & 0.33                        & 0.31                                   &                       & 0.42                          & 0.29                          & 0.36                                    &                       & 0.13                          & 0.05                         & 0.09                                    &                       & 0.35                           & 0.40                           & 0.37                                     \\
GESSVDD-Sb-GR-max                  &                       & 0.34                        & 0.31                        & 0.32                                   &                       & 0.35                          & 0.23                          & 0.29                                    &                       & 0.33                          & 0.31                         & \textbf{0.32}                           &                       & 0.27                           & 0.45                           & 0.36                                     \\
GESSVDD-Sb-SR-max                  &                       & 0.32                        & 0.24                        & 0.28                                   & \textbf{}             & 0.22                          & 0.24                          & 0.23                                    &                       & 0.30                          & 0.22                         & 0.26                                    &                       & 0.43                           & 0.44                           & 0.44                                     \\
GESSVDD-Sb-$\mathcal{S}$-min                   &                       & 0.41                        & 0.43                        & 0.42                                   &                       & \textbf{0.44}                 & 0.19                          & 0.32                                    &                       & 0.19                          & 0.19                         & 0.19                                    &                       & 0.35                           & 0.20                           & 0.28                                     \\
GESSVDD-Sb-GR-min                  &                       & 0.45                        & 0.33                        & 0.39                                   &                       & 0.32                          & 0.18                          & 0.25                                    &                       & 0.21                          & 0.32                         & 0.26                                    &                       & 0.33                           & 0.33                           & 0.33                                     \\
GESSVDD-Sb-SR-min                  &                       & 0.46                        & 0.32                        & 0.39                                   & \textbf{}             & 0.25                          & 0.13                          & 0.19                                    &                       & \textbf{0.36}                 & 0.16                         & 0.26                                    &                       & 0.43                           & 0.44                           & 0.43                                     \\
GESSVDD-Sw-$\mathcal{S}$-max                   &                       & 0.43                        & 0.38                        & 0.41                                   &                       & 0.00                          & 0.12                          & 0.06                                    &                       & 0.13                          & 0.00                         & 0.06                                    &                       & 0.36                           & 0.34                           & 0.35                                     \\
GESSVDD-Sw-GR-max                  &                       & 0.16                        & 0.32                        & 0.24                                   &                       & 0.36                          & \textbf{0.39}                 & \textbf{0.38}                           &                       & 0.07                          & 0.11                         & 0.09                                    &                       & 0.42                           & 0.39                           & 0.41                                     \\
GESSVDD-Sw-SR-max                  &                       & 0.43                        & 0.40                        & 0.41                                   &                       & 0.11                          & 0.06                          & 0.08                                    &                       & 0.00                          & 0.06                         & 0.03                                    &                       & 0.43                           & 0.41                           & 0.42                                     \\
GESSVDD-Sw-$\mathcal{S}$-min                   &                       & 0.34                        & 0.46                        & 0.40                                   &                       & 0.00                          & 0.16                          & 0.08                                    &                       & 0.00                          & 0.15                         & 0.08                                    &                       & 0.44                           & 0.43                           & 0.43                                     \\
GESSVDD-Sw-GR-min                  &                       & 0.40                        & 0.47                        & 0.44                                   &                       & 0.18                          & 0.29                          & 0.24                                    &                       & 0.04                          & 0.05                         & 0.05                                    &                       & 0.45                           & 0.44                           & 0.45                                     \\
GESSVDD-Sw-SR-min                  &                       & 0.44                        & 0.30                        & 0.37                                   &                       & 0.03                          & 0.19                          & 0.11                                    &                       & 0.00                          & 0.00                         & 0.00                                    &                       & 0.41                           & 0.33                           & 0.37                                     \\
GESSVDD-kNN-$\mathcal{S}$-max                  &                       & 0.44                        & 0.47                        & 0.45                                   &                       & 0.00                          & 0.00                          & 0.00                                    &                       & 0.14                          & 0.04                         & 0.09                                    &                       & 0.45                           & 0.45                           & 0.45                                     \\
GESSVDD-kNN-GR-max                 &                       & 0.43                        & \textbf{0.57}               & 0.50                                   &                       & 0.28                          & 0.19                          & 0.24                                    &                       & 0.12                          & 0.00                         & 0.06                                    &                       & 0.44                           & 0.38                           & 0.41                                     \\
GESSVDD-kNN-SR-max                 &                       & 0.42                        & 0.43                        & 0.42                                   &                       & 0.04                          & 0.03                          & 0.03                                    &                       & 0.00                          & 0.00                         & 0.00                                    &                       & 0.47                           & 0.40                           & 0.44                                     \\
GESSVDD-kNN-$\mathcal{S}$-min                  &                       & 0.49                        & 0.51                        & 0.50                                   &                       & 0.03                          & 0.00                          & 0.02                                    &                       & 0.03                          & 0.00                         & 0.02                                    &                       & 0.46                           & 0.43                           & 0.44                                     \\
GESSVDD-kNN-GR-min                 &                       & 0.47                        & 0.44                        & 0.46                                   &                       & 0.00                          & 0.00                          & 0.00                                    &                       & 0.00                          & 0.00                         & 0.00                                    &                       & 0.48                           & 0.43                           & 0.46                                     \\
GESSVDD-kNN-SR-min                 &                       & 0.47                        & 0.50                        & 0.49                                   &                       & 0.04                          & 0.07                          & 0.06                                    &                       & 0.06                          & 0.00                         & 0.03                                    &                       & 0.45                           & 0.38                           & 0.41                                     \\
GESSVDD-PCA-$\mathcal{S}$-max                  &                       & 0.37                        & 0.31                        & 0.34                                   &                       & 0.00                          & 0.03                          & 0.01                                    & \textbf{}             & 0.00                          & 0.00                         & 0.00                                    &                       & \textbf{0.49}                  & 0.36                           & 0.42                                     \\
GESSVDD-PCA-GR-max                 &                       & 0.47                        & 0.30                        & 0.39                                   &                       & 0.30                          & 0.34                          & 0.32                                    &                       & 0.13                          & 0.16                         & 0.14                                    &                       & 0.45                           & 0.36                           & 0.40                                     \\
GESSVDD-PCA-SR-max                 &                       & 0.46                        & 0.42                        & 0.44                                   &                       & 0.19                          & 0.05                          & 0.12                                    &                       & 0.06                          & 0.09                         & 0.08                                    &                       & 0.34                           & 0.41                           & 0.37                                     \\
GESSVDD-PCA-$\mathcal{S}$-min                  &                       & 0.24                        & 0.27                        & 0.25                                   &                       & 0.06                          & 0.11                          & 0.08                                    & \textbf{}             & 0.00                          & 0.00                         & 0.00                                    &                       & 0.43                           & 0.39                           & 0.41                                     \\
GESSVDD-PCA-GR-min                 &                       & 0.30                        & 0.28                        & 0.29                                   &                       & 0.28                          & 0.22                          & 0.25                                    &                       & 0.00                          & 0.04                         & 0.02                                    &                       & 0.46                           & \textbf{0.47}                  & \textbf{0.47}                            \\
GESSVDD-PCA-SR-min                 &                       & 0.44                        & 0.44                        & 0.44                                   &                       & 0.10                          & 0.07                          & 0.09                                    &                       & 0.03                          & 0.00                         & 0.02                                    &                       & 0.37                           & 0.41                           & 0.39                                     \\
GESSVDD-I-$\mathcal{S}$-max                    &                       & 0.38                        & 0.39                        & 0.38                                   &                       & 0.06                          & 0.03                          & 0.04                                    & \textbf{}             & 0.04                          & 0.00                         & 0.02                                    &                       & 0.45                           & 0.38                           & 0.42                                     \\
GESSVDD-I-GR-max                   &                       & 0.42                        & 0.28                        & 0.35                                   &                       & 0.37                          & 0.35                          & 0.36                                    &                       & 0.03                          & 0.05                         & 0.04                                    &                       & 0.48                           & 0.31                           & 0.39                                     \\
GESSVDD-I-SR-max                   &                       & 0.49                        & 0.46                        & 0.47                                   &                       & 0.13                          & 0.03                          & 0.08                                    &                       & 0.00                          & 0.00                         & 0.00                                    &                       & 0.43                           & 0.44                           & 0.43                                     \\
GESSVDD-I-$\mathcal{S}$-min                    &                       & 0.44                        & 0.35                        & 0.39                                   &                       & 0.00                          & 0.25                          & 0.12                                    & \textbf{}             & 0.08                          & 0.00                         & 0.04                                    &                       & 0.36                           & 0.39                           & 0.38                                     \\
GESSVDD-I-GR-min (ESSVDD)          &                       & 0.38                        & 0.37                        & 0.37                                   &                       & 0.37                          & 0.15                          & 0.26                                    &                       & 0.00                          & 0.11                         & 0.06                                    &                       & 0.38                           & 0.46                           & 0.42                                     \\
GESSVDD-I-SR-min                   &                       & 0.46                        & 0.44                        & 0.45                                   &                       & 0.15                          & 0.14                          & 0.14                                    &                       & 0.00                          & 0.08                         & 0.04                                    &                       & 0.42                           & 0.45                           & 0.44                                     \\
GESSVDD-0-$\mathcal{S}$-max                    &                       & 0.35                        & 0.34                        & 0.34                                   &                       & 0.24                          & 0.21                          & 0.22                                    &                       & 0.26                          & 0.32                         & 0.29                                    &                       & 0.39                           & 0.36                           & 0.38                                     \\
GESSVDD-0-GR-max                   &                       & 0.45                        & 0.46                        & 0.46                                   &                       & 0.06                          & 0.22                          & 0.14                                    &                       & 0.06                          & \textbf{0.44}                & 0.25                                    &                       & 0.46                           & 0.36                           & 0.41                                     \\
GESSVDD-0-$\mathcal{S}$-min                    &                       & 0.32                        & 0.41                        & 0.37                                   &                       & 0.14                          & 0.16                          & 0.15                                    &                       & \textbf{0.36}                 & 0.09                         & 0.23                                    &                       & 0.34                           & 0.40                           & 0.37                                     \\
GESSVDD-0-GR-min (SSVDD)           &                       & \textbf{0.53}               & 0.49                        & \textbf{0.51}                          &                       & 0.12                          & 0.24                          & 0.18                                    &                       & 0.00                          & 0.09                         & 0.04                                    &                       & 0.37                           & \textbf{0.47}                  & 0.42                                    
\end{tabular}
\end{table*}
\clearpage

\section{GESSVDD TPR TNR FPR FNR results}\label{tpretc}

\subsection{TPR TNR FPR FNR results for \textbf{linear} data description}

\subsubsection{TPR results \textbf{linear} data description}
\begin{table*}[h!]
  \footnotesize\setlength{\tabcolsep}{4.9pt}\renewcommand{\arraystretch}{1.20}
  \centering \scriptsize
         \caption{\textit{TPR} results for \textbf{linear} data description}
\begin{tabular}{llllllllllllll}
Dataset                            &                       & \multicolumn{4}{c}{Seeds}                                                &                       & \multicolumn{3}{c}{Qualitative bankruptcy}               &                       & \multicolumn{3}{c}{Somerville happiness}                 \\ \cline{1-1} \cline{3-6} \cline{8-10} \cline{12-14} 
\multicolumn{1}{|l|}{Target class} & \multicolumn{1}{l|}{} & S-K          & S-R          & S-C      & \multicolumn{1}{l|}{Av.} & \multicolumn{1}{l|}{} & QB-B            & QB-N        & \multicolumn{1}{l|}{Av.} & \multicolumn{1}{l|}{} & SH-H         & SH-U      & \multicolumn{1}{l|}{Av.} \\ \cline{1-1} \cline{3-6} \cline{8-10} \cline{12-14} 
GESSVDD-Sb-$\mathcal{S}$-max                   &                       & 0.77          & 0.76          & 0.88          & 0.80                     &                       & 0.76          & 0.46          & 0.61                     &                       & \textbf{0.83} & 0.76          & 0.80                     \\
GESSVDD-Sb-GR-max                  &                       & 0.90          & 0.62          & 0.65          & 0.72                     &                       & 0.83          & 0.44          & 0.63                     &                       & 0.73          & 0.73          & 0.73                     \\
GESSVDD-Sb-SR-max                  &                       & 0.82          & 0.86          & 0.96          & 0.88                     &                       & 0.58          & 0.46          & 0.52                     &                       & 0.61          & 0.68          & 0.65                     \\
GESSVDD-Sb-$\mathcal{S}$-min                   &                       & 0.81          & 0.75          & 0.92          & 0.83                     &                       & 0.79          & 0.59          & 0.69                     &                       & 0.68          & 0.68          & 0.68                     \\
\end{tabular}
\end{table*}
\begin{table*}[h!]
  \footnotesize\setlength{\tabcolsep}{4.9pt}\renewcommand{\arraystretch}{1.20}
  \centering \scriptsize
         \caption{\textit{TPR} results for \textbf{linear} data description}
\begin{tabular}{llllllllllllll}
Dataset                            &                       & \multicolumn{4}{c}{Iris}                                                 &                       & \multicolumn{3}{c}{Ionosphere}                           &                       & \multicolumn{3}{c}{Sonar}                        \\ \cline{1-1} \cline{3-6} \cline{8-10} \cline{12-14} 
\multicolumn{1}{|l|}{Target class} & \multicolumn{1}{l|}{} & I-S        & S-VC    & S-V     & \multicolumn{1}{l|}{Av.} & \multicolumn{1}{l|}{} & I-B           & I-G          & \multicolumn{1}{l|}{Av.} & \multicolumn{1}{l|}{} & S-R          & S-M & \multicolumn{1}{l|}{Av.} \\ \cline{1-1} \cline{3-6} \cline{8-10} \cline{12-14} 
GESSVDD-Sb-$\mathcal{S}$-max                   &                       & 0.85          & 0.88          & 0.79          & 0.84                     &                       & 0.68          & 0.68          & 0.68                     &                       & 0.62          & 0.31  & 0.46                     \\
GESSVDD-Sb-GR-max                  &                       & 0.87          & 0.89          & 0.85          & 0.87                     &                       & 0.80          & 0.96          & 0.88                     &                       & 0.86          & 0.75  & 0.80                     \\
GESSVDD-Sb-SR-max                  &                       & 0.87          & 0.83          & 0.85          & 0.85                     &                       & 0.58          & 0.21          & 0.39                     &                       & 0.59          & 0.69  & 0.64                     \\
GESSVDD-Sb-$\mathcal{S}$-min                   &                       & 0.83          & 0.84          & 0.79          & 0.82                     &                       & 0.71          & 0.49          & 0.60                     &                       & 0.52          & 0.27  & 0.39                     \\
\end{tabular}
\end{table*}

\begin{table*}[h!]
  \footnotesize\setlength{\tabcolsep}{4.5pt}\renewcommand{\arraystretch}{1.20}
  \centering \scriptsize
         \caption{\textit{TPR} results for \textbf{linear} data description}
\begin{tabular}{lllllllllllllllll}
Dataset                            &                       & \multicolumn{11}{c}{MNIST}                                                                                                                                                               & \multicolumn{1}{c}{}  & \multicolumn{3}{c}{Liver}                                \\ \cline{1-1} \cline{3-13} \cline{15-17} 
\multicolumn{1}{|l|}{Target class} & \multicolumn{1}{l|}{} & 0             & 1             & 2             & 3             & 4             & 5             & 6             & 7             & 8             & 9             & \multicolumn{1}{l|}{Av.} & \multicolumn{1}{l|}{} & DP            & DA            & \multicolumn{1}{l|}{Av.} \\ \cline{1-1} \cline{3-13} \cline{15-17} 
GESSVDD-Sb-$\mathcal{S}$-max                   &                       & 0.89          & 0.69          & 0.33          & 0.30          & 0.46          & 0.65          & 0.78          & 0.64          & 0.79          & 0.65          & 0.62                     &                       & 0.92          & 0.63          & 0.78                     \\
GESSVDD-Sb-GR-max                  &                       & 0.84          & 0.82          & \textbf{0.98} & 0.43          & 0.69          & 0.82          & 0.41          & 0.12          & 0.64          & 0.90          & 0.67                     &                       & 0.51          & 0.60          & 0.56                     \\
GESSVDD-Sb-SR-max                  &                       & 0.64          & 0.47          & 0.62          & 0.32          & 0.69          & 0.37          & 0.15          & 0.69          & 0.45          & 0.51          & 0.49                     &                       & 0.53          & 0.80          & 0.67                     \\
\end{tabular}
\end{table*}
\begin{table*}[h!]
  \footnotesize\setlength{\tabcolsep}{4.9pt}\renewcommand{\arraystretch}{1.20}
  \centering \scriptsize
         \caption{\textit{TPR} results for \textbf{linear} data description}
%
} \\ \cline{1-1} \cline{3-5} \cline{7-9} \cline{11-13} \cline{15-17} 
\multicolumn{1}{|l|}{Target class} & \multicolumn{1}{l|}{} & DP                          & DA                          & \multicolumn{1}{l|}{Av.}               & \multicolumn{1}{l|}{} & DP                            & DA                            & \multicolumn{1}{l|}{Av.}                & \multicolumn{1}{l|}{} & DP                            & DA                           & \multicolumn{1}{l|}{Av.}                & \multicolumn{1}{l|}{} & DP                             & DA                             & \multicolumn{1}{l|}{Av.}                 \\ \cline{1-1} \cline{3-5} \cline{7-9} \cline{11-13} \cline{15-17} 
GESSVDD-Sb-$\mathcal{S}$-max                   &                       & 0.74                        & 0.57                        & 0.65                                   &                       & 0.06                          & 0.08                          & 0.07                                    &                       & 0.68                          & 0.58                         & 0.63                                    &                       & 0.76                           & 0.69                           & 0.73                                     \\
GESSVDD-Sb-GR-max                  &                       & 0.64                        & 0.44                        & 0.54                                   &                       & 0.07                          & 0.08                          & 0.07                                    &                       & 0.80                          & 0.66                         & 0.73                                    &                       & 0.56                           & 0.57                           & 0.56                                     \\
GESSVDD-Sb-SR-max                  &                       & 0.65                        & 0.57                        & 0.61                                   & \textbf{}             & 0.04                          & 0.04                          & 0.04                                    &                       & 0.60                          & 0.40                         & 0.50                                    &                       & 0.63                           & 0.55                           & 0.59                                     \\
GESSVDD-Sb-$\mathcal{S}$-min                   &                       & 0.72                        & 0.48                        & 0.60                                   &                       & 0.08                          & 0.12                          & 0.10                                    &                       & 0.70                          & 0.64                         & 0.67                                    &                       & 0.70                           & 0.75                           & 0.73                                     \\
GESSVDD-Sb-GR-min                  &                       & 0.58                        & 0.70                        & 0.64                                   &                       & 0.05                          & 0.09                          & 0.07                                    &                       & 0.68                          & 0.33                         & 0.50                                    &                       & 0.54                           & 0.39                           & 0.46                                     \\
GESSVDD-Sb-SR-min                  &                       & 0.48                        & 0.63                        & 0.55                                   & \textbf{}             & 0.12                          & 0.04                          & 0.08                                    &                       & 0.75                          & 0.53                         & 0.64                                    &                       & 0.70                           & 0.48                           & 0.59                                     \\
GESSVDD-Sw-$\mathcal{S}$-max                   &                       & \textbf{0.92}               & 0.90                        & 0.91                                   &                       & 0.06                          & 0.04                          & 0.05                                    &                       & \textbf{1.00}                 & \textbf{1.00}                & \textbf{1.00}                           &                       & 0.94                           & 0.92                           & 0.93                                     \\
GESSVDD-Sw-GR-max                  &                       & 0.89                        & 0.86                        & 0.88                                   &                       & 0.14                          & 0.12                          & 0.13                                    &                       & \textbf{1.00}                 & \textbf{1.00}                & \textbf{1.00}                           &                       & 0.90                           & 0.90                           & 0.90                                     \\
GESSVDD-Sw-SR-max                  &                       & 0.84                        & 0.83                        & 0.83                                   &                       & 0.00                          & 0.00                          & 0.00                                    &                       & \textbf{1.00}                 & \textbf{1.00}                & \textbf{1.00}                           &                       & 0.84                           & 0.84                           & 0.84                                     \\
GESSVDD-Sw-$\mathcal{S}$-min                   &                       & 0.86                        & 0.90                        & 0.88                                   &                       & 0.03                          & 0.05                          & 0.04                                    &                       & 0.99                          & \textbf{1.00}                & \textbf{1.00}                           &                       & 0.88                           & 0.82                           & 0.85                                     \\
GESSVDD-Sw-GR-min                  &                       & 0.83                        & 0.82                        & 0.83                                   &                       & 0.15                          & 0.13                          & 0.14                                    &                       & \textbf{1.00}                 & \textbf{1.00}                & \textbf{1.00}                           &                       & 0.84                           & 0.82                           & 0.83                                     \\
GESSVDD-Sw-SR-min                  &                       & 0.86                        & 0.85                        & 0.85                                   &                       & 0.00                          & 0.00                          & 0.00                                    &                       & 0.95                          & \textbf{1.00}                & 0.98                                    &                       & 0.90                           & 0.84                           & 0.87                                     \\
GESSVDD-kNN-$\mathcal{S}$-max                  &                       & 0.75                        & 0.80                        & 0.77                                   &                       & 0.01                          & 0.03                          & 0.02                                    &                       & 0.94                          & \textbf{1.00}                & 0.97                                    &                       & 0.86                           & 0.82                           & 0.84                                     \\
GESSVDD-kNN-GR-max                 &                       & 0.78                        & 0.82                        & 0.80                                   &                       & 0.04                          & 0.07                          & 0.05                                    &                       & 0.94                          & 0.96                         & 0.95                                    &                       & 0.62                           & 0.63                           & 0.62                                     \\
GESSVDD-kNN-SR-max                 &                       & 0.74                        & 0.83                        & 0.78                                   &                       & 0.00                          & 0.01                          & 0.00                                    &                       & \textbf{1.00}                 & \textbf{1.00}                & \textbf{1.00}                           &                       & 0.88                           & 0.73                           & 0.80                                     \\
GESSVDD-kNN-$\mathcal{S}$-min                  &                       & 0.80                        & 0.75                        & 0.78                                   &                       & 0.02                          & 0.00                          & 0.01                                    &                       & \textbf{1.00}                 & \textbf{1.00}                & \textbf{1.00}                           &                       & 0.76                           & 0.76                           & 0.76                                     \\
GESSVDD-kNN-GR-min                 &                       & 0.75                        & 0.78                        & 0.76                                   &                       & 0.02                          & 0.00                          & 0.01                                    &                       & 0.88                          & 0.96                         & 0.92                                    &                       & 0.63                           & 0.77                           & 0.70                                     \\
GESSVDD-kNN-SR-min                 &                       & 0.75                        & 0.83                        & 0.79                                   &                       & 0.00                          & 0.05                          & 0.03                                    &                       & \textbf{1.00}                 & \textbf{1.00}                & \textbf{1.00}                           &                       & 0.70                           & 0.64                           & 0.67                                     \\
GESSVDD-PCA-$\mathcal{S}$-max                  &                       & 0.88                        & 0.88                        & 0.88                                   &                       & 0.01                          & 0.00                          & 0.01                                    & \textbf{}             & 0.99                          & \textbf{1.00}                & \textbf{1.00}                           &                       & 0.87                           & \textbf{0.93}                  & 0.90                                     \\
GESSVDD-PCA-GR-max                 &                       & 0.83                        & 0.85                        & 0.84                                   &                       & 0.00                          & 0.00                          & 0.00                                    &                       & \textbf{1.00}                 & \textbf{1.00}                & \textbf{1.00}                           &                       & 0.85                           & 0.83                           & 0.84                                     \\
GESSVDD-PCA-SR-max                 &                       & 0.85                        & 0.84                        & 0.85                                   &                       & 0.00                          & 0.02                          & 0.01                                    &                       & \textbf{1.00}                 & \textbf{1.00}                & \textbf{1.00}                           &                       & 0.88                           & 0.80                           & 0.84                                     \\
GESSVDD-PCA-$\mathcal{S}$-min                  &                       & 0.85                        & 0.84                        & 0.84                                   &                       & 0.00                          & 0.05                          & 0.03                                    & \textbf{}             & \textbf{1.00}                 & \textbf{1.00}                & \textbf{1.00}                           &                       & 0.90                           & 0.79                           & 0.84                                     \\
GESSVDD-PCA-GR-min                 &                       & 0.82                        & 0.83                        & 0.83                                   &                       & 0.10                          & 0.00                          & 0.05                                    &                       & \textbf{1.00}                 & \textbf{1.00}                & \textbf{1.00}                           &                       & 0.82                           & 0.80                           & 0.81                                     \\
GESSVDD-PCA-SR-min                 &                       & 0.86                        & 0.86                        & 0.86                                   &                       & 0.03                          & 0.03                          & 0.03                                    &                       & \textbf{1.00}                 & \textbf{1.00}                & \textbf{1.00}                           &                       & 0.86                           & 0.88                           & 0.87                                     \\
GESSVDD-I-$\mathcal{S}$-max                    &                       & 0.86                        & 0.86                        & 0.86                                   &                       & 0.02                          & 0.02                          & 0.02                                    & \textbf{}             & 0.98                          & \textbf{1.00}                & 0.99                                    &                       & 0.89                           & 0.85                           & 0.87                                     \\
GESSVDD-I-GR-max                   &                       & 0.83                        & 0.85                        & 0.84                                   &                       & 0.14                          & 0.06                          & 0.10                                    &                       & \textbf{1.00}                 & \textbf{1.00}                & \textbf{1.00}                           &                       & 0.89                           & 0.83                           & 0.86                                     \\
GESSVDD-I-SR-max                   &                       & 0.85                        & 0.85                        & 0.85                                   &                       & 0.00                          & 0.00                          & 0.00                                    &                       & \textbf{1.00}                 & \textbf{1.00}                & \textbf{1.00}                           &                       & 0.90                           & 0.86                           & 0.88                                     \\
GESSVDD-I-$\mathcal{S}$-min                    &                       & 0.8                         & 0.84                        & 0.82                                   &                       & 0.03                          & 0.00                          & 0.01                                    & \textbf{}             & \textbf{1.00}                 & \textbf{1.00}                & \textbf{1.00}                           &                       & 0.81                           & 0.80                           & 0.81                                     \\
GESSVDD-I-GR-min (ESSVDD)          &                       & 0.86                        & 0.83                        & 0.85                                   &                       & 0.00                          & 0.00                          & 0.00                                    &                       & \textbf{1.00}                 & \textbf{1.00}                & \textbf{1.00}                           &                       & 0.85                           & 0.80                           & 0.83                                     \\
GESSVDD-I-SR-min                   &                       & 0.83                        & 0.89                        & 0.86                                   &                       & 0.00                          & 0.00                          & 0.00                                    &                       & \textbf{1.00}                 & \textbf{1.00}                & \textbf{1.00}                           &                       & 0.88                           & 0.92                           & 0.90                                     \\
GESSVDD-0-$\mathcal{S}$-max                    &                       & \textbf{0.92}               & \textbf{0.92}               & \textbf{0.92}                          &                       & 0.03                          & 0.08                          & 0.06                                    &                       & \textbf{1.00}                 & \textbf{1.00}                & \textbf{1.00}                           &                       & \textbf{0.96}                  & 0.92                           & \textbf{0.94}                            \\
GESSVDD-0-GR-max                   &                       & \textbf{0.92}               & 0.89                        & 0.91                                   &                       & 0.06                          & 0.11                          & 0.09                                    &                       & \textbf{1.00}                 & \textbf{1.00}                & \textbf{1.00}                           &                       & 0.95                           & \textbf{0.93}                  & \textbf{0.94}                            \\
GESSVDD-0-$\mathcal{S}$-min                    &                       & 0.84                        & 0.88                        & 0.86                                   &                       & 0.07                          & 0.00                          & 0.04                                    &                       & \textbf{1.00}                 & \textbf{1.00}                & \textbf{1.00}                           &                       & 0.91                           & 0.85                           & 0.88                                     \\
GESSVDD-0-GR-min (SSVDD)           &                       & 0.78                        & 0.84                        & 0.81                                   &                       & 0.00                          & 0.10                          & 0.05                                    &                       & \textbf{1.00}                 & \textbf{1.00}                & \textbf{1.00}                           &                       & 0.87                           & 0.82                           & 0.85                                     \\
ESVDD                              &                       & 0.76                        & 0.80                        & 0.78                                   &                       & 0.00                          & 0.00                          & 0.00                                    &                       & \textbf{1.00}                 & \textbf{1.00}                & \textbf{1.00}                           &                       & 0.82                           & 0.75                           & 0.79                                     \\
SVDD                               &                       & 0.59                        & 0.48                        & 0.54                                   &                       & \textbf{0.53}                 & \textbf{0.73}                 & \textbf{0.63}                           &                       & 0.67                          & 0.26                         & 0.47                                    &                       & 0.54                           & 0.49                           & 0.51                                     \\
OCSVM                              &                       & 0.90                        & 0.85                        & 0.87                                   &                       & 0.00                          & 0.00                          & 0.00                                    &                       & \textbf{1.00}                 & \textbf{1.00}                & \textbf{1.00}                           &                       & 0.87                           & 0.85                           & 0.86                                    
\end{tabular}
\end{table*}
\clearpage

 \subsubsection{TNR results \textbf{linear} data description}
\begin{table*}[h!]
  \footnotesize\setlength{\tabcolsep}{4.9pt}\renewcommand{\arraystretch}{1.20}
  \centering \scriptsize
         \caption{\textit{TNR} results for \textbf{linear} data description}
\begin{tabular}{llllllllllllll}
Dataset                            &                       & \multicolumn{4}{c}{Seeds}                                                &                       & \multicolumn{3}{c}{Qualitative bankruptcy}               &                       & \multicolumn{3}{c}{Somerville happiness}                 \\ \cline{1-1} \cline{3-6} \cline{8-10} \cline{12-14} 
\multicolumn{1}{|l|}{Target class} & \multicolumn{1}{l|}{} & S-K          & S-R          & S-C      & \multicolumn{1}{l|}{Av.} & \multicolumn{1}{l|}{} & QB-B            & QB-N        & \multicolumn{1}{l|}{Av.} & \multicolumn{1}{l|}{} & SH-H         & SH-U      & \multicolumn{1}{l|}{Av.} \\ \cline{1-1} \cline{3-6} \cline{8-10} \cline{12-14} 
GESSVDD-Sb-$\mathcal{S}$-max                   &                       & 0.67          & 0.50          & 0.84          & 0.67                     &                       & 0.62          & 0.59          & 0.60                     &                       & 0.17          & 0.15          & 0.16                     \\
GESSVDD-Sb-GR-max                  &                       & 0.63          & 0.95          & 0.90          & 0.83                     &                       & 0.57          & 0.53          & 0.55                     &                       & 0.33          & 0.37          & 0.35                     \\
GESSVDD-Sb-SR-max                  &                       & 0.73          & 0.94          & 0.87          & 0.85                     &                       & 0.62          & 0.28          & 0.45                     &                       & \textbf{0.42} & 0.20          & 0.31                     \\
GESSVDD-Sb-$\mathcal{S}$-min                   &                       & 0.73          & 0.97          & 0.79          & 0.83                     &                       & 0.49          & 0.50          & 0.50                     &                       & 0.37          & 0.21          & 0.29                     \\
\end{tabular}
\end{table*}
\begin{table*}[h!]
  \footnotesize\setlength{\tabcolsep}{4.9pt}\renewcommand{\arraystretch}{1.30}
  \centering \scriptsize
         \caption{\textit{TNR} results for \textbf{linear} data description}
\begin{tabular}{llllllllllllll}
Dataset                            &                       & \multicolumn{4}{c}{Iris}                                                 &                       & \multicolumn{3}{c}{Ionosphere}                           &                       & \multicolumn{3}{c}{Sonar}                                \\ \cline{1-1} \cline{3-6} \cline{8-10} \cline{12-14} 
\multicolumn{1}{|l|}{Target class} & \multicolumn{1}{l|}{} & I-S        & S-VC    & S-V     & \multicolumn{1}{l|}{Av.} & \multicolumn{1}{l|}{} & I-B           & I-G          & \multicolumn{1}{l|}{Av.} & \multicolumn{1}{l|}{} & S-R          & S-M         & \multicolumn{1}{l|}{Av.} \\ \cline{1-1} \cline{3-6} \cline{8-10} \cline{12-14} 
GESSVDD-Sb-$\mathcal{S}$-max                   &                       & 0.99          & 0.85          & 0.85          & 0.90                     &                       & 0.24          & 0.44          & 0.34                     &                       & 0.44          & 0.72          & 0.58                     \\
GESSVDD-Sb-GR-max                  &                       & 0.94          & 0.86          & 0.92          & 0.91                     &                       & 0.23          & 0.82          & 0.52                     &                       & 0.34          & 0.34          & 0.34                     \\
GESSVDD-Sb-SR-max                  &                       & 0.85          & 0.94          & 0.83          & 0.87                     &                       & 0.44          & 0.54          & 0.49                     &                       & 0.20          & 0.46          & 0.33                     \\
GESSVDD-Sb-$\mathcal{S}$-min                   &                       & 0.99          & 0.83          & 0.84          & 0.89                     &                       & 0.38          & 0.59          & 0.48                     &                       & 0.53          & 0.76          & 0.65                     \\
\end{tabular}
\end{table*}

\begin{table*}[h!]
  \footnotesize\setlength{\tabcolsep}{4.5pt}\renewcommand{\arraystretch}{1.20}
  \centering \scriptsize
         \caption{\textit{TNR} results for \textbf{linear} data description}
\begin{tabular}{lllllllllllllllll}
Dataset                            &                       & \multicolumn{11}{c}{MNIST}                                                                                                                                                               & \multicolumn{1}{c}{}  & \multicolumn{3}{c}{Liver}                                \\ \cline{1-1} \cline{3-13} \cline{15-17} 
\multicolumn{1}{|l|}{Target class} & \multicolumn{1}{l|}{} & 0             & 1             & 2             & 3             & 4             & 5             & 6             & 7             & 8             & 9             & \multicolumn{1}{l|}{Av.} & \multicolumn{1}{l|}{} & DP            & DA            & \multicolumn{1}{l|}{Av.} \\ \cline{1-1} \cline{3-13} \cline{15-17} 
GESSVDD-Sb-$\mathcal{S}$-max                   &                       & 0.31          & 0.70          & 0.74          & 0.69          & 0.69          & 0.40          & 0.47          & 0.43          & 0.38          & 0.56          & 0.54                     &                       & 0.16          & 0.46          & 0.31                     \\
GESSVDD-Sb-GR-max                  &                       & 0.45          & 0.53          & 0.08          & 0.70          & 0.47          & 0.22          & 0.64          & 0.89          & 0.46          & 0.28          & 0.47                     &                       & 0.54          & 0.44          & 0.49                     \\
GESSVDD-Sb-SR-max                  &                       & 0.49          & 0.89          & 0.51          & 0.70          & 0.62          & 0.77          & 0.22          & 0.62          & 0.69          & 0.77          & 0.63                     &                       & 0.52          & 0.21          & 0.37                     \\
\end{tabular}
\end{table*}
\begin{table*}[h!]
  \footnotesize\setlength{\tabcolsep}{4.9pt}\renewcommand{\arraystretch}{1.20}
  \centering \scriptsize
         \caption{\textit{TNR} results for \textbf{linear} data description}
%
} \\ \cline{1-1} \cline{3-5} \cline{7-9} \cline{11-13} \cline{15-17} 
\multicolumn{1}{|l|}{Target class} & \multicolumn{1}{l|}{} & DP                          & DA                          & \multicolumn{1}{l|}{Av.}               & \multicolumn{1}{l|}{} & DP                            & DA                            & \multicolumn{1}{l|}{Av.}                & \multicolumn{1}{l|}{} & DP                            & DA                           & \multicolumn{1}{l|}{Av.}                & \multicolumn{1}{l|}{} & DP                             & DA                             & \multicolumn{1}{l|}{Av.}                 \\ \cline{1-1} \cline{3-5} \cline{7-9} \cline{11-13} \cline{15-17} 
GESSVDD-Sb-$\mathcal{S}$-max                   &                       & 0.46                        & 0.60                        & 0.53                                   &                       & 0.96                          & 0.87                          & 0.91                                    &                       & 0.22                          & 0.21                         & 0.22                                    &                       & 0.28                           & 0.31                           & 0.30                                     \\
GESSVDD-Sb-GR-max                  &                       & 0.58                        & \textbf{0.83}               & \textbf{0.70}                          &                       & 0.94                          & 0.97                          & 0.95                                    &                       & 0.17                          & 0.38                         & 0.27                                    &                       & 0.41                           & 0.40                           & 0.41                                     \\
GESSVDD-Sb-SR-max                  &                       & 0.56                        & 0.39                        & 0.47                                   & \textbf{}             & 0.98                          & 0.98                          & 0.98                                    &                       & 0.48                          & 0.51                         & 0.50                                    &                       & 0.42                           & 0.42                           & 0.42                                     \\
GESSVDD-Sb-$\mathcal{S}$-min                   &                       & 0.46                        & 0.75                        & 0.60                                   &                       & 0.99                          & 0.86                          & 0.92                                    &                       & 0.18                          & 0.23                         & 0.21                                    &                       & 0.29                           & 0.31                           & 0.30                                     \\
GESSVDD-Sb-GR-min                  &                       & 0.70                        & 0.69                        & \textbf{0.70}                          &                       & 0.95                          & 0.93                          & 0.94                                    &                       & 0.29                          & 0.59                         & 0.44                                    &                       & 0.40                           & \textbf{0.63}                  & \textbf{0.52}                            \\
GESSVDD-Sb-SR-min                  &                       & \textbf{0.75}               & 0.57                        & 0.66                                   & \textbf{}             & 0.90                          & 0.97                          & 0.94                                    &                       & 0.27                          & 0.39                         & 0.33                                    &                       & 0.32                           & 0.49                           & 0.40                                     \\
GESSVDD-Sw-$\mathcal{S}$-max                   &                       & 0.29                        & 0.28                        & 0.28                                   &                       & 0.96                          & 0.95                          & 0.96                                    &                       & 0.00                          & 0.00                         & 0.00                                    &                       & 0.09                           & 0.18                           & 0.13                                     \\
GESSVDD-Sw-GR-max                  &                       & 0.35                        & 0.51                        & 0.43                                   &                       & 0.90                          & 0.89                          & 0.89                                    &                       & 0.00                          & 0.00                         & 0.00                                    &                       & 0.12                           & 0.18                           & 0.15                                     \\
GESSVDD-Sw-SR-max                  &                       & 0.34                        & 0.42                        & 0.38                                   &                       & 0.99                          & \textbf{1.00}                 & \textbf{1.00}                           &                       & 0.00                          & 0.00                         & 0.00                                    &                       & 0.14                           & 0.22                           & 0.18                                     \\
GESSVDD-Sw-$\mathcal{S}$-min                   &                       & 0.31                        & 0.42                        & 0.37                                   &                       & 0.98                          & 0.92                          & 0.95                                    &                       & 0.00                          & 0.00                         & 0.00                                    &                       & 0.11                           & 0.23                           & 0.17                                     \\
GESSVDD-Sw-GR-min                  &                       & 0.46                        & 0.46                        & 0.46                                   &                       & 0.89                          & 0.89                          & 0.89                                    &                       & 0.00                          & 0.00                         & 0.00                                    &                       & 0.14                           & 0.17                           & 0.15                                     \\
GESSVDD-Sw-SR-min                  &                       & 0.30                        & 0.41                        & 0.35                                   &                       & \textbf{1.00}                 & \textbf{1.00}                 & \textbf{1.00}                           &                       & 0.00                          & 0.00                         & 0.00                                    &                       & 0.10                           & 0.14                           & 0.12                                     \\
GESSVDD-kNN-$\mathcal{S}$-max                  &                       & 0.42                        & 0.40                        & 0.41                                   &                       & \textbf{1.00}                 & 0.95                          & 0.98                                    &                       & 0.01                          & 0.00                         & 0.00                                    &                       & 0.22                           & 0.24                           & 0.23                                     \\
GESSVDD-kNN-GR-max                 &                       & 0.38                        & 0.44                        & 0.41                                   &                       & 0.99                          & 0.97                          & 0.98                                    &                       & 0.08                          & 0.04                         & 0.06                                    &                       & 0.40                           & 0.36                           & 0.38                                     \\
GESSVDD-kNN-SR-max                 &                       & 0.35                        & 0.41                        & 0.38                                   &                       & \textbf{1.00}                 & \textbf{1.00}                 & \textbf{1.00}                           &                       & 0.00                          & 0.00                         & 0.00                                    &                       & 0.18                           & 0.27                           & 0.22                                     \\
GESSVDD-kNN-$\mathcal{S}$-min                  &                       & 0.37                        & 0.45                        & 0.41                                   &                       & 0.99                          & \textbf{1.00}                 & 0.99                                    &                       & 0.03                          & 0.00                         & 0.01                                    &                       & 0.30                           & 0.25                           & 0.28                                     \\
GESSVDD-kNN-GR-min                 &                       & 0.40                        & 0.49                        & 0.45                                   &                       & \textbf{1.00}                 & \textbf{1.00}                 & \textbf{1.00}                           &                       & 0.23                          & 0.04                         & 0.14                                    &                       & 0.30                           & 0.26                           & 0.28                                     \\
GESSVDD-kNN-SR-min                 &                       & 0.39                        & 0.41                        & 0.40                                   &                       & \textbf{1.00}                 & 0.96                          & 0.98                                    &                       & 0.00                          & 0.00                         & 0.00                                    &                       & 0.33                           & 0.38                           & 0.35                                     \\
GESSVDD-PCA-$\mathcal{S}$-max                  &                       & 0.29                        & 0.39                        & 0.34                                   &                       & 0.98                          & \textbf{1.00}                 & 0.99                                    & \textbf{}             & 0.00                          & 0.00                         & 0.00                                    &                       & 0.13                           & 0.16                           & 0.14                                     \\
GESSVDD-PCA-GR-max                 &                       & 0.34                        & 0.34                        & 0.34                                   &                       & \textbf{1.00}                 & \textbf{1.00}                 & \textbf{1.00}                           &                       & 0.00                          & 0.00                         & 0.00                                    &                       & 0.18                           & 0.24                           & 0.21                                     \\
GESSVDD-PCA-SR-max                 &                       & 0.28                        & 0.35                        & 0.32                                   &                       & \textbf{1.00}                 & \textbf{1.00}                 & \textbf{1.00}                           &                       & 0.00                          & 0.00                         & 0.00                                    &                       & 0.13                           & 0.20                           & 0.17                                     \\
GESSVDD-PCA-$\mathcal{S}$-min                  &                       & 0.36                        & 0.52                        & 0.44                                   &                       & 0.99                          & 0.98                          & 0.99                                    & \textbf{}             & 0.00                          & 0.00                         & 0.00                                    &                       & 0.10                           & 0.26                           & 0.18                                     \\
GESSVDD-PCA-GR-min                 &                       & 0.37                        & 0.49                        & 0.43                                   &                       & 0.95                          & \textbf{1.00}                 & 0.98                                    &                       & 0.00                          & 0.00                         & 0.00                                    &                       & 0.19                           & 0.25                           & 0.22                                     \\
GESSVDD-PCA-SR-min                 &                       & 0.34                        & 0.36                        & 0.35                                   &                       & 0.97                          & \textbf{1.00}                 & 0.98                                    &                       & 0.00                          & 0.00                         & 0.00                                    &                       & 0.12                           & 0.16                           & 0.14                                     \\
GESSVDD-I-$\mathcal{S}$-max                    &                       & 0.34                        & 0.35                        & 0.34                                   &                       & \textbf{1.00}                 & 0.99                          & 0.99                                    & \textbf{}             & 0.00                          & 0.00                         & 0.00                                    &                       & 0.08                           & 0.15                           & 0.11                                     \\
GESSVDD-I-GR-max                   &                       & 0.39                        & 0.45                        & 0.42                                   &                       & 0.90                          & 0.94                          & 0.92                                    &                       & 0.00                          & 0.00                         & 0.00                                    &                       & 0.12                           & 0.21                           & 0.17                                     \\
GESSVDD-I-SR-max                   &                       & 0.28                        & 0.39                        & 0.33                                   &                       & \textbf{1.00}                 & \textbf{1.00}                 & \textbf{1.00}                           &                       & 0.00                          & 0.00                         & 0.00                                    &                       & 0.20                           & 0.16                           & 0.18                                     \\
GESSVDD-I-$\mathcal{S}$-min                    &                       & 0.4                         & 0.40                        & 0.40                                   &                       & 0.99                          & \textbf{1.00}                 & \textbf{1.00}                           & \textbf{}             & 0.00                          & 0.00                         & 0.00                                    &                       & 0.18                           & 0.24                           & 0.21                                     \\
GESSVDD-I-GR-min (ESSVDD)          &                       & 0.34                        & 0.58                        & 0.46                                   &                       & \textbf{1.00}                 & \textbf{1.00}                 & \textbf{1.00}                           &                       & 0.00                          & 0.00                         & 0.00                                    &                       & 0.15                           & 0.20                           & 0.17                                     \\
GESSVDD-I-SR-min                   &                       & 0.31                        & 0.28                        & 0.30                                   &                       & \textbf{1.00}                 & \textbf{1.00}                 & \textbf{1.00}                           &                       & 0.00                          & 0.00                         & 0.00                                    &                       & 0.14                           & 0.13                           & 0.13                                     \\
GESSVDD-0-$\mathcal{S}$-max                    &                       & 0.26                        & 0.26                        & 0.26                                   &                       & 0.97                          & 0.94                          & 0.95                                    &                       & 0.00                          & 0.00                         & 0.00                                    &                       & 0.05                           & 0.10                           & 0.07                                     \\
GESSVDD-0-GR-max                   &                       & 0.34                        & 0.47                        & 0.40                                   &                       & 0.95                          & 0.88                          & 0.91                                    &                       & 0.00                          & 0.00                         & 0.00                                    &                       & 0.07                           & 0.11                           & 0.09                                     \\
GESSVDD-0-$\mathcal{S}$-min                    &                       & 0.34                        & 0.45                        & 0.40                                   &                       & 0.97                          & \textbf{1.00}                 & 0.98                                    &                       & 0.00                          & 0.00                         & 0.00                                    &                       & 0.12                           & 0.19                           & 0.15                                     \\
GESSVDD-0-GR-min (SSVDD)           &                       & 0.46                        & 0.47                        & 0.46                                   &                       & \textbf{1.00}                 & 0.90                          & 0.95                                    &                       & 0.00                          & 0.00                         & 0.00                                    &                       & 0.15                           & 0.20                           & 0.18                                     \\
ESVDD                              &                       & 0.42                        & 0.48                        & 0.45                                   &                       & \textbf{1.00}                 & \textbf{1.00}                 & \textbf{1.00}                           &                       & 0.00                          & 0.00                         & 0.00                                    &                       & 0.23                           & 0.32                           & 0.27                                     \\
SVDD                               &                       & 0.40                        & 0.28                        & 0.34                                   &                       & 0.37                          & 0.31                          & 0.34                                    &                       & \textbf{0.54}                 & \textbf{0.73}                & \textbf{0.63}                           &                       & \textbf{0.44}                  & 0.55                           & 0.50                                     \\
OCSVM                              &                       & 0.37                        & 0.47                        & 0.42                                   &                       & \textbf{1.00}                 & \textbf{1.00}                 & \textbf{1.00}                           &                       & 0.00                          & 0.00                         & 0.00                                    &                       & 0.14                           & 0.20                           & 0.17                                    
\end{tabular}
\end{table*}
\clearpage

\subsubsection{FPR results \textbf{linear} data description}
\begin{table*}[h!]
  \footnotesize\setlength{\tabcolsep}{4.9pt}\renewcommand{\arraystretch}{1.20}
  \centering \scriptsize
         \caption{\textit{FPR} results for \textbf{linear} data description}
\begin{tabular}{llllllllllllll}
Dataset                            &                       & \multicolumn{4}{c}{Seeds}                                                &                       & \multicolumn{3}{c}{Qualitative bankruptcy}               &                       & \multicolumn{3}{c}{Somerville happiness}                 \\ \cline{1-1} \cline{3-6} \cline{8-10} \cline{12-14} 
\multicolumn{1}{|l|}{Target class} & \multicolumn{1}{l|}{} & S-K          & S-R          & S-C      & \multicolumn{1}{l|}{Av.} & \multicolumn{1}{l|}{} & QB-B            & QB-N        & \multicolumn{1}{l|}{Av.} & \multicolumn{1}{l|}{} & SH-H         & SH-U      & \multicolumn{1}{l|}{Av.} \\ \cline{1-1} \cline{3-6} \cline{8-10} \cline{12-14} 
GESSVDD-Sb-$\mathcal{S}$-max                   &                       & 0.33          & \textbf{0.50} & 0.16          & 0.33                     &                       & 0.38          & 0.41          & 0.40                     &                       & \textbf{0.83} & \textbf{0.85} & \textbf{0.84}            \\
GESSVDD-Sb-GR-max                  &                       & 0.37          & 0.05          & 0.10          & 0.17                     &                       & 0.43          & 0.47          & 0.45                     &                       & 0.67          & 0.63          & 0.65                     \\
GESSVDD-Sb-SR-max                  &                       & 0.27          & 0.06          & 0.13          & 0.15                     &                       & 0.38          & 0.72          & 0.55                     &                       & 0.58          & 0.80          & 0.69                     \\
GESSVDD-Sb-$\mathcal{S}$-min                   &                       & 0.27          & 0.03          & 0.21          & 0.17                     &                       & 0.51          & 0.50          & 0.50                     &                       & 0.63          & 0.79          & 0.71                     \\
\end{tabular}
\end{table*}
\begin{table*}[h!]
  \footnotesize\setlength{\tabcolsep}{4.9pt}\renewcommand{\arraystretch}{1.20}
  \centering \scriptsize
         \caption{\textit{FPR} results for \textbf{linear} data description}
\begin{tabular}{llllllllllllll}
Dataset                            &                       & \multicolumn{4}{c}{Iris}                                                 &                       & \multicolumn{3}{c}{Ionosphere}                           &                       & \multicolumn{3}{c}{Sonar}                                \\ \cline{1-1} \cline{3-6} \cline{8-10} \cline{12-14} 
\multicolumn{1}{|l|}{Target class} & \multicolumn{1}{l|}{} & I-S        & S-VC    & S-V     & \multicolumn{1}{l|}{Av.} & \multicolumn{1}{l|}{} & I-B           & I-G          & \multicolumn{1}{l|}{Av.} & \multicolumn{1}{l|}{} & S-R          & S-M         & \multicolumn{1}{l|}{Av.} \\ \cline{1-1} \cline{3-6} \cline{8-10} \cline{12-14} 
GESSVDD-Sb-$\mathcal{S}$-max                   &                       & 0.01          & 0.15          & 0.15          & 0.10                     &                       & 0.76          & 0.56          & 0.66                     &                       & 0.56          & 0.28          & 0.42                     \\
GESSVDD-Sb-GR-max                  &                       & 0.06          & 0.14          & 0.08          & 0.09                     &                       & 0.77          & 0.18          & 0.48                     &                       & 0.66          & 0.66          & 0.66                     \\
GESSVDD-Sb-SR-max                  &                       & 0.15          & 0.06          & 0.17          & 0.13                     &                       & 0.56          & 0.46          & 0.51                     &                       & \textbf{0.80} & 0.54          & 0.67                     \\
GESSVDD-Sb-$\mathcal{S}$-min                   &                       & 0.01          & 0.17          & 0.16          & 0.11                     &                       & 0.62          & 0.41          & 0.52                     &                       & 0.47          & 0.24          & 0.35                     \\
\end{tabular}
\end{table*}

\begin{table*}[h!]
  \footnotesize\setlength{\tabcolsep}{4.5pt}\renewcommand{\arraystretch}{1.20}
  \centering \scriptsize
         \caption{\textit{FPR} results for \textbf{linear} data description}
\begin{tabular}{lllllllllllllllll}
Dataset                            &                       & \multicolumn{11}{c}{MNIST}                                                                                                                                                               & \multicolumn{1}{c}{}  & \multicolumn{3}{c}{Liver}                                \\ \cline{1-1} \cline{3-13} \cline{15-17} 
\multicolumn{1}{|l|}{Target class} & \multicolumn{1}{l|}{} & 0             & 1             & 2             & 3             & 4             & 5             & 6             & 7             & 8             & 9             & \multicolumn{1}{l|}{Av.} & \multicolumn{1}{l|}{} & DP            & DA            & \multicolumn{1}{l|}{Av.} \\ \cline{1-1} \cline{3-13} \cline{15-17} 
GESSVDD-Sb-$\mathcal{S}$-max                   &                       & 0.69          & 0.30          & 0.26          & 0.31          & 0.31          & 0.60          & 0.53          & 0.57          & 0.62          & 0.44          & 0.46                     &                       & 0.84          & 0.54          & 0.69                     \\
GESSVDD-Sb-GR-max                  &                       & 0.55          & 0.47          & \textbf{0.92} & 0.30          & 0.53          & 0.78          & 0.36          & 0.11          & 0.54          & 0.72          & 0.53                     &                       & 0.46          & 0.56          & 0.51                     \\
GESSVDD-Sb-SR-max                  &                       & 0.51          & 0.11          & 0.49          & 0.30          & 0.38          & 0.23          & 0.78          & 0.38          & 0.31          & 0.23          & 0.37                     &                       & 0.48          & 0.79          & 0.63                     \\
\end{tabular}
\end{table*}
\begin{table*}[h!]
  \footnotesize\setlength{\tabcolsep}{4.9pt}\renewcommand{\arraystretch}{1.20}
  \centering \scriptsize
         \caption{\textit{FPR} results for \textbf{linear} data description}
%
} \\ \cline{1-1} \cline{3-5} \cline{7-9} \cline{11-13} \cline{15-17} 
\multicolumn{1}{|l|}{Target class} & \multicolumn{1}{l|}{} & DP                          & DA                          & \multicolumn{1}{l|}{Av.}               & \multicolumn{1}{l|}{} & DP                            & DA                            & \multicolumn{1}{l|}{Av.}                & \multicolumn{1}{l|}{} & DP                            & DA                           & \multicolumn{1}{l|}{Av.}                & \multicolumn{1}{l|}{} & DP                             & DA                             & \multicolumn{1}{l|}{Av.}                 \\ \cline{1-1} \cline{3-5} \cline{7-9} \cline{11-13} \cline{15-17} 
GESSVDD-Sb-$\mathcal{S}$-max                   &                       & 0.54                        & 0.40                        & 0.47                                   &                       & 0.04                          & 0.13                          & 0.09                                    &                       & 0.78                          & 0.79                         & 0.78                                    &                       & 0.72                           & 0.69                           & 0.70                                     \\
GESSVDD-Sb-GR-max                  &                       & 0.42                        & 0.17                        & 0.30                                   &                       & 0.06                          & 0.03                          & 0.05                                    &                       & 0.83                          & 0.62                         & 0.73                                    &                       & 0.59                           & 0.60                           & 0.59                                     \\
GESSVDD-Sb-SR-max                  &                       & 0.44                        & 0.61                        & 0.53                                   & \textbf{}             & 0.02                          & 0.02                          & 0.02                                    &                       & 0.52                          & 0.49                         & 0.50                                    &                       & 0.58                           & 0.58                           & 0.58                                     \\
GESSVDD-Sb-$\mathcal{S}$-min                   &                       & 0.54                        & 0.25                        & 0.40                                   &                       & 0.01                          & 0.14                          & 0.08                                    &                       & 0.82                          & 0.77                         & 0.79                                    &                       & 0.71                           & 0.69                           & 0.70                                     \\
GESSVDD-Sb-GR-min                  &                       & 0.30                        & 0.31                        & 0.30                                   &                       & 0.05                          & 0.07                          & 0.06                                    &                       & 0.71                          & 0.41                         & 0.56                                    &                       & 0.60                           & 0.37                           & 0.48                                     \\
GESSVDD-Sb-SR-min                  &                       & 0.25                        & 0.43                        & 0.34                                   & \textbf{}             & 0.10                          & 0.03                          & 0.06                                    &                       & 0.73                          & 0.61                         & 0.67                                    &                       & 0.68                           & 0.51                           & 0.60                                     \\
GESSVDD-Sw-$\mathcal{S}$-max                   &                       & 0.71                        & 0.72                        & 0.72                                   &                       & 0.04                          & 0.05                          & 0.04                                    &                       & \textbf{1.00}                 & \textbf{1.00}                & \textbf{1.00}                           &                       & 0.91                           & 0.82                           & 0.87                                     \\
GESSVDD-Sw-GR-max                  &                       & 0.65                        & 0.49                        & 0.57                                   &                       & 0.10                          & 0.11                          & 0.11                                    &                       & \textbf{1.00}                 & \textbf{1.00}                & \textbf{1.00}                           &                       & 0.88                           & 0.82                           & 0.85                                     \\
GESSVDD-Sw-SR-max                  &                       & 0.66                        & 0.58                        & 0.62                                   &                       & 0.01                          & 0.00                          & 0.00                                    &                       & \textbf{1.00}                 & \textbf{1.00}                & \textbf{1.00}                           &                       & 0.86                           & 0.78                           & 0.82                                     \\
GESSVDD-Sw-$\mathcal{S}$-min                   &                       & 0.69                        & 0.58                        & 0.63                                   &                       & 0.02                          & 0.08                          & 0.05                                    &                       & \textbf{1.00}                 & \textbf{1.00}                & \textbf{1.00}                           &                       & 0.89                           & 0.77                           & 0.83                                     \\
GESSVDD-Sw-GR-min                  &                       & 0.54                        & 0.54                        & 0.54                                   &                       & 0.11                          & 0.11                          & 0.11                                    &                       & \textbf{1.00}                 & \textbf{1.00}                & \textbf{1.00}                           &                       & 0.86                           & 0.83                           & 0.85                                     \\
GESSVDD-Sw-SR-min                  &                       & 0.70                        & 0.59                        & 0.65                                   &                       & 0.00                          & 0.00                          & 0.00                                    &                       & \textbf{1.00}                 & \textbf{1.00}                & \textbf{1.00}                           &                       & 0.90                           & 0.86                           & 0.88                                     \\
GESSVDD-kNN-$\mathcal{S}$-max                  &                       & 0.58                        & 0.60                        & 0.59                                   &                       & 0.00                          & 0.05                          & 0.02                                    &                       & 0.99                          & \textbf{1.00}                & \textbf{1.00}                           &                       & 0.78                           & 0.76                           & 0.77                                     \\
GESSVDD-kNN-GR-max                 &                       & 0.62                        & 0.56                        & 0.59                                   &                       & 0.01                          & 0.03                          & 0.02                                    &                       & 0.92                          & 0.96                         & 0.94                                    &                       & 0.60                           & 0.64                           & 0.62                                     \\
GESSVDD-kNN-SR-max                 &                       & 0.65                        & 0.59                        & 0.62                                   &                       & 0.00                          & 0.00                          & 0.00                                    &                       & \textbf{1.00}                 & \textbf{1.00}                & \textbf{1.00}                           &                       & 0.82                           & 0.73                           & 0.78                                     \\
GESSVDD-kNN-$\mathcal{S}$-min                  &                       & 0.63                        & 0.55                        & 0.59                                   &                       & 0.01                          & 0.00                          & 0.01                                    &                       & 0.97                          & \textbf{1.00}                & 0.99                                    &                       & 0.70                           & 0.75                           & 0.72                                     \\
GESSVDD-kNN-GR-min                 &                       & 0.60                        & 0.51                        & 0.55                                   &                       & 0.00                          & 0.00                          & 0.00                                    &                       & 0.77                          & 0.96                         & 0.86                                    &                       & 0.70                           & 0.74                           & 0.72                                     \\
GESSVDD-kNN-SR-min                 &                       & 0.61                        & 0.59                        & 0.60                                   &                       & 0.00                          & 0.04                          & 0.02                                    &                       & \textbf{1.00}                 & \textbf{1.00}                & \textbf{1.00}                           &                       & 0.67                           & 0.62                           & 0.65                                     \\
GESSVDD-PCA-$\mathcal{S}$-max                  &                       & 0.71                        & 0.61                        & 0.66                                   &                       & 0.02                          & 0.00                          & 0.01                                    & \textbf{}             & \textbf{1.00}                 & \textbf{1.00}                & \textbf{1.00}                           &                       & 0.87                           & 0.84                           & 0.86                                     \\
GESSVDD-PCA-GR-max                 &                       & 0.66                        & 0.66                        & 0.66                                   &                       & 0.00                          & 0.00                          & 0.00                                    &                       & \textbf{1.00}                 & \textbf{1.00}                & \textbf{1.00}                           &                       & 0.82                           & 0.76                           & 0.79                                     \\
GESSVDD-PCA-SR-max                 &                       & 0.72                        & 0.65                        & 0.68                                   &                       & 0.00                          & 0.00                          & 0.00                                    &                       & \textbf{1.00}                 & \textbf{1.00}                & \textbf{1.00}                           &                       & 0.87                           & 0.80                           & 0.83                                     \\
GESSVDD-PCA-$\mathcal{S}$-min                  &                       & 0.64                        & 0.48                        & 0.56                                   &                       & 0.01                          & 0.02                          & 0.01                                    & \textbf{}             & \textbf{1.00}                 & \textbf{1.00}                & \textbf{1.00}                           &                       & 0.90                           & 0.74                           & 0.82                                     \\
GESSVDD-PCA-GR-min                 &                       & 0.63                        & 0.51                        & 0.57                                   &                       & 0.05                          & 0.00                          & 0.02                                    &                       & \textbf{1.00}                 & \textbf{1.00}                & \textbf{1.00}                           &                       & 0.81                           & 0.75                           & 0.78                                     \\
GESSVDD-PCA-SR-min                 &                       & 0.66                        & 0.64                        & 0.65                                   &                       & 0.03                          & 0.00                          & 0.02                                    &                       & \textbf{1.00}                 & \textbf{1.00}                & \textbf{1.00}                           &                       & 0.88                           & 0.84                           & 0.86                                     \\
GESSVDD-I-$\mathcal{S}$-max                    &                       & 0.66                        & 0.65                        & 0.66                                   &                       & 0.00                          & 0.01                          & 0.01                                    & \textbf{}             & \textbf{1.00}                 & \textbf{1.00}                & \textbf{1.00}                           &                       & 0.92                           & 0.85                           & 0.89                                     \\
GESSVDD-I-GR-max                   &                       & 0.61                        & 0.55                        & 0.58                                   &                       & 0.10                          & 0.06                          & 0.08                                    &                       & \textbf{1.00}                 & \textbf{1.00}                & \textbf{1.00}                           &                       & 0.88                           & 0.79                           & 0.83                                     \\
GESSVDD-I-SR-max                   &                       & 0.72                        & 0.61                        & 0.67                                   &                       & 0.00                          & 0.00                          & 0.00                                    &                       & \textbf{1.00}                 & \textbf{1.00}                & \textbf{1.00}                           &                       & 0.80                           & 0.84                           & 0.82                                     \\
GESSVDD-I-$\mathcal{S}$-min                    &                       & 0.6                         & 0.60                        & 0.60                                   &                       & 0.01                          & 0.00                          & 0.00                                    & \textbf{}             & \textbf{1.00}                 & \textbf{1.00}                & \textbf{1.00}                           &                       & 0.82                           & 0.76                           & 0.79                                     \\
GESSVDD-I-GR-min (ESSVDD)          &                       & 0.66                        & 0.42                        & 0.54                                   &                       & 0.00                          & 0.00                          & 0.00                                    &                       & \textbf{1.00}                 & \textbf{1.00}                & \textbf{1.00}                           &                       & 0.85                           & 0.80                           & 0.83                                     \\
GESSVDD-I-SR-min                   &                       & 0.69                        & 0.72                        & 0.70                                   &                       & 0.00                          & 0.00                          & 0.00                                    &                       & \textbf{1.00}                 & \textbf{1.00}                & \textbf{1.00}                           &                       & 0.86                           & 0.87                           & 0.87                                     \\
GESSVDD-0-$\mathcal{S}$-max                    &                       & \textbf{0.74}               & \textbf{0.74}               & \textbf{0.74}                          &                       & 0.03                          & 0.06                          & 0.05                                    &                       & \textbf{1.00}                 & \textbf{1.00}                & \textbf{1.00}                           &                       & \textbf{0.95}                  & \textbf{0.90}                  & \textbf{0.93}                            \\
GESSVDD-0-GR-max                   &                       & 0.66                        & 0.53                        & 0.60                                   &                       & 0.05                          & 0.12                          & 0.09                                    &                       & \textbf{1.00}                 & \textbf{1.00}                & \textbf{1.00}                           &                       & 0.93                           & 0.89                           & 0.91                                     \\
GESSVDD-0-$\mathcal{S}$-min                    &                       & 0.66                        & 0.55                        & 0.60                                   &                       & 0.03                          & 0.00                          & 0.02                                    &                       & \textbf{1.00}                 & \textbf{1.00}                & \textbf{1.00}                           &                       & 0.88                           & 0.81                           & 0.85                                     \\
GESSVDD-0-GR-min (SSVDD)           &                       & 0.54                        & 0.53                        & 0.54                                   &                       & 0.00                          & 0.10                          & 0.05                                    &                       & \textbf{1.00}                 & \textbf{1.00}                & \textbf{1.00}                           &                       & 0.85                           & 0.80                           & 0.82                                     \\
ESVDD                              &                       & 0.58                        & 0.52                        & 0.55                                   &                       & 0.00                          & 0.00                          & 0.00                                    &                       & \textbf{1.00}                 & \textbf{1.00}                & \textbf{1.00}                           &                       & 0.77                           & 0.68                           & 0.73                                     \\
SVDD                               &                       & 0.60                        & 0.72                        & 0.66                                   &                       & \textbf{0.63}                 & \textbf{0.69}                 & \textbf{0.66}                           &                       & 0.46                          & 0.27                         & 0.37                                    &                       & 0.56                           & 0.45                           & 0.50                                     \\
OCSVM                              &                       & 0.63                        & 0.53                        & 0.58                                   &                       & 0.00                          & 0.00                          & 0.00                                    &                       & \textbf{1.00}                 & \textbf{1.00}                & \textbf{1.00}                           &                       & 0.86                           & 0.80                           & 0.83                                    
\end{tabular}
\end{table*}
\clearpage
 
 \subsubsection{FNR result \textbf{linear} data description}
\begin{table*}[h!]
  \footnotesize\setlength{\tabcolsep}{4.9pt}\renewcommand{\arraystretch}{1.20}
  \centering \scriptsize
         \caption{\textit{FNR} results for \textbf{linear} data description}
\begin{tabular}{llllllllllllll}
Dataset                            &                       & \multicolumn{4}{c}{Seeds}                                                &                       & \multicolumn{3}{c}{Qualitative bankruptcy}               &                       & \multicolumn{3}{c}{Somerville happiness}                 \\ \cline{1-1} \cline{3-6} \cline{8-10} \cline{12-14} 
\multicolumn{1}{|l|}{Target class} & \multicolumn{1}{l|}{} & S-K          & S-R          & S-C      & \multicolumn{1}{l|}{Av.} & \multicolumn{1}{l|}{} & QB-B            & QB-N        & \multicolumn{1}{l|}{Av.} & \multicolumn{1}{l|}{} & SH-H         & SH-U      & \multicolumn{1}{l|}{Av.} \\ \cline{1-1} \cline{3-6} \cline{8-10} \cline{12-14} 
GESSVDD-Sb-$\mathcal{S}$-max                   &                       & 0.23          & 0.24          & 0.12          & 0.20                     &                       & 0.24          & 0.54          & 0.39                     &                       & 0.17          & 0.24          & 0.20                     \\
GESSVDD-Sb-GR-max                  &                       & 0.10          & 0.38          & 0.35          & 0.28                     &                       & 0.17          & 0.56          & 0.37                     &                       & 0.27          & 0.27          & 0.27                     \\
GESSVDD-Sb-SR-max                  &                       & 0.18          & 0.14          & 0.04          & 0.12                     &                       & 0.42          & 0.54          & 0.48                     &                       & 0.39          & 0.32          & 0.35                     \\
GESSVDD-Sb-$\mathcal{S}$-min                   &                       & 0.19          & 0.25          & 0.08          & 0.17                     &                       & 0.21          & 0.41          & 0.31                     &                       & 0.32          & 0.32          & 0.32                     \\
\end{tabular}
\end{table*}
\begin{table*}[h!]
  \footnotesize\setlength{\tabcolsep}{4.9pt}\renewcommand{\arraystretch}{1.20}
  \centering \scriptsize
         \caption{\textit{FNR} results for \textbf{linear} data description}
\begin{tabular}{llllllllllllll}
Dataset                            &                       & \multicolumn{4}{c}{Iris}                                                 &                       & \multicolumn{3}{c}{Ionosphere}                           &                       & \multicolumn{3}{c}{Sonar}                                \\ \cline{1-1} \cline{3-6} \cline{8-10} \cline{12-14} 
\multicolumn{1}{|l|}{Target class} & \multicolumn{1}{l|}{} & I-S        & S-VC    & S-V     & \multicolumn{1}{l|}{Av.} & \multicolumn{1}{l|}{} & I-B           & I-G          & \multicolumn{1}{l|}{Av.} & \multicolumn{1}{l|}{} & S-R          & S-M         & \multicolumn{1}{l|}{Av.} \\ \cline{1-1} \cline{3-6} \cline{8-10} \cline{12-14} 
GESSVDD-Sb-$\mathcal{S}$-max                   &                       & 0.15          & 0.12          & 0.21          & 0.16                     &                       & 0.32          & 0.32          & 0.32                     &                       & 0.38          & 0.69          & 0.54                     \\
GESSVDD-Sb-GR-max                  &                       & 0.13          & 0.11          & 0.15          & 0.13                     &                       & 0.20          & 0.04          & 0.12                     &                       & 0.14          & 0.25          & 0.20                     \\
GESSVDD-Sb-SR-max                  &                       & 0.13          & 0.17          & 0.15          & 0.15                     &                       & 0.42          & \textbf{0.79} & \textbf{0.61}            &                       & 0.41          & 0.31          & 0.36                     \\
GESSVDD-Sb-$\mathcal{S}$-min                   &                       & 0.17          & 0.16          & 0.21          & 0.18                     &                       & 0.29          & 0.51          & 0.40                     &                       & 0.48          & 0.73          & 0.61                     \\
\end{tabular}
\end{table*}

\begin{table*}[h!]
  \footnotesize\setlength{\tabcolsep}{4.5pt}\renewcommand{\arraystretch}{1.20}
  \centering \scriptsize
         \caption{\textit{FNR} results for \textbf{linear} data description}
\begin{tabular}{lllllllllllllllll}
Dataset                            &                       & \multicolumn{11}{c}{MNIST}                                                                                                                                                               & \multicolumn{1}{c}{}  & \multicolumn{3}{c}{Liver}                                \\ \cline{1-1} \cline{3-13} \cline{15-17} 
\multicolumn{1}{|l|}{Target class} & \multicolumn{1}{l|}{} & 0             & 1             & 2             & 3             & 4             & 5             & 6             & 7             & 8             & 9             & \multicolumn{1}{l|}{Av.} & \multicolumn{1}{l|}{} & DP            & DA            & \multicolumn{1}{l|}{Av.} \\ \cline{1-1} \cline{3-13} \cline{15-17} 
GESSVDD-Sb-$\mathcal{S}$-max                   &                       & 0.11          & 0.31          & 0.67          & 0.70          & 0.54          & 0.35          & 0.22          & 0.36          & 0.21          & 0.35          & 0.38                     &                       & 0.08          & 0.37          & 0.22                     \\
GESSVDD-Sb-GR-max                  &                       & 0.16          & 0.18          & 0.02          & 0.57          & 0.31          & 0.18          & 0.59          & 0.88          & 0.36          & 0.10          & 0.33                     &                       & 0.49          & 0.40          & 0.44                     \\
GESSVDD-Sb-SR-max                  &                       & 0.36          & \textbf{0.53} & 0.38          & 0.68          & 0.31          & 0.63          & 0.85          & 0.31          & 0.55          & 0.49          & 0.51                     &                       & 0.47          & 0.20          & 0.33                     \\
\end{tabular}
\end{table*}
\begin{table*}[h!]
  \footnotesize\setlength{\tabcolsep}{4.9pt}\renewcommand{\arraystretch}{1.20}
  \centering \scriptsize
         \caption{\textit{FNR} results for \textbf{linear} data description}
%
} \\ \cline{1-1} \cline{3-5} \cline{7-9} \cline{11-13} \cline{15-17} 
\multicolumn{1}{|l|}{Target class} & \multicolumn{1}{l|}{} & DP                          & DA                          & \multicolumn{1}{l|}{Av.}               & \multicolumn{1}{l|}{} & DP                            & DA                            & \multicolumn{1}{l|}{Av.}                & \multicolumn{1}{l|}{} & DP                            & DA                           & \multicolumn{1}{l|}{Av.}                & \multicolumn{1}{l|}{} & DP                             & DA                             & \multicolumn{1}{l|}{Av.}                 \\ \cline{1-1} \cline{3-5} \cline{7-9} \cline{11-13} \cline{15-17} 
GESSVDD-Sb-$\mathcal{S}$-max                   &                       & 0.26                        & 0.43                        & 0.35                                   &                       & 0.94                          & 0.92                          & 0.93                                    &                       & 0.32                          & 0.42                         & 0.37                                    &                       & 0.24                           & 0.31                           & 0.27                                     \\
GESSVDD-Sb-GR-max                  &                       & 0.36                        & \textbf{0.56}               & \textbf{0.46}                          &                       & 0.93                          & 0.92                          & 0.93                                    &                       & 0.20                          & 0.34                         & 0.27                                    &                       & 0.44                           & 0.43                           & 0.44                                     \\
GESSVDD-Sb-SR-max                  &                       & 0.35                        & 0.43                        & 0.39                                   & \textbf{}             & 0.96                          & 0.96                          & 0.96                                    &                       & \textbf{0.40}                 & 0.60                         & 0.50                                    &                       & 0.37                           & 0.45                           & 0.41                                     \\
GESSVDD-Sb-$\mathcal{S}$-min                   &                       & 0.28                        & 0.52                        & 0.40                                   &                       & 0.92                          & 0.88                          & 0.90                                    &                       & 0.30                          & 0.36                         & 0.33                                    &                       & 0.30                           & 0.25                           & 0.27                                     \\
GESSVDD-Sb-GR-min                  &                       & 0.42                        & 0.30                        & 0.36                                   &                       & 0.95                          & 0.91                          & 0.93                                    &                       & 0.32                          & 0.67                         & 0.50                                    &                       & \textbf{0.46}                  & \textbf{0.61}                  & \textbf{0.54}                            \\
GESSVDD-Sb-SR-min                  &                       & \textbf{0.52}               & 0.37                        & 0.45                                   & \textbf{}             & 0.88                          & 0.96                          & 0.92                                    &                       & 0.25                          & 0.47                         & 0.36                                    &                       & 0.30                           & 0.52                           & 0.41                                     \\
GESSVDD-Sw-$\mathcal{S}$-max                   &                       & 0.08                        & 0.10                        & 0.09                                   &                       & 0.94                          & 0.96                          & 0.95                                    &                       & 0.00                          & 0.00                         & 0.00                                    &                       & 0.06                           & 0.08                           & 0.07                                     \\
GESSVDD-Sw-GR-max                  &                       & 0.11                        & 0.14                        & 0.12                                   &                       & 0.86                          & 0.88                          & 0.87                                    &                       & 0.00                          & 0.00                         & 0.00                                    &                       & 0.10                           & 0.10                           & 0.10                                     \\
GESSVDD-Sw-SR-max                  &                       & 0.16                        & 0.17                        & 0.17                                   &                       & \textbf{1.00}                 & \textbf{1.00}                 & \textbf{1.00}                           &                       & 0.00                          & 0.00                         & 0.00                                    &                       & 0.16                           & 0.16                           & 0.16                                     \\
GESSVDD-Sw-$\mathcal{S}$-min                   &                       & 0.14                        & 0.10                        & 0.12                                   &                       & 0.97                          & 0.95                          & 0.96                                    &                       & 0.01                          & 0.00                         & 0.00                                    &                       & 0.12                           & 0.18                           & 0.15                                     \\
GESSVDD-Sw-GR-min                  &                       & 0.17                        & 0.18                        & 0.17                                   &                       & 0.85                          & 0.87                          & 0.86                                    &                       & 0.00                          & 0.00                         & 0.00                                    &                       & 0.16                           & 0.18                           & 0.17                                     \\
GESSVDD-Sw-SR-min                  &                       & 0.14                        & 0.15                        & 0.15                                   &                       & \textbf{1.00}                 & \textbf{1.00}                 & \textbf{1.00}                           &                       & 0.05                          & 0.00                         & 0.02                                    &                       & 0.10                           & 0.16                           & 0.13                                     \\
GESSVDD-kNN-$\mathcal{S}$-max                  &                       & 0.25                        & 0.20                        & 0.23                                   &                       & 0.99                          & 0.97                          & 0.98                                    &                       & 0.06                          & 0.00                         & 0.03                                    &                       & 0.14                           & 0.18                           & 0.16                                     \\
GESSVDD-kNN-GR-max                 &                       & 0.22                        & 0.18                        & 0.20                                   &                       & 0.96                          & 0.93                          & 0.95                                    &                       & 0.06                          & 0.04                         & 0.05                                    &                       & 0.38                           & 0.37                           & 0.38                                     \\
GESSVDD-kNN-SR-max                 &                       & 0.26                        & 0.17                        & 0.22                                   &                       & \textbf{1.00}                 & 0.99                          & \textbf{1.00}                           &                       & 0.00                          & 0.00                         & 0.00                                    &                       & 0.12                           & 0.27                           & 0.20                                     \\
GESSVDD-kNN-$\mathcal{S}$-min                  &                       & 0.20                        & 0.25                        & 0.22                                   &                       & 0.98                          & \textbf{1.00}                 & 0.99                                    &                       & 0.00                          & 0.00                         & 0.00                                    &                       & 0.24                           & 0.24                           & 0.24                                     \\
GESSVDD-kNN-GR-min                 &                       & 0.25                        & 0.22                        & 0.24                                   &                       & 0.98                          & \textbf{1.00}                 & 0.99                                    &                       & 0.12                          & 0.04                         & 0.08                                    &                       & 0.37                           & 0.23                           & 0.30                                     \\
GESSVDD-kNN-SR-min                 &                       & 0.25                        & 0.17                        & 0.21                                   &                       & \textbf{1.00}                 & 0.95                          & 0.97                                    &                       & 0.00                          & 0.00                         & 0.00                                    &                       & 0.30                           & 0.36                           & 0.33                                     \\
GESSVDD-PCA-$\mathcal{S}$-max                  &                       & 0.12                        & 0.12                        & 0.12                                   &                       & 0.99                          & \textbf{1.00}                 & 0.99                                    & \textbf{}             & 0.01                          & 0.00                         & 0.00                                    &                       & 0.13                           & 0.07                           & 0.10                                     \\
GESSVDD-PCA-GR-max                 &                       & 0.17                        & 0.15                        & 0.16                                   &                       & \textbf{1.00}                 & \textbf{1.00}                 & \textbf{1.00}                           &                       & 0.00                          & 0.00                         & 0.00                                    &                       & 0.15                           & 0.17                           & 0.16                                     \\
GESSVDD-PCA-SR-max                 &                       & 0.15                        & 0.16                        & 0.15                                   &                       & \textbf{1.00}                 & 0.98                          & 0.99                                    &                       & 0.00                          & 0.00                         & 0.00                                    &                       & 0.12                           & 0.20                           & 0.16                                     \\
GESSVDD-PCA-$\mathcal{S}$-min                  &                       & 0.15                        & 0.16                        & 0.16                                   &                       & \textbf{1.00}                 & 0.95                          & 0.97                                    & \textbf{}             & 0.00                          & 0.00                         & 0.00                                    &                       & 0.10                           & 0.21                           & 0.16                                     \\
GESSVDD-PCA-GR-min                 &                       & 0.18                        & 0.17                        & 0.17                                   &                       & 0.90                          & \textbf{1.00}                 & 0.95                                    &                       & 0.00                          & 0.00                         & 0.00                                    &                       & 0.18                           & 0.20                           & 0.19                                     \\
GESSVDD-PCA-SR-min                 &                       & 0.14                        & 0.14                        & 0.14                                   &                       & 0.97                          & 0.97                          & 0.97                                    &                       & 0.00                          & 0.00                         & 0.00                                    &                       & 0.14                           & 0.12                           & 0.13                                     \\
GESSVDD-I-$\mathcal{S}$-max                    &                       & 0.14                        & 0.14                        & 0.14                                   &                       & 0.98                          & 0.98                          & 0.98                                    & \textbf{}             & 0.02                          & 0.00                         & 0.01                                    &                       & 0.11                           & 0.15                           & 0.13                                     \\
GESSVDD-I-GR-max                   &                       & 0.17                        & 0.15                        & 0.16                                   &                       & 0.86                          & 0.94                          & 0.90                                    &                       & 0.00                          & 0.00                         & 0.00                                    &                       & 0.11                           & 0.17                           & 0.14                                     \\
GESSVDD-I-SR-max                   &                       & 0.15                        & 0.15                        & 0.15                                   &                       & \textbf{1.00}                 & \textbf{1.00}                 & \textbf{1.00}                           &                       & 0.00                          & 0.00                         & 0.00                                    &                       & 0.10                           & 0.14                           & 0.12                                     \\
GESSVDD-I-$\mathcal{S}$-min                    &                       & 0.2                         & 0.16                        & 0.18                                   &                       & 0.97                          & \textbf{1.00}                 & 0.99                                    & \textbf{}             & 0.00                          & 0.00                         & 0.00                                    &                       & 0.19                           & 0.20                           & 0.19                                     \\
GESSVDD-I-GR-min (ESSVDD)          &                       & 0.14                        & 0.17                        & 0.15                                   &                       & \textbf{1.00}                 & \textbf{1.00}                 & \textbf{1.00}                           &                       & 0.00                          & 0.00                         & 0.00                                    &                       & 0.15                           & 0.20                           & 0.17                                     \\
GESSVDD-I-SR-min                   &                       & 0.17                        & 0.11                        & 0.14                                   &                       & \textbf{1.00}                 & \textbf{1.00}                 & \textbf{1.00}                           &                       & 0.00                          & 0.00                         & 0.00                                    &                       & 0.12                           & 0.08                           & 0.10                                     \\
GESSVDD-0-$\mathcal{S}$-max                    &                       & 0.08                        & 0.08                        & 0.08                                   &                       & 0.97                          & 0.92                          & 0.94                                    &                       & 0.00                          & 0.00                         & 0.00                                    &                       & 0.04                           & 0.08                           & 0.06                                     \\
GESSVDD-0-GR-max                   &                       & 0.08                        & 0.11                        & 0.09                                   &                       & 0.94                          & 0.89                          & 0.91                                    &                       & 0.00                          & 0.00                         & 0.00                                    &                       & 0.05                           & 0.07                           & 0.06                                     \\
GESSVDD-0-$\mathcal{S}$-min                    &                       & 0.16                        & 0.12                        & 0.14                                   &                       & 0.93                          & \textbf{1.00}                 & 0.96                                    &                       & 0.00                          & 0.00                         & 0.00                                    &                       & 0.09                           & 0.15                           & 0.12                                     \\
GESSVDD-0-GR-min (SSVDD)           &                       & 0.22                        & 0.16                        & 0.19                                   &                       & \textbf{1.00}                 & 0.90                          & 0.95                                    &                       & 0.00                          & 0.00                         & 0.00                                    &                       & 0.13                           & 0.18                           & 0.15                                     \\
ESVDD                              &                       & 0.24                        & 0.20                        & 0.22                                   &                       & \textbf{1.00}                 & \textbf{1.00}                 & \textbf{1.00}                           &                       & 0.00                          & 0.00                         & 0.00                                    &                       & 0.18                           & 0.25                           & 0.21                                     \\
SVDD                               &                       & 0.41                        & 0.52                        & \textbf{0.46}                          &                       & 0.47                          & 0.27                          & 0.37                                    &                       & 0.33                          & \textbf{0.74}                & \textbf{0.53}                           &                       & \textbf{0.46}                  & 0.51                           & 0.49                                     \\
OCSVM                              &                       & 0.10                        & 0.15                        & 0.13                                   &                       & \textbf{1.00}                 & \textbf{1.00}                 & \textbf{1.00}                           &                       & 0.00                          & 0.00                         & 0.00                                    &                       & 0.13                           & 0.15                           & 0.14                                    
\end{tabular}
\end{table*}
\clearpage

\subsection{TPR TNR FPR FNR results for \textbf{non-linear} data description}

\subsubsection{TPR results \textbf{non-linear} data description}
\begin{table*}[h!]
  \footnotesize\setlength{\tabcolsep}{4.9pt}\renewcommand{\arraystretch}{1.20}
  \centering \scriptsize
         \caption{\textit{TPR} results for \textbf{non-linear} data description}
\begin{tabular}{llllllllllllll}
Dataset                            &                       & \multicolumn{4}{c}{Seeds}                                                &                       & \multicolumn{3}{c}{Qualitative bankruptcy}               &                       & \multicolumn{3}{c}{Somerville happiness}                 \\ \cline{1-1} \cline{3-6} \cline{8-10} \cline{12-14} 
\multicolumn{1}{|l|}{Target class} & \multicolumn{1}{l|}{} & S-K          & S-R          & S-C      & \multicolumn{1}{l|}{Av.} & \multicolumn{1}{l|}{} & QB-B            & QB-N        & \multicolumn{1}{l|}{Av.} & \multicolumn{1}{l|}{} & SH-H         & SH-U      & \multicolumn{1}{l|}{Av.} \\ \cline{1-1} \cline{3-6} \cline{8-10} \cline{12-14} 
GESSVDD-Sb-$\mathcal{S}$-max                   &                       & 0.86          & 0.94          & 0.89          & 0.90                     &                       & 0.76          & 0.46          & 0.61                     &                       & \textbf{0.78} & 0.77          & 0.78                     \\
GESSVDD-Sb-GR-max                  &                       & 0.78          & 0.64          & 0.92          & 0.78                     &                       & 0.93          & 0.59          & 0.76                     &                       & 0.74          & 0.66          & 0.70                     \\
GESSVDD-Sb-SR-max                  &                       & 0.78          & 0.74          & 0.97          & 0.83                     &                       & 0.70          & 0.45          & 0.58                     &                       & 0.47          & 0.49          & 0.48                     \\
GESSVDD-Sb-$\mathcal{S}$-min                   &                       & 0.86          & 0.90          & 0.81          & 0.85                     &                       & 0.76          & 0.51          & 0.63                     &                       & 0.62          & 0.72          & 0.67                     \\
\end{tabular}
\end{table*}
\begin{table*}[h!]
  \footnotesize\setlength{\tabcolsep}{4.9pt}\renewcommand{\arraystretch}{1.20}
  \centering \scriptsize
         \caption{\textit{TPR} results for \textbf{non-linear} data description}
\begin{tabular}{llllllllllllll}
Dataset                            &                       & \multicolumn{4}{c}{Iris}                                                 &                       & \multicolumn{3}{c}{Ionosphere}                           &                       & \multicolumn{3}{c}{Sonar}                                \\ \cline{1-1} \cline{3-6} \cline{8-10} \cline{12-14} 
\multicolumn{1}{|l|}{Target class} & \multicolumn{1}{l|}{} & I-S        & S-VC    & S-V     & \multicolumn{1}{l|}{Av.} & \multicolumn{1}{l|}{} & I-B           & I-G          & \multicolumn{1}{l|}{Av.} & \multicolumn{1}{l|}{} & S-R          & S-M         & \multicolumn{1}{l|}{Av.} \\ \cline{1-1} \cline{3-6} \cline{8-10} \cline{12-14} 
GESSVDD-Sb-$\mathcal{S}$-max                   &                       & 0.92          & 0.83          & 0.87          & 0.87                     &                       & 0.58          & 0.73          & 0.66                     &                       & 0.63          & 0.64          & 0.64                     \\
GESSVDD-Sb-GR-max                  &                       & 0.91          & 0.83          & 0.85          & 0.86                     &                       & 0.80          & 0.96          & 0.88                     &                       & 0.86          & 0.69          & 0.78                     \\
GESSVDD-Sb-SR-max                  &                       & 0.91          & 0.85          & 0.81          & 0.86                     &                       & 0.81          & 0.27          & 0.54                     &                       & 0.76          & 0.42          & 0.59                     \\
GESSVDD-Sb-$\mathcal{S}$-min                   &                       & 0.92          & 0.87          & 0.89          & 0.89                     &                       & 0.57          & 0.53          & 0.55                     &                       & 0.59          & 0.48          & 0.54                     \\
\end{tabular}
\end{table*}

\begin{table*}[h!]
  \footnotesize\setlength{\tabcolsep}{4.5pt}\renewcommand{\arraystretch}{1.20}
  \centering \scriptsize
         \caption{\textit{TPR} results for \textbf{non-linear} data description}
\begin{tabular}{lllllllllllllllll}
Dataset                            &                       & \multicolumn{11}{c}{MNIST}                                                                                                                                                               & \multicolumn{1}{c}{}  & \multicolumn{3}{c}{Liver}                                \\ \cline{1-1} \cline{3-13} \cline{15-17} 
\multicolumn{1}{|l|}{Target class} & \multicolumn{1}{l|}{} & 0             & 1             & 2             & 3             & 4             & 5             & 6             & 7             & 8             & 9             & \multicolumn{1}{l|}{Av.} & \multicolumn{1}{l|}{} & DP            & DA            & \multicolumn{1}{l|}{Av.} \\ \cline{1-1} \cline{3-13} \cline{15-17} 
GESSVDD-Sb-$\mathcal{S}$-max                   &                       & 0.16          & 0.16          & 0.06          & 0.08          & 0.39          & 0.29          & 0.31          & 0.00          & 0.12          & 0.09          & 0.17                     &                       & 0.37          & 0.35          & 0.36                     \\
GESSVDD-Sb-GR-max                  &                       & 0.62          & 0.32          & 0.48          & 0.41          & 0.20          & 0.33          & 0.21          & 0.38          & 0.11          & 0.58          & 0.36                     &                       & 0.29          & 0.35          & 0.32                     \\
GESSVDD-Sb-SR-max                  &                       & 0.01          & 0.59          & 0.26          & 0.67          & 0.05          & 0.04          & 0.10          & 0.26          & 0.11          & 0.47          & 0.26                     &                       & 0.29          & 0.27          & 0.28                     \\
\end{tabular}
\end{table*}
\begin{table*}[h!]
  \footnotesize\setlength{\tabcolsep}{4.9pt}\renewcommand{\arraystretch}{1.20}
  \centering \scriptsize
         \caption{\textit{TPR} results for \textbf{non-linear} data description in the proposed framework with added noise}
%
} \\ \cline{1-1} \cline{3-5} \cline{7-9} \cline{11-13} \cline{15-17} 
\multicolumn{1}{|l|}{Target class} & \multicolumn{1}{l|}{} & DP                          & DA                          & \multicolumn{1}{l|}{Av.}               & \multicolumn{1}{l|}{} & DP                            & DA                            & \multicolumn{1}{l|}{Av.}                & \multicolumn{1}{l|}{} & DP                            & DA                           & \multicolumn{1}{l|}{Av.}                & \multicolumn{1}{l|}{} & DP                             & DA                             & \multicolumn{1}{l|}{Av.}                 \\ \cline{1-1} \cline{3-5} \cline{7-9} \cline{11-13} \cline{15-17} 
GESSVDD-Sb-$\mathcal{S}$-max                   &                       & 0.29                        & 0.29                        & 0.29                                   &                       & 0.42                          & 0.29                          & 0.36                                    &                       & 0.13                          & 0.05                         & 0.09                                    &                       & 0.35                           & 0.40                           & 0.37                                     \\
GESSVDD-Sb-GR-max                  &                       & 0.34                        & 0.34                        & 0.34                                   &                       & 0.35                          & 0.23                          & 0.29                                    &                       & 0.33                          & 0.31                         & 0.32                                    &                       & 0.27                           & 0.45                           & 0.36                                     \\
GESSVDD-Sb-SR-max                  &                       & 0.32                        & 0.32                        & 0.32                                   & \textbf{}             & 0.22                          & 0.24                          & 0.23                                    &                       & 0.30                          & 0.22                         & 0.26                                    &                       & 0.43                           & 0.44                           & 0.44                                     \\
GESSVDD-Sb-$\mathcal{S}$-min                   &                       & 0.41                        & 0.41                        & 0.41                                   &                       & \textbf{0.44}                 & 0.19                          & 0.32                                    &                       & 0.19                          & 0.19                         & 0.19                                    &                       & 0.35                           & 0.20                           & 0.28                                     \\
GESSVDD-Sb-GR-min                  &                       & 0.45                        & 0.45                        & 0.45                                   &                       & 0.32                          & 0.18                          & 0.25                                    &                       & 0.21                          & 0.32                         & 0.26                                    &                       & 0.33                           & 0.33                           & 0.33                                     \\
GESSVDD-Sb-SR-min                  &                       & 0.46                        & 0.46                        & 0.46                                   & \textbf{}             & 0.25                          & 0.13                          & 0.19                                    &                       & 0.36                          & 0.16                         & 0.26                                    &                       & 0.43                           & 0.44                           & 0.43                                     \\
GESSVDD-Sw-$\mathcal{S}$-max                   &                       & 0.43                        & 0.43                        & 0.43                                   &                       & 0.00                          & 0.12                          & 0.06                                    &                       & 0.13                          & 0.00                         & 0.06                                    &                       & 0.36                           & 0.34                           & 0.35                                     \\
GESSVDD-Sw-GR-max                  &                       & 0.16                        & 0.16                        & 0.16                                   &                       & 0.36                          & \textbf{0.39}                 & \textbf{0.38}                           &                       & 0.07                          & 0.11                         & 0.09                                    &                       & 0.42                           & 0.39                           & 0.41                                     \\
GESSVDD-Sw-SR-max                  &                       & 0.43                        & 0.43                        & 0.43                                   &                       & 0.11                          & 0.06                          & 0.08                                    &                       & 0.00                          & 0.06                         & 0.03                                    &                       & 0.43                           & 0.41                           & 0.42                                     \\
GESSVDD-Sw-$\mathcal{S}$-min                   &                       & 0.34                        & 0.34                        & 0.34                                   &                       & 0.00                          & 0.16                          & 0.08                                    &                       & 0.00                          & 0.15                         & 0.08                                    &                       & 0.44                           & 0.43                           & 0.43                                     \\
GESSVDD-Sw-GR-min                  &                       & 0.40                        & 0.40                        & 0.40                                   &                       & 0.18                          & 0.29                          & 0.24                                    &                       & 0.04                          & 0.05                         & 0.05                                    &                       & 0.45                           & 0.44                           & 0.45                                     \\
GESSVDD-Sw-SR-min                  &                       & 0.44                        & 0.44                        & 0.44                                   &                       & 0.03                          & 0.19                          & 0.11                                    &                       & 0.00                          & 0.00                         & 0.00                                    &                       & 0.41                           & 0.33                           & 0.37                                     \\
GESSVDD-kNN-$\mathcal{S}$-max                  &                       & 0.44                        & 0.44                        & 0.44                                   &                       & 0.00                          & 0.00                          & 0.00                                    &                       & 0.14                          & 0.04                         & 0.09                                    &                       & 0.45                           & 0.45                           & 0.45                                     \\
GESSVDD-kNN-GR-max                 &                       & 0.43                        & 0.43                        & 0.43                                   &                       & 0.28                          & 0.19                          & 0.24                                    &                       & 0.12                          & 0.00                         & 0.06                                    &                       & 0.44                           & 0.38                           & 0.41                                     \\
GESSVDD-kNN-SR-max                 &                       & 0.42                        & 0.42                        & 0.42                                   &                       & 0.04                          & 0.03                          & 0.03                                    &                       & 0.00                          & 0.00                         & 0.00                                    &                       & 0.47                           & 0.40                           & 0.44                                     \\
GESSVDD-kNN-$\mathcal{S}$-min                  &                       & 0.49                        & 0.49                        & 0.49                                   &                       & 0.03                          & 0.00                          & 0.02                                    &                       & 0.03                          & 0.00                         & 0.02                                    &                       & 0.46                           & 0.43                           & 0.44                                     \\
GESSVDD-kNN-GR-min                 &                       & 0.47                        & 0.47                        & 0.47                                   &                       & 0.00                          & 0.00                          & 0.00                                    &                       & 0.00                          & 0.00                         & 0.00                                    &                       & 0.48                           & 0.43                           & 0.46                                     \\
GESSVDD-kNN-SR-min                 &                       & 0.47                        & 0.47                        & 0.47                                   &                       & 0.04                          & 0.07                          & 0.06                                    &                       & 0.06                          & 0.00                         & 0.03                                    &                       & 0.45                           & 0.38                           & 0.41                                     \\
GESSVDD-PCA-$\mathcal{S}$-max                  &                       & 0.37                        & 0.37                        & 0.37                                   &                       & 0.00                          & 0.03                          & 0.01                                    & \textbf{}             & 0.00                          & 0.00                         & 0.00                                    &                       & 0.49                           & 0.36                           & 0.42                                     \\
GESSVDD-PCA-GR-max                 &                       & 0.47                        & 0.47                        & 0.47                                   &                       & 0.30                          & 0.34                          & 0.32                                    &                       & 0.13                          & 0.16                         & 0.14                                    &                       & 0.45                           & 0.36                           & 0.40                                     \\
GESSVDD-PCA-SR-max                 &                       & 0.46                        & 0.46                        & 0.46                                   &                       & 0.19                          & 0.05                          & 0.12                                    &                       & 0.06                          & 0.09                         & 0.08                                    &                       & 0.34                           & 0.41                           & 0.37                                     \\
GESSVDD-PCA-$\mathcal{S}$-min                  &                       & 0.24                        & 0.24                        & 0.24                                   &                       & 0.06                          & 0.11                          & 0.08                                    & \textbf{}             & 0.00                          & 0.00                         & 0.00                                    &                       & 0.43                           & 0.39                           & 0.41                                     \\
GESSVDD-PCA-GR-min                 &                       & 0.30                        & 0.30                        & 0.30                                   &                       & 0.28                          & 0.22                          & 0.25                                    &                       & 0.00                          & 0.04                         & 0.02                                    &                       & 0.46                           & 0.47                           & 0.47                                     \\
GESSVDD-PCA-SR-min                 &                       & 0.44                        & 0.44                        & 0.44                                   &                       & 0.10                          & 0.07                          & 0.09                                    &                       & 0.03                          & 0.00                         & 0.02                                    &                       & 0.37                           & 0.41                           & 0.39                                     \\
GESSVDD-I-$\mathcal{S}$-max                    &                       & 0.38                        & 0.38                        & 0.38                                   &                       & 0.06                          & 0.03                          & 0.04                                    & \textbf{}             & 0.04                          & 0.00                         & 0.02                                    &                       & 0.45                           & 0.38                           & 0.42                                     \\
GESSVDD-I-GR-max                   &                       & 0.42                        & 0.42                        & 0.42                                   &                       & 0.37                          & 0.35                          & 0.36                                    &                       & 0.03                          & 0.05                         & 0.04                                    &                       & 0.48                           & 0.31                           & 0.39                                     \\
GESSVDD-I-SR-max                   &                       & 0.49                        & 0.49                        & 0.49                                   &                       & 0.13                          & 0.03                          & 0.08                                    &                       & 0.00                          & 0.00                         & 0.00                                    &                       & 0.43                           & 0.44                           & 0.43                                     \\
GESSVDD-I-$\mathcal{S}$-min                    &                       & 0.44                        & 0.44                        & 0.44                                   &                       & 0.00                          & 0.25                          & 0.12                                    & \textbf{}             & 0.08                          & 0.00                         & 0.04                                    &                       & 0.36                           & 0.39                           & 0.38                                     \\
GESSVDD-I-GR-min (ESSVDD)          &                       & 0.38                        & 0.38                        & 0.38                                   &                       & 0.37                          & 0.15                          & 0.26                                    &                       & 0.00                          & 0.11                         & 0.06                                    &                       & 0.38                           & 0.46                           & 0.42                                     \\
GESSVDD-I-SR-min                   &                       & 0.46                        & 0.46                        & 0.46                                   &                       & 0.15                          & 0.14                          & 0.14                                    &                       & 0.00                          & 0.08                         & 0.04                                    &                       & 0.42                           & 0.45                           & 0.44                                     \\
GESSVDD-0-$\mathcal{S}$-max                    &                       & 0.35                        & 0.35                        & 0.35                                   &                       & 0.24                          & 0.21                          & 0.22                                    &                       & 0.26                          & 0.32                         & 0.29                                    &                       & 0.39                           & 0.36                           & 0.38                                     \\
GESSVDD-0-GR-max                   &                       & 0.45                        & 0.45                        & 0.45                                   &                       & 0.06                          & 0.22                          & 0.14                                    &                       & 0.06                          & 0.44                         & 0.25                                    &                       & 0.46                           & 0.36                           & 0.41                                     \\
GESSVDD-0-$\mathcal{S}$-min                    &                       & 0.32                        & 0.32                        & 0.32                                   &                       & 0.14                          & 0.16                          & 0.15                                    &                       & 0.36                          & 0.09                         & 0.23                                    &                       & 0.34                           & 0.40                           & 0.37                                     \\
GESSVDD-0-GR-min (SSVDD)           &                       & 0.53                        & 0.53                        & 0.53                                   &                       & 0.12                          & 0.24                          & 0.18                                    &                       & 0.00                          & 0.09                         & 0.04                                    &                       & 0.37                           & 0.47                           & 0.42                                     \\
ESVDD                              &                       & 0.34                        & 0.34                        & 0.34                                   &                       & 0.00                          & 0.00                          & 0.00                                    &                       & 0.00                          & 0.03                         & 0.02                                    &                       & 0.49                           & 0.52                           & 0.51                                     \\
SVDD                               &                       & 0.53                        & 0.53                        & 0.53                                   &                       & 0.07                          & 0.15                          & 0.11                                    &                       & 0.22                          & 0.07                         & 0.15                                    &                       & 0.47                           & 0.45                           & 0.46                                     \\
OCSVM                              &                       & 0.20                        & 0.20                        & 0.20                                   &                       & 0.00                          & 0.00                          & 0.00                                    &                       & 0.00                          & 0.07                         & 0.03                                    &                       & 0.50                           & 0.49                           & 0.50                                     \\
GESVDD-PCA                         &                       & 0.68                        & 0.68                        & 0.68                                   &                       & 0.00                          & 0.00                          & 0.00                                    &                       & 0.00                          & 0.00                         & 0.00                                    &                       & 0.51                           & \textbf{0.53}                  & \textbf{0.52}                            \\
GESVDD-Sw                          &                       & 0.68                        & 0.68                        & 0.68                                   &                       & 0.00                          & 0.00                          & 0.00                                    &                       & 0.00                          & 0.00                         & 0.00                                    &                       & \textbf{0.53}                  & 0.52                           & \textbf{0.52}                            \\
GESVDD-kNN                         &                       & \textbf{0.70}               & \textbf{0.70}               & \textbf{0.70}                          &                       & 0.00                          & 0.00                          & 0.00                                    &                       & 0.00                          & 0.00                         & 0.00                                    &                       & 0.48                           & 0.52                           & 0.50                                     \\
GESVM-PCA                          &                       & 0.66                        & 0.66                        & 0.66                                   &                       & 0.00                          & 0.00                          & 0.00                                    &                       & 0.00                          & 0.00                         & 0.00                                    &                       & 0.49                           & 0.49                           & 0.49                                     \\
GESVM-Sw                           &                       & 0.67                        & 0.67                        & 0.67                                   &                       & 0.00                          & 0.00                          & 0.00                                    &                       & 0.00                          & 0.05                         & 0.03                                    &                       & 0.50                           & 0.48                           & 0.49                                     \\
GESVM-kNN                          &                       & 0.60                        & 0.60                        & 0.60                                   &                       & 0.00                          & 0.00                          & 0.00                                    &                       & \textbf{1.00}                 & \textbf{1.00}                & \textbf{1.00}                           &                       & 0.45                           & 0.45                           & 0.45                                    
\end{tabular}
\end{table*}

\clearpage
\subsubsection{TNR results \textbf{non-linear} data description}
\begin{table*}[h!]
  \footnotesize\setlength{\tabcolsep}{4.9pt}\renewcommand{\arraystretch}{1.20}
  \centering \scriptsize
         \caption{\textit{TNR} results for \textbf{non-linear} data description}
\begin{tabular}{llllllllllllll}
Dataset                            &                       & \multicolumn{4}{c}{Seeds}                                                &                       & \multicolumn{3}{c}{Qualitative bankruptcy}               &                       & \multicolumn{3}{c}{Somerville happiness}                 \\ \cline{1-1} \cline{3-6} \cline{8-10} \cline{12-14} 
\multicolumn{1}{|l|}{Target class} & \multicolumn{1}{l|}{} & S-K          & S-R          & S-C      & \multicolumn{1}{l|}{Av.} & \multicolumn{1}{l|}{} & QB-B            & QB-N        & \multicolumn{1}{l|}{Av.} & \multicolumn{1}{l|}{} & SH-H         & SH-U      & \multicolumn{1}{l|}{Av.} \\ \cline{1-1} \cline{3-6} \cline{8-10} \cline{12-14} 
GESSVDD-Sb-$\mathcal{S}$-max                   &                       & 0.69          & 0.89          & 0.88          & 0.82                     &                       & 0.50          & 0.70          & 0.60                     &                       & 0.19          & 0.16          & 0.17                     \\
GESSVDD-Sb-GR-max                  &                       & 0.89          & 0.95          & 0.86          & 0.90                     &                       & 0.74          & 0.53          & 0.63                     &                       & 0.25          & 0.34          & 0.30                     \\
GESSVDD-Sb-SR-max                  &                       & 0.78          & 0.90          & 0.89          & 0.85                     &                       & 0.48          & 0.41          & 0.45                     &                       & 0.61          & 0.24          & 0.43                     \\
GESSVDD-Sb-$\mathcal{S}$-min                   &                       & 0.74          & 0.92          & 0.88          & 0.85                     &                       & 0.53          & 0.61          & 0.57                     &                       & 0.45          & 0.26          & 0.36                     \\
\end{tabular}
\end{table*}
\begin{table*}[h!]
  \footnotesize\setlength{\tabcolsep}{4.9pt}\renewcommand{\arraystretch}{1.20}
  \centering \scriptsize
         \caption{\textit{TNR} results for \textbf{non-linear} data description}
\begin{tabular}{llllllllllllll}
Dataset                            &                       & \multicolumn{4}{c}{Iris}                                                 &                       & \multicolumn{3}{c}{Ionosphere}                           &                       & \multicolumn{3}{c}{Sonar}                                \\ \cline{1-1} \cline{3-6} \cline{8-10} \cline{12-14} 
\multicolumn{1}{|l|}{Target class} & \multicolumn{1}{l|}{} & I-S        & S-VC    & S-V     & \multicolumn{1}{l|}{Av.} & \multicolumn{1}{l|}{} & I-B           & I-G          & \multicolumn{1}{l|}{Av.} & \multicolumn{1}{l|}{} & S-R          & S-M         & \multicolumn{1}{l|}{Av.} \\ \cline{1-1} \cline{3-6} \cline{8-10} \cline{12-14} 
GESSVDD-Sb-$\mathcal{S}$-max                   &                       & 0.96          & 0.94          & 0.73          & 0.88                     &                       & 0.61          & 0.57          & 0.59                     &                       & 0.34          & 0.41          & 0.37                     \\
GESSVDD-Sb-GR-max                  &                       & \textbf{1.00} & 0.84          & 0.95          & 0.93                     &                       & 0.37          & 0.80          & 0.58                     &                       & 0.33          & 0.32          & 0.32                     \\
GESSVDD-Sb-SR-max                  &                       & 0.75          & 0.85          & 0.89          & 0.83                     &                       & 0.36          & 0.66          & 0.51                     &                       & 0.27          & 0.41          & 0.34                     \\
GESSVDD-Sb-$\mathcal{S}$-min                   &                       & 0.93          & 0.93          & 0.85          & 0.90                     &                       & 0.61          & 0.71          & 0.66                     &                       & 0.42          & 0.56          & 0.49                     \\
\end{tabular}
\end{table*}

\begin{table*}[h!]
  \footnotesize\setlength{\tabcolsep}{4.5pt}\renewcommand{\arraystretch}{1.20}
  \centering \scriptsize
         \caption{\textit{TNR} results for \textbf{non-linear} data description}
\begin{tabular}{lllllllllllllllll}
Dataset                            &                       & \multicolumn{11}{c}{MNIST}                                                                                                                                                               & \multicolumn{1}{c}{}  & \multicolumn{3}{c}{Liver}                                \\ \cline{1-1} \cline{3-13} \cline{15-17} 
\multicolumn{1}{|l|}{Target class} & \multicolumn{1}{l|}{} & 0             & 1             & 2             & 3             & 4             & 5             & 6             & 7             & 8             & 9             & \multicolumn{1}{l|}{Av.} & \multicolumn{1}{l|}{} & DP            & DA            & \multicolumn{1}{l|}{Av.} \\ \cline{1-1} \cline{3-13} \cline{15-17} 
GESSVDD-Sb-$\mathcal{S}$-max                   &                       & 0.16          & 0.16          & 0.06          & 0.08          & 0.39          & 0.29          & 0.31          & 0.00          & 0.12          & 0.09          & 0.17                     &                       & 0.37          & 0.35          & 0.36                     \\
GESSVDD-Sb-GR-max                  &                       & 0.62          & 0.32          & 0.48          & 0.41          & 0.20          & 0.33          & 0.21          & 0.38          & 0.11          & 0.58          & 0.36                     &                       & 0.29          & 0.35          & 0.32                     \\
GESSVDD-Sb-SR-max                  &                       & 0.01          & 0.59          & 0.26          & 0.67          & 0.05          & 0.04          & 0.10          & 0.26          & 0.11          & 0.47          & 0.26                     &                       & 0.29          & 0.27          & 0.28                     \\
\end{tabular}
\end{table*}
\begin{table*}[h!]
  \footnotesize\setlength{\tabcolsep}{4.9pt}\renewcommand{\arraystretch}{1.20}
  \centering \scriptsize
         \caption{\textit{TNR} results for \textbf{non-linear} data description in the proposed framework with added noise}
%
} \\ \cline{1-1} \cline{3-5} \cline{7-9} \cline{11-13} \cline{15-17} 
\multicolumn{1}{|l|}{Target class} & \multicolumn{1}{l|}{} & DP                          & DA                          & \multicolumn{1}{l|}{Av.}               & \multicolumn{1}{l|}{} & DP                            & DA                            & \multicolumn{1}{l|}{Av.}                & \multicolumn{1}{l|}{} & DP                            & DA                           & \multicolumn{1}{l|}{Av.}                & \multicolumn{1}{l|}{} & DP                             & DA                             & \multicolumn{1}{l|}{Av.}                 \\ \cline{1-1} \cline{3-5} \cline{7-9} \cline{11-13} \cline{15-17} 
GESSVDD-Sb-$\mathcal{S}$-max                   &                       & 0.29                        & 0.29                        & 0.29                                   &                       & 0.42                          & 0.29                          & 0.36                                    &                       & 0.13                          & 0.05                         & 0.09                                    &                       & 0.35                           & 0.40                           & 0.37                                     \\
GESSVDD-Sb-GR-max                  &                       & 0.34                        & 0.34                        & 0.34                                   &                       & 0.35                          & 0.23                          & 0.29                                    &                       & 0.33                          & 0.31                         & \textbf{0.32}                           &                       & 0.27                           & 0.45                           & 0.36                                     \\
GESSVDD-Sb-SR-max                  &                       & 0.32                        & 0.32                        & 0.32                                   & \textbf{}             & 0.22                          & 0.24                          & 0.23                                    &                       & 0.30                          & 0.22                         & 0.26                                    &                       & 0.43                           & 0.44                           & 0.44                                     \\
GESSVDD-Sb-$\mathcal{S}$-min                   &                       & 0.41                        & 0.41                        & 0.41                                   &                       & 0.44                          & 0.19                          & 0.32                                    &                       & 0.19                          & 0.19                         & 0.19                                    &                       & 0.35                           & 0.20                           & 0.28                                     \\
GESSVDD-Sb-GR-min                  &                       & 0.45                        & 0.45                        & 0.45                                   &                       & 0.32                          & 0.18                          & 0.25                                    &                       & 0.21                          & 0.32                         & 0.26                                    &                       & 0.33                           & 0.33                           & 0.33                                     \\
GESSVDD-Sb-SR-min                  &                       & 0.46                        & 0.46                        & 0.46                                   & \textbf{}             & 0.25                          & 0.13                          & 0.19                                    &                       & \textbf{0.36}                 & 0.16                         & 0.26                                    &                       & 0.43                           & 0.44                           & 0.43                                     \\
GESSVDD-Sw-$\mathcal{S}$-max                   &                       & 0.43                        & 0.43                        & 0.43                                   &                       & 0.00                          & 0.12                          & 0.06                                    &                       & 0.13                          & 0.00                         & 0.06                                    &                       & 0.36                           & 0.34                           & 0.35                                     \\
GESSVDD-Sw-GR-max                  &                       & 0.16                        & 0.16                        & 0.16                                   &                       & 0.36                          & 0.39                          & 0.38                                    &                       & 0.07                          & 0.11                         & 0.09                                    &                       & 0.42                           & 0.39                           & 0.41                                     \\
GESSVDD-Sw-SR-max                  &                       & 0.43                        & 0.43                        & 0.43                                   &                       & 0.11                          & 0.06                          & 0.08                                    &                       & 0.00                          & 0.06                         & 0.03                                    &                       & 0.43                           & 0.41                           & 0.42                                     \\
GESSVDD-Sw-$\mathcal{S}$-min                   &                       & 0.34                        & 0.34                        & 0.34                                   &                       & 0.00                          & 0.16                          & 0.08                                    &                       & 0.00                          & 0.15                         & 0.08                                    &                       & 0.44                           & 0.43                           & 0.43                                     \\
GESSVDD-Sw-GR-min                  &                       & 0.40                        & 0.40                        & 0.40                                   &                       & 0.18                          & 0.29                          & 0.24                                    &                       & 0.04                          & 0.05                         & 0.05                                    &                       & 0.45                           & 0.44                           & 0.45                                     \\
GESSVDD-Sw-SR-min                  &                       & 0.44                        & 0.44                        & 0.44                                   &                       & 0.03                          & 0.19                          & 0.11                                    &                       & 0.00                          & 0.00                         & 0.00                                    &                       & 0.41                           & 0.33                           & 0.37                                     \\
GESSVDD-kNN-$\mathcal{S}$-max                  &                       & 0.44                        & 0.44                        & 0.44                                   &                       & 0.00                          & 0.00                          & 0.00                                    &                       & 0.14                          & 0.04                         & 0.09                                    &                       & 0.45                           & 0.45                           & 0.45                                     \\
GESSVDD-kNN-GR-max                 &                       & 0.43                        & 0.43                        & 0.43                                   &                       & 0.28                          & 0.19                          & 0.24                                    &                       & 0.12                          & 0.00                         & 0.06                                    &                       & 0.44                           & 0.38                           & 0.41                                     \\
GESSVDD-kNN-SR-max                 &                       & 0.42                        & 0.42                        & 0.42                                   &                       & 0.04                          & 0.03                          & 0.03                                    &                       & 0.00                          & 0.00                         & 0.00                                    &                       & 0.47                           & 0.40                           & 0.44                                     \\
GESSVDD-kNN-$\mathcal{S}$-min                  &                       & 0.49                        & 0.49                        & 0.49                                   &                       & 0.03                          & 0.00                          & 0.02                                    &                       & 0.03                          & 0.00                         & 0.02                                    &                       & 0.46                           & 0.43                           & 0.44                                     \\
GESSVDD-kNN-GR-min                 &                       & 0.47                        & 0.47                        & 0.47                                   &                       & 0.00                          & 0.00                          & 0.00                                    &                       & 0.00                          & 0.00                         & 0.00                                    &                       & 0.48                           & 0.43                           & 0.46                                     \\
GESSVDD-kNN-SR-min                 &                       & 0.47                        & 0.47                        & 0.47                                   &                       & 0.04                          & 0.07                          & 0.06                                    &                       & 0.06                          & 0.00                         & 0.03                                    &                       & 0.45                           & 0.38                           & 0.41                                     \\
GESSVDD-PCA-$\mathcal{S}$-max                  &                       & 0.37                        & 0.37                        & 0.37                                   &                       & 0.00                          & 0.03                          & 0.01                                    & \textbf{}             & 0.00                          & 0.00                         & 0.00                                    &                       & 0.49                           & 0.36                           & 0.42                                     \\
GESSVDD-PCA-GR-max                 &                       & 0.47                        & 0.47                        & 0.47                                   &                       & 0.30                          & 0.34                          & 0.32                                    &                       & 0.13                          & 0.16                         & 0.14                                    &                       & 0.45                           & 0.36                           & 0.40                                     \\
GESSVDD-PCA-SR-max                 &                       & 0.46                        & 0.46                        & 0.46                                   &                       & 0.19                          & 0.05                          & 0.12                                    &                       & 0.06                          & 0.09                         & 0.08                                    &                       & 0.34                           & 0.41                           & 0.37                                     \\
GESSVDD-PCA-$\mathcal{S}$-min                  &                       & 0.24                        & 0.24                        & 0.24                                   &                       & 0.06                          & 0.11                          & 0.08                                    & \textbf{}             & 0.00                          & 0.00                         & 0.00                                    &                       & 0.43                           & 0.39                           & 0.41                                     \\
GESSVDD-PCA-GR-min                 &                       & 0.30                        & 0.30                        & 0.30                                   &                       & 0.28                          & 0.22                          & 0.25                                    &                       & 0.00                          & 0.04                         & 0.02                                    &                       & 0.46                           & 0.47                           & 0.47                                     \\
GESSVDD-PCA-SR-min                 &                       & 0.44                        & 0.44                        & 0.44                                   &                       & 0.10                          & 0.07                          & 0.09                                    &                       & 0.03                          & 0.00                         & 0.02                                    &                       & 0.37                           & 0.41                           & 0.39                                     \\
GESSVDD-I-$\mathcal{S}$-max                    &                       & 0.38                        & 0.38                        & 0.38                                   &                       & 0.06                          & 0.03                          & 0.04                                    & \textbf{}             & 0.04                          & 0.00                         & 0.02                                    &                       & 0.45                           & 0.38                           & 0.42                                     \\
GESSVDD-I-GR-max                   &                       & 0.42                        & 0.42                        & 0.42                                   &                       & 0.37                          & 0.35                          & 0.36                                    &                       & 0.03                          & 0.05                         & 0.04                                    &                       & 0.48                           & 0.31                           & 0.39                                     \\
GESSVDD-I-SR-max                   &                       & 0.49                        & 0.49                        & 0.49                                   &                       & 0.13                          & 0.03                          & 0.08                                    &                       & 0.00                          & 0.00                         & 0.00                                    &                       & 0.43                           & 0.44                           & 0.43                                     \\
GESSVDD-I-$\mathcal{S}$-min                    &                       & 0.44                        & 0.44                        & 0.44                                   &                       & 0.00                          & 0.25                          & 0.12                                    & \textbf{}             & 0.08                          & 0.00                         & 0.04                                    &                       & 0.36                           & 0.39                           & 0.38                                     \\
GESSVDD-I-GR-min (ESSVDD)          &                       & 0.38                        & 0.38                        & 0.38                                   &                       & 0.37                          & 0.15                          & 0.26                                    &                       & 0.00                          & 0.11                         & 0.06                                    &                       & 0.38                           & 0.46                           & 0.42                                     \\
GESSVDD-I-SR-min                   &                       & 0.46                        & 0.46                        & 0.46                                   &                       & 0.15                          & 0.14                          & 0.14                                    &                       & 0.00                          & 0.08                         & 0.04                                    &                       & 0.42                           & 0.45                           & 0.44                                     \\
GESSVDD-0-$\mathcal{S}$-max                    &                       & 0.35                        & 0.35                        & 0.35                                   &                       & 0.24                          & 0.21                          & 0.22                                    &                       & 0.26                          & 0.32                         & 0.29                                    &                       & 0.39                           & 0.36                           & 0.38                                     \\
GESSVDD-0-GR-max                   &                       & 0.45                        & 0.45                        & 0.45                                   &                       & 0.06                          & 0.22                          & 0.14                                    &                       & 0.06                          & \textbf{0.44}                & 0.25                                    &                       & 0.46                           & 0.36                           & 0.41                                     \\
GESSVDD-0-$\mathcal{S}$-min                    &                       & 0.32                        & 0.32                        & 0.32                                   &                       & 0.14                          & 0.16                          & 0.15                                    &                       & \textbf{0.36}                 & 0.09                         & 0.23                                    &                       & 0.34                           & 0.40                           & 0.37                                     \\
GESSVDD-0-GR-min (SSVDD)           &                       & 0.53                        & 0.53                        & 0.53                                   &                       & 0.12                          & 0.24                          & 0.18                                    &                       & 0.00                          & 0.09                         & 0.04                                    &                       & 0.37                           & 0.47                           & 0.42                                     \\
ESVDD                              &                       & 0.34                        & 0.34                        & 0.34                                   &                       & 0.00                          & 0.00                          & 0.00                                    &                       & 0.00                          & 0.03                         & 0.02                                    &                       & 0.49                           & 0.52                           & 0.51                                     \\
SVDD                               &                       & 0.53                        & 0.53                        & 0.53                                   &                       & 0.07                          & 0.15                          & 0.11                                    &                       & 0.22                          & 0.07                         & 0.15                                    &                       & 0.47                           & 0.45                           & 0.46                                     \\
OCSVM                              &                       & 0.20                        & 0.20                        & 0.20                                   &                       & 0.00                          & 0.00                          & 0.00                                    &                       & 0.00                          & 0.07                         & 0.03                                    &                       & 0.50                           & 0.49                           & 0.50                                     \\
GESVDD-PCA                         &                       & 0.68                        & 0.68                        & 0.68                                   &                       & 0.00                          & 0.00                          & 0.00                                    &                       & 0.00                          & 0.00                         & 0.00                                    &                       & 0.51                           & 0.53                           & 0.52                                     \\
GESVDD-Sw                          &                       & 0.68                        & 0.68                        & 0.68                                   &                       & 0.00                          & 0.00                          & 0.00                                    &                       & 0.00                          & 0.00                         & 0.00                                    &                       & 0.53                           & 0.52                           & 0.52                                     \\
GESVDD-kNN                         &                       & 0.70                        & 0.70                        & 0.70                                   &                       & 0.00                          & 0.00                          & 0.00                                    &                       & 0.00                          & 0.00                         & 0.00                                    &                       & 0.48                           & 0.52                           & 0.50                                     \\
GESVM-PCA                          &                       & 0.66                        & 0.66                        & 0.66                                   &                       & 0.00                          & 0.00                          & 0.00                                    &                       & 0.00                          & 0.00                         & 0.00                                    &                       & 0.49                           & 0.49                           & 0.49                                     \\
GESVM-Sw                           &                       & 0.67                        & 0.67                        & 0.67                                   &                       & 0.00                          & 0.00                          & 0.00                                    &                       & 0.00                          & 0.05                         & 0.03                                    &                       & 0.50                           & 0.48                           & 0.49                                     \\
GESVM-kNN                          &                       & \textbf{0.76}               & \textbf{0.76}               & \textbf{0.76}                          &                       & \textbf{1.00}                 & \textbf{1.00}                 & \textbf{1.00}                           &                       & 0.00                          & 0.00                         & 0.00                                    &                       & \textbf{0.56}                  & \textbf{0.56}                  & \textbf{0.56}                           
\end{tabular}
\end{table*}

\clearpage
\subsubsection{FPR results \textbf{non-linear} data description}
\begin{table*}[h!]
  \footnotesize\setlength{\tabcolsep}{4.9pt}\renewcommand{\arraystretch}{1.20}
  \centering \scriptsize
         \caption{\textit{FPR} results for \textbf{non-linear} data description}
\begin{tabular}{llllllllllllll}
Dataset                            &                       & \multicolumn{4}{c}{Seeds}                                                &                       & \multicolumn{3}{c}{Qualitative bankruptcy}               &                       & \multicolumn{3}{c}{Somerville happiness}                 \\ \cline{1-1} \cline{3-6} \cline{8-10} \cline{12-14} 
\multicolumn{1}{|l|}{Target class} & \multicolumn{1}{l|}{} & S-K          & S-R          & S-C      & \multicolumn{1}{l|}{Av.} & \multicolumn{1}{l|}{} & QB-B            & QB-N        & \multicolumn{1}{l|}{Av.} & \multicolumn{1}{l|}{} & SH-H         & SH-U      & \multicolumn{1}{l|}{Av.} \\ \cline{1-1} \cline{3-6} \cline{8-10} \cline{12-14} 
GESSVDD-Sb-$\mathcal{S}$-max                   &                       & 0.31          & 0.11          & 0.12          & 0.18                     &                       & 0.50          & 0.30          & 0.40                     &                       & \textbf{0.81} & \textbf{0.84} & \textbf{0.83}            \\
GESSVDD-Sb-GR-max                  &                       & 0.11          & 0.05          & 0.14          & 0.10                     &                       & 0.26          & 0.47          & 0.37                     &                       & 0.75          & 0.66          & 0.70                     \\
GESSVDD-Sb-SR-max                  &                       & 0.22          & 0.10          & 0.11          & 0.15                     &                       & 0.52          & 0.59          & 0.55                     &                       & 0.39          & 0.76          & 0.57                     \\
GESSVDD-Sb-$\mathcal{S}$-min                   &                       & 0.26          & 0.08          & 0.12          & 0.15                     &                       & 0.47          & 0.39          & 0.43                     &                       & 0.55          & 0.74          & 0.64                     \\
\end{tabular}
\end{table*}
\begin{table*}[h!]
  \footnotesize\setlength{\tabcolsep}{4.9pt}\renewcommand{\arraystretch}{1.20}
  \centering \scriptsize
         \caption{\textit{FPR} results for \textbf{non-linear} data description}
\begin{tabular}{llllllllllllll}
Dataset                            &                       & \multicolumn{4}{c}{Iris}                                                 &                       & \multicolumn{3}{c}{Ionosphere}                           &                       & \multicolumn{3}{c}{Sonar}                                \\ \cline{1-1} \cline{3-6} \cline{8-10} \cline{12-14} 
\multicolumn{1}{|l|}{Target class} & \multicolumn{1}{l|}{} & I-S        & S-VC    & S-V     & \multicolumn{1}{l|}{Av.} & \multicolumn{1}{l|}{} & I-B           & I-G          & \multicolumn{1}{l|}{Av.} & \multicolumn{1}{l|}{} & S-R          & S-M         & \multicolumn{1}{l|}{Av.} \\ \cline{1-1} \cline{3-6} \cline{8-10} \cline{12-14} 
GESSVDD-Sb-$\mathcal{S}$-max                   &                       & 0.04          & 0.06          & \textbf{0.27} & 0.12                     &                       & 0.39          & \textbf{0.43} & 0.41                     &                       & 0.66          & 0.59          & 0.63                     \\
GESSVDD-Sb-GR-max                  &                       & 0.00          & 0.16          & 0.05          & 0.07                     &                       & 0.63          & 0.20          & 0.42                     &                       & 0.67          & 0.68          & 0.68                     \\
GESSVDD-Sb-SR-max                  &                       & 0.25          & 0.15          & 0.11          & 0.17                     &                       & 0.64          & 0.34          & 0.49                     &                       & 0.73          & 0.59          & 0.66                     \\
GESSVDD-Sb-$\mathcal{S}$-min                   &                       & 0.07          & 0.07          & 0.15          & 0.10                     &                       & 0.39          & 0.29          & 0.34                     &                       & 0.58          & 0.44          & 0.51                     \\
\end{tabular}
\end{table*}
\begin{table*}[h!]
  \footnotesize\setlength{\tabcolsep}{4.5pt}\renewcommand{\arraystretch}{1.20}
  \centering \scriptsize
         \caption{\textit{FPR} results for \textbf{non-linear} data description}
\begin{tabular}{lllllllllllllllll}
Dataset                            &                       & \multicolumn{11}{c}{MNIST}                                                                                                                                                               & \multicolumn{1}{c}{}  & \multicolumn{3}{c}{Liver}                                \\ \cline{1-1} \cline{3-13} \cline{15-17} 
\multicolumn{1}{|l|}{Target class} & \multicolumn{1}{l|}{} & 0             & 1             & 2             & 3             & 4             & 5             & 6             & 7             & 8             & 9             & \multicolumn{1}{l|}{Av.} & \multicolumn{1}{l|}{} & DP            & DA            & \multicolumn{1}{l|}{Av.} \\ \cline{1-1} \cline{3-13} \cline{15-17} 
GESSVDD-Sb-$\mathcal{S}$-max                   &                       & 0.84          & 0.84          & 0.94          & 0.92          & 0.61          & 0.71          & 0.69          & \textbf{1.00} & 0.88          & 0.91          & 0.83                     &                       & 0.63          & 0.65          & 0.64                     \\
GESSVDD-Sb-GR-max                  &                       & 0.38          & 0.68          & 0.52          & 0.59          & 0.80          & 0.67          & 0.79          & 0.62          & 0.89          & 0.42          & 0.64                     &                       & 0.71          & 0.65          & 0.68                     \\
GESSVDD-Sb-SR-max                  &                       & 0.99          & 0.41          & 0.74          & 0.33          & 0.95          & 0.96          & 0.90          & 0.74          & 0.89          & 0.53          & 0.74                     &                       & 0.71          & 0.73          & 0.72                     \\
\end{tabular}
\end{table*}

\begin{table*}[h!]
  \footnotesize\setlength{\tabcolsep}{4.9pt}\renewcommand{\arraystretch}{1.20}
  \centering \scriptsize
         \caption{\textit{FPR} results for \textbf{non-linear} data description in the proposed framework with added noise}
%
} \\ \cline{1-1} \cline{3-5} \cline{7-9} \cline{11-13} \cline{15-17} 
\multicolumn{1}{|l|}{Target class} & \multicolumn{1}{l|}{} & DP                          & DA                          & \multicolumn{1}{l|}{Av.}               & \multicolumn{1}{l|}{} & DP                            & DA                            & \multicolumn{1}{l|}{Av.}                & \multicolumn{1}{l|}{} & DP                            & DA                           & \multicolumn{1}{l|}{Av.}                & \multicolumn{1}{l|}{} & DP                             & DA                             & \multicolumn{1}{l|}{Av.}                 \\ \cline{1-1} \cline{3-5} \cline{7-9} \cline{11-13} \cline{15-17} 
GESSVDD-Sb-$\mathcal{S}$-max                   &                       & 0.71                        & 0.71                        & 0.71                                   &                       & 0.58                          & 0.71                          & 0.64                                    &                       & 0.87                          & 0.95                         & 0.91                                    &                       & 0.65                           & 0.60                           & 0.63                                     \\
GESSVDD-Sb-GR-max                  &                       & 0.66                        & 0.66                        & 0.66                                   &                       & 0.65                          & 0.77                          & 0.71                                    &                       & 0.67                          & 0.69                         & 0.68                                    &                       & \textbf{0.73}                  & 0.55                           & 0.64                                     \\
GESSVDD-Sb-SR-max                  &                       & 0.68                        & 0.68                        & 0.68                                   & \textbf{}             & 0.78                          & 0.76                          & 0.77                                    &                       & 0.70                          & 0.78                         & 0.74                                    &                       & 0.57                           & 0.56                           & 0.56                                     \\
GESSVDD-Sb-$\mathcal{S}$-min                   &                       & 0.59                        & 0.59                        & 0.59                                   &                       & 0.56                          & 0.81                          & 0.68                                    &                       & 0.81                          & 0.81                         & 0.81                                    &                       & 0.65                           & \textbf{0.80}                  & \textbf{0.72}                            \\
GESSVDD-Sb-GR-min                  &                       & 0.55                        & 0.55                        & 0.55                                   &                       & 0.68                          & 0.82                          & 0.75                                    &                       & 0.79                          & 0.68                         & 0.74                                    &                       & 0.67                           & 0.67                           & 0.67                                     \\
GESSVDD-Sb-SR-min                  &                       & 0.54                        & 0.54                        & 0.54                                   & \textbf{}             & 0.75                          & 0.87                          & 0.81                                    &                       & 0.64                          & 0.84                         & 0.74                                    &                       & 0.57                           & 0.56                           & 0.57                                     \\
GESSVDD-Sw-$\mathcal{S}$-max                   &                       & 0.57                        & 0.57                        & 0.57                                   &                       & \textbf{1.00}                 & 0.88                          & 0.94                                    &                       & 0.87                          & \textbf{1.00}                & 0.94                                    &                       & 0.64                           & 0.66                           & 0.65                                     \\
GESSVDD-Sw-GR-max                  &                       & \textbf{0.84}               & \textbf{0.84}               & \textbf{0.84}                          &                       & 0.64                          & 0.61                          & 0.62                                    &                       & 0.93                          & 0.89                         & 0.91                                    &                       & 0.58                           & 0.61                           & 0.59                                     \\
GESSVDD-Sw-SR-max                  &                       & 0.57                        & 0.57                        & 0.57                                   &                       & 0.89                          & 0.94                          & 0.92                                    &                       & \textbf{1.00}                 & 0.94                         & 0.97                                    &                       & 0.57                           & 0.59                           & 0.58                                     \\
GESSVDD-Sw-$\mathcal{S}$-min                   &                       & 0.66                        & 0.66                        & 0.66                                   &                       & \textbf{1.00}                 & 0.84                          & 0.92                                    &                       & \textbf{1.00}                 & 0.85                         & 0.92                                    &                       & 0.56                           & 0.57                           & 0.57                                     \\
GESSVDD-Sw-GR-min                  &                       & 0.60                        & 0.60                        & 0.60                                   &                       & 0.82                          & 0.71                          & 0.76                                    &                       & 0.96                          & 0.95                         & 0.95                                    &                       & 0.55                           & 0.56                           & 0.55                                     \\
GESSVDD-Sw-SR-min                  &                       & 0.56                        & 0.56                        & 0.56                                   &                       & 0.97                          & 0.81                          & 0.89                                    &                       & \textbf{1.00}                 & \textbf{1.00}                & \textbf{1.00}                           &                       & 0.59                           & 0.67                           & 0.63                                     \\
GESSVDD-kNN-$\mathcal{S}$-max                  &                       & 0.56                        & 0.56                        & 0.56                                   &                       & \textbf{1.00}                 & \textbf{1.00}                 & \textbf{1.00}                           &                       & 0.86                          & 0.96                         & 0.91                                    &                       & 0.55                           & 0.55                           & 0.55                                     \\
GESSVDD-kNN-GR-max                 &                       & 0.57                        & 0.57                        & 0.57                                   &                       & 0.72                          & 0.81                          & 0.76                                    &                       & 0.88                          & \textbf{1.00}                & 0.94                                    &                       & 0.56                           & 0.62                           & 0.59                                     \\
GESSVDD-kNN-SR-max                 &                       & 0.58                        & 0.58                        & 0.58                                   &                       & 0.96                          & 0.97                          & 0.97                                    &                       & \textbf{1.00}                 & \textbf{1.00}                & \textbf{1.00}                           &                       & 0.53                           & 0.60                           & 0.56                                     \\
GESSVDD-kNN-$\mathcal{S}$-min                  &                       & 0.51                        & 0.51                        & 0.51                                   &                       & 0.97                          & \textbf{1.00}                 & 0.98                                    &                       & 0.97                          & \textbf{1.00}                & 0.98                                    &                       & 0.54                           & 0.57                           & 0.56                                     \\
GESSVDD-kNN-GR-min                 &                       & 0.53                        & 0.53                        & 0.53                                   &                       & \textbf{1.00}                 & \textbf{1.00}                 & \textbf{1.00}                           & \textbf{}             & \textbf{1.00}                 & \textbf{1.00}                & \textbf{1.00}                           &                       & 0.52                           & 0.57                           & 0.54                                     \\
GESSVDD-kNN-SR-min                 &                       & 0.53                        & 0.53                        & 0.53                                   &                       & 0.96                          & 0.93                          & 0.94                                    &                       & 0.94                          & \textbf{1.00}                & 0.97                                    &                       & 0.55                           & 0.62                           & 0.59                                     \\
GESSVDD-PCA-$\mathcal{S}$-max                  &                       & 0.63                        & 0.63                        & 0.63                                   &                       & \textbf{1.00}                 & 0.97                          & 0.99                                    & \textbf{}             & \textbf{1.00}                 & \textbf{1.00}                & \textbf{1.00}                           &                       & 0.51                           & 0.64                           & 0.58                                     \\
GESSVDD-PCA-GR-max                 &                       & 0.53                        & 0.53                        & 0.53                                   &                       & 0.70                          & 0.66                          & 0.68                                    &                       & 0.87                          & 0.84                         & 0.86                                    &                       & 0.55                           & 0.64                           & 0.60                                     \\
GESSVDD-PCA-SR-max                 &                       & 0.54                        & 0.54                        & 0.54                                   &                       & 0.81                          & 0.95                          & 0.88                                    &                       & 0.94                          & 0.91                         & 0.92                                    &                       & 0.66                           & 0.59                           & 0.63                                     \\
GESSVDD-PCA-$\mathcal{S}$-min                  &                       & 0.76                        & 0.76                        & 0.76                                   &                       & 0.94                          & 0.89                          & 0.92                                    & \textbf{}             & \textbf{1.00}                 & \textbf{1.00}                & \textbf{1.00}                           &                       & 0.57                           & 0.61                           & 0.59                                     \\
GESSVDD-PCA-GR-min                 &                       & 0.70                        & 0.70                        & 0.70                                   &                       & 0.72                          & 0.78                          & 0.75                                    &                       & \textbf{1.00}                 & 0.96                         & 0.98                                    &                       & 0.54                           & 0.53                           & 0.53                                     \\
GESSVDD-PCA-SR-min                 &                       & 0.56                        & 0.56                        & 0.56                                   &                       & 0.90                          & 0.93                          & 0.91                                    &                       & 0.97                          & \textbf{1.00}                & 0.98                                    &                       & 0.63                           & 0.59                           & 0.61                                     \\
GESSVDD-I-$\mathcal{S}$-max                    &                       & 0.62                        & 0.62                        & 0.62                                   &                       & 0.94                          & 0.97                          & 0.96                                    & \textbf{}             & 0.96                          & \textbf{1.00}                & 0.98                                    &                       & 0.55                           & 0.62                           & 0.58                                     \\
GESSVDD-I-GR-max                   &                       & 0.58                        & 0.58                        & 0.58                                   &                       & 0.63                          & 0.65                          & 0.64                                    &                       & 0.97                          & 0.95                         & 0.96                                    &                       & 0.52                           & 0.69                           & 0.61                                     \\
GESSVDD-I-SR-max                   &                       & 0.51                        & 0.51                        & 0.51                                   &                       & 0.87                          & 0.97                          & 0.92                                    &                       & \textbf{1.00}                 & \textbf{1.00}                & \textbf{1.00}                           &                       & 0.57                           & 0.56                           & 0.57                                     \\
GESSVDD-I-$\mathcal{S}$-min                    &                       & 0.56                        & 0.56                        & 0.56                                   &                       & \textbf{1.00}                 & 0.75                          & 0.88                                    & \textbf{}             & 0.92                          & \textbf{1.00}                & 0.96                                    &                       & 0.64                           & 0.61                           & 0.62                                     \\
GESSVDD-I-GR-min (ESSVDD)          &                       & 0.62                        & 0.62                        & 0.62                                   &                       & 0.63                          & 0.85                          & 0.74                                    &                       & \textbf{1.00}                 & 0.89                         & 0.94                                    &                       & 0.62                           & 0.54                           & 0.58                                     \\
GESSVDD-I-SR-min                   &                       & 0.54                        & 0.54                        & 0.54                                   &                       & 0.85                          & 0.86                          & 0.86                                    &                       & \textbf{1.00}                 & 0.92                         & 0.96                                    &                       & 0.58                           & 0.55                           & 0.56                                     \\
GESSVDD-0-$\mathcal{S}$-max                    &                       & 0.65                        & 0.65                        & 0.65                                   &                       & 0.76                          & 0.79                          & 0.78                                    &                       & 0.74                          & 0.68                         & 0.71                                    &                       & 0.61                           & 0.64                           & 0.62                                     \\
GESSVDD-0-GR-max                   &                       & 0.55                        & 0.55                        & 0.55                                   &                       & 0.94                          & 0.78                          & 0.86                                    &                       & 0.94                          & 0.56                         & 0.75                                    &                       & 0.54                           & 0.64                           & 0.59                                     \\
GESSVDD-0-$\mathcal{S}$-min                    &                       & 0.68                        & 0.68                        & 0.68                                   &                       & 0.86                          & 0.84                          & 0.85                                    &                       & 0.64                          & 0.91                         & 0.77                                    &                       & 0.66                           & 0.60                           & 0.63                                     \\
GESSVDD-0-GR-min (SSVDD)           &                       & 0.47                        & 0.47                        & 0.47                                   &                       & 0.88                          & 0.76                          & 0.82                                    &                       & \textbf{1.00}                 & 0.91                         & 0.96                                    &                       & 0.63                           & 0.53                           & 0.58                                     \\
ESVDD                              &                       & 0.66                        & 0.66                        & 0.66                                   &                       & \textbf{1.00}                 & \textbf{1.00}                 & \textbf{1.00}                           &                       & \textbf{1.00}                 & 0.97                         & 0.98                                    &                       & 0.51                           & 0.48                           & 0.49                                     \\
SVDD                               &                       & 0.47                        & 0.47                        & 0.47                                   &                       & 0.93                          & 0.85                          & 0.89                                    &                       & 0.78                          & 0.93                         & 0.85                                    &                       & 0.53                           & 0.55                           & 0.54                                     \\
OCSVM                              &                       & 0.80                        & 0.80                        & 0.80                                   &                       & \textbf{1.00}                 & \textbf{1.00}                 & \textbf{1.00}                           &                       & \textbf{1.00}                 & 0.93                         & 0.97                                    &                       & 0.50                           & 0.51                           & 0.50                                     \\
GESVDD-PCA                         &                       & 0.32                        & 0.32                        & 0.32                                   &                       & \textbf{1.00}                 & \textbf{1.00}                 & \textbf{1.00}                           &                       & \textbf{1.00}                 & \textbf{1.00}                & \textbf{1.00}                           &                       & 0.49                           & 0.47                           & 0.48                                     \\
GESVDD-Sw                          &                       & 0.32                        & 0.32                        & 0.32                                   &                       & \textbf{1.00}                 & \textbf{1.00}                 & \textbf{1.00}                           &                       & \textbf{1.00}                 & \textbf{1.00}                & \textbf{1.00}                           &                       & 0.47                           & 0.48                           & 0.48                                     \\
GESVDD-kNN                         &                       & 0.30                        & 0.30                        & 0.30                                   &                       & \textbf{1.00}                 & \textbf{1.00}                 & \textbf{1.00}                           &                       & \textbf{1.00}                 & \textbf{1.00}                & \textbf{1.00}                           &                       & 0.52                           & 0.48                           & 0.50                                     \\
GESVM-PCA                          &                       & 0.34                        & 0.34                        & 0.34                                   &                       & \textbf{1.00}                 & \textbf{1.00}                 & \textbf{1.00}                           &                       & \textbf{1.00}                 & \textbf{1.00}                & \textbf{1.00}                           &                       & 0.51                           & 0.51                           & 0.51                                     \\
GESVM-Sw                           &                       & 0.33                        & 0.33                        & 0.33                                   &                       & \textbf{1.00}                 & \textbf{1.00}                 & \textbf{1.00}                           &                       & \textbf{1.00}                 & 0.95                         & 0.97                                    &                       & 0.50                           & 0.52                           & 0.51                                     \\
GESVM-kNN                          &                       & 0.24                        & 0.24                        & 0.24                                   &                       & 0.00                          & 0.00                          & 0.00                                    &                       & \textbf{1.00}                 & \textbf{1.00}                & \textbf{1.00}                           &                       & 0.44                           & 0.44                           & 0.44                                    
\end{tabular}
\end{table*}

\clearpage
\subsubsection{FNR results \textbf{non-linear} data description}
\begin{table*}[h!]
  \footnotesize\setlength{\tabcolsep}{4.9pt}\renewcommand{\arraystretch}{1.20}
  \centering \scriptsize
         \caption{\textit{FNR} results for \textbf{non-linear} data description}
\begin{tabular}{llllllllllllll}
Dataset                            &                       & \multicolumn{4}{c}{Seeds}                                                &                       & \multicolumn{3}{c}{Qualitative bankruptcy}               &                       & \multicolumn{3}{c}{Somerville happiness}                 \\ \cline{1-1} \cline{3-6} \cline{8-10} \cline{12-14} 
\multicolumn{1}{|l|}{Target class} & \multicolumn{1}{l|}{} & S-K          & S-R          & S-C      & \multicolumn{1}{l|}{Av.} & \multicolumn{1}{l|}{} & QB-B            & QB-N        & \multicolumn{1}{l|}{Av.} & \multicolumn{1}{l|}{} & SH-H         & SH-U      & \multicolumn{1}{l|}{Av.} \\ \cline{1-1} \cline{3-6} \cline{8-10} \cline{12-14} 
GESSVDD-Sb-$\mathcal{S}$-max                   &                       & 0.14          & 0.06          & 0.11          & 0.10                     &                       & 0.24          & 0.54          & 0.39                     &                       & 0.22          & 0.23          & 0.22                     \\
GESSVDD-Sb-GR-max                  &                       & 0.22          & 0.36          & 0.08          & 0.22                     &                       & 0.07          & 0.41          & 0.24                     &                       & 0.26          & 0.34          & 0.30                     \\
GESSVDD-Sb-SR-max                  &                       & 0.22          & 0.26          & 0.03          & 0.17                     &                       & 0.30          & 0.55          & 0.42                     &                       & 0.53          & 0.51          & 0.52                     \\
GESSVDD-Sb-$\mathcal{S}$-min                   &                       & 0.14          & 0.10          & 0.19          & 0.15                     &                       & 0.24          & 0.49          & 0.37                     &                       & 0.38          & 0.28          & 0.33                     \\
\end{tabular}
\end{table*}
\begin{table*}[h!]
  \footnotesize\setlength{\tabcolsep}{4.9pt}\renewcommand{\arraystretch}{1.20}
  \centering \scriptsize
         \caption{\textit{FNR} results for \textbf{non-linear} data description}
%
\end{table*}
\begin{table*}[h!]
  \footnotesize\setlength{\tabcolsep}{4.5pt}\renewcommand{\arraystretch}{1.3}
  \centering \scriptsize
         \caption{\textit{FNR} results for \textbf{non-linear} data description}
%
\end{table*}

\begin{table*}[h!]
  \footnotesize\setlength{\tabcolsep}{4.9pt}\renewcommand{\arraystretch}{1.20}
  \centering \scriptsize
         \caption{\textit{FNR} results for \textbf{non-linear} data description in the proposed framework with added noise}
%
} \\ \cline{1-1} \cline{3-5} \cline{7-9} \cline{11-13} \cline{15-17} 
\multicolumn{1}{|l|}{Target class} & \multicolumn{1}{l|}{} & DP                          & DA                          & \multicolumn{1}{l|}{Av.}               & \multicolumn{1}{l|}{} & DP                            & DA                            & \multicolumn{1}{l|}{Av.}                & \multicolumn{1}{l|}{} & DP                            & DA                           & \multicolumn{1}{l|}{Av.}                & \multicolumn{1}{l|}{} & DP                             & DA                             & \multicolumn{1}{l|}{Av.}                 \\ \cline{1-1} \cline{3-5} \cline{7-9} \cline{11-13} \cline{15-17} 
GESSVDD-Sb-$\mathcal{S}$-max                   &                       & 0.71                        & 0.71                        & 0.71                                   &                       & 0.58                          & 0.71                          & 0.64                                    &                       & 0.87                          & 0.95                         & 0.91                                    &                       & 0.65                           & 0.60                           & 0.63                                     \\
GESSVDD-Sb-GR-max                  &                       & 0.66                        & 0.66                        & 0.66                                   &                       & 0.65                          & 0.77                          & 0.71                                    &                       & 0.67                          & 0.69                         & 0.68                                    &                       & \textbf{0.73}                  & 0.55                           & 0.64                                     \\
GESSVDD-Sb-SR-max                  &                       & 0.68                        & 0.68                        & 0.68                                   & \textbf{}             & 0.78                          & 0.76                          & 0.77                                    &                       & 0.70                          & 0.78                         & 0.74                                    &                       & 0.57                           & 0.56                           & 0.56                                     \\
GESSVDD-Sb-$\mathcal{S}$-min                   &                       & 0.59                        & 0.59                        & 0.59                                   &                       & 0.56                          & 0.81                          & 0.68                                    &                       & 0.81                          & 0.81                         & 0.81                                    &                       & 0.65                           & \textbf{0.80}                  & \textbf{0.72}                            \\
GESSVDD-Sb-GR-min                  &                       & 0.55                        & 0.55                        & 0.55                                   &                       & 0.68                          & 0.82                          & 0.75                                    &                       & 0.79                          & 0.68                         & 0.74                                    &                       & 0.67                           & 0.67                           & 0.67                                     \\
GESSVDD-Sb-SR-min                  &                       & 0.54                        & 0.54                        & 0.54                                   & \textbf{}             & 0.75                          & 0.87                          & 0.81                                    &                       & 0.64                          & 0.84                         & 0.74                                    &                       & 0.57                           & 0.56                           & 0.57                                     \\
GESSVDD-Sw-$\mathcal{S}$-max                   &                       & 0.57                        & 0.57                        & 0.57                                   &                       & \textbf{1.00}                 & 0.88                          & 0.94                                    &                       & 0.87                          & \textbf{1.00}                & 0.94                                    &                       & 0.64                           & 0.66                           & 0.65                                     \\
GESSVDD-Sw-GR-max                  &                       & \textbf{0.84}               & \textbf{0.84}               & \textbf{0.84}                          &                       & 0.64                          & 0.61                          & 0.62                                    &                       & 0.93                          & 0.89                         & 0.91                                    &                       & 0.58                           & 0.61                           & 0.59                                     \\
GESSVDD-Sw-SR-max                  &                       & 0.57                        & 0.57                        & 0.57                                   &                       & 0.89                          & 0.94                          & 0.92                                    &                       & \textbf{1.00}                 & 0.94                         & 0.97                                    &                       & 0.57                           & 0.59                           & 0.58                                     \\
GESSVDD-Sw-$\mathcal{S}$-min                   &                       & 0.66                        & 0.66                        & 0.66                                   &                       & \textbf{1.00}                 & 0.84                          & 0.92                                    &                       & \textbf{1.00}                 & 0.85                         & 0.92                                    &                       & 0.56                           & 0.57                           & 0.57                                     \\
GESSVDD-Sw-GR-min                  &                       & 0.60                        & 0.60                        & 0.60                                   &                       & 0.82                          & 0.71                          & 0.76                                    &                       & 0.96                          & 0.95                         & 0.95                                    &                       & 0.55                           & 0.56                           & 0.55                                     \\
GESSVDD-Sw-SR-min                  &                       & 0.56                        & 0.56                        & 0.56                                   &                       & 0.97                          & 0.81                          & 0.89                                    &                       & \textbf{1.00}                 & \textbf{1.00}                & \textbf{1.00}                           &                       & 0.59                           & 0.67                           & 0.63                                     \\
GESSVDD-kNN-$\mathcal{S}$-max                  &                       & 0.56                        & 0.56                        & 0.56                                   &                       & \textbf{1.00}                 & \textbf{1.00}                 & \textbf{1.00}                           &                       & 0.86                          & 0.96                         & 0.91                                    &                       & 0.55                           & 0.55                           & 0.55                                     \\
GESSVDD-kNN-GR-max                 &                       & 0.57                        & 0.57                        & 0.57                                   &                       & 0.72                          & 0.81                          & 0.76                                    &                       & 0.88                          & \textbf{1.00}                & 0.94                                    &                       & 0.56                           & 0.62                           & 0.59                                     \\
GESSVDD-kNN-SR-max                 &                       & 0.58                        & 0.58                        & 0.58                                   &                       & 0.96                          & 0.97                          & 0.97                                    &                       & \textbf{1.00}                 & \textbf{1.00}                & \textbf{1.00}                           &                       & 0.53                           & 0.60                           & 0.56                                     \\
GESSVDD-kNN-$\mathcal{S}$-min                  &                       & 0.51                        & 0.51                        & 0.51                                   &                       & 0.97                          & \textbf{1.00}                 & 0.98                                    &                       & 0.97                          & \textbf{1.00}                & 0.98                                    &                       & 0.54                           & 0.57                           & 0.56                                     \\
GESSVDD-kNN-GR-min                 &                       & 0.53                        & 0.53                        & 0.53                                   &                       & \textbf{1.00}                 & \textbf{1.00}                 & \textbf{1.00}                           & \textbf{}             & \textbf{1.00}                 & \textbf{1.00}                & \textbf{1.00}                           &                       & 0.52                           & 0.57                           & 0.54                                     \\
GESSVDD-kNN-SR-min                 &                       & 0.53                        & 0.53                        & 0.53                                   &                       & 0.96                          & 0.93                          & 0.94                                    &                       & 0.94                          & \textbf{1.00}                & 0.97                                    &                       & 0.55                           & 0.62                           & 0.59                                     \\
GESSVDD-PCA-$\mathcal{S}$-max                  &                       & 0.63                        & 0.63                        & 0.63                                   &                       & \textbf{1.00}                 & 0.97                          & 0.99                                    & \textbf{}             & \textbf{1.00}                 & \textbf{1.00}                & \textbf{1.00}                           &                       & 0.51                           & 0.64                           & 0.58                                     \\
GESSVDD-PCA-GR-max                 &                       & 0.53                        & 0.53                        & 0.53                                   &                       & 0.70                          & 0.66                          & 0.68                                    &                       & 0.87                          & 0.84                         & 0.86                                    &                       & 0.55                           & 0.64                           & 0.60                                     \\
GESSVDD-PCA-SR-max                 &                       & 0.54                        & 0.54                        & 0.54                                   &                       & 0.81                          & 0.95                          & 0.88                                    &                       & 0.94                          & 0.91                         & 0.92                                    &                       & 0.66                           & 0.59                           & 0.63                                     \\
GESSVDD-PCA-$\mathcal{S}$-min                  &                       & 0.76                        & 0.76                        & 0.76                                   &                       & 0.94                          & 0.89                          & 0.92                                    & \textbf{}             & \textbf{1.00}                 & \textbf{1.00}                & \textbf{1.00}                           &                       & 0.57                           & 0.61                           & 0.59                                     \\
GESSVDD-PCA-GR-min                 &                       & 0.70                        & 0.70                        & 0.70                                   &                       & 0.72                          & 0.78                          & 0.75                                    &                       & \textbf{1.00}                 & 0.96                         & 0.98                                    &                       & 0.54                           & 0.53                           & 0.53                                     \\
GESSVDD-PCA-SR-min                 &                       & 0.56                        & 0.56                        & 0.56                                   &                       & 0.90                          & 0.93                          & 0.91                                    &                       & 0.97                          & \textbf{1.00}                & 0.98                                    &                       & 0.63                           & 0.59                           & 0.61                                     \\
GESSVDD-I-$\mathcal{S}$-max                    &                       & 0.62                        & 0.62                        & 0.62                                   &                       & 0.94                          & 0.97                          & 0.96                                    & \textbf{}             & 0.96                          & \textbf{1.00}                & 0.98                                    &                       & 0.55                           & 0.62                           & 0.58                                     \\
GESSVDD-I-GR-max                   &                       & 0.58                        & 0.58                        & 0.58                                   &                       & 0.63                          & 0.65                          & 0.64                                    &                       & 0.97                          & 0.95                         & 0.96                                    &                       & 0.52                           & 0.69                           & 0.61                                     \\
GESSVDD-I-SR-max                   &                       & 0.51                        & 0.51                        & 0.51                                   &                       & 0.87                          & 0.97                          & 0.92                                    &                       & \textbf{1.00}                 & \textbf{1.00}                & \textbf{1.00}                           &                       & 0.57                           & 0.56                           & 0.57                                     \\
GESSVDD-I-$\mathcal{S}$-min                    &                       & 0.56                        & 0.56                        & 0.56                                   &                       & \textbf{1.00}                 & 0.75                          & 0.88                                    & \textbf{}             & 0.92                          & \textbf{1.00}                & 0.96                                    &                       & 0.64                           & 0.61                           & 0.62                                     \\
GESSVDD-I-GR-min (ESSVDD)          &                       & 0.62                        & 0.62                        & 0.62                                   &                       & 0.63                          & 0.85                          & 0.74                                    &                       & \textbf{1.00}                 & 0.89                         & 0.94                                    &                       & 0.62                           & 0.54                           & 0.58                                     \\
GESSVDD-I-SR-min                   &                       & 0.54                        & 0.54                        & 0.54                                   &                       & 0.85                          & 0.86                          & 0.86                                    &                       & \textbf{1.00}                 & 0.92                         & 0.96                                    &                       & 0.58                           & 0.55                           & 0.56                                     \\
GESSVDD-0-$\mathcal{S}$-max                    &                       & 0.65                        & 0.65                        & 0.65                                   &                       & 0.76                          & 0.79                          & 0.78                                    &                       & 0.74                          & 0.68                         & 0.71                                    &                       & 0.61                           & 0.64                           & 0.62                                     \\
GESSVDD-0-GR-max                   &                       & 0.55                        & 0.55                        & 0.55                                   &                       & 0.94                          & 0.78                          & 0.86                                    &                       & 0.94                          & 0.56                         & 0.75                                    &                       & 0.54                           & 0.64                           & 0.59                                     \\
GESSVDD-0-$\mathcal{S}$-min                    &                       & 0.68                        & 0.68                        & 0.68                                   &                       & 0.86                          & 0.84                          & 0.85                                    &                       & 0.64                          & 0.91                         & 0.77                                    &                       & 0.66                           & 0.60                           & 0.63                                     \\
GESSVDD-0-GR-min (SSVDD)           &                       & 0.47                        & 0.47                        & 0.47                                   &                       & 0.88                          & 0.76                          & 0.82                                    &                       & \textbf{1.00}                 & 0.91                         & 0.96                                    &                       & 0.63                           & 0.53                           & 0.58                                     \\
ESVDD                              &                       & 0.66                        & 0.66                        & 0.66                                   &                       & \textbf{1.00}                 & \textbf{1.00}                 & \textbf{1.00}                           &                       & \textbf{1.00}                 & 0.97                         & 0.98                                    &                       & 0.51                           & 0.48                           & 0.49                                     \\
SVDD                               &                       & 0.47                        & 0.47                        & 0.47                                   &                       & 0.93                          & 0.85                          & 0.89                                    &                       & 0.78                          & 0.93                         & 0.85                                    &                       & 0.53                           & 0.55                           & 0.54                                     \\
OCSVM                              &                       & 0.80                        & 0.80                        & 0.80                                   &                       & \textbf{1.00}                 & \textbf{1.00}                 & \textbf{1.00}                           &                       & \textbf{1.00}                 & 0.93                         & 0.97                                    &                       & 0.50                           & 0.51                           & 0.50                                     \\
GESVDD-PCA                         &                       & 0.32                        & 0.32                        & 0.32                                   &                       & \textbf{1.00}                 & \textbf{1.00}                 & \textbf{1.00}                           &                       & \textbf{1.00}                 & \textbf{1.00}                & \textbf{1.00}                           &                       & 0.49                           & 0.47                           & 0.48                                     \\
GESVDD-Sw                          &                       & 0.32                        & 0.32                        & 0.32                                   &                       & \textbf{1.00}                 & \textbf{1.00}                 & \textbf{1.00}                           &                       & \textbf{1.00}                 & \textbf{1.00}                & \textbf{1.00}                           &                       & 0.47                           & 0.48                           & 0.48                                     \\
GESVDD-kNN                         &                       & 0.30                        & 0.30                        & 0.30                                   &                       & \textbf{1.00}                 & \textbf{1.00}                 & \textbf{1.00}                           &                       & \textbf{1.00}                 & \textbf{1.00}                & \textbf{1.00}                           &                       & 0.52                           & 0.48                           & 0.50                                     \\
GESVM-PCA                          &                       & 0.34                        & 0.34                        & 0.34                                   &                       & \textbf{1.00}                 & \textbf{1.00}                 & \textbf{1.00}                           &                       & \textbf{1.00}                 & \textbf{1.00}                & \textbf{1.00}                           &                       & 0.51                           & 0.51                           & 0.51                                     \\
GESVM-Sw                           &                       & 0.33                        & 0.33                        & 0.33                                   &                       & \textbf{1.00}                 & \textbf{1.00}                 & \textbf{1.00}                           &                       & \textbf{1.00}                 & 0.95                         & 0.97                                    &                       & 0.50                           & 0.52                           & 0.51                                     \\
GESVM-kNN                          &                       & 0.40                        & 0.40                        & 0.40                                   &                       & \textbf{1.00}                 & \textbf{1.00}                 & \textbf{1.00}                           &                       & 0.00                          & 0.00                         & 0.00                                    &                       & 0.55                           & 0.55                           & 0.55                                    
\end{tabular}
\end{table*}

\clearpage
\section{\textit{Standard deviation} of the \textit{Gmean} results for \textbf{linear} methods}\label{sdlinear}

\begin{table*}[h!]
  \footnotesize\setlength{\tabcolsep}{2.8pt}\renewcommand{\arraystretch}{1.20}
  \centering \scriptsize
         \caption{\textit{Standard deviation} of the \textit{Gmean} results for \textbf{linear} data description over Seeds, Qualitative bankruptcy, Somerville happiness and Liver datasets}

\end{table*}
\begin{table*}[h!]
  \footnotesize\setlength{\tabcolsep}{4.9pt}\renewcommand{\arraystretch}{1.30}
  \centering \scriptsize
         \caption{\textit{Standard deviation} of the \textit{Gmean} results for \textbf{linear} data description over Iris, ionosphere, and Sonar datasets}
%
\end{table*}
\begin{table*}[h!]
  \footnotesize\setlength{\tabcolsep}{4.9pt}\renewcommand{\arraystretch}{1.20}
  \centering \scriptsize
         \caption{\textit{Standard deviation} of the \textit{Gmean} results for \textbf{linear} data description over Heart dataset and its manually created corrupted versions}
%
} \\ \cline{1-1} \cline{3-5} \cline{7-9} \cline{11-13} \cline{15-17} 
\multicolumn{1}{|l|}{Target class} & \multicolumn{1}{l|}{} & DP                       & DA                       & \multicolumn{1}{l|}{Av.}                     & \multicolumn{1}{l|}{} & DP                         & DA                         & \multicolumn{1}{l|}{Av.}                      & \multicolumn{1}{l|}{} & DP                         & DA                        & \multicolumn{1}{l|}{Av.}                      & \multicolumn{1}{l|}{} & DP                          & DA                          & \multicolumn{1}{l|}{Av.}                       \\ \cline{1-1} \cline{3-5} \cline{7-9} \cline{11-13} \cline{15-17} 
GESSVDD-Sb-$\mathcal{S}$-max                   &                       & 0.12                     & 0.25                     & 0.19                                         &                       & 0.14                       & 0.14                       & 0.14                                          &                       & 0.16                       & 0.09                      & 0.13                                          &                       & 0.12                        & 0.12                        & 0.12                                           \\
GESSVDD-Sb-GR-max                  &                       & 0.17                     & 0.32                     & 0.25                                         &                       & 0.18                       & 0.16                       & 0.17                                          &                       & 0.23                       & 0.22                      & 0.23                                          &                       & 0.10                        & 0.14                        & 0.12                                           \\
GESSVDD-Sb-SR-max                  &                       & 0.18                     & 0.15                     & 0.16                                         & \textbf{}             & 0.14                       & 0.17                       & 0.15                                          &                       & 0.24                       & 0.21                      & 0.22                                          &                       & 0.06                        & 0.14                        & 0.10                                           \\
GESSVDD-Sb-$\mathcal{S}$-min                   &                       & 0.18                     & 0.20                     & 0.19                                         &                       & 0.11                       & 0.18                       & 0.15                                          &                       & 0.11                       & 0.13                      & 0.12                                          &                       & 0.14                        & 0.11                        & 0.12                                           \\
GESSVDD-Sb-GR-min                  &                       & 0.16                     & 0.04                     & 0.10                                         &                       & 0.16                       & 0.17                       & 0.16                                          &                       & 0.21                       & 0.21                      & 0.21                                          &                       & 0.13                        & 0.09                        & 0.11                                           \\
GESSVDD-Sb-SR-min                  &                       & 0.11                     & 0.15                     & 0.13                                         & \textbf{}             & 0.18                       & 0.17                       & 0.17                                          &                       & 0.27                       & 0.08                      & 0.18                                          &                       & 0.08                        & 0.11                        & 0.09                                           \\
GESSVDD-Sw-$\mathcal{S}$-max                   &                       & 0.08                     & 0.09                     & 0.09                                         &                       & 0.20                       & 0.14                       & 0.17                                          &                       & 0.00                       & 0.00                      & 0.00                                          &                       & 0.12                        & 0.08                        & 0.10                                           \\
GESSVDD-Sw-GR-max                  &                       & 0.15                     & 0.04                     & 0.10                                         &                       & 0.28                       & 0.26                       & 0.27                                          &                       & 0.00                       & 0.00                      & 0.00                                          &                       & 0.17                        & 0.05                        & 0.11                                           \\
GESSVDD-Sw-SR-max                  &                       & 0.11                     & 0.12                     & 0.12                                         &                       & 0.06                       & 0.00                       & 0.03                                          &                       & 0.00                       & 0.00                      & 0.00                                          &                       & 0.08                        & 0.05                        & 0.06                                           \\
GESSVDD-Sw-$\mathcal{S}$-min                   &                       & 0.16                     & 0.08                     & 0.12                                         &                       & 0.14                       & 0.17                       & 0.16                                          &                       & 0.00                       & 0.00                      & 0.00                                          &                       & 0.05                        & 0.07                        & 0.06                                           \\
GESSVDD-Sw-GR-min                  &                       & 0.09                     & 0.09                     & 0.09                                         &                       & 0.28                       & 0.27                       & 0.28                                          &                       & 0.00                       & 0.00                      & 0.00                                          &                       & 0.08                        & 0.07                        & 0.08                                           \\
GESSVDD-Sw-SR-min                  &                       & 0.06                     & 0.06                     & 0.06                                         &                       & 0.00                       & 0.06                       & 0.03                                          &                       & 0.06                       & 0.00                      & 0.03                                          &                       & 0.04                        & 0.07                        & 0.05                                           \\
GESSVDD-kNN-$\mathcal{S}$-max                  &                       & 0.06                     & 0.13                     & 0.10                                         &                       & 0.09                       & 0.13                       & 0.11                                          &                       & 0.08                       & 0.00                      & 0.04                                          &                       & 0.15                        & 0.07                        & 0.11                                           \\
GESSVDD-kNN-GR-max                 &                       & 0.09                     & 0.09                     & 0.09                                         &                       & 0.19                       & 0.19                       & 0.19                                          &                       & 0.24                       & 0.19                      & 0.21                                          &                       & 0.13                        & 0.07                        & 0.10                                           \\
GESSVDD-kNN-SR-max                 &                       & 0.15                     & 0.11                     & 0.13                                         &                       & 0.00                       & 0.08                       & 0.04                                          &                       & 0.00                       & 0.00                      & 0.00                                          &                       & 0.11                        & 0.09                        & 0.10                                           \\
GESSVDD-kNN-$\mathcal{S}$-min                  &                       & 0.11                     & 0.16                     & 0.14                                         &                       & 0.11                       & 0.06                       & 0.09                                          &                       & 0.17                       & 0.00                      & 0.08                                          &                       & 0.10                        & 0.09                        & 0.09                                           \\
GESSVDD-kNN-GR-min                 &                       & 0.09                     & 0.09                     & 0.09                                         &                       & 0.13                       & 0.00                       & 0.06                                          &                       & 0.30                       & 0.19                      & 0.24                                          &                       & 0.05                        & 0.07                        & 0.06                                           \\
GESSVDD-kNN-SR-min                 &                       & 0.06                     & 0.12                     & 0.09                                         &                       & 0.00                       & 0.19                       & 0.10                                          &                       & 0.00                       & 0.00                      & 0.00                                          &                       & 0.10                        & 0.06                        & 0.08                                           \\
GESSVDD-PCA-$\mathcal{S}$-max                  &                       & 0.12                     & 0.10                     & 0.11                                         &                       & 0.11                       & 0.00                       & 0.05                                          & \textbf{}             & 0.00                       & 0.00                      & 0.00                                          &                       & 0.08                        & 0.07                        & 0.07                                           \\
GESSVDD-PCA-GR-max                 &                       & 0.07                     & 0.06                     & 0.07                                         &                       & 0.00                       & 0.00                       & 0.00                                          &                       & 0.00                       & 0.00                      & 0.00                                          &                       & 0.11                        & 0.07                        & 0.09                                           \\
GESSVDD-PCA-SR-max                 &                       & 0.10                     & 0.06                     & 0.08                                         &                       & 0.00                       & 0.06                       & 0.03                                          &                       & 0.00                       & 0.00                      & 0.00                                          &                       & 0.13                        & 0.08                        & 0.10                                           \\
GESSVDD-PCA-$\mathcal{S}$-min                  &                       & 0.13                     & 0.08                     & 0.11                                         &                       & 0.00                       & 0.19                       & 0.09                                          & \textbf{}             & 0.00                       & 0.00                      & 0.00                                          &                       & 0.05                        & 0.06                        & 0.05                                           \\
GESSVDD-PCA-GR-min                 &                       & 0.05                     & 0.15                     & 0.10                                         &                       & 0.27                       & 0.00                       & 0.14                                          &                       & 0.00                       & 0.00                      & 0.00                                          &                       & 0.09                        & 0.03                        & 0.06                                           \\
GESSVDD-PCA-SR-min                 &                       & 0.10                     & 0.09                     & 0.09                                         &                       & 0.16                       & 0.12                       & 0.14                                          &                       & 0.00                       & 0.00                      & 0.00                                          &                       & 0.11                        & 0.10                        & 0.10                                           \\
GESSVDD-I-$\mathcal{S}$-max                    &                       & 0.11                     & 0.10                     & 0.11                                         &                       & 0.13                       & 0.12                       & 0.13                                          & \textbf{}             & 0.00                       & 0.00                      & 0.00                                          &                       & 0.11                        & 0.07                        & 0.09                                           \\
GESSVDD-I-GR-max                   &                       & 0.11                     & 0.15                     & 0.13                                         &                       & 0.28                       & 0.21                       & 0.25                                          &                       & 0.00                       & 0.00                      & 0.00                                          &                       & 0.12                        & 0.04                        & 0.08                                           \\
GESSVDD-I-SR-max                   &                       & 0.09                     & 0.08                     & 0.09                                         &                       & 0.06                       & 0.06                       & 0.06                                          &                       & 0.00                       & 0.00                      & 0.00                                          &                       & 0.12                        & 0.08                        & 0.10                                           \\
GESSVDD-I-$\mathcal{S}$-min                    &                       & 0.06                     & 0.14                     & 0.10                                         &                       & 0.15                       & 0.06                       & 0.11                                          & \textbf{}             & 0.00                       & 0.00                      & 0.00                                          &                       & 0.04                        & 0.06                        & 0.05                                           \\
GESSVDD-I-GR-min (ESSVDD)          &                       & 0.11                     & 0.08                     & 0.09                                         &                       & 0.00                       & 0.00                       & 0.00                                          &                       & 0.00                       & 0.00                      & 0.00                                          &                       & 0.09                        & 0.08                        & 0.08                                           \\
GESSVDD-I-SR-min                   &                       & 0.10                     & 0.12                     & 0.11                                         &                       & 0.06                       & 0.00                       & 0.03                                          &                       & 0.00                       & 0.00                      & 0.00                                          &                       & 0.04                        & 0.15                        & 0.09                                           \\
GESSVDD-0-$\mathcal{S}$-max                    &                       & 0.11                     & 0.14                     & 0.13                                         &                       & 0.15                       & 0.21                       & 0.18                                          &                       & 0.00                       & 0.00                      & 0.00                                          &                       & 0.19                        & 0.10                        & 0.14                                           \\
GESSVDD-0-GR-max                   &                       & 0.12                     & 0.09                     & 0.10                                         &                       & 0.21                       & 0.24                       & 0.23                                          &                       & 0.00                       & 0.00                      & 0.00                                          &                       & 0.09                        & 0.08                        & 0.09                                           \\
GESSVDD-0-$\mathcal{S}$-min                    &                       & 0.07                     & 0.04                     & 0.05                                         &                       & 0.20                       & 0.00                       & 0.10                                          &                       & 0.00                       & 0.00                      & 0.00                                          &                       & 0.11                        & 0.05                        & 0.08                                           \\
GESSVDD-0-GR-min (SSVDD)           &                       & 0.10                     & 0.11                     & 0.10                                         &                       & 0.00                       & 0.24                       & 0.12                                          &                       & 0.00                       & 0.00                      & 0.00                                          &                       & 0.09                        & 0.05                        & 0.07                                           \\
ESVDD                              &                       & 0.07                     & 0.05                     & 0.06                                         &                       & 0.00                       & 0.00                       & 0.00                                          &                       & 0.00                       & 0.00                      & 0.00                                          &                       & 0.09                        & 0.03                        & 0.06                                           \\
SVDD                               &                       & 0.17                     & 0.10                     & 0.14                                         &                       & 0.11                       & 0.08                       & 0.10                                          &                       & 0.31                       & 0.09                      & 0.20                                          &                       & 0.09                        & 0.09                        & 0.09                                           \\
OCSVM                              &                       & 0.08                     & 0.06                     & 0.07                                         &                       & 0.00                       & 0.00                       & 0.00                                          &                       & 0.00                       & 0.00                      & 0.00                                          &                       & 0.09                        & 0.06                        & 0.07                                          
\end{tabular}
\end{table*}

\clearpage

\section{\textit{Standard deviation} of the \textit{Gmean} results for \textbf{non-linear} methods}\label{sdnonlinear}

\begin{table*}[h!]
  \footnotesize\setlength{\tabcolsep}{2.8pt}\renewcommand{\arraystretch}{1.20}
  \centering \scriptsize

         \caption{  \textit{Standard deviation} of the \textit{Gmean} results for \textbf{non-linear} data description over Seeds, Qualitative bankruptcy, Somerville happiness and Liver datasets}

\end{table*}
\begin{table*}[h!]
  \footnotesize\setlength{\tabcolsep}{3.0pt}\renewcommand{\arraystretch}{1.20}
  \centering \scriptsize
         \caption{\textit{Standard deviation} of the \textit{Gmean} results for \textbf{non-linear} data description over Iris, ionosphere, and Sonar datasets}
%
\end{table*}
\begin{table*}[h!]
  \footnotesize\setlength{\tabcolsep}{4.9pt}\renewcommand{\arraystretch}{1.20}
  \centering \scriptsize
         \caption{\textit{Standard deviation} of the \textit{Gmean} results for \textbf{non-linear} data description over Heart dataset and its manually created corrupted versions}
%
} \\ \cline{1-1} \cline{3-5} \cline{7-9} \cline{11-13} \cline{15-17} 
\multicolumn{1}{|l|}{Target class} & \multicolumn{1}{l|}{} & DP                       & DA                       & \multicolumn{1}{l|}{Av.}                     & \multicolumn{1}{l|}{} & DP                         & DA                         & \multicolumn{1}{l|}{Av.}                      & \multicolumn{1}{l|}{} & DP                         & DA                        & \multicolumn{1}{l|}{Av.}                      & \multicolumn{1}{l|}{} & DP                          & DA                          & \multicolumn{1}{l|}{Av.}                       \\ \cline{1-1} \cline{3-5} \cline{7-9} \cline{11-13} \cline{15-17} 
GESSVDD-Sb-$\mathcal{S}$-max                   &                       & 0.21                     & 0.20                     & 0.20                                         &                       & 0.10                       & 0.16                       & 0.13                                          &                       & 0.21                       & 0.11                      & 0.16                                          &                       & 0.13                        & 0.08                        & 0.10                                           \\
GESSVDD-Sb-GR-max                  &                       & 0.20                     & 0.17                     & 0.19                                         &                       & 0.07                       & 0.15                       & 0.11                                          &                       & 0.26                       & 0.27                      & 0.26                                          &                       & 0.20                        & 0.06                        & 0.13                                           \\
GESSVDD-Sb-SR-max                  &                       & 0.20                     & 0.25                     & 0.23                                         & \textbf{}             & 0.21                       & 0.23                       & 0.22                                          &                       & 0.22                       & 0.18                      & 0.20                                          &                       & 0.08                        & 0.06                        & 0.07                                           \\
GESSVDD-Sb-$\mathcal{S}$-min                   &                       & 0.12                     & 0.09                     & 0.10                                         &                       & 0.11                       & 0.12                       & 0.12                                          &                       & 0.20                       & 0.30                      & 0.25                                          &                       & 0.12                        & 0.14                        & 0.13                                           \\
GESSVDD-Sb-GR-min                  &                       & 0.15                     & 0.25                     & 0.20                                         &                       & 0.20                       & 0.18                       & 0.19                                          &                       & 0.23                       & 0.31                      & 0.27                                          &                       & 0.19                        & 0.11                        & 0.15                                           \\
GESSVDD-Sb-SR-min                  &                       & 0.18                     & 0.23                     & 0.21                                         & \textbf{}             & 0.24                       & 0.18                       & 0.21                                          &                       & 0.18                       & 0.16                      & 0.17                                          &                       & 0.08                        & 0.07                        & 0.08                                           \\
GESSVDD-Sw-$\mathcal{S}$-max                   &                       & 0.17                     & 0.17                     & 0.17                                         &                       & 0.00                       & 0.18                       & 0.09                                          &                       & 0.29                       & 0.00                      & 0.14                                          &                       & 0.16                        & 0.12                        & 0.14                                           \\
GESSVDD-Sw-GR-max                  &                       & 0.23                     & 0.12                     & 0.17                                         &                       & 0.07                       & 0.23                       & 0.15                                          &                       & 0.16                       & 0.17                      & 0.16                                          &                       & 0.10                        & 0.14                        & 0.12                                           \\
GESSVDD-Sw-SR-max                  &                       & 0.08                     & 0.05                     & 0.06                                         &                       & 0.17                       & 0.08                       & 0.12                                          &                       & 0.00                       & 0.13                      & 0.06                                          &                       & 0.12                        & 0.03                        & 0.07                                           \\
GESSVDD-Sw-$\mathcal{S}$-min                   &                       & 0.10                     & 0.14                     & 0.12                                         &                       & 0.00                       & 0.15                       & 0.07                                          &                       & 0.00                       & 0.23                      & 0.11                                          &                       & 0.10                        & 0.10                        & 0.10                                           \\
GESSVDD-Sw-GR-min                  &                       & 0.13                     & 0.13                     & 0.13                                         &                       & 0.13                       & 0.21                       & 0.17                                          &                       & 0.09                       & 0.12                      & 0.11                                          &                       & 0.05                        & 0.13                        & 0.09                                           \\
GESSVDD-Sw-SR-min                  &                       & 0.10                     & 0.10                     & 0.10                                         &                       & 0.06                       & 0.21                       & 0.14                                          &                       & 0.00                       & 0.00                      & 0.00                                          &                       & 0.06                        & 0.20                        & 0.13                                           \\
GESSVDD-kNN-$\mathcal{S}$-max                  &                       & 0.13                     & 0.10                     & 0.11                                         &                       & 0.00                       & 0.00                       & 0.00                                          &                       & 0.21                       & 0.10                      & 0.15                                          &                       & 0.09                        & 0.06                        & 0.08                                           \\
GESSVDD-kNN-GR-max                 &                       & 0.09                     & 0.20                     & 0.14                                         &                       & 0.20                       & 0.27                       & 0.23                                          &                       & 0.26                       & 0.00                      & 0.13                                          &                       & 0.08                        & 0.11                        & 0.09                                           \\
GESSVDD-kNN-SR-max                 &                       & 0.11                     & 0.07                     & 0.09                                         &                       & 0.09                       & 0.06                       & 0.08                                          &                       & 0.00                       & 0.00                      & 0.00                                          &                       & 0.11                        & 0.05                        & 0.08                                           \\
GESSVDD-kNN-$\mathcal{S}$-min                  &                       & 0.11                     & 0.10                     & 0.11                                         &                       & 0.07                       & 0.00                       & 0.03                                          &                       & 0.07                       & 0.00                      & 0.03                                          &                       & 0.10                        & 0.09                        & 0.10                                           \\
GESSVDD-kNN-GR-min                 &                       & 0.07                     & 0.07                     & 0.07                                         &                       & 0.00                       & 0.00                       & 0.00                                          &                       & 0.00                       & 0.00                      & 0.00                                          &                       & 0.07                        & 0.14                        & 0.10                                           \\
GESSVDD-kNN-SR-min                 &                       & 0.09                     & 0.12                     & 0.10                                         &                       & 0.09                       & 0.16                       & 0.12                                          &                       & 0.08                       & 0.00                      & 0.04                                          &                       & 0.07                        & 0.10                        & 0.08                                           \\
GESSVDD-PCA-$\mathcal{S}$-max                  &                       & 0.16                     & 0.10                     & 0.13                                         &                       & 0.00                       & 0.06                       & 0.03                                          & \textbf{}             & 0.00                       & 0.00                      & 0.00                                          &                       & 0.09                        & 0.21                        & 0.15                                           \\
GESSVDD-PCA-GR-max                 &                       & 0.05                     & 0.18                     & 0.12                                         &                       & 0.22                       & 0.22                       & 0.22                                          &                       & 0.29                       & 0.23                      & 0.26                                          &                       & 0.12                        & 0.14                        & 0.13                                           \\
GESSVDD-PCA-SR-max                 &                       & 0.11                     & 0.23                     & 0.17                                         &                       & 0.18                       & 0.11                       & 0.15                                          &                       & 0.08                       & 0.21                      & 0.15                                          &                       & 0.11                        & 0.05                        & 0.08                                           \\
GESSVDD-PCA-$\mathcal{S}$-min                  &                       & 0.20                     & 0.16                     & 0.18                                         &                       & 0.13                       & 0.10                       & 0.11                                          & \textbf{}             & 0.00                       & 0.00                      & 0.00                                          &                       & 0.07                        & 0.23                        & 0.15                                           \\
GESSVDD-PCA-GR-min                 &                       & 0.09                     & 0.30                     & 0.20                                         &                       & 0.12                       & 0.15                       & 0.14                                          &                       & 0.00                       & 0.08                      & 0.04                                          &                       & 0.07                        & 0.09                        & 0.08                                           \\
GESSVDD-PCA-SR-min                 &                       & 0.13                     & 0.13                     & 0.13                                         &                       & 0.10                       & 0.10                       & 0.10                                          &                       & 0.07                       & 0.00                      & 0.03                                          &                       & 0.16                        & 0.03                        & 0.09                                           \\
GESSVDD-I-$\mathcal{S}$-max                    &                       & 0.11                     & 0.09                     & 0.10                                         &                       & 0.08                       & 0.06                       & 0.07                                          & \textbf{}             & 0.09                       & 0.00                      & 0.05                                          &                       & 0.09                        & 0.13                        & 0.11                                           \\
GESSVDD-I-GR-max                   &                       & 0.12                     & 0.17                     & 0.15                                         &                       & 0.15                       & 0.23                       & 0.19                                          &                       & 0.07                       & 0.11                      & 0.09                                          &                       & 0.08                        & 0.10                        & 0.09                                           \\
GESSVDD-I-SR-max                   &                       & 0.09                     & 0.26                     & 0.18                                         &                       & 0.09                       & 0.07                       & 0.08                                          &                       & 0.00                       & 0.00                      & 0.00                                          &                       & 0.11                        & 0.04                        & 0.08                                           \\
GESSVDD-I-$\mathcal{S}$-min                    &                       & 0.07                     & 0.22                     & 0.15                                         &                       & 0.00                       & 0.22                       & 0.11                                          & \textbf{}             & 0.18                       & 0.00                      & 0.09                                          &                       & 0.14                        & 0.08                        & 0.11                                           \\
GESSVDD-I-GR-min (ESSVDD)          &                       & 0.12                     & 0.12                     & 0.12                                         &                       & 0.11                       & 0.15                       & 0.13                                          &                       & 0.00                       & 0.18                      & 0.09                                          &                       & 0.09                        & 0.05                        & 0.07                                           \\
GESSVDD-I-SR-min                   &                       & 0.09                     & 0.18                     & 0.13                                         &                       & 0.22                       & 0.17                       & 0.20                                          &                       & 0.00                       & 0.19                      & 0.09                                          &                       & 0.14                        & 0.04                        & 0.09                                           \\
GESSVDD-0-$\mathcal{S}$-max                    &                       & 0.11                     & 0.13                     & 0.12                                         &                       & 0.10                       & 0.22                       & 0.16                                          &                       & 0.29                       & 0.24                      & 0.26                                          &                       & 0.09                        & 0.17                        & 0.13                                           \\
GESSVDD-0-GR-max                   &                       & 0.10                     & 0.14                     & 0.12                                         &                       & 0.13                       & 0.13                       & 0.13                                          &                       & 0.14                       & 0.21                      & 0.17                                          &                       & 0.11                        & 0.20                        & 0.16                                           \\
GESSVDD-0-$\mathcal{S}$-min                    &                       & 0.10                     & 0.21                     & 0.15                                         &                       & 0.18                       & 0.16                       & 0.17                                          &                       & 0.15                       & 0.13                      & 0.14                                          &                       & 0.20                        & 0.17                        & 0.19                                           \\
GESSVDD-0-GR-min (SSVDD)           &                       & 0.18                     & 0.09                     & 0.14                                         &                       & 0.19                       & 0.26                       & 0.22                                          &                       & 0.00                       & 0.20                      & 0.10                                          &                       & 0.21                        & 0.06                        & 0.14                                           \\
ESVDD                              &                       & 0.04                     & 0.07                     & 0.05                                         &                       & 0.00                       & 0.00                       & 0.00                                          &                       & 0.00                       & 0.07                      & 0.03                                          &                       & 0.12                        & 0.04                        & 0.08                                           \\
SVDD                               &                       & 0.07                     & 0.04                     & 0.06                                         &                       & 0.15                       & 0.21                       & 0.18                                          &                       & 0.35                       & 0.10                      & 0.22                                          &                       & 0.06                        & 0.18                        & 0.12                                           \\
OCSVM                              &                       & 0.19                     & 0.16                     & 0.18                                         &                       & 0.00                       & 0.00                       & 0.00                                          &                       & 0.00                       & 0.16                      & 0.08                                          &                       & 0.08                        & 0.14                        & 0.11                                           \\
GESVDD-PCA                         &                       & 0.03                     & 0.07                     & 0.05                                         &                       & 0.00                       & 0.00                       & 0.00                                          &                       & 0.00                       & 0.00                      & 0.00                                          &                       & 0.06                        & 0.07                        & 0.06                                           \\
GESVDD-Sw                          &                       & 0.03                     & 0.03                     & 0.03                                         &                       & 0.00                       & 0.00                       & 0.00                                          &                       & 0.00                       & 0.00                      & 0.00                                          &                       & 0.07                        & 0.07                        & 0.07                                           \\
GESVDD-kNN                         &                       & 0.03                     & 0.05                     & 0.04                                         &                       & 0.00                       & 0.00                       & 0.00                                          &                       & 0.00                       & 0.00                      & 0.00                                          &                       & 0.03                        & 0.08                        & 0.05                                           \\
GESVM-PCA                          &                       & 0.02                     & 0.05                     & 0.04                                         &                       & 0.00                       & 0.00                       & 0.00                                          &                       & 0.00                       & 0.00                      & 0.00                                          &                       & 0.04                        & 0.07                        & 0.06                                           \\
GESVM-Sw                           &                       & 0.02                     & 0.06                     & 0.04                                         &                       & 0.00                       & 0.00                       & 0.00                                          &                       & 0.00                       & 0.12                      & 0.06                                          &                       & 0.07                        & 0.06                        & 0.07                                           \\
GESVM-kNN                          &                       & 0.03                     & 0.04                     & 0.03                                         &                       & 0.00                       & 0.00                       & 0.00                                          &                       & 0.00                       & 0.07                      & 0.03                                          &                       & 0.06                        & 0.07                        & 0.07                                          
\end{tabular}
\end{table*}
\clearpage

\section{Sensitivity analysis}\label{sensitivityanalysis}
\label{sensitivityAnslysis}
\begin{figure}[h!]
	\centering
	\includegraphics[scale=0.38]{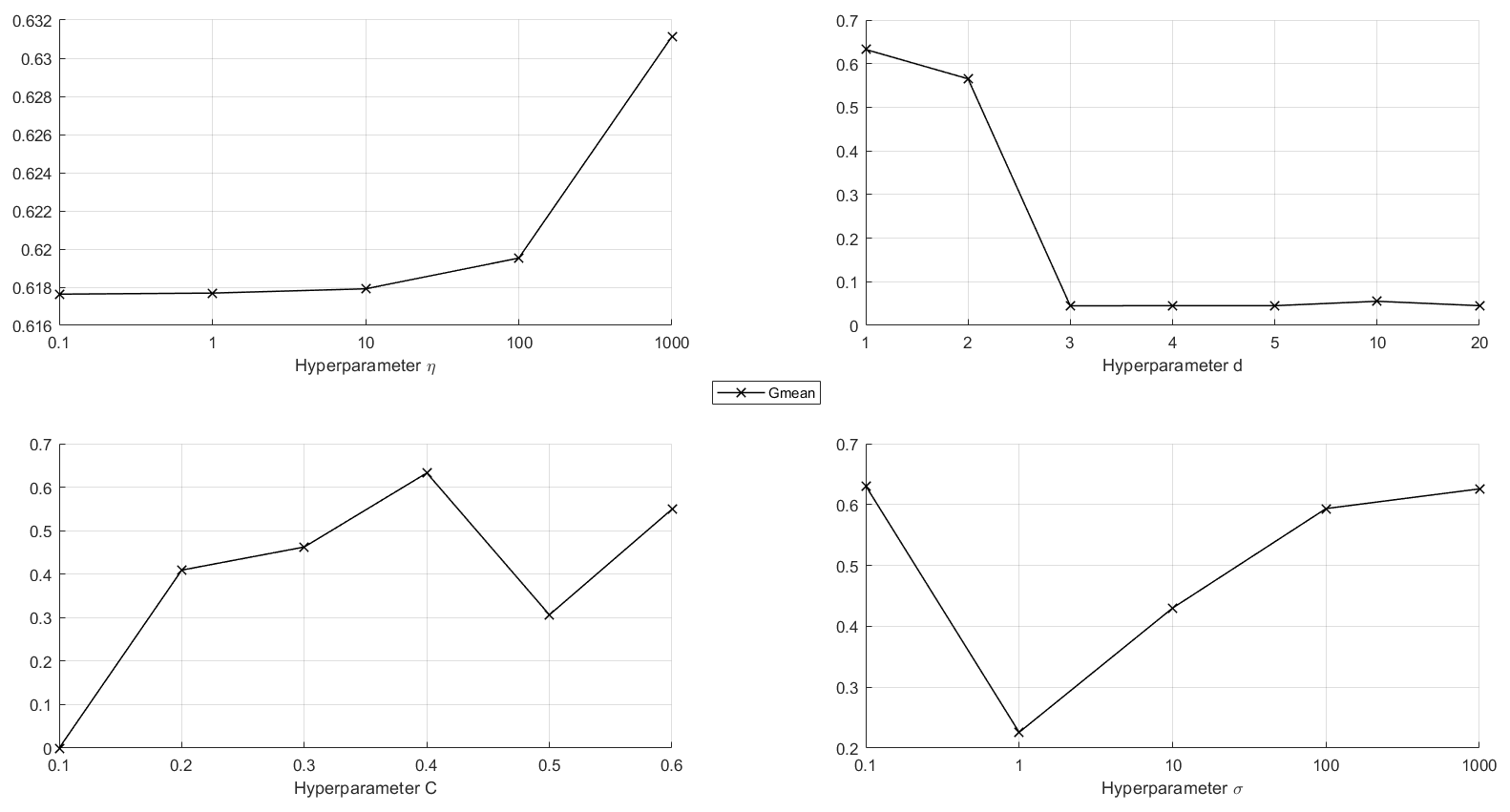}
	\caption{Hyperparameters sensitivity analysis for GESSVDD-Sb-GR-min on MNIST dataset (target class=0)}
\end{figure}

\begin{figure}
	\centering
	\includegraphics[scale=0.38]{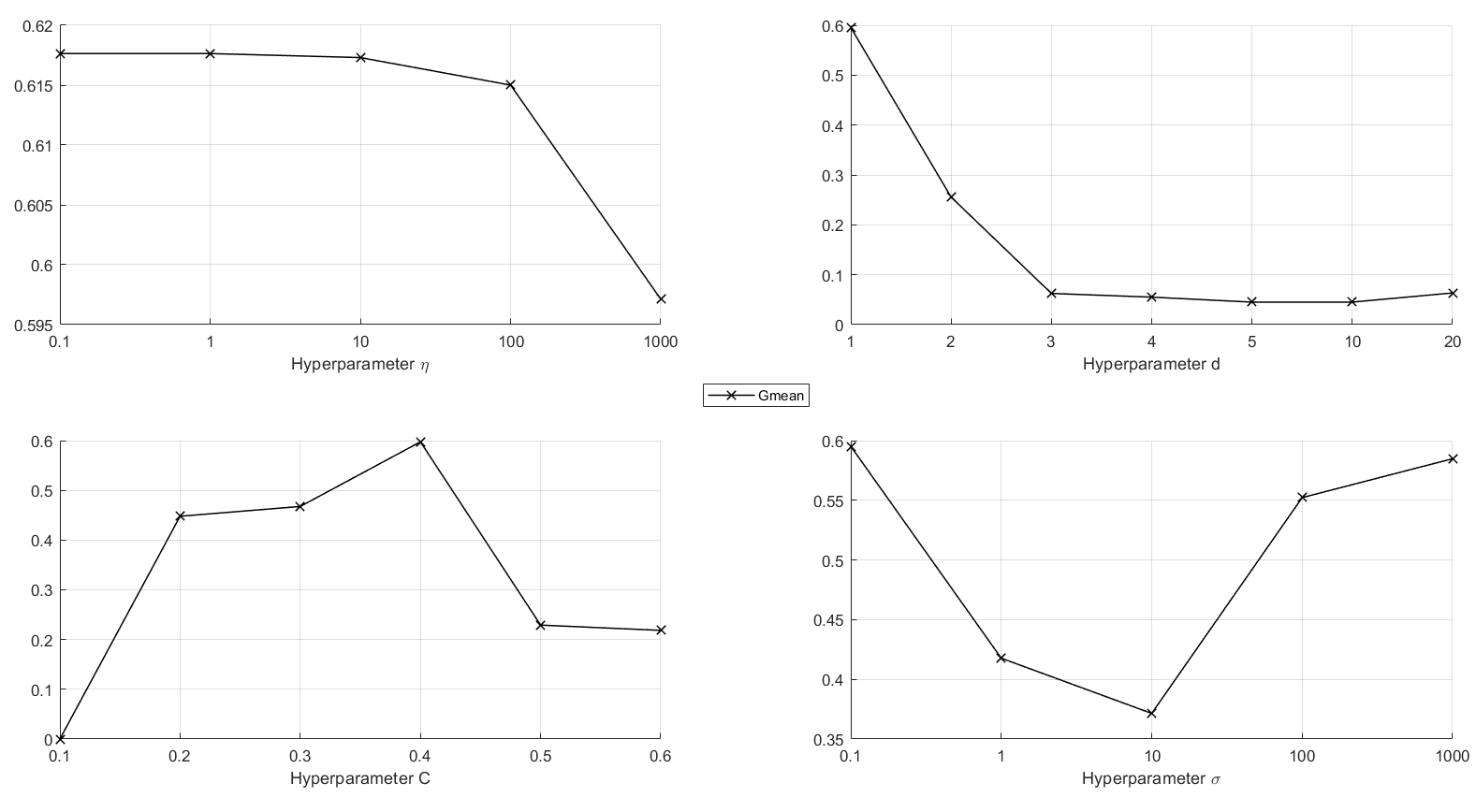}
	\caption{Hyperparameters sensitivity analysis for GESSVDD-Sb-GR-max on MNIST dataset (target class=0) }
\end{figure}

\begin{figure}
	\centering
	\includegraphics[scale=0.38]{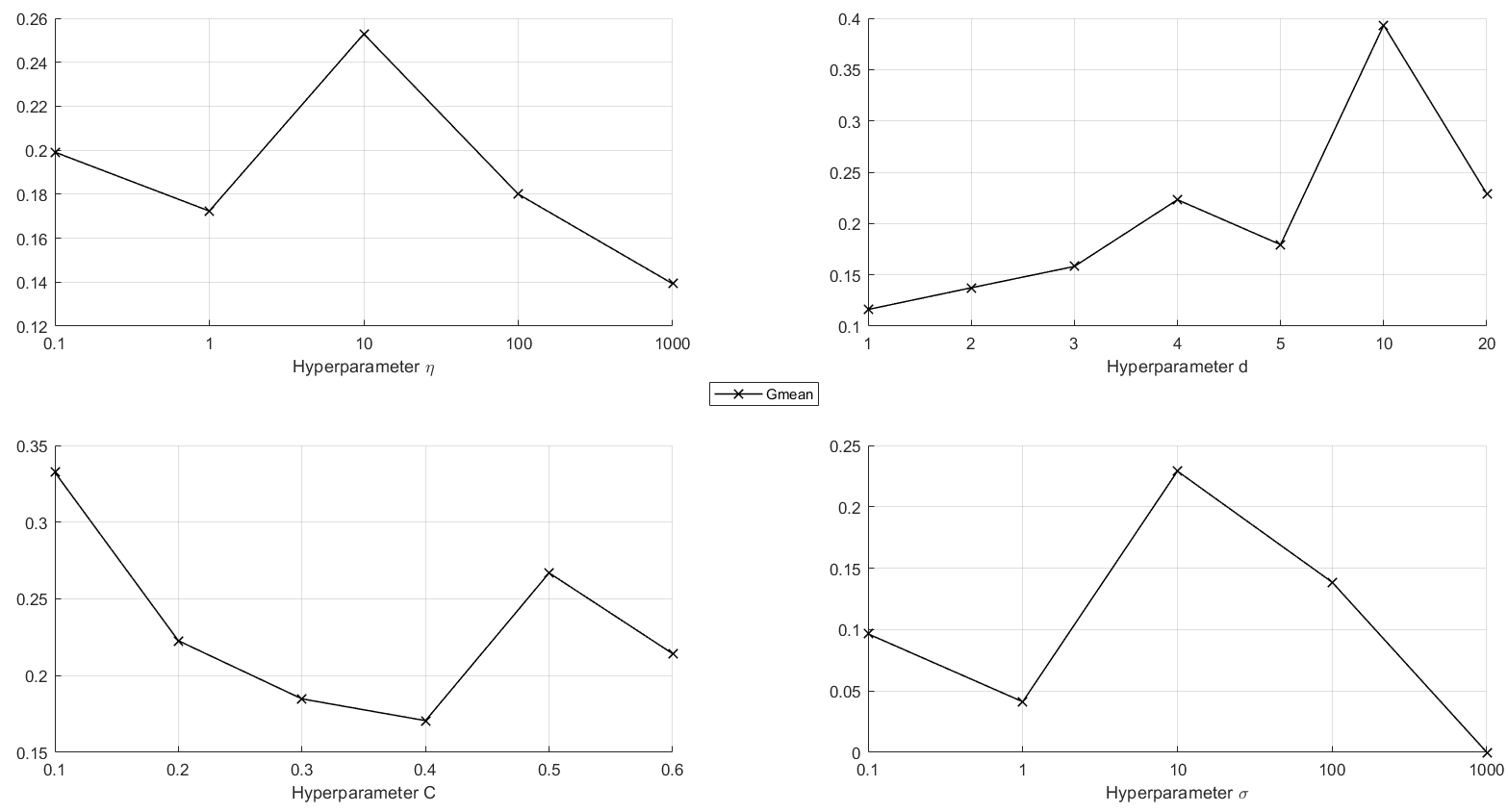}
	\caption{Hyperparameters sensitivity analysis for GESSVDD-Sb-$\mathcal{S}$-min on MNIST dataset (target class=0)}
\end{figure}

\begin{figure}
	\centering
	\includegraphics[scale=0.38]{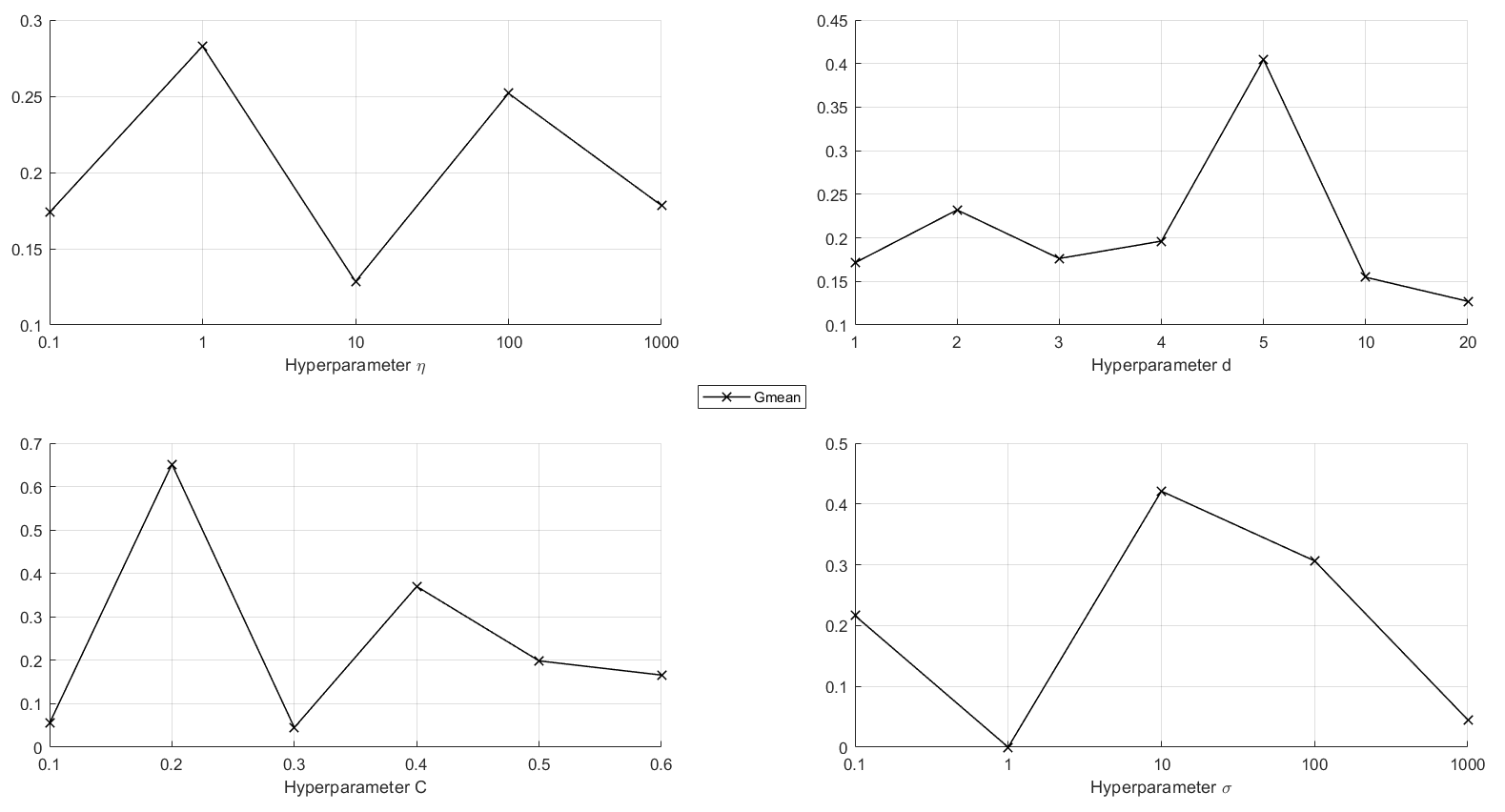}
	\caption{Hyperparameters sensitivity analysis for GESSVDD-Sb-$\mathcal{S}$-max on MNIST dataset (target class=0)}
\end{figure}

\begin{figure}[h]
	\centering
	\includegraphics[scale=0.38]{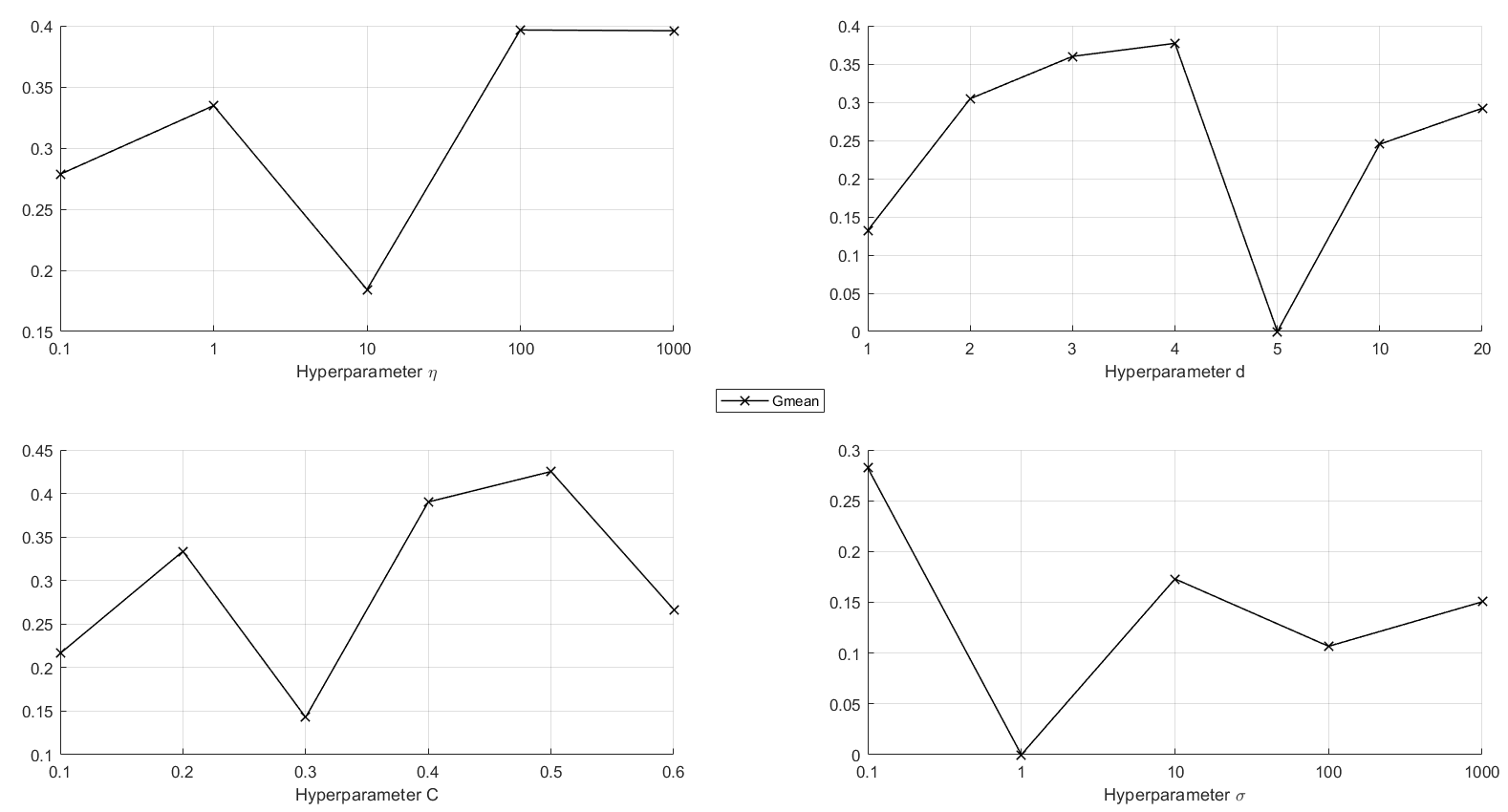}
	\caption{Hyperparameters sensitivity analysis for GESSVDD-Sb-SR-min on MNIST dataset (target class=0)}
\end{figure}

\begin{figure}
	\centering
	\includegraphics[scale=0.38]{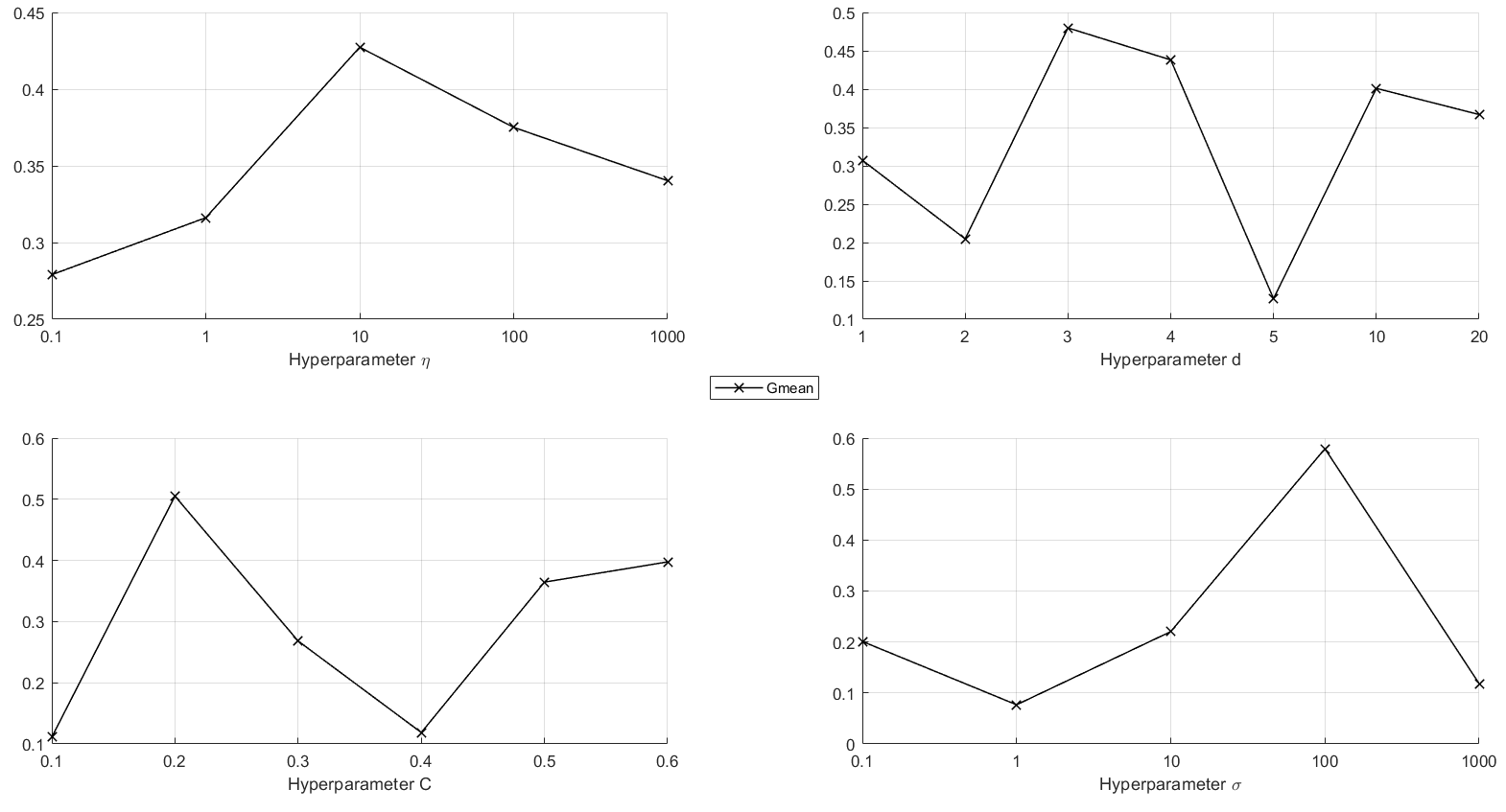}
	\caption{Hyperparameters sensitivity analysis for GESSVDD-Sb-SR-max on MNIST dataset (target class=0)}
\end{figure}

\begin{figure}
	\centering
	\includegraphics[scale=0.38]{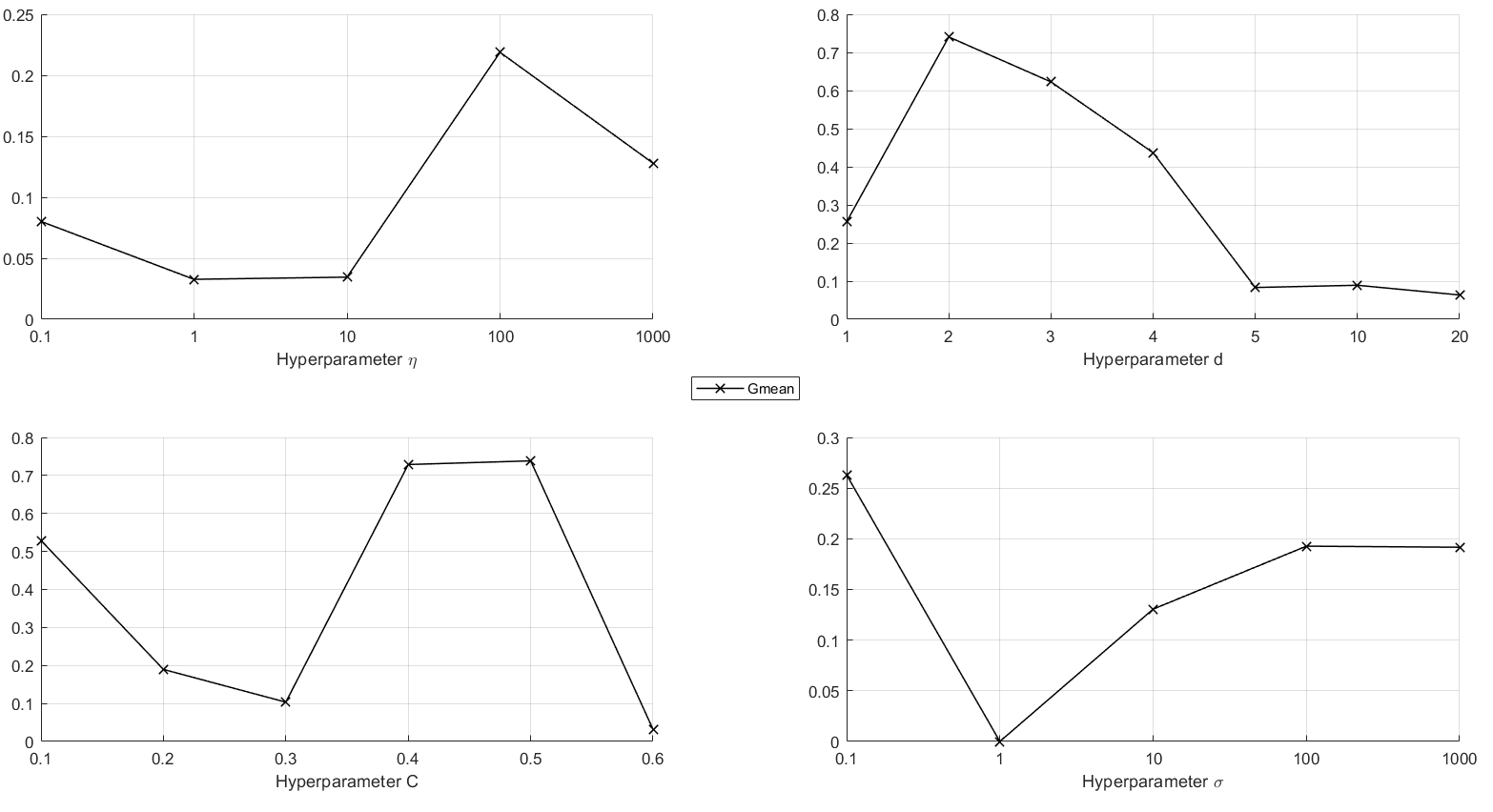}
	\caption{Hyperparameters sensitivity analysis for GESSVDD-Sw-GR-min on MNIST dataset (target class=0)}
\end{figure}

\begin{figure}
	\centering
	\includegraphics[scale=0.38]{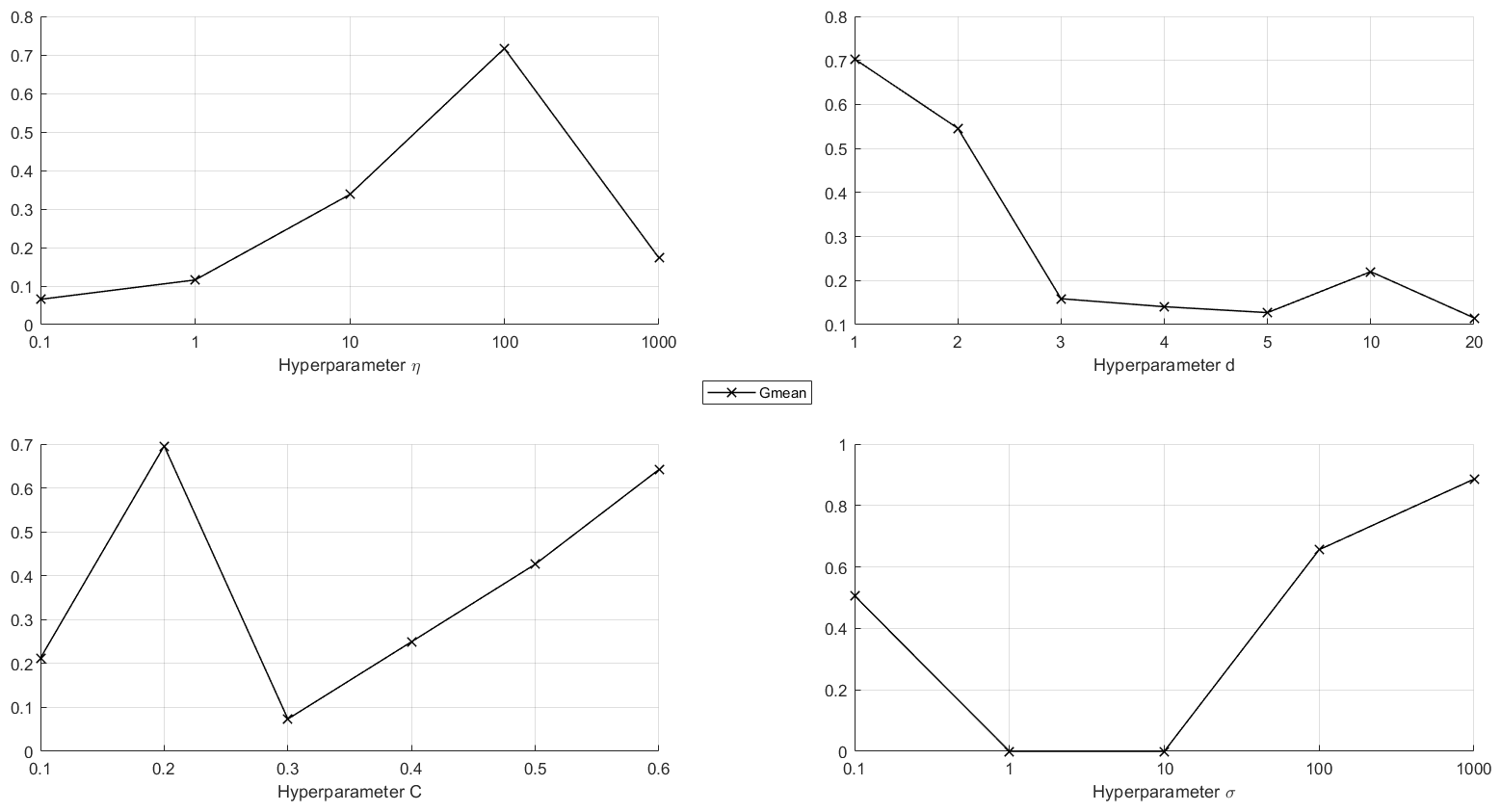}
	\caption{Hyperparameters sensitivity analysis for GESSVDD-Sw-GR-max on MNIST dataset (target class=0)}
\end{figure}

\begin{figure}
	\centering
	\includegraphics[scale=0.38]{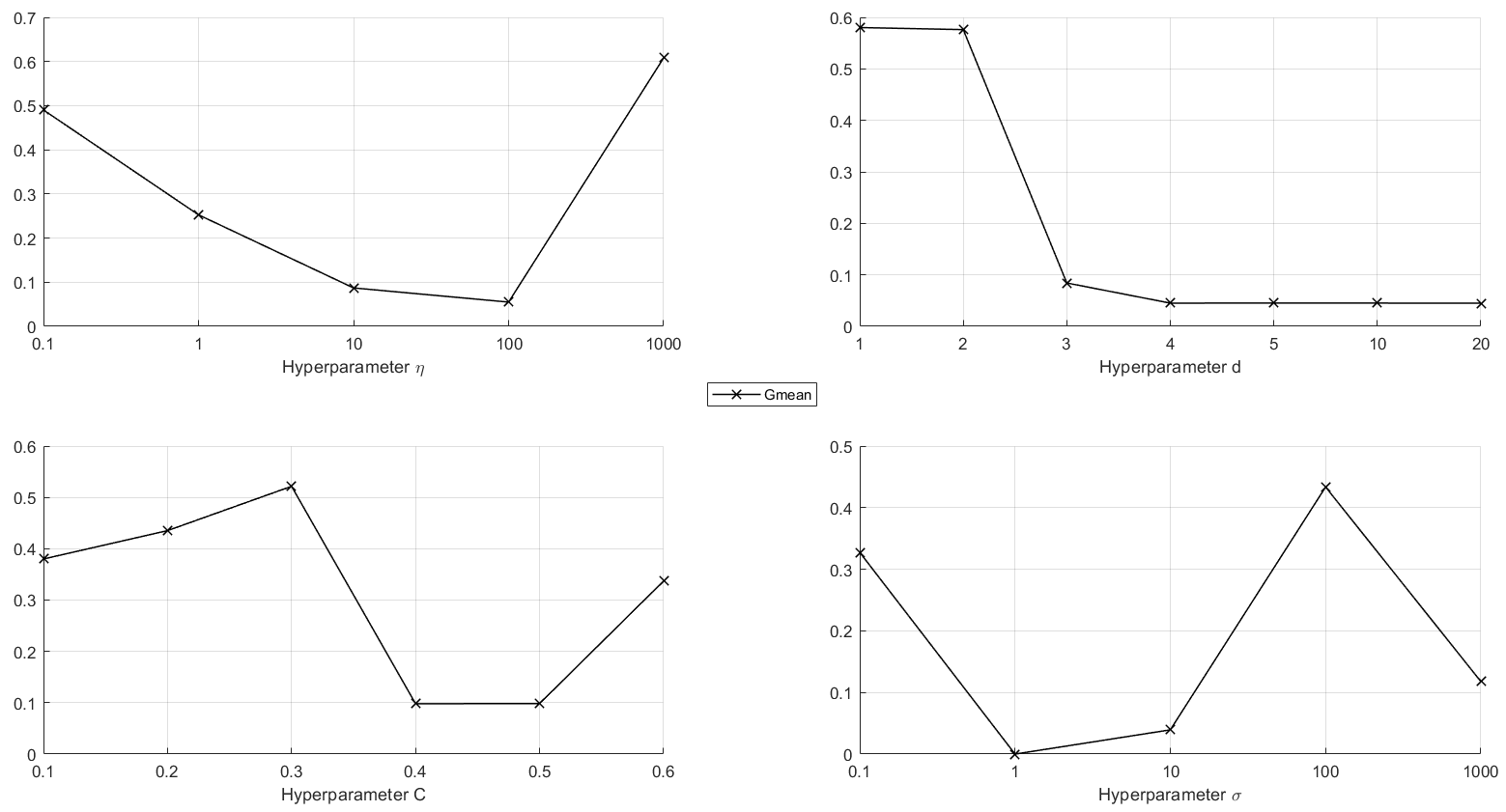}
	\caption{Hyperparameters sensitivity analysis for GESSVDD-Sw-$\mathcal{S}$-min on MNIST dataset (target class=0)}
\end{figure}

\begin{figure}
	\centering
	\includegraphics[scale=0.38]{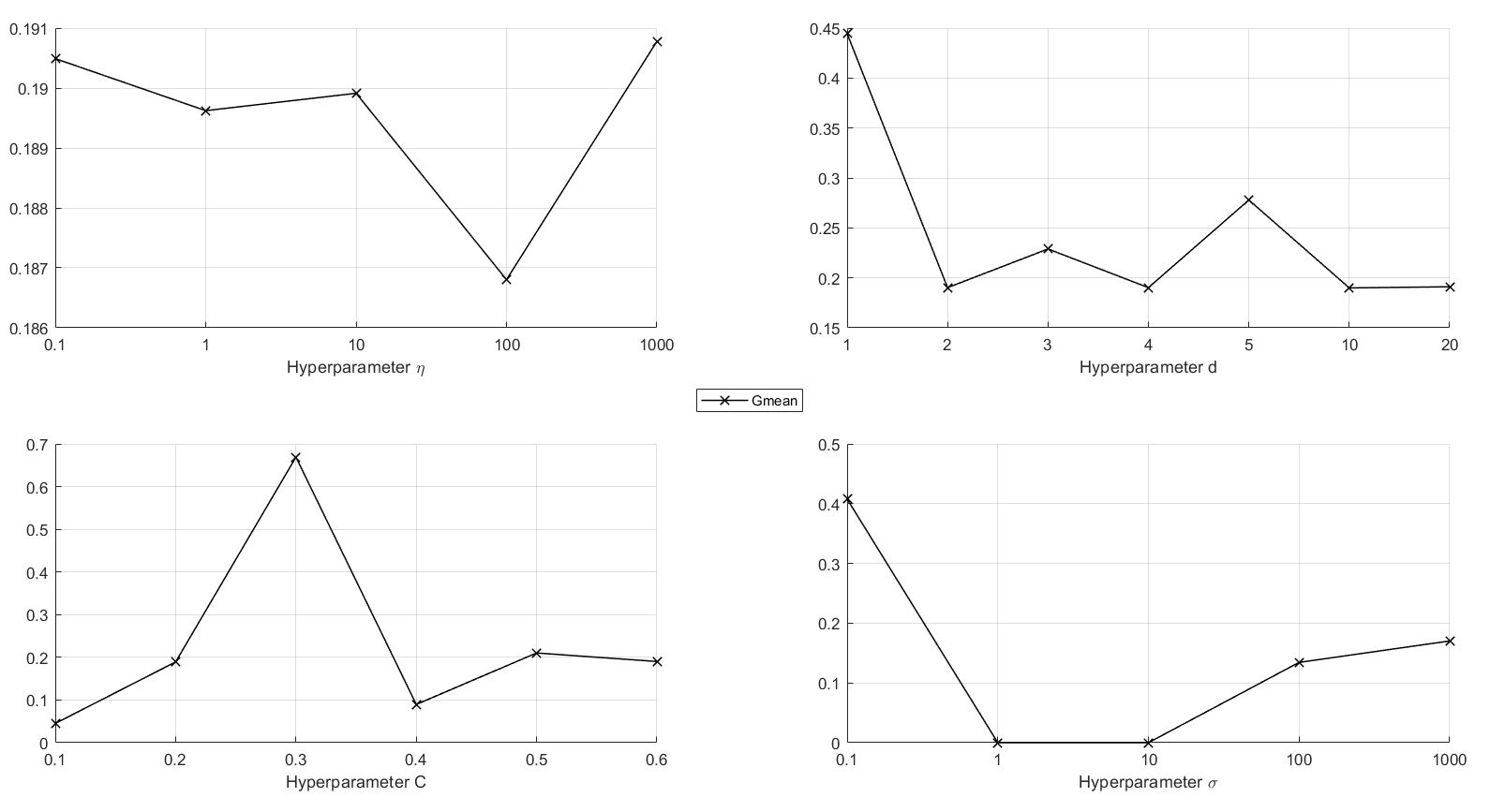}
	\caption{Hyperparameters sensitivity analysis for GESSVDD-Sw-$\mathcal{S}$-max on MNIST dataset (target class=0)}
\end{figure}

\begin{figure}
	\centering
	\includegraphics[scale=0.38]{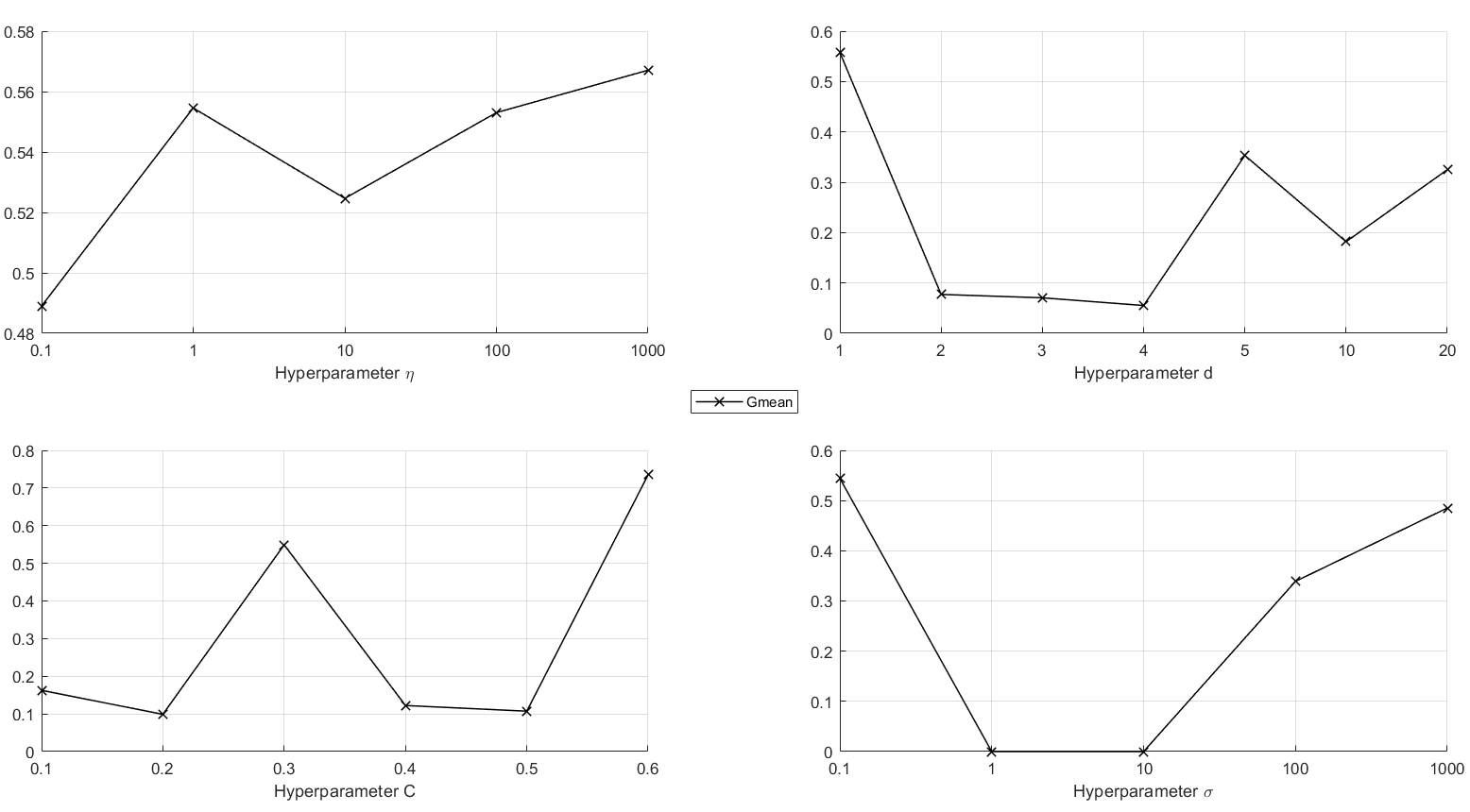}
	\caption{Hyperparameters sensitivity analysis for GESSVDD-Sw-SR-min on MNIST dataset (target class=0)}
\end{figure}

\begin{figure}
	\centering
	\includegraphics[scale=0.38]{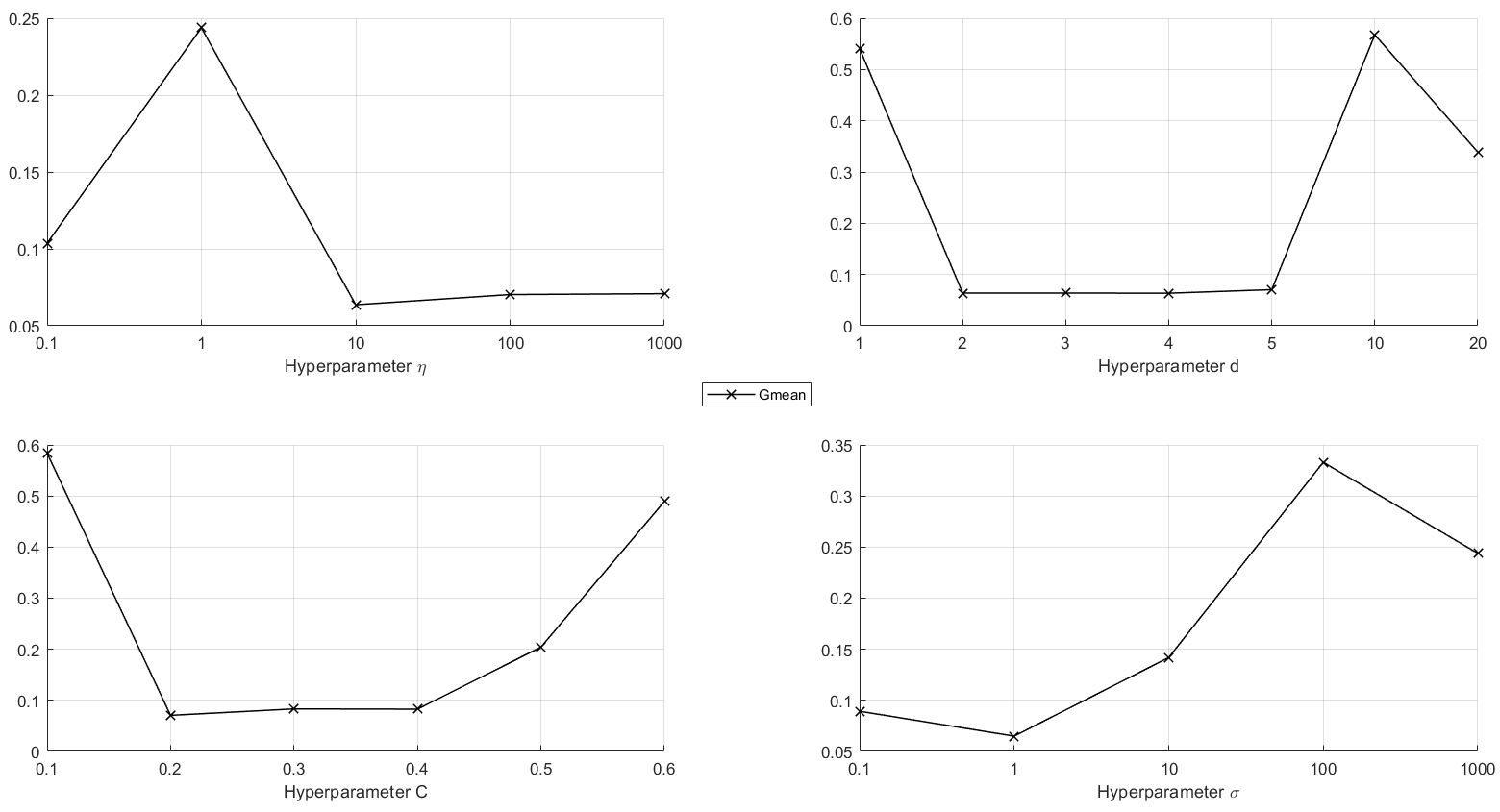}
	\caption{Hyperparameters sensitivity analysis for GESSVDD-Sw-SR-max on MNIST dataset (target class=0)}
\end{figure}

\begin{figure}
	\centering
	\includegraphics[scale=0.38]{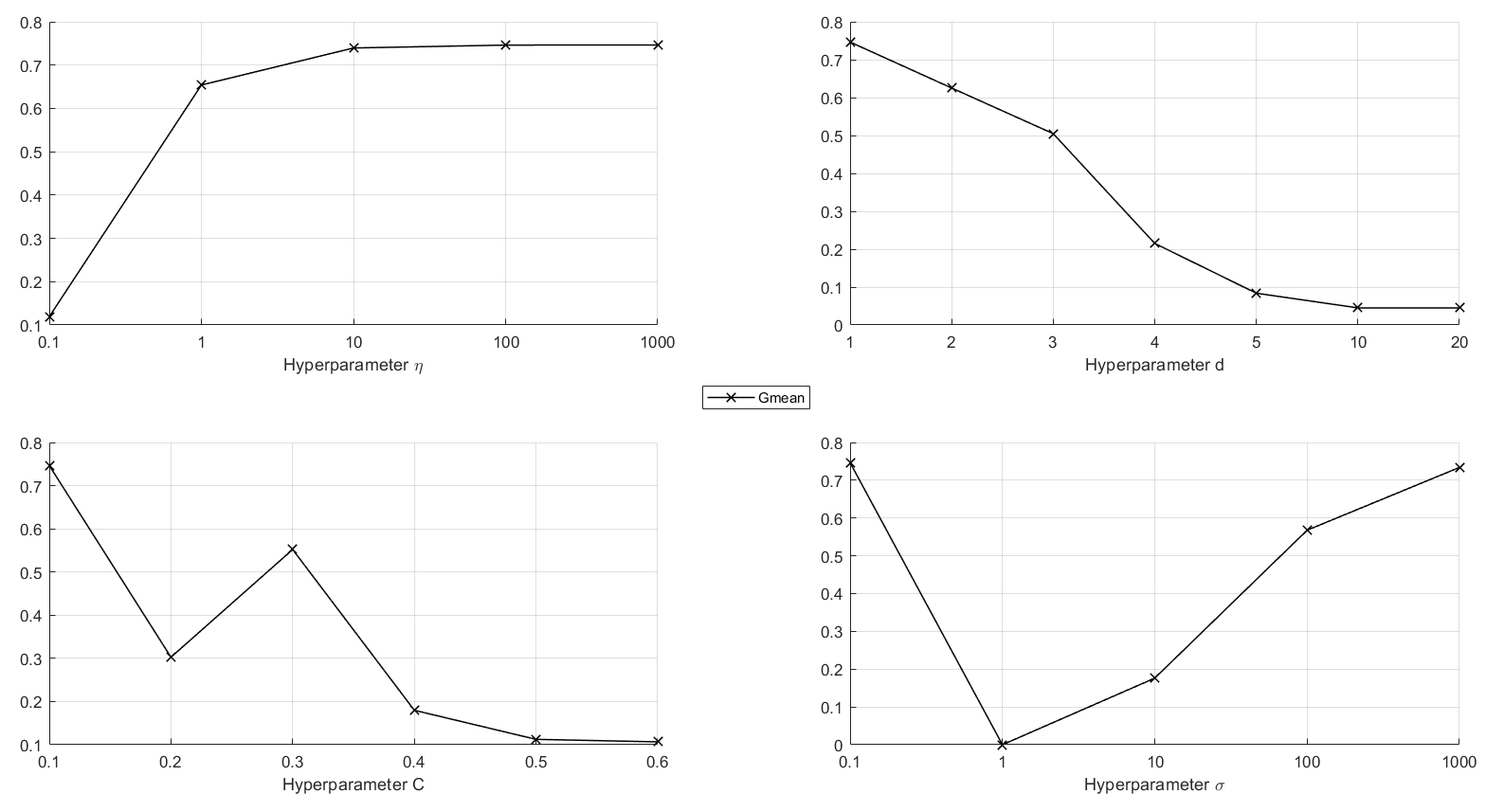}
	\caption{Hyperparameters sensitivity analysis for GESSVDD-PCA-GR-min}
\end{figure}
\begin{figure}
	\centering
	\includegraphics[scale=0.38]{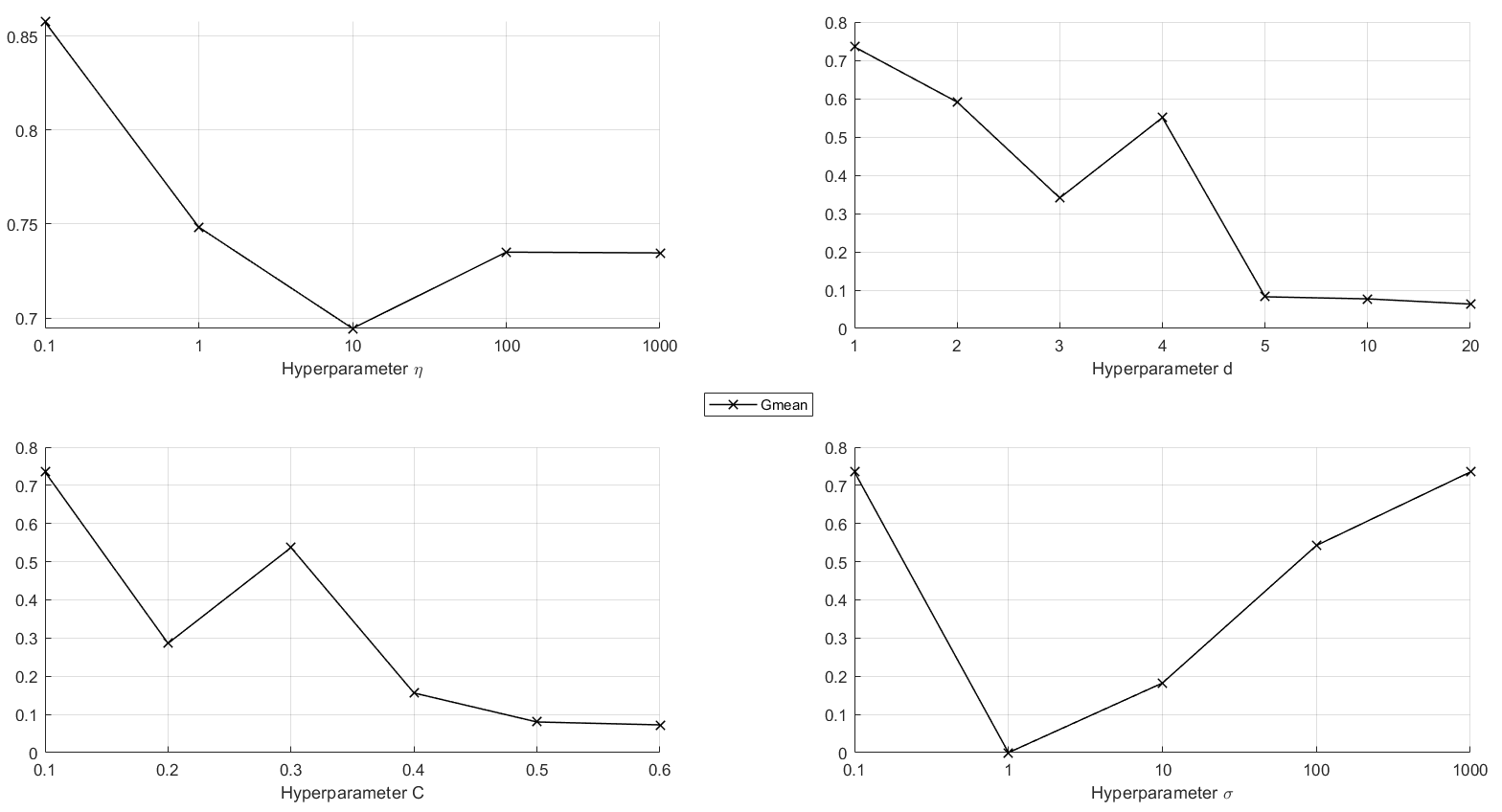}
	\caption{Hyperparameters sensitivity analysis for GESSVDD-PCA-GR-max on MNIST dataset (target class=0)}
\end{figure}

\begin{figure}
	\centering
	\includegraphics[scale=0.38]{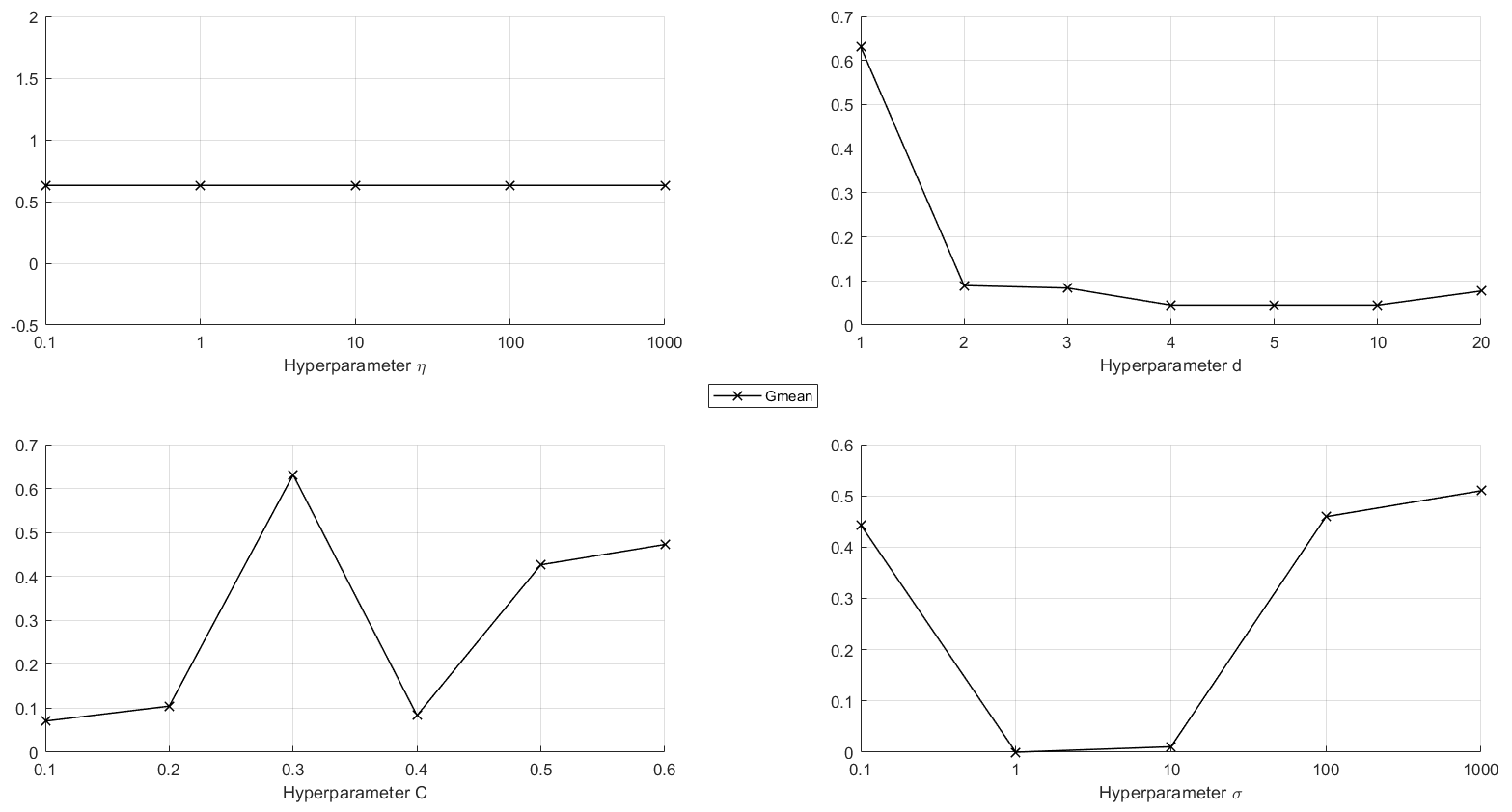}
	\caption{Hyperparameters sensitivity analysis for GESSVDD-PCA-$\mathcal{S}$-min on MNIST dataset (target class=0)}
\end{figure}

\begin{figure}
	\centering
	\includegraphics[scale=0.38]{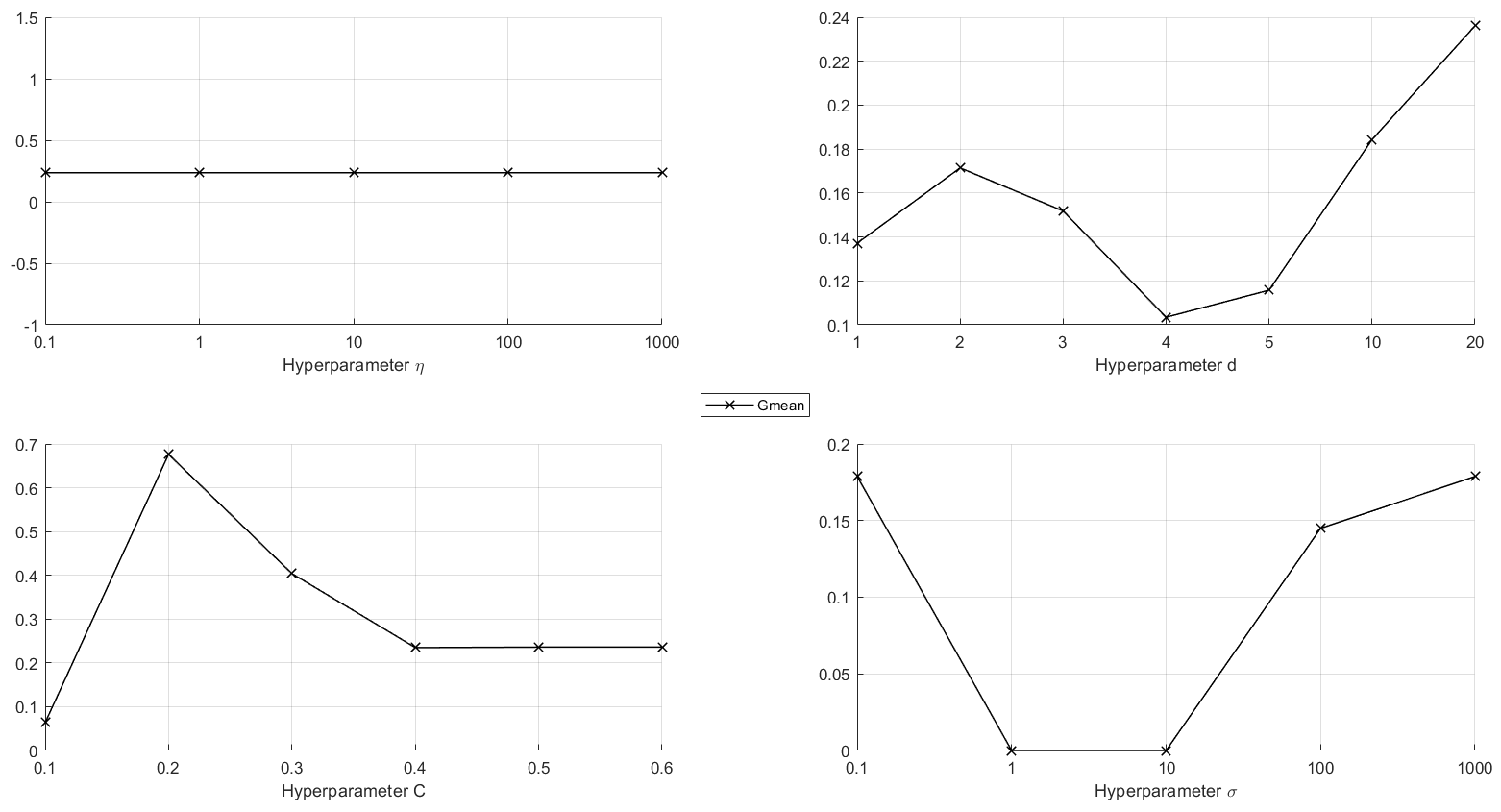}
	\caption{Hyperparameters sensitivity analysis for GESSVDD-PCA-$\mathcal{S}$-max on MNIST dataset (target class=0)}
\end{figure}

\begin{figure}
	\centering
	\includegraphics[scale=0.38]{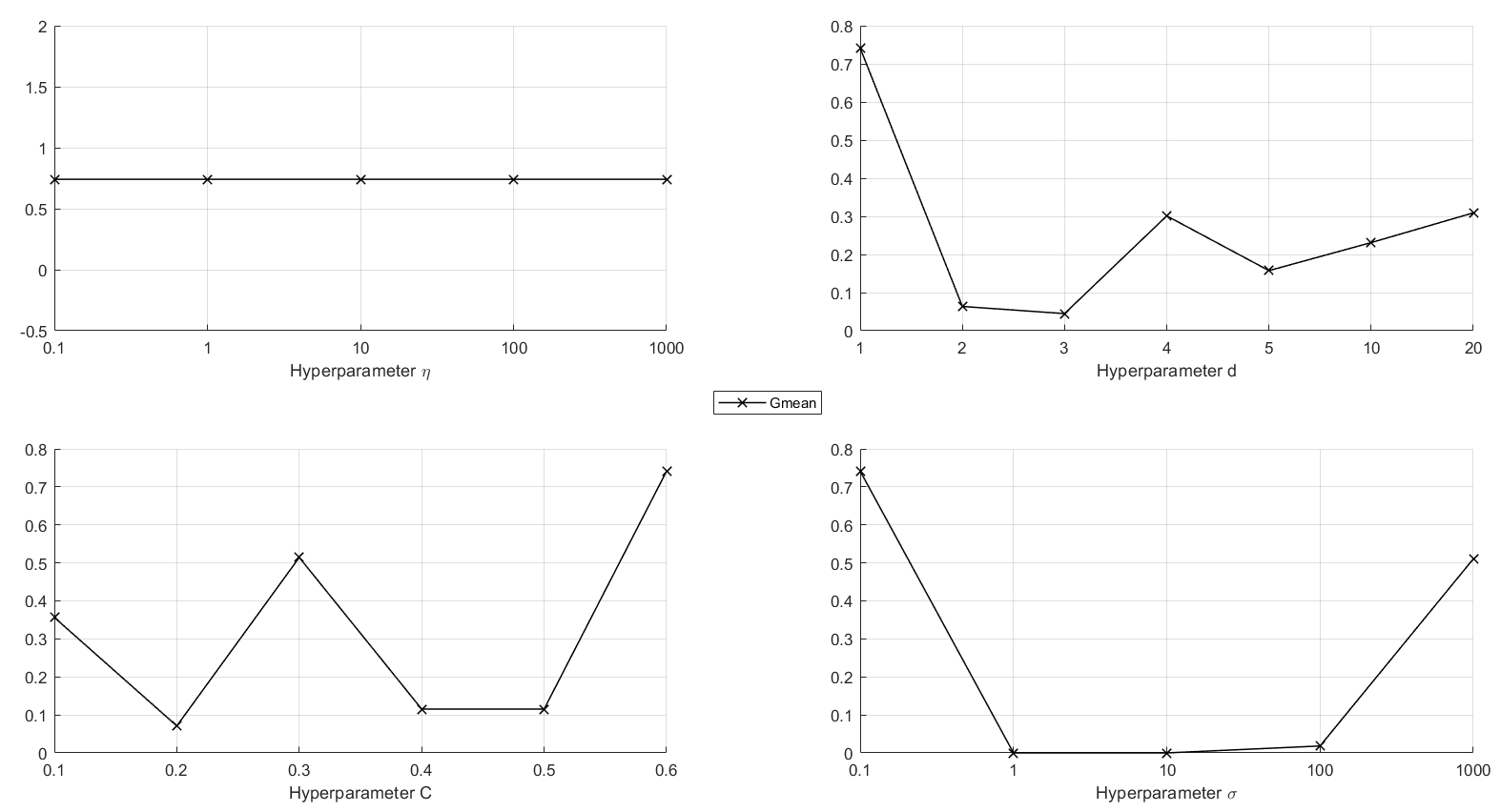}
	\caption{Hyperparameters sensitivity analysis for GESSVDD-SR-PCA-min on MNIST dataset (target class=0)}
\end{figure}
\begin{figure}
	\centering
	\includegraphics[scale=0.38]{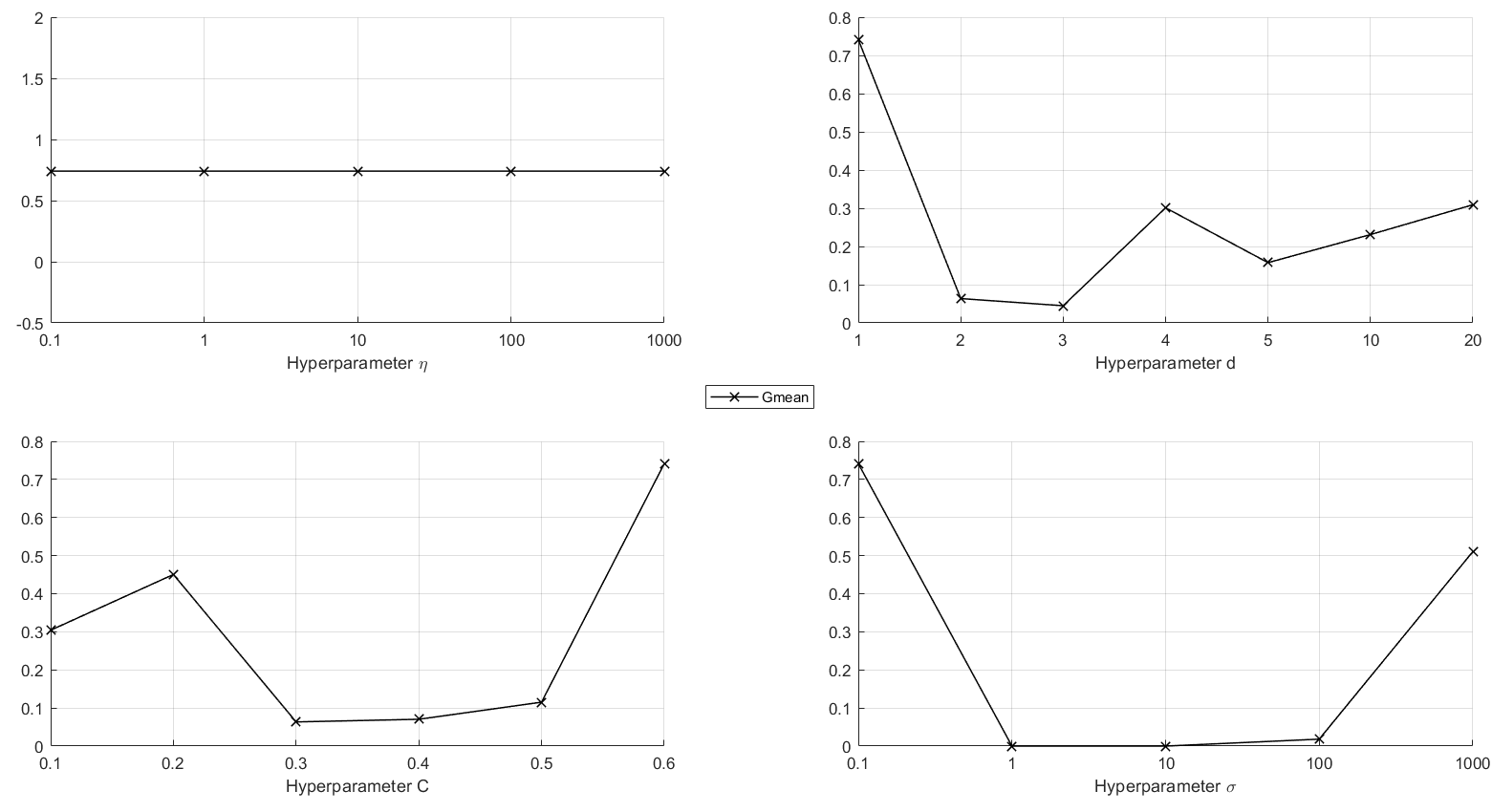}
	\caption{Hyperparameters sensitivity analysis for GESSVDD-SR-PCA-max on MNIST dataset (target class=0)}
\end{figure}
\begin{figure}
	\centering
	\includegraphics[scale=0.38]{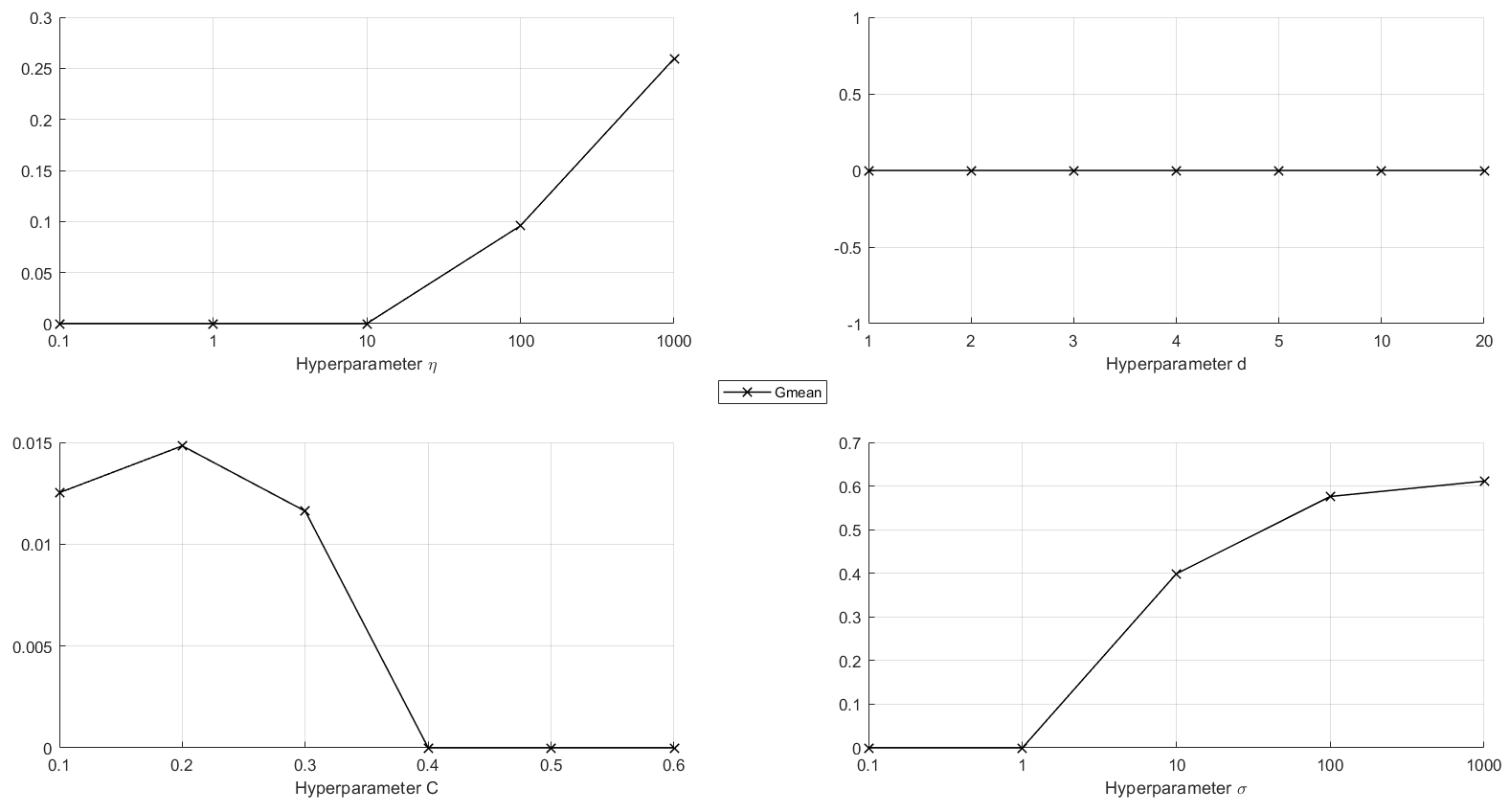}
	\caption{Hyperparameters sensitivity analysis for GESSVDD-GR-kNN-min on MNIST dataset (target class=0)}
\end{figure}
\begin{figure}
	\centering
	\includegraphics[scale=0.38]{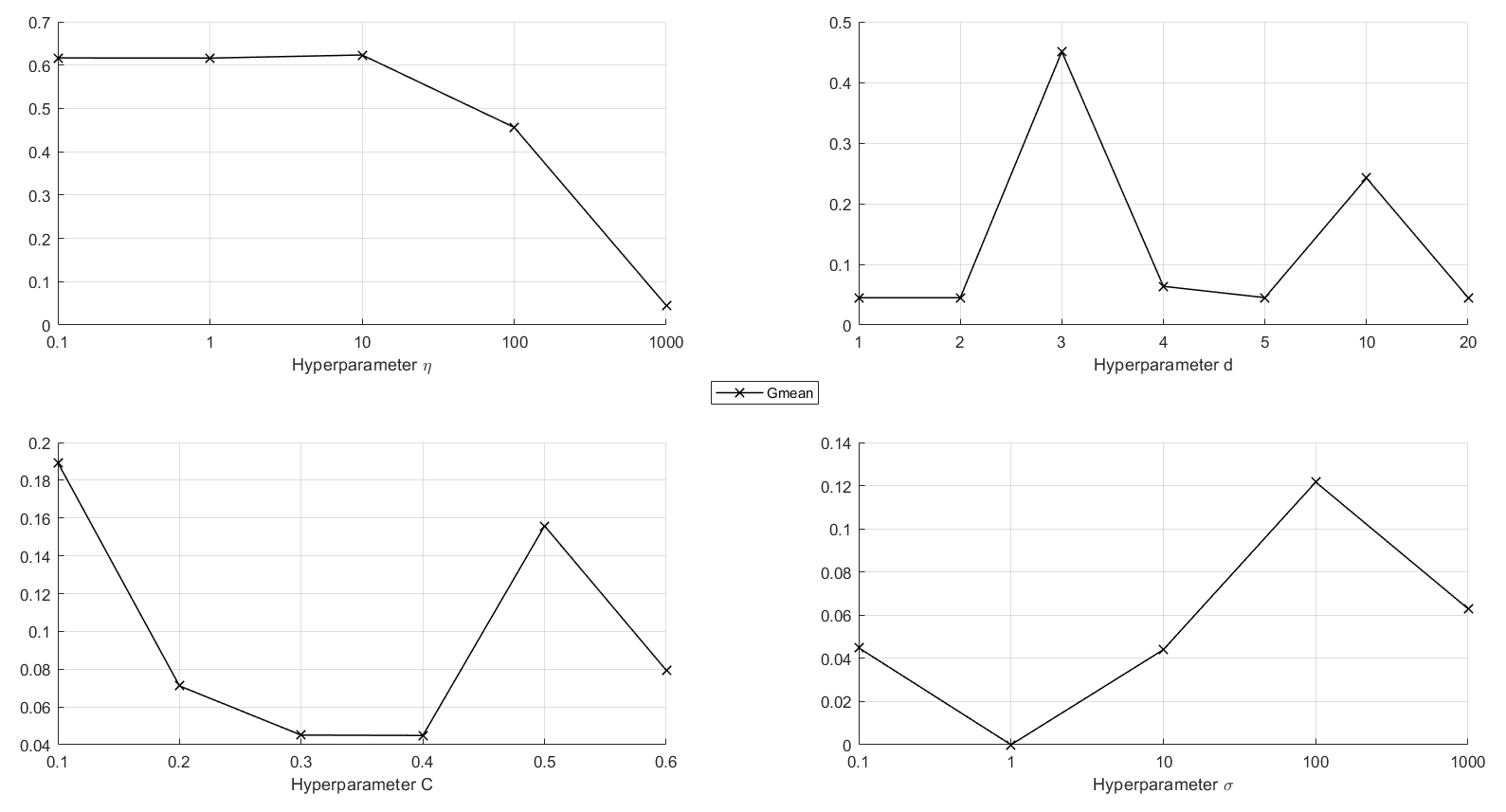}
	\caption{Hyperparameters sensitivity analysis for GESSVDD-GR-max on MNIST dataset (target class=0)}
\end{figure}
\begin{figure}
	\centering
	\includegraphics[scale=0.38]{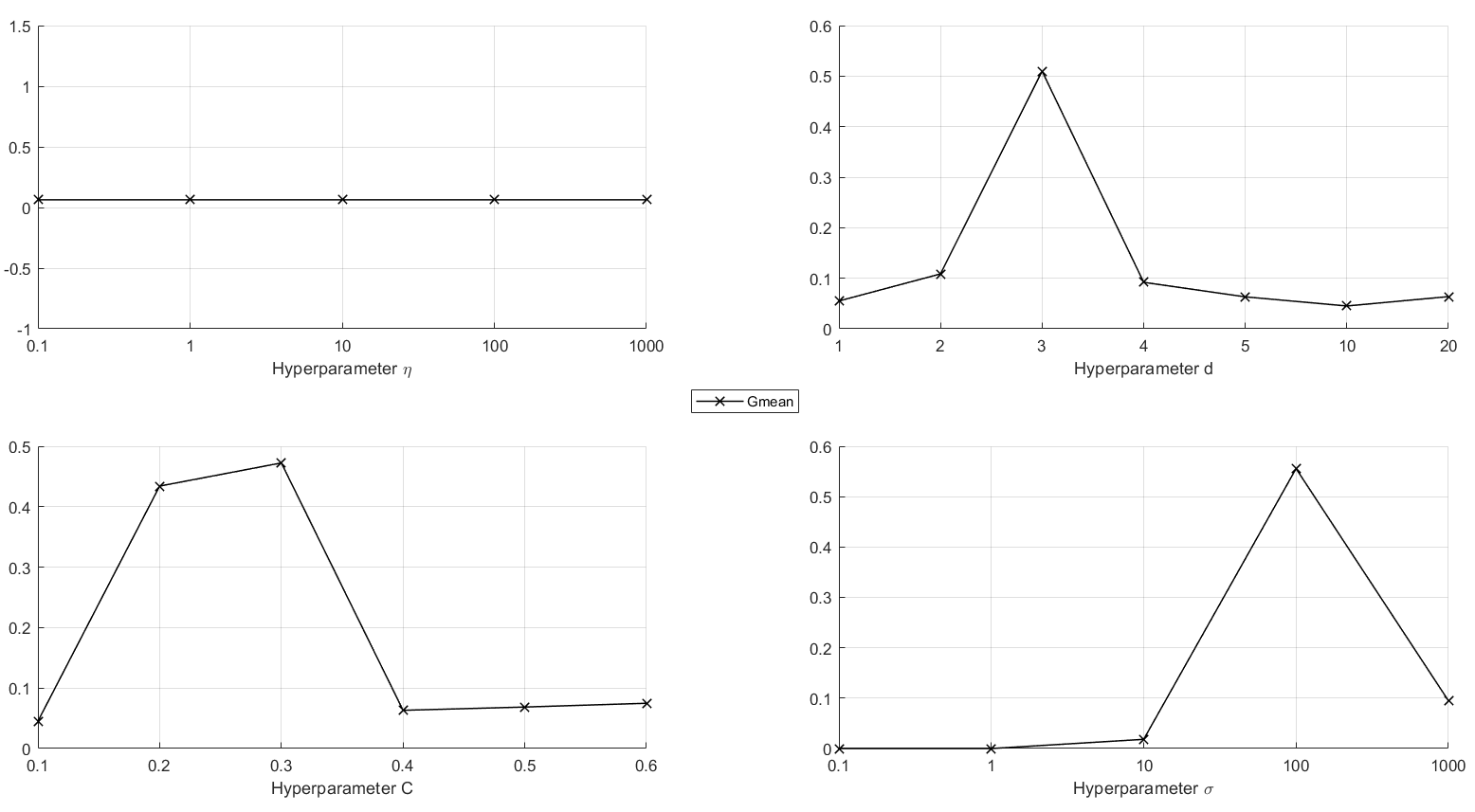}
	\caption{Hyperparameters sensitivity analysis for GESSVDD-$\mathcal{S}$-kNN-min on MNIST dataset (target class=0)}
\end{figure}
\begin{figure}
	\centering
	\includegraphics[scale=0.38]{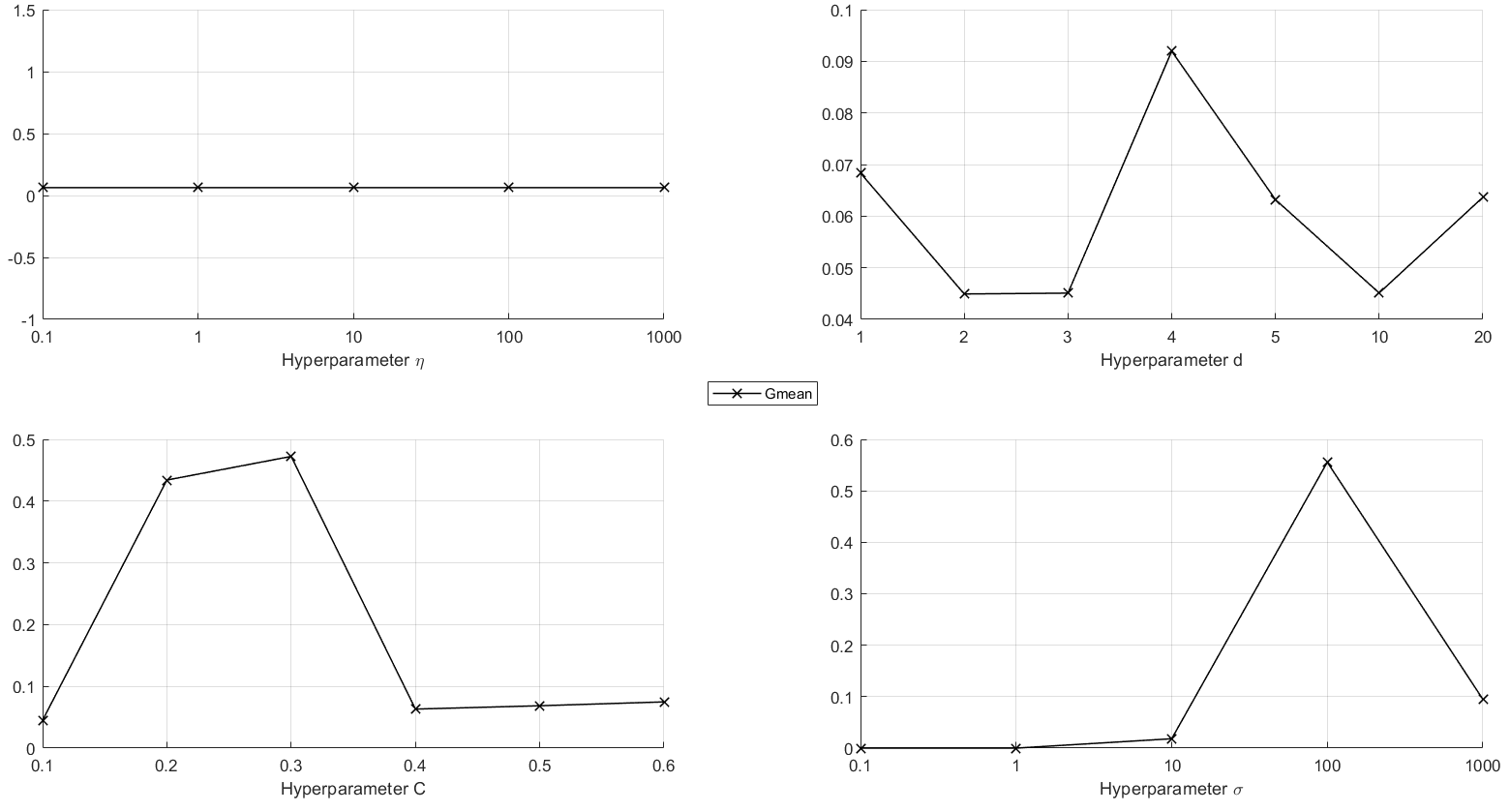}
	\caption{Hyperparameters sensitivity analysis for GESSVDD-$\mathcal{S}$-kNN-max on MNIST dataset (target class=0)}
\end{figure}
\begin{figure}
	\centering
	\includegraphics[scale=0.38]{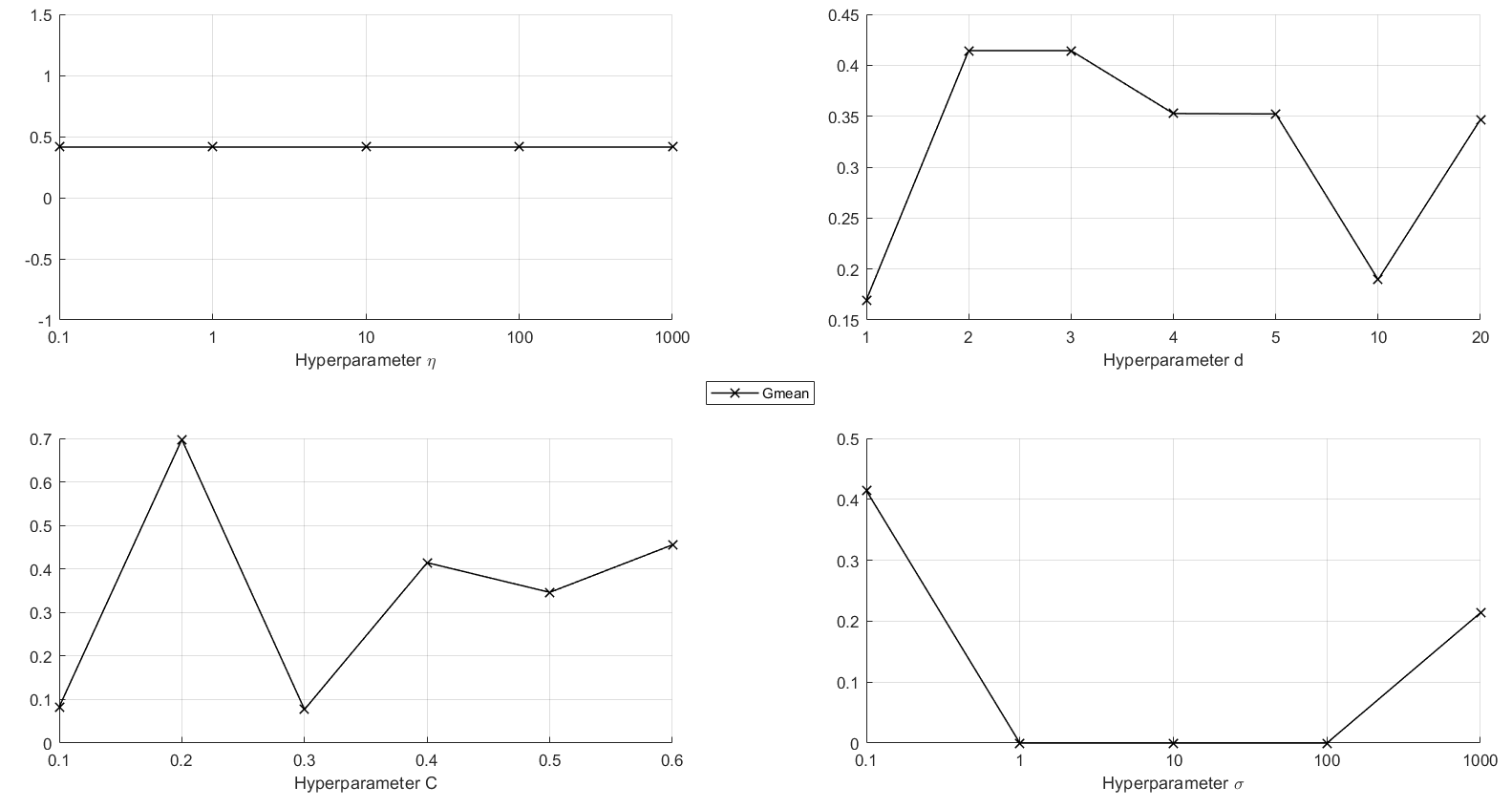}
	\caption{Hyperparameters sensitivity analysis for GESSVDD-SR-kNN-min on MNIST dataset (target class=0)}
\end{figure}
\begin{figure}
	\centering
	\includegraphics[scale=0.38]{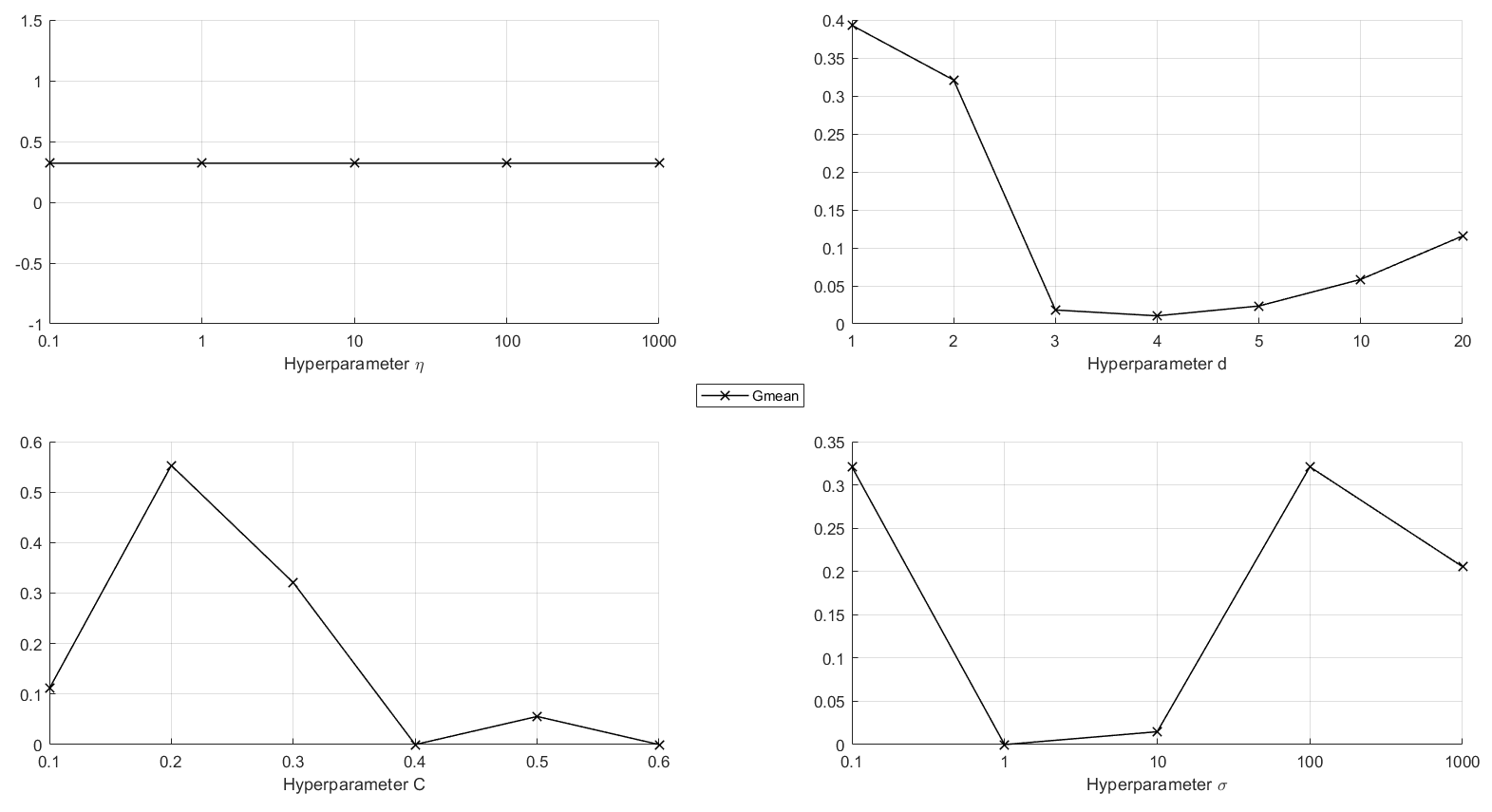}
	\caption{Hyperparameters sensitivity analysis for GESSVDD-SR-kNN-max on MNIST dataset (target class=0)}
\end{figure}
\begin{figure}
	\centering
	\includegraphics[scale=0.38]{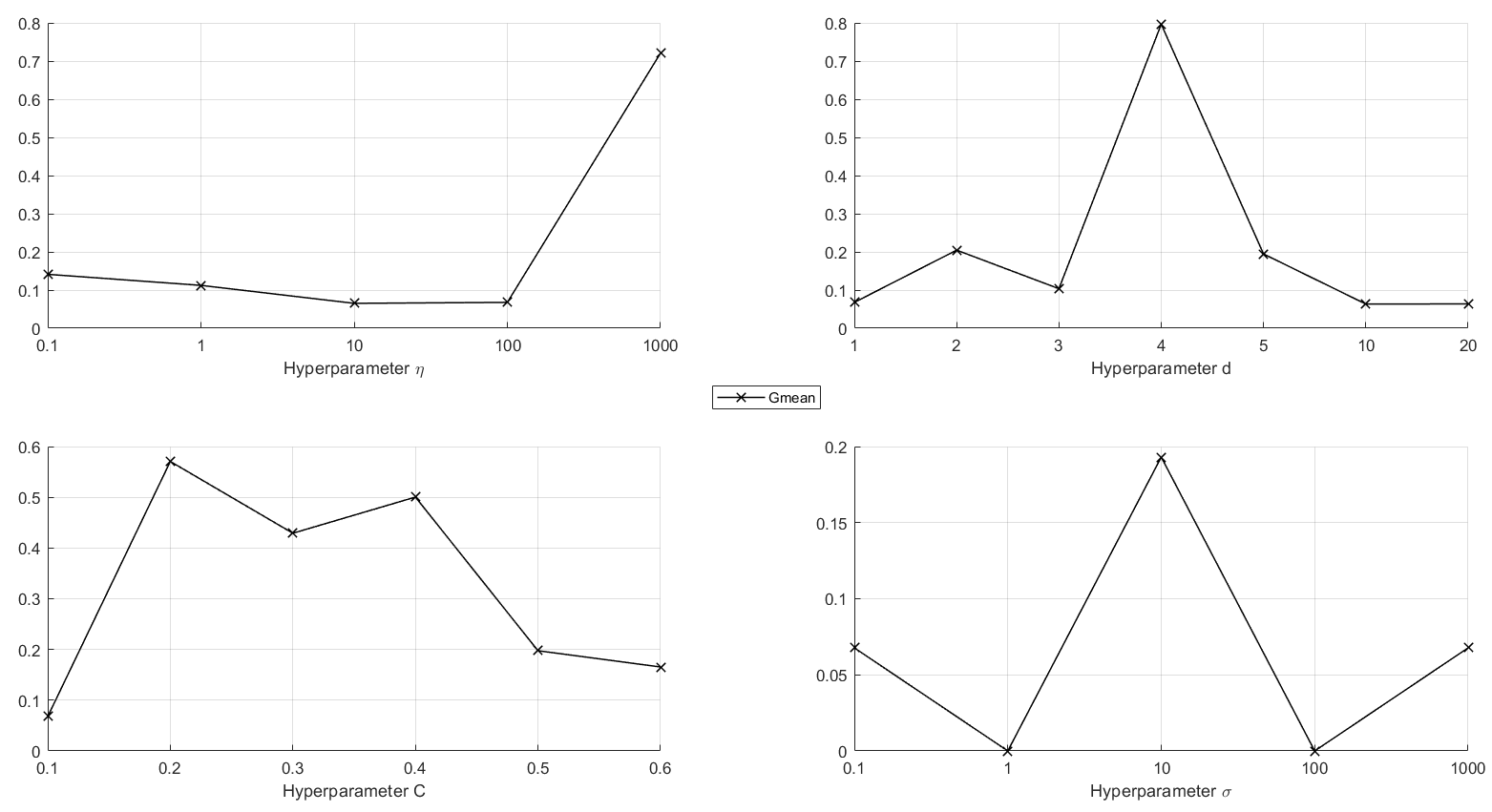}
	\caption{Hyperparameters sensitivity analysis for GESSVDD-GR-I-min on MNIST dataset (target class=0)}
\end{figure}
\begin{figure}
	\centering
	\includegraphics[scale=0.38]{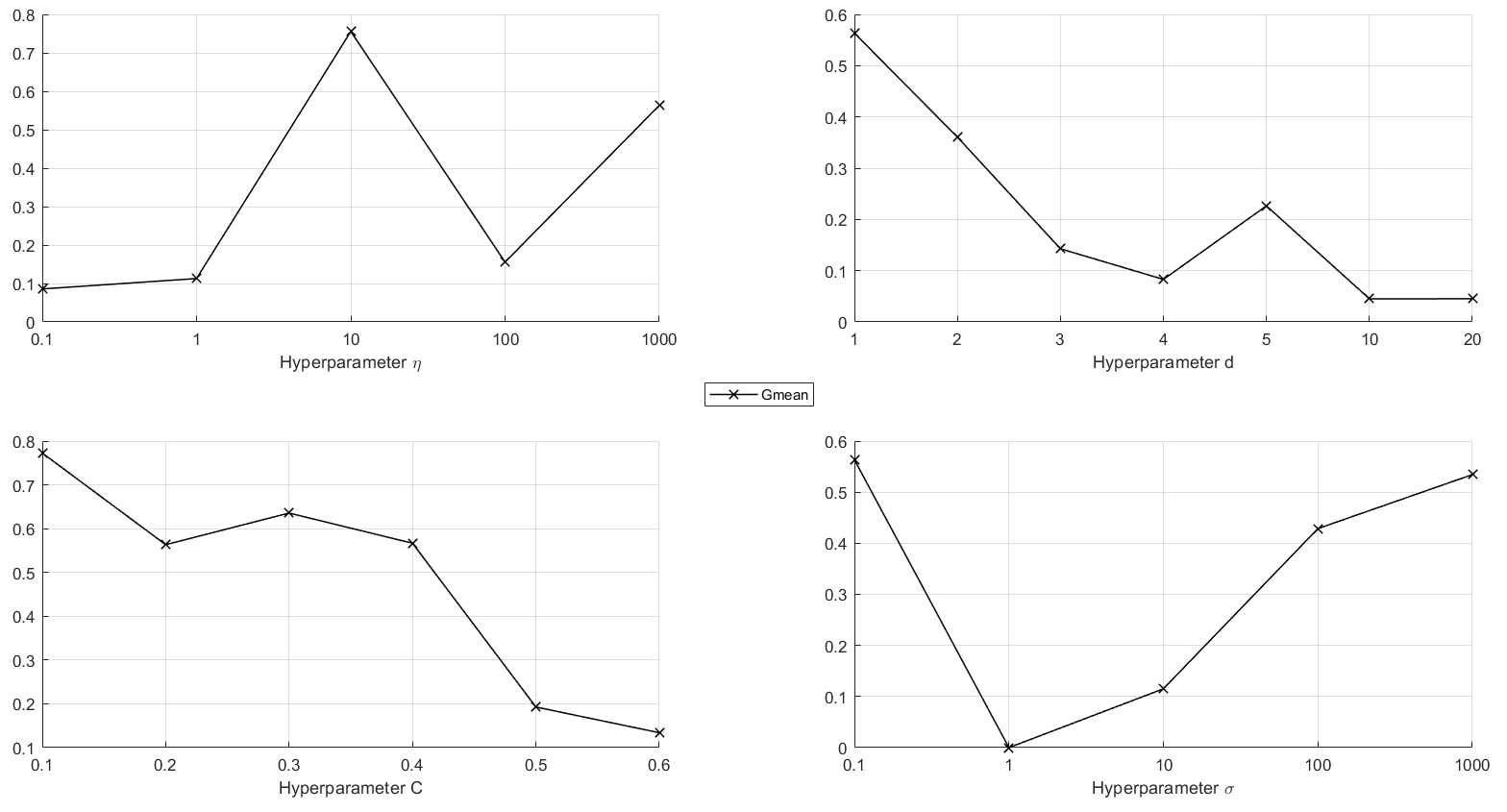}
	\caption{Hyperparameters sensitivity analysis for GESSVDD-GR-I-max on MNIST dataset (target class=0)}
\end{figure}
\begin{figure}
	\centering
	\includegraphics[scale=0.38]{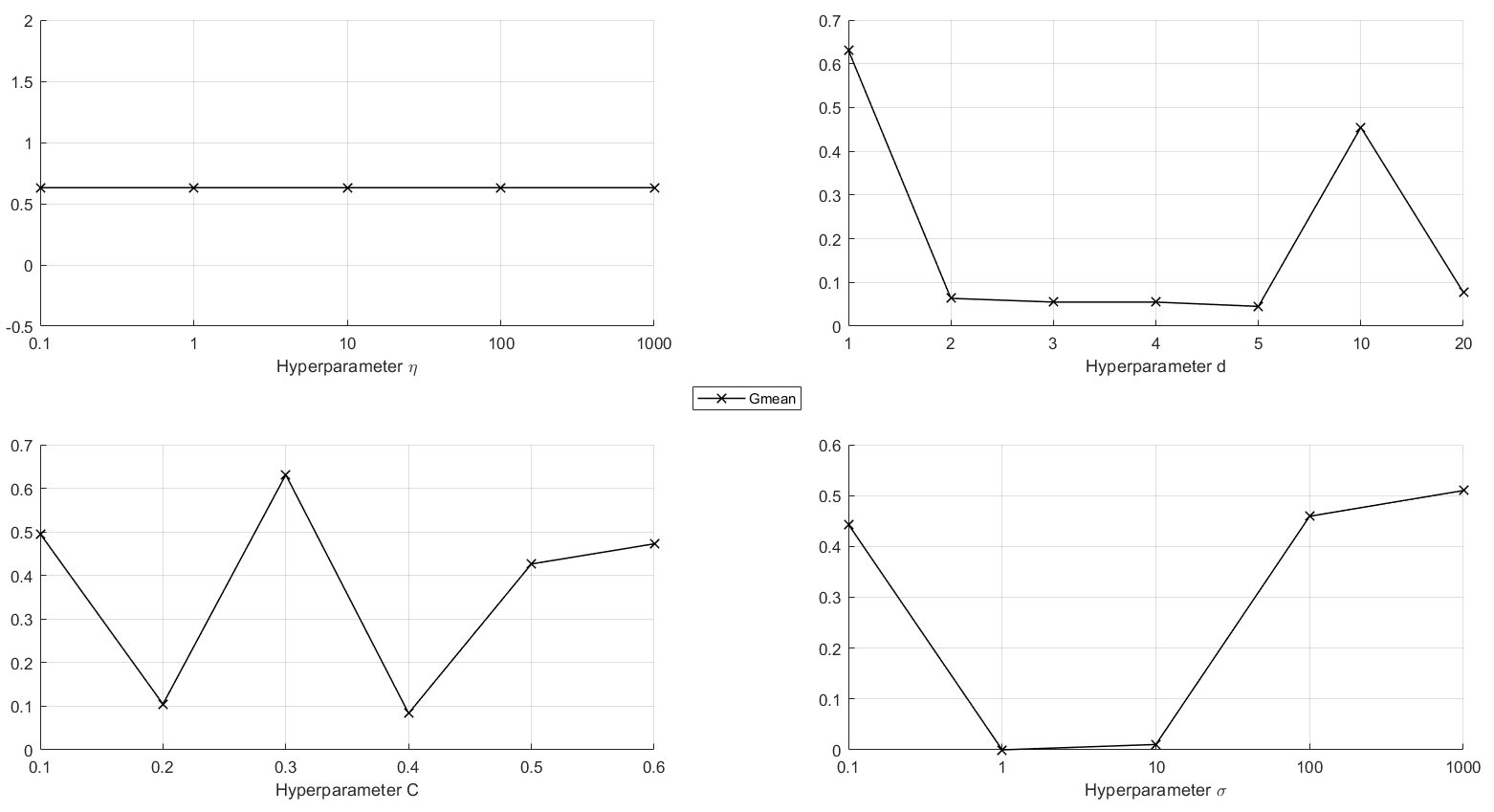}
	\caption{Hyperparameters sensitivity analysis for GESSVDD-$\mathcal{S}$-I-min on MNIST dataset (target class=0)}
\end{figure}
\begin{figure}
	\centering
	\includegraphics[scale=0.38]{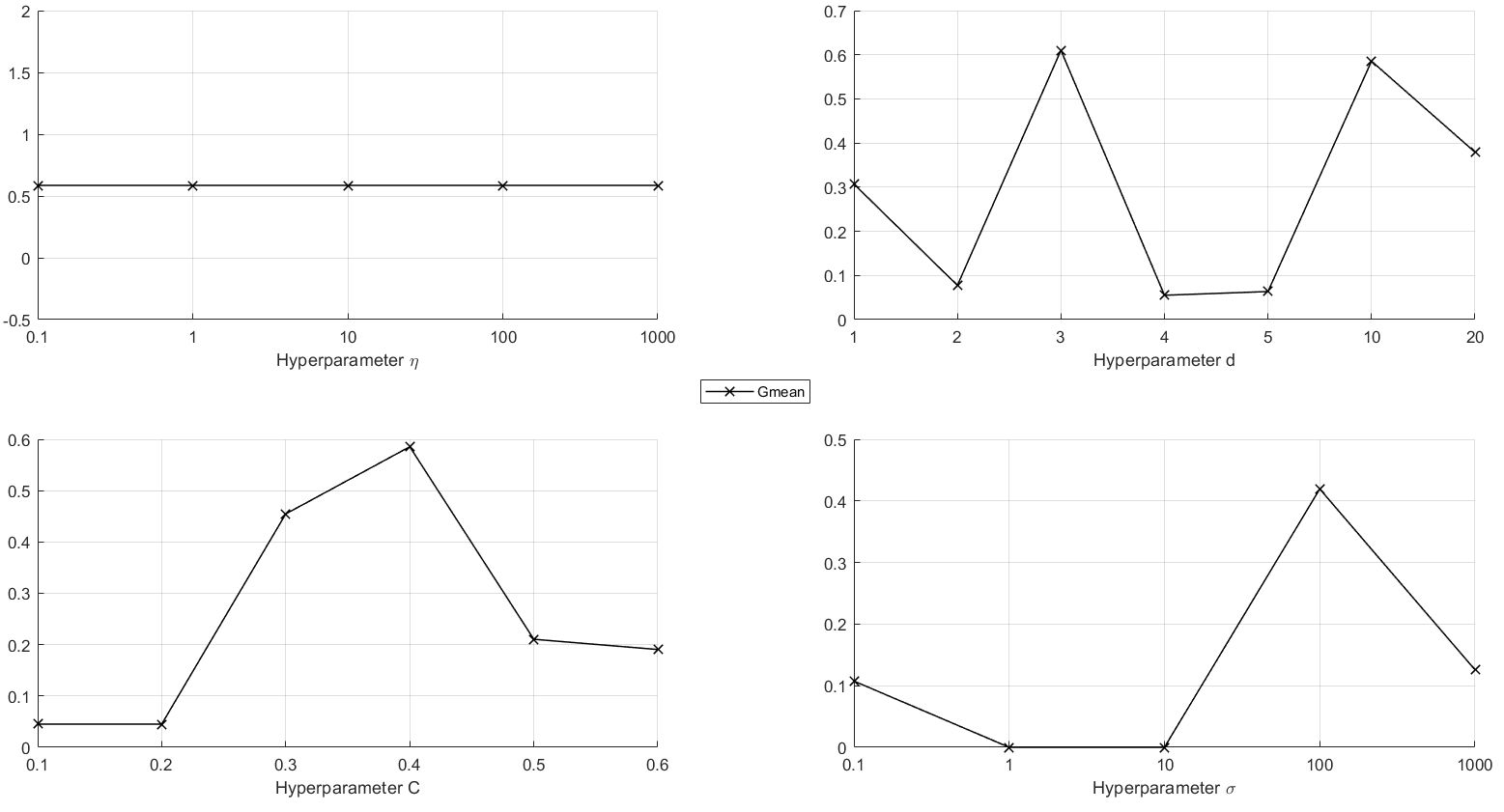}
	\caption{Hyperparameters sensitivity analysis for GESSVDD-$\mathcal{S}$-I-max on MNIST dataset (target class=0)}
\end{figure}
\begin{figure}
	\centering
	\includegraphics[scale=0.38]{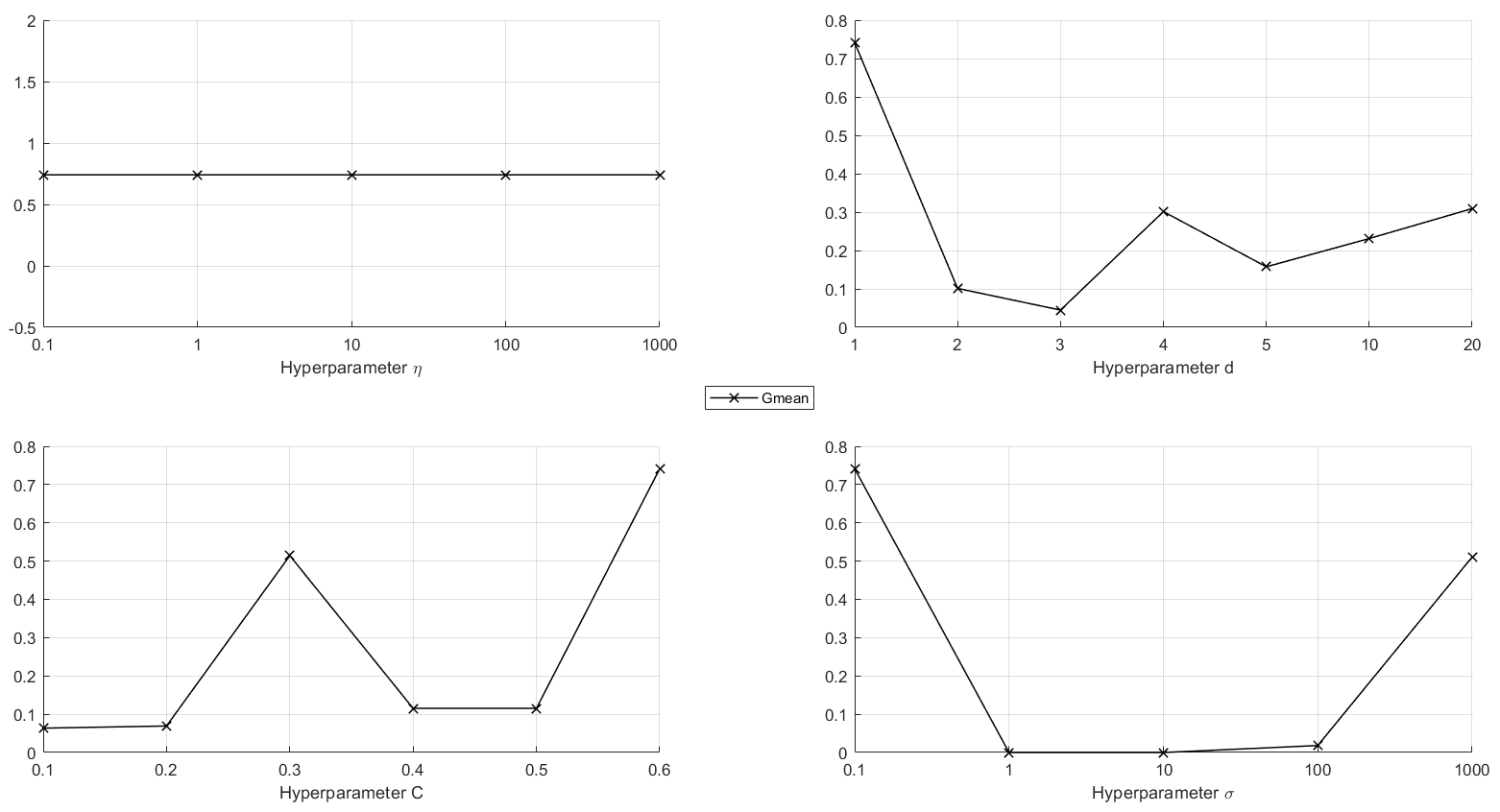}
	\caption{Hyperparameters sensitivity analysis for GESSVDD-SR-I-min on MNIST dataset (target class=0)}
\end{figure}
\begin{figure}
	\centering
	\includegraphics[scale=0.38]{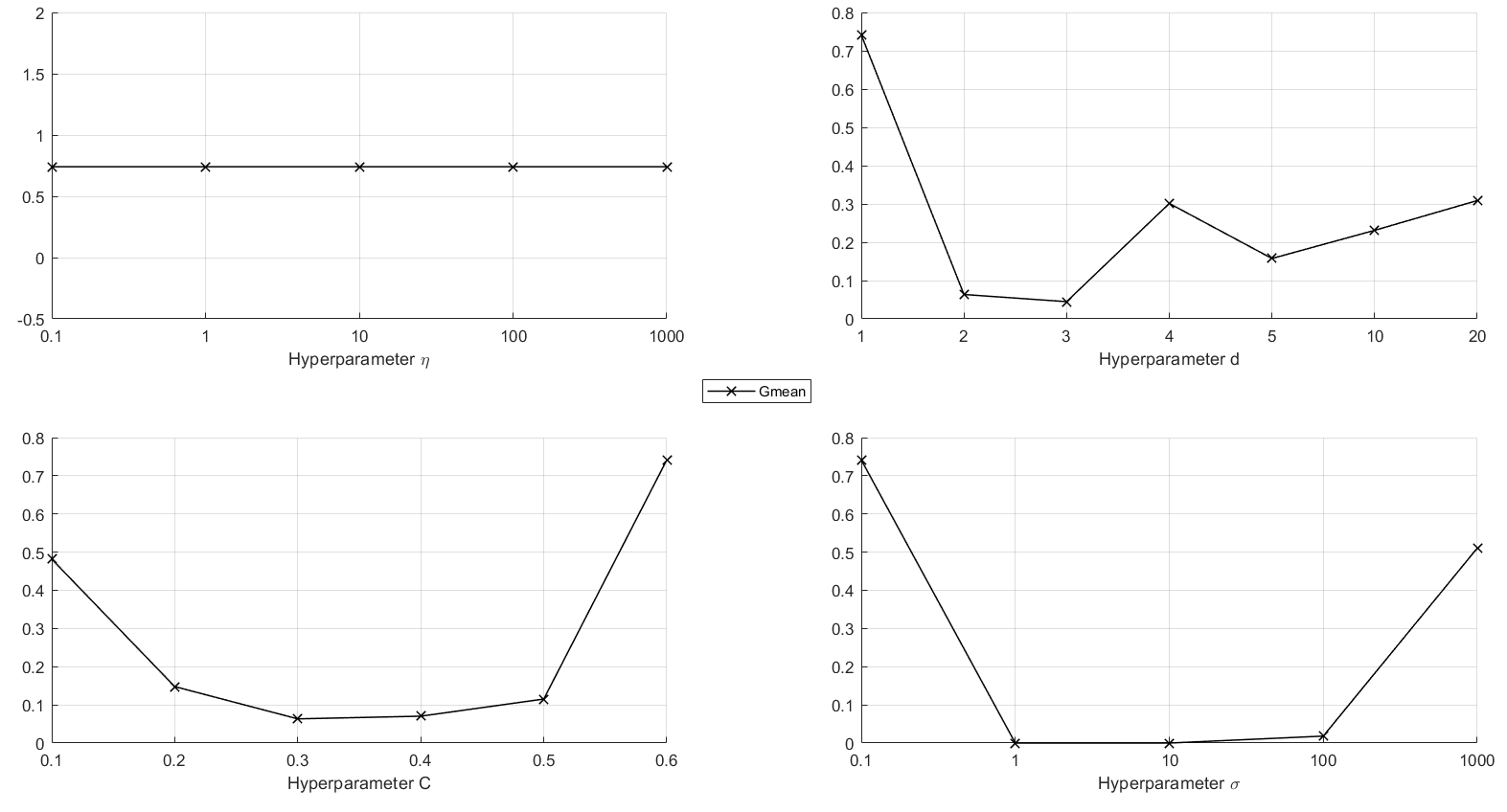}
	\caption{Hyperparameters sensitivity analysis for GESSVDD-SR-I-max on MNIST dataset (target class=0)}
\end{figure}
\begin{figure}
	\centering
	\includegraphics[scale=0.38]{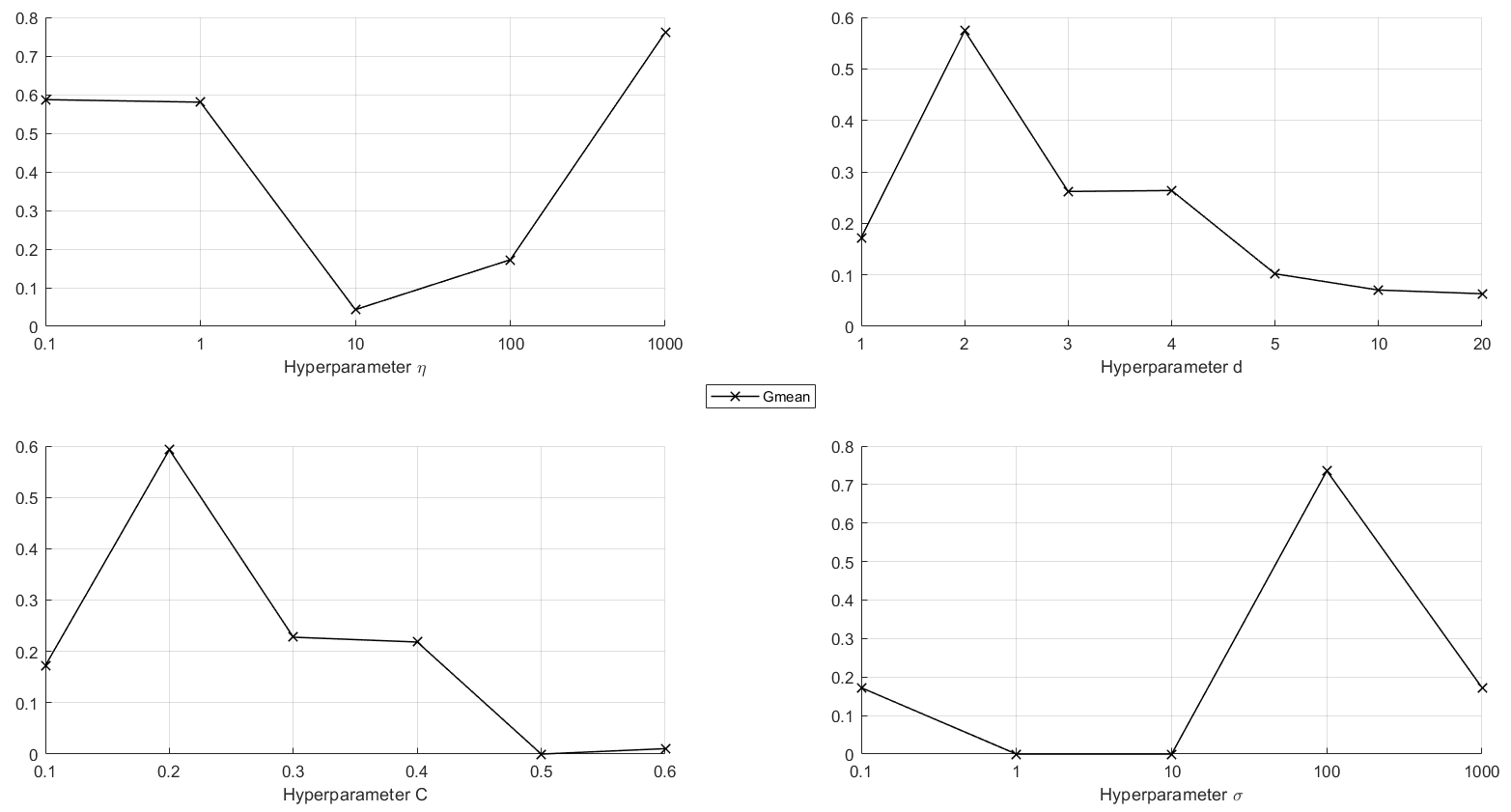}
	\caption{Hyperparameters sensitivity analysis for GESSVDD-GR-0-min on MNIST dataset (target class=0)}
\end{figure}
\begin{figure}
	\centering
	\includegraphics[scale=0.38]{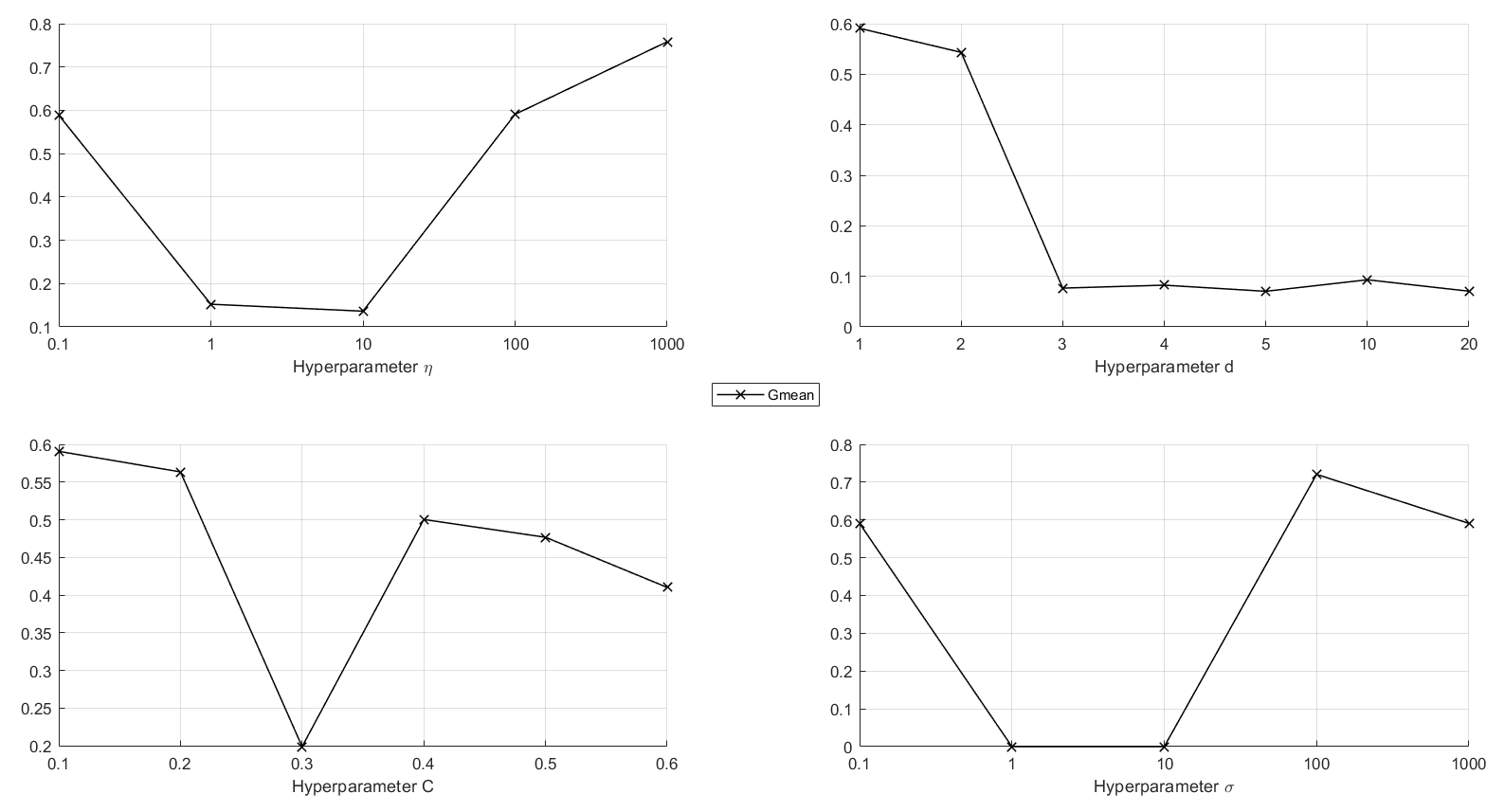}
	\caption{Hyperparameters sensitivity analysis for GESSVDD-GR-0-max on MNIST dataset (target class=0)}
\end{figure}
\begin{figure}
	\centering
	\includegraphics[scale=0.38]{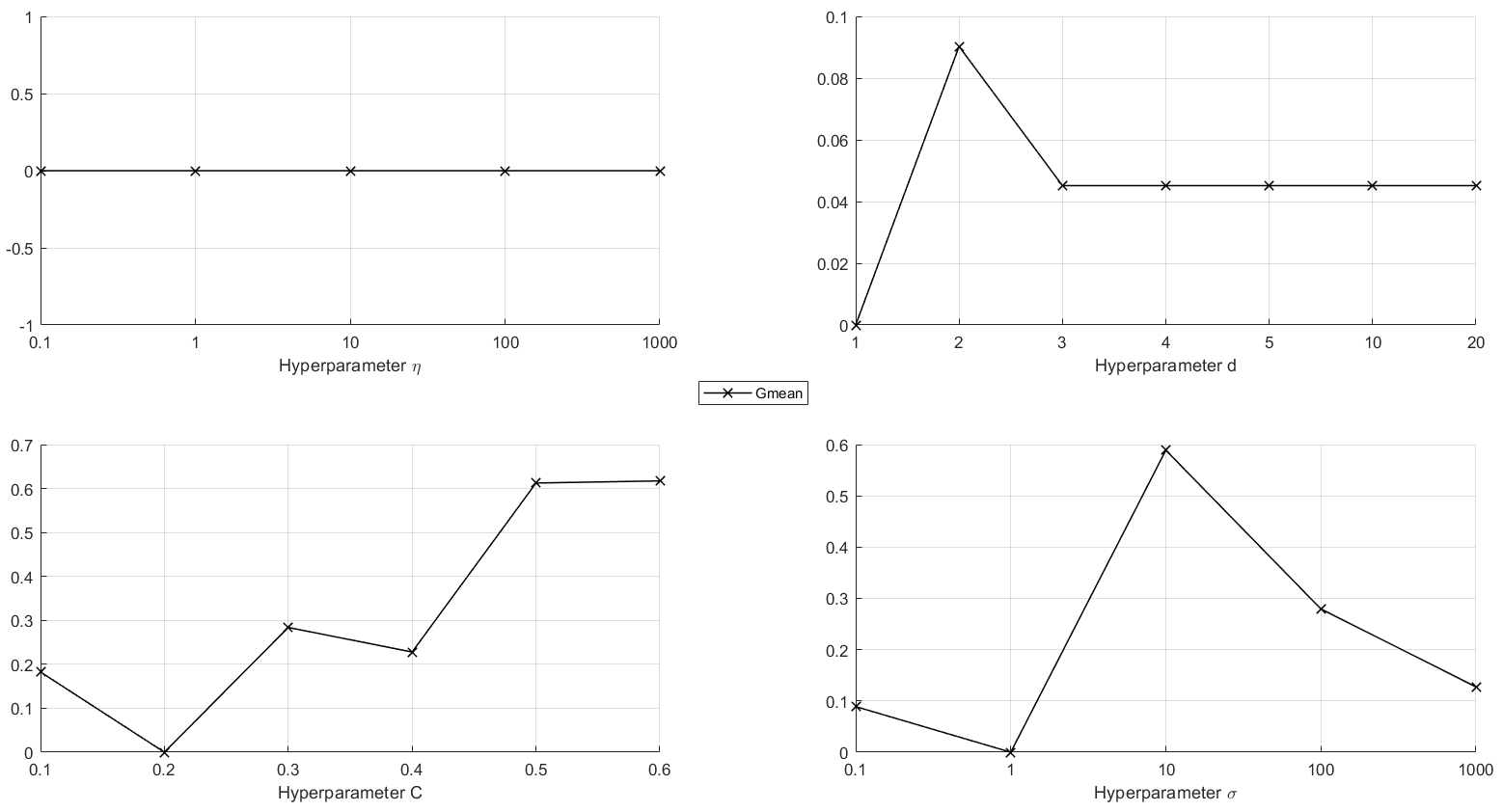}
	\caption{Hyperparameters sensitivity analysis for GESSVDD-$\mathcal{S}$-0-min on MNIST dataset (target class=0)}
\end{figure}
\begin{figure}
	\centering
	\includegraphics[scale=0.38]{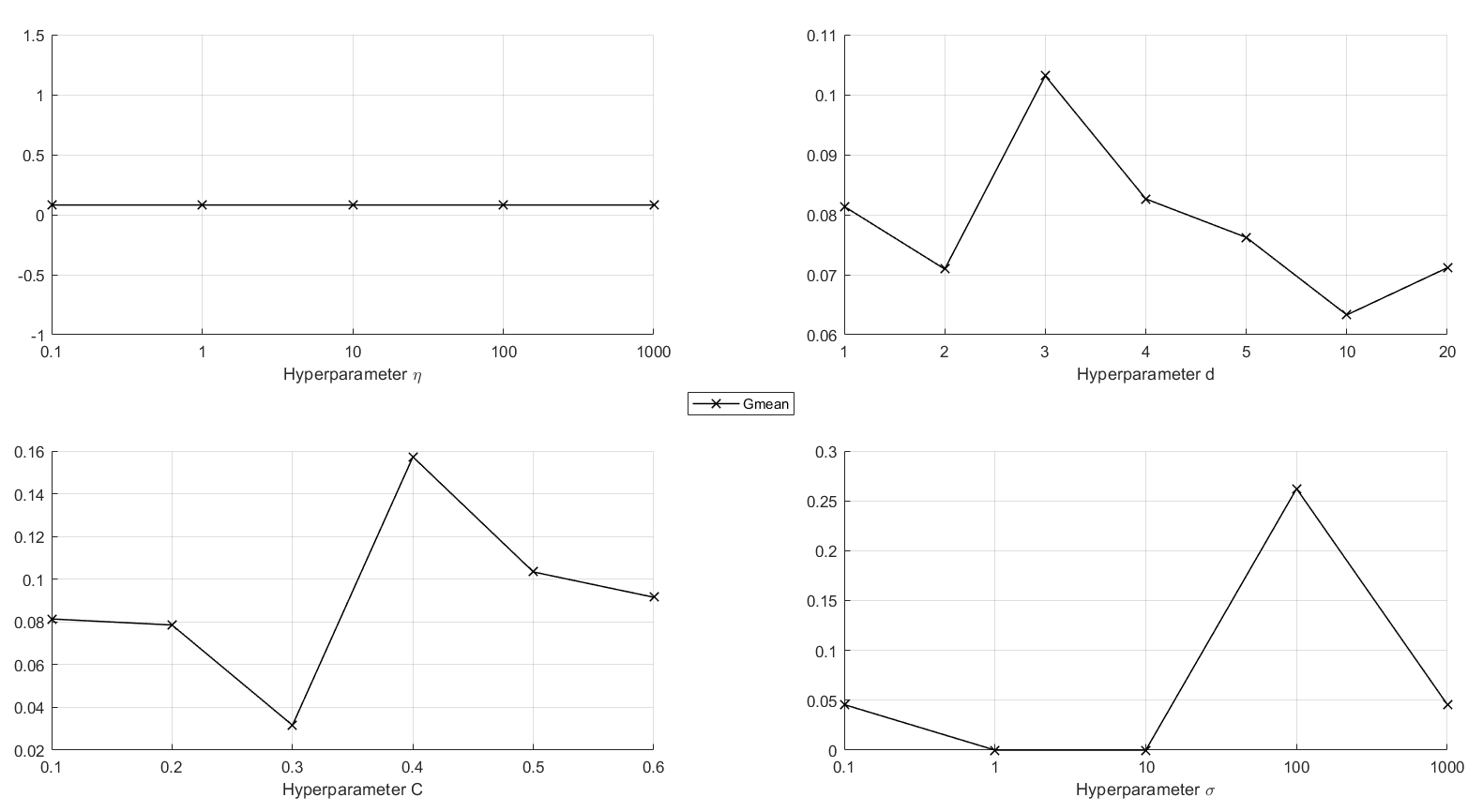}
	\caption{Hyperparameters sensitivity analysis for GESSVDD-$\mathcal{S}$-0-max on MNIST dataset (target class=0)}
\end{figure}
\end{document}